\documentclass[11pt, a4paper, logo, one column]{craftjarvis}

\pdftrailerid{redacted}

\makeatletter
\renewcommand\bibentry[1]{\nocite{#1}{\frenchspacing\@nameuse{BR@r@#1\@extra@b@citeb}}}
\makeatother

\usepackage[utf8]{inputenc} 
\usepackage[T1]{fontenc}    
\usepackage{url}            
\usepackage{booktabs}       
\usepackage{amsfonts}       
\usepackage{nicefrac}       
\usepackage{microtype}      
\usepackage{subfigure}
\usepackage{graphicx}

\usepackage{microtype}

\usepackage{booktabs} 

\usepackage{algorithm} 
\usepackage{algpseudocode} 
\usepackage{hyperref}

\usepackage{amsmath,amsfonts,bm}

\def\eqref#1{equation~\ref{#1}}

\def\1{\bm{1}}

\def\vx{{\bm{x}}}

\DeclareMathAlphabet{\mathsfit}{\encodingdefault}{\sfdefault}{m}{sl}
\SetMathAlphabet{\mathsfit}{bold}{\encodingdefault}{\sfdefault}{bx}{n}

\newcommand{\E}{\mathbb{E}}

\newcommand{\R}{\mathbb{R}}

\usepackage{mathtools}
\usepackage{amsthm}
\usepackage{enumerate}   

\usepackage[capitalize,noabbrev]{cleveref}
\usepackage[normalem]{ulem}
\usepackage{booktabs}
\theoremstyle{plain}

\theoremstyle{definition}

\theoremstyle{remark}

\usepackage{hhline}
\usepackage{diagbox}
\usepackage{arydshln}

\usepackage{ragged2e}
\usepackage{makecell}
\usepackage{tcolorbox}
\usepackage{CJKutf8}
\usepackage[utf8]{inputenc}

\usepackage{booktabs}
\usepackage{wrapfig}
\usepackage{graphicx}
\usepackage{subcaption}

\usepackage{amsmath}
\usepackage{amssymb}
\usepackage{multirow}
\usepackage{color, colortbl}
\usepackage{subcaption}
\usepackage{pifont}
\usepackage{todonotes}
\usepackage{bbding}
\usepackage{wasysym}
\usepackage{color}
\usepackage{tabularray}

\usepackage{xcolor}
\definecolor{google_yellow}{HTML}{ffbc32}
\definecolor{google_red}{HTML}{f4433c}
\definecolor{google_blue}{HTML}{2d85f0}
\definecolor{Gray}{gray}{0.9}
\definecolor{mygreen}{rgb}{0.0, 0.5, 0.0}
\definecolor{myred}{rgb}{0.8, 0.25, 0.33}
\definecolor{myblue}{rgb}{0.19, 0.55, 0.91}
\definecolor{uclablue}{rgb}{0.15, 0.45, 0.68}
\definecolor{boxgreen}{rgb}{0.02, 0.66, 0.02}
\definecolor{boxred}{rgb}{0.66, 0.1, 0.1}
\definecolor{boxblue}{rgb}{0.01, 0.01, 0.73}
\definecolor{mygray}{gray}{0.4}
\usepackage[toc,page,header]{appendix}
\usepackage{minitoc}

\usepackage{listings}
\newcommand\Tstrut{\rule{0pt}{2.4ex}}         
\newcommand\Bstrut{\rule[-0.9ex]{0pt}{0pt}}   
\usepackage{amssymb}
\usepackage{xspace}

\usepackage[authoryear, sort&compress, round]{natbib} 

\title{CLoG: Benchmarking {C}ontinual {L}earning {o}f Image {G}eneration Models}

\reportnumber{} 

\renewcommand{\today}{May 2024} 

\author[1]{Haotian~Zhang$^\star$}
\author[1]{Junting~Zhou$^\star$}
\author[1]{Haowei~Lin$^\star$}
\author[1]{Hang~Ye$^\star$}
\author[1]{Jianhua~Zhu$^\star$}
\author[1]{\\Zihao~Wang}
\author[1]{Liangcai~Gao}
\author[1]{Yizhou~Wang}
\author[1]{Yitao~Liang}

\affil[1]{Peking~University}

\begin{abstract}
Continual Learning (CL) poses a significant challenge in Artificial Intelligence, aiming to mirror the human ability to incrementally acquire knowledge and skills. While extensive research has focused on CL within the context of classification tasks, the advent of increasingly powerful generative models necessitates the exploration of \underline{C}ontinual \underline{L}earning \underline{o}f \underline{G}enerative models (CLoG). This paper advocates for shifting the research focus from classification-based CL to CLoG. We systematically identify the unique challenges presented by CLoG compared to traditional classification-based CL. We adapt three types of existing CL methodologies—replay-based, regularization-based, and parameter-isolation-based methods—to generative tasks and introduce comprehensive benchmarks for CLoG that feature great diversity and broad task coverage. Our benchmarks and results yield intriguing insights that can be valuable for developing future CLoG methods. Additionally, we will release a codebase designed to facilitate easy benchmarking and experimentation in CLoG publicly at \url{https://github.com/linhaowei1/CLoG}. We believe that shifting the research focus to CLoG will benefit the continual learning community and illuminate the path for next-generation AI-generated content (AIGC) in a lifelong learning paradigm.
\end{abstract}

\begin{document}

\correspondingauthor{Yitao Liang<yitaol@pku.edu.cn>, Haowei Lin <linhaowei@pku.edu.cn>\\
$^\star$Lead Authors. 
}


\maketitle

\section{Introduction}

\begin{figure}[t]
  \centering
  \includegraphics[width=\textwidth]{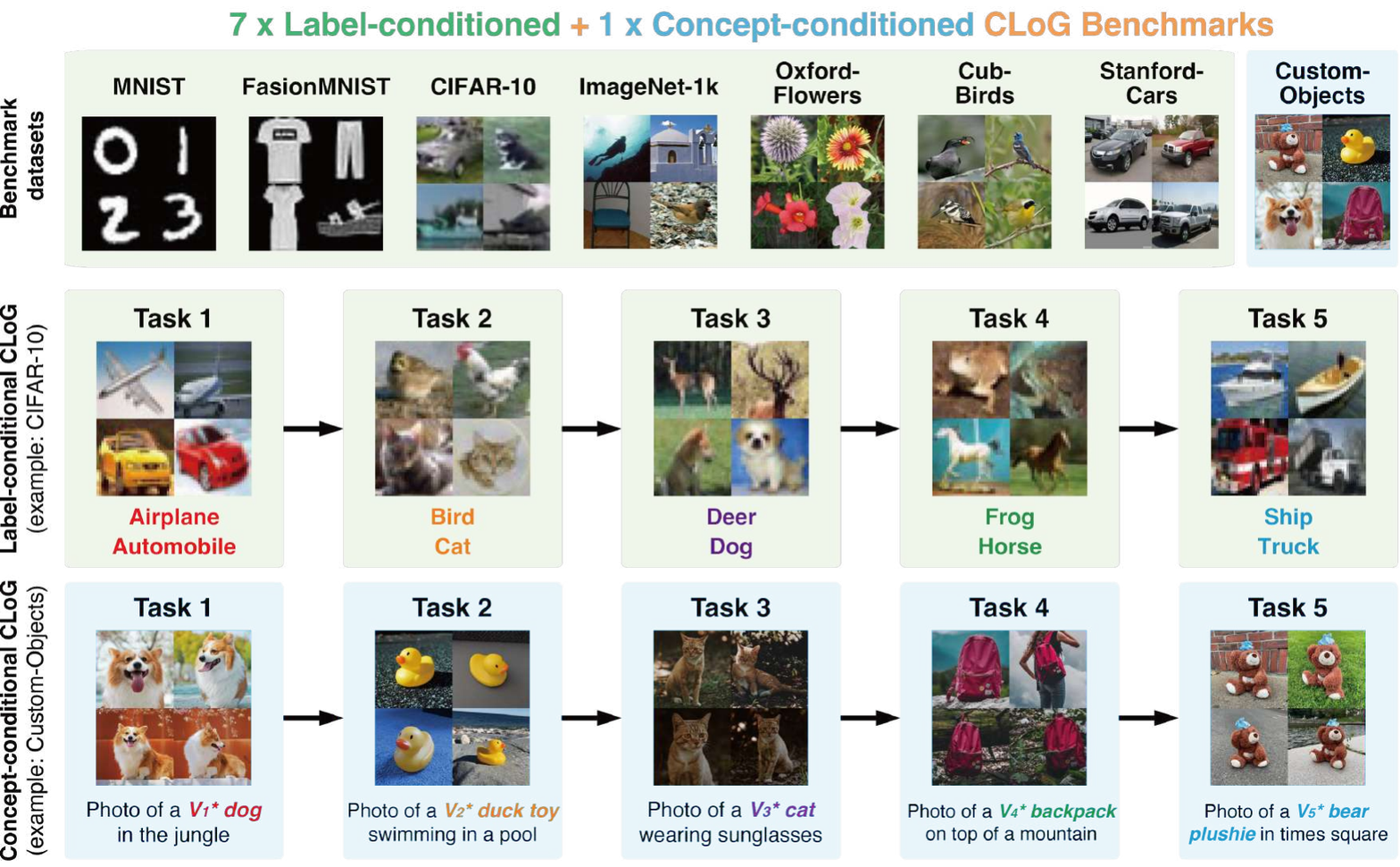}
  \caption{\textbf{Overview of benchmarks.} Seven label-conditioned and one concept-conditioned CLoG benchmarks are studied, with details presented in~\cref{tab.dataset_comparison} and~\cref{sec:task_selection}. \emph{Label-conditioned CLoG} learns a sequence of generation tasks conditioned on label indices. \emph{Concept-conditional CLoG} learns to synthesize a sequence of concepts (denoted as $V_i^*$ for the $i$th concept) given arbitrary text prompts.}
  \label{fig:main1}
\end{figure}

\begin{figure}[t]
  \centering
  \includegraphics[width=\textwidth]{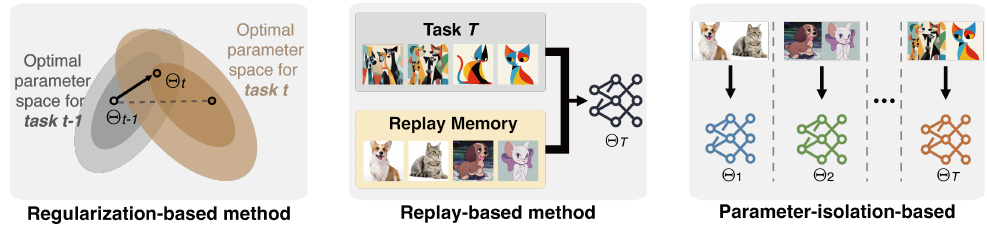}
  \caption{\textbf{Overview of baselines.} Three types of CL baselines are adapted to CLoG, which include \emph{regularization-based}, \emph{replay-based}, and \emph{parameter-isolation-based} methods, resulting in a total of twelve different CLoG baselines. The detailed information on the baselines are in~\Cref{sec:baseline_setup}.}
  \vspace{-2ex}
  \label{fig:main2}
\end{figure}

The development of Artificial Intelligence Generated Content (AIGC) marks a paradigm shift from classification-based applications, such as image recognition~\citep{alexnet, vgg, googlenet, resnet, vit} and text classification~\citep{hdltex, Jiang2016TextCB, rcnn, minaee2021deep, kalchbrenner2014convolutional, yang-etal-2016-hierarchical}, to powerful generative models. Advances in generative learning, including GANs~\citep{goodfellow2014generative, radford2016unsupervised, karras2019stylebased, park2019semantic, zhu2020unpaired} and diffusion models~\citep{sohldickstein2015deep, ho2020denoising, song2021scorebased}, have enabled AI to create novel content, such as images~\citep{isola2018imagetoimage, ho2020denoising, dhariwal2021diffusion, stablediffusion, ramesh2021zeroshot, saharia2022photorealistic}, music~\citep{huang2018music, oord2016wavenet, Yu_2021, donahue2019adversarial}, video~\citep{singer2023makeavideo, anonymous2022imagen, xing2023simda, lu2023vdt, wang2024lavie}, and molecules~\citep{xu2022geodiff, NEURIPS2022_994545b2}. This shift significantly broadens AI's impact across various fields in both industry and daily life.

Continual learning (CL), which involves AI systems incrementally mastering a sequence of tasks $\mathcal T^{(1)}, \mathcal T^{(2)}, …, \mathcal T^{(T)}$~\citep{clsurveybingliu, clsurveyMat}, is a crucial and well-regarded challenge in AI research. A main assumption of continual learning is that once a task is learned, its training data becomes inaccessible, leading to a phenomenon known as \emph{catastrophic forgetting} (CF)~\citep{CF}. Catastrophic forgetting is characterized by a decline in performance on previously learned tasks due to updates in model parameters during the acquisition of new tasks~\citep{CF, clsurveybingliu, clsurveyMat}. Recent advancements have significantly mitigated CF in various continual learning settings~\citep{serrà2018overcoming, supsup,sun2020lamal, ke2021continual, madotto-etal-2021-continual, ke2022continual}, however, most existing CL approaches and frameworks have been tailored primarily for \emph{classification-based models}~\citep{van2019three}. Given the rising importance of generative models, we believe that \emph{now is an opportune time to pivot the research focus of CL community towards the Continual Learning of Generative models (\textbf{CLoG})}. Compared to classification-based CL that often learns a sequence of categorical distributions ($i.e.$, classification tasks) via standard classifier architecture (encoder + classification head)~\citep{yan2021dynamically,wang2022beef,lin2024class}, CLoG typically necessitates the use of sophisticated generative models, such as VAE~\citep{kingma2013auto}, GAN~\citep{goodfellow2014generative}, or score-based models~\citep{song2020score}, to model the complicated data distributions. The diverse model architectures, which are generally more complex to optimize or incrementally expand during continual learning, make CLoG a more challenging setting than classification-based CL.

The earliest related work of CLoG proposes \emph{generative replay} (GR)~\citep{dgr} for classification-based CL. GR utilizes a continually trained generator to synthesize data from previous tasks, thereby preventing forgetting when training a continual learning classifier. The primary goal of GR methods, however, is to enhance classification performance rather than the quality of the generated data. In past years, pioneering studies in CLoG have emerged~\citep{Zhai2019LifelongGC,Cong2020GANMW,sun2024create}, but they often lacked a unified evaluation protocol and tested on distinctive distinct tasks, making the comparison difficult. Additionally, the diversity in model architectures, data processing techniques, training pipelines, and evaluation metrics complicates fair comparisons between different CLoG methods. This contrasts with classification-based CL, where these choices are relatively standardized~\citep{lomonaco2021avalanche,de2021continual,wang2023comprehensive}.

In this paper, we establish a foundational framework for studying CLoG. Initially, we define the problem of CLoG and delve deeply into it by leveraging insights from the existing research on classification-based CL (\cref{sec:from_traditional_cl_to_clog}). We then meticulously develop benchmarks for CLoG, focusing on task selection (\cref{sec:task_selection}), baseline setup (\cref{sec:baseline_setup}), metrics design (\cref{sec:metrics_design}), and training specifics (\cref{sec:training_specifics}). Our benchmarks are designed to encompass a broad spectrum of tasks and scenarios, which include varying image resolutions, knowledge transfer capabilities, sizes of training sets, and input conditions, etc. We maintain a clean baseline set by adapting representative CL methods to CLoG, employ unified evaluation metrics, and enhance the efficiency of evaluation process by focusing only on the specifics crucial for CL. Our benchmarks provide valuable insights and are intended to inspire further advancements in CLoG methodologies. We will also release our extensible codebase for the benefit of CLoG research to the public.

In addition to establishing a foundational framework for CLoG, this paper also aims to reflect on current CL research. The shift from classification-based CL to CLoG is driven by the emergence of generative foundation models (e.g., Sora~\citep{videoworldsimulators2024}, GPT-4~\citep{Achiam2023GPT4TR}, Gemini~\citep{team2023gemini}), which integrate task-specific models into a unified generalist framework. As traditional CL methods primarily focus on discrete tasks, the development of CL methods that are tailored for foundation models presents an urgent and relevant challenge.
Additionally, we discuss the potential applications of CLoG in diverse modalities and outline several promising directions for future advancements in CLoG methodologies.

\section{From Traditional CL to CLoG}
\label{sec:from_traditional_cl_to_clog}

In this section, we offer an overview of continual learning (CL). We will discuss the ``three-types'' of CL methods: regularization-based, replay-based, and parameter-isolation-based methods in~\cref{sec:traditional_cl_review}. Then we will contextualize the approach and insights into continual learning for generative models (CLoG) within the existing framework of CL research, as detailed in Section \ref{sec:formulate_clog}.

\subsection{Continual Learning}
\label{sec:traditional_cl_review}

\paragraph{Formulation of general CL.} CL learns a sequence of tasks $\mathcal T^{(1)}, \mathcal T^{(2)},\cdots, \mathcal T^{(T)}$ incrementally. Each task $\mathcal T^{(t)}$ has an input space $\mathcal X^{(t)}$, an output space $\mathcal Y^{(t)}$, and a training set $\mathcal D^{(t)}=\{(\boldsymbol{x}_j^{(t)},\boldsymbol{y}_j^{(t)})\}_{j=1}^{|\mathcal D^{(t)}|}$ drawn $i.i.d.$ from distribution $\mathcal P_{\mathcal X^{(t)}\mathcal Y^{(t)}}$. The goal of continual learning is to learn a function $f:\cup_{t=1}^T\mathcal X^{(t)}\to \cup_{t=1}^T\mathcal Y^{(t)}$ that can achieve good performance on each task $\mathcal T^{(t)}$.

A main assumption of CL is that once a task is learned, its training data $\mathcal D^{(t)}$ is no longer accessible (or with limited access). This assumption simulates the learning process of humans but also causes \emph{catastrophic forgetting} for machine learning models, which refers to performance degradation of previous tasks due to parameter updates in learning each new task~\citep{McCloskey1989}. The existing CL methods which aim to prevent forgetting can be roughly categorized into three types:

(1) \textbf{Regularization-based Methods}: The idea of this family is to add regularization to penalize changes to important parameters learned for previous tasks in learning a new task~\citep{ewc,si,mas,RWalk,LwF}.

(2) \textbf{Replay-based Methods}: These methods store a small subset of training data from previous tasks~\citep{lopez2017gradient,chaudhry2019tiny,riemer2018learning,rebuffi2017icarl,kemker2017fearnet} or learn a data generator to synthesize pseudo data~\citep{dgr,ganmemory,mergan,zhu2021prototype} of previous tasks. The saved data, the synthesized data, and the new task data are both used in training. 

(3) \textbf{Parameter-isolation Based Methods}: These methods allocate task-specific parameters to prevent subsequent tasks from interfering the previously learned parameters~\citep{wortsman2020supermasks,serra2018overcoming,fernando2017pathnet,rusu2016progressive,aljundi2017expert,masana2021ternary,lin2024class}.

\subsection{Continual learning of generative models (CLoG)}
\label{sec:formulate_clog}


\paragraph{Formulation of CLoG.} CLoG learns a sequence of generation tasks $\mathcal T^{(1)}, \mathcal T^{(2)},\cdots, \mathcal T^{(T)}$ incrementally. Each task $\mathcal T^{(t)}$ has an input space $\mathcal X^{(t)}$ (generation conditions) and output space $\mathcal Y$ (generation targets), and a training set $\mathcal D^{(t)}=\{(\bm x_j^{(t)}, \bm y_j^{(t)})\}_{j=1}^{|\mathcal D^{(t)}|}$ drawn $i.i.d.$ from distribution $\mathcal P_{\mathcal X^{(t)}\mathcal Y^{(t)}}$. The goal of CLoG is to learn a mapping $f:\cup_{t=1}^T \mathcal X^{(t)}\to \cup_{t=1}^T \mathcal Y^{(t)}$ that can achieve good performance on each task $\mathcal T^{(t)}$. The generation conditions can be in various forms such as text~\citep{controlgan, zhang2023adding}, images~\citep{Zhai2021HyperLifelongGANSL, zhang2023adding}, or label indices~\citep{ho2021classifierfree}, while the generation targets can be various modalities such as images~\citep{ramesh2021zeroshot, saharia2022photorealistic}, audio~\citep{huang2018music, oord2016wavenet}, or 3D objects~\citep{shi2023deep, zeng2022lion}. As an initial step towards CLoG, we only focus on image generation conditioned on text or label indices in this paper.

\paragraph{Comparison between CLoG and classification-based CL.} The key difference between classification and generative tasks lies in the input space $\mathcal X$ and output space $\mathcal Y$. In image generation, the input $\bm x$ may be some label conditions ($e.g.$, one-hot class) or instructions (text or images), and the output $\boldsymbol{y}\in \R^{C\times H\times W}$ should be images ($C,H,W$ denote the number of channels, height, and width). 
CLoG is more challenging as its output space inherently possesses a significantly larger cardinality, while the output of classification-based CL is typically limited to discrete class indices. Thus the model architectures also diverge; classification-based CL typically requires only a simple mechanism to model a categorical distribution, such as a linear mapping or MLP head~\citep{popescu2009multilayer}, while CLoG necessitates the use of more sophisticated generative models, such as VAE~\citep{kingma2013auto}, GAN~\citep{goodfellow2014generative}, or score-based models~\citep{song2020score}. These generative models are generally more complex to optimize and incrementally expand within a continual learning framework compared to classification architectures~\citep{roth2017stabilizing,song2020improved}.

\paragraph{Remarks on text generation in CL.} CL of text generation tasks~\citep{shi2024continual,wu2024continual} has garnered increasing interest in the so-called “post-LLM era”~\citep{lin2024selecting}. Including CL of text generation tasks in CLoG is logical, given their generative nature. However, text generation typically operates under the framework of ``next token prediction''~\citep{radford2019language}, which substantially differs from the typical probabilistic generative models. Specifically, text generation models the conditional data distribution, $\mathbf{P}(\bm{y}|\bm{x})$, through autoregressive generation of the form $\prod_{j=1}^{|\bm{y}|} \mathbf{P}(\bm{y}_j|\bm{y}_1, \bm{y}_2, \cdots, \bm{y}_{j-1}, \bm{x})$, with $\bm{y}_0$ representing the ``start of sentence'' token and $\bm y_j$ the $j$th token of $\bm y$. This approach simplifies the modeling of complex data distributions into predicting a sequence of categorical distributions over the vocabulary space. Existing studies have shown that text generation tasks tend to exhibit less forgetting when integrated into CL frameworks, indicating a potentially smoother adaptation to continual learning~\citep{yin2022contintin,shao2023class,cao2024generative}. While we advocate for the inclusion of text generation in CLoG due to its alignment with CLoG’s formulation, the primary focus of this paper remains on general probabilistic generative modeling due to its broader applications and the insufficient attention it has received in research.

\section{Benchmark Design}

In this section, we show how to design a foundational benchmark for the study of CLoG. Our design framework is structured around four key research questions, which are addressed in each subsection:
\begin{itemize}
    \item[($Q_1$)] \emph{How can we select CLoG tasks that are both diverse and representative?}
    \item[($Q_2$)] \emph{How should we establish baselines, considering the current advancements in CL research?}
    \item[($Q_3$)] \emph{What metrics should be employed to effectively and efficiently assess the CLoG methods?}
    \item[($Q_4$)] \emph{How can we implement different CLoG methods fairly, given the variety of training specifics?} 
\end{itemize}

\subsection{Task selection}
\label{sec:task_selection}

The fundamental criterion for choosing tasks for our benchmark is to ensure they are \emph{diverse} and \emph{representative}. Diversity is crucial for evaluating the CLoG methods \underline{across various dimensions}, making them relevant for different real-world applications. Selecting representative tasks is essential for \underline{efficiency} since executing numerous redundant tasks can be resource-intensive and unproductive.

\paragraph{Datasets} We follow traditional CL literature~\citep{rebuffi2017icarl,buzzega2020dark,kim2022multi,lin2024class,wang2022beef} to split a publicly available dataset into a sequence of tasks for CLoG. The \emph{training datasets} are summarized in~\cref{tab.dataset_comparison} and introduced as follows. \textbf{MNIST}~\citep{lecun2010mnist} contains 60,000 grayscale images of handwritten digits (0-9) in a $28\times28$ pixel format. We resize the images to $32\times32$ resolution for image generation. \textbf{FasionMNIST~\citep{fashionmnist}} consists of 60,000 grayscale images across 10 fashion categories, such as shirts, dresses, and shoes. The images are also resized from $28\times 28$ to $32\times 32$ for image generation.
\textbf{CIFAR-10~\citep{cifar10}} consists of 60,000 colored images sized at $32\times 32$ pixels, divided into 10 classes including airplane, automobile, bird, cat, deer, dog, frog, horse, ship, and truck.
\textbf{ImageNet-1k~\citep{imagenet15russakovsky}} is the most commonly used subset of ImageNet, which spans 1000 classes and contains 1,281,167 training images. We use the down-sampled $64\times64$ version following~\citet{chrabaszcz2017downsampled}.
\textbf{Oxford-Flower~\citep{nilsback2008automated}} consists of 102 flower categories that commonly occur in the United Kingdom. Each class consists of between 40 and 258 images, with a total of 7,169 images. The images are resized to $128\times 128$ for generation.
\textbf{CUB-Birds~\citep{wah2011caltech}} contains 11,788 images of 200 subcategories belonging to birds, which is a widely-used dataset for fine-grained visual categorization task. We resize the images to $128\times 128$ for generation.
\textbf{Stanford-Cars~\citep{krause20133d}} consists of 196 classes of cars with a total of 8,144 images. The images are also resized to $128\times 128$ for image generation.
\textbf{Custom-Objects~\citep{sun2024create}} contains 5 customized concepts from users with paired text-image demonstrations. Each concept has 5 demonstrations with $512\times 512$ image resolution. The task is to generate the customized concepts given arbitrary text conditions.

\paragraph{Task sequence} We partitioned MNIST, FashionMNIST, and CIFAR-10 into five tasks, assigning two classes to each task. ImageNet-1k was divided into 20 tasks with 50 classes per task, Oxford-Flower into five tasks with 20 categories per task, CUB-Birds into 10 tasks with 20 categories per task, Stanford-Cars into 14 tasks with 14 classes per task, and Custom-Objects into five tasks with one object per task. Following the random class order protocol in~\citet{rebuffi2017icarl}, we  generate five different class orders for each experiment and report their averaged metrics over five random orders. For a fair comparison, the class orderings are fixed in our experiments (see~\cref{app:random_class_ordering}). It is important to note that one dataset can be segmented into varying numbers of tasks~\citep{lin2024class} or without requiring uniformity in class~\citep{hemati2024continual}. These customized CL settings can be explored in the future, and our current benchmark focuses on addressing more fundamental challenges in CLoG for now.


\begin{table}[tbp]
    \centering
    \caption{The detailed configurations of eight CLoG benchmarks studied in this paper.}
    \label{tab.dataset_comparison}
    \resizebox{\textwidth}{!}{
        \begin{tabular}{ccccc}
            \toprule
            \multirow{2}{*}{\textbf{Dataset}} & \textbf{Image} & \textbf{\#Training Images} & \multirow{2}{*}{\textbf{\#Tasks}} & \multirow{2}{*}{\textbf{Description of Each Task}}  \\
             & \textbf{Resolution} & \textbf{per Task} & & \\ 
            \midrule
            MNIST~\citep{lecun2010mnist} & $32\times 32$ & 12,000 & 5 & Conditional generation of 2 classes of handwritten digits\\
            FashionMNIST~\citep{fashionmnist} & $32\times 32$ & 12,000 & 5 & Conditional generation of 2 classes of fashion products \\
            CIFAR-10~\citep{cifar10} & $32\times 32$ & 10,000 & 5 & Conditional generation of 2 classes of common objects\\
            ImageNet-1k~\citep{imagenet15russakovsky} & $64\times 64$ & $\sim$64,000 & 20 & Conditional generation of 50 classes of ImageNet images \\
            Oxford-Flower~\citep{nilsback2008automated} & $128\times128$ &$\sim$1,400 & 5 & Conditional generation of 20 categories of flowers\\
            CUB-Birds~\citep{wah2011caltech} & $128\times128$ & $\sim$1,200 & 10 & Conditional generation of 20 species of birds\\
            Stanford-Cars~\citep{krause20133d} &  $128\times128$ & $\sim$600 & 14 & Conditional generation of 14 classes of cars\\ 
            Custom-Objects~\citep{sun2024create} & $512\times 512$ & 5 & 5 & Generate a customized object given text conditions\\
            \bottomrule
        \end{tabular}
    }
    \label{tab:tasks}
\vspace{-2ex}
\end{table}

\subsection{Baseline setup}

\label{sec:baseline_setup}
To establish the baselines, we adapt the three types of CL techniques ($i.e.$, regularization-based, replay-based, parameter-isolation-based) to CLoG. It is noted that there are several classification-based CL or CLoG methods that combine multiple techniques, but we did not include these baselines in our set as many of their basic components can be unified into the three types of methods (see~\Cref{sec:related_work}). To facilitate a deeper understanding of each type of the method, we selected twelve representative baselines and two types of generative models ($i.e.$, GAN, Diffusion Models) to initiate our analysis.

\paragraph{Adapted Baselines} (1) \textbf{Naive Continual Learning (NCL)} means continually training the same model without any CL techniques to deal with forgetting, which is the simplest baseline in CL.
(2) \textbf{Non-Continual Learning (Non-CL)} means pooling the data from all tasks together and training only one model for all tasks. This is not under a CL setting but its performance can be viewed as an upper bound for CL baselines.
(3) \textbf{Ensemble} trains a separate model for each task. This baseline is forgetting-free, but the memory consumption is huge when more tasks arrive, and there is no knowledge transfer between different tasks.
(4) \textbf{Experience Replay (ER)~\citep{lopez2017gradient}} directly combines replay samples and current task samples in training batches to train the model. The replay data is saved by reservoir sampling~\citep{chaudhry2019tiny,riemer2018learning}.
(5) \textbf{Generative Replay (GR)~\citep{dgr}} replaces the replay samples used in ER with generative replay samples. When training a new task in CLoG, the model is copied and the replay samples are generated via the copied model.
(6) \textbf{Knowledge Distillation (KD)~\citep{KD}} is a regularization-based method in CL. The model is copied as a fixed teacher model before learning the new task. An $\ell_2$ auxiliary loss between the new and old model outputs is added to the NCL objective.
(7) \textbf{L2~\citep{l2}} is also a regularization-based method and copies the model before learning a new task. An $\ell_2$ distance between the current and copied network parameters is added as an auxiliary loss.
(8) \textbf{Elastic Weight Consolidation (EWC)~\citep{ewc}} is also a regularization technique that reweights the $\ell_2$ loss for different parameters. The weights are based on the degree of overlap between the two tasks’ Fisher matrices.
(9) \textbf{Synapse Intelligence (SI)~\citep{si}} is a regularization method that is similar to EWC, while the parameter weights are computed by measuring the parameter updating trajectory during training.
(10) \textbf{Memory Aware Synapses (MAS)~\citep{mas}} is also a regularization-based method. It measures the importance of parameters by the magnitude of the gradient and penalizes changes to parameters that are essential to previous tasks.
(11) \textbf{Averaged Gradient Episodic Memory (A-GEM)~\citep{chaudhry2018efficient}} is a regularization-based method that exploits replay data. It prevents the loss increasing on replay samples by gradient projection.
(12) \textbf{C-LoRA}~\citep{smith2024continual} is a parameter-isolation-based CL method that was first designed for concept-conditional CLoG. It overcomes forgetting by learning task-specific LoRA~\citep{hu2022lora} upon a pre-trained backbone. We adapt it to from-scratch-training by fully training the backbone on the first task and adopting LoRA tuning in the subsequent tasks.

\paragraph{Generative models} We apply CLoG methods on two representative generative models: Generative Adversarial Networks (GAN)~\citep{goodfellow2014generative} and Diffusion Models~\citep{ddpm}. We introduce them in~\cref{app:generative_models}.

\subsection{Metrics design}
\label{sec:metrics_design}
We found that the existing CLoG literature used many distinct metrics (refer to~\cref{sec:related_work} for details), which makes the comparison between different works hard. In this paper, we provide unified metric choices for evaluating CLoG. Suppose $m(f,\mathcal T)$ is a metric to evaluate the generation quality of a generative model $f$ on a task $\mathcal T$, then we extend the metrics from classification-based CL to evaluate the performance of CLoG. When learning the task sequence $\mathcal T^{(1)},\mathcal T^{(2)},\cdots, \mathcal T^{(T)}$, we denote the model after learning $\mathcal T^{(i)}$ as $f^{(i)}$, then we define the CLoG metrics as follows:

\paragraph{Average Incremental Quality (AIQ)~\citep{douillard2020podnet,hou2019learning}} We first define the average quality (AQ) when the model just learns the $t$-th task $\mathcal T^{(t)}$ as $AQ^{(t)} = \frac1t\sum_{i=1}^t m(f^{(t)}, \mathcal T^{(i)})$. Then the \emph{average incremental quality} (AIQ) is defined to evaluate the historical performance as $AIQ = \frac{1}T\sum_{t=1}^T AQ^{(t)}$.

\paragraph{Average Final Quality (AFQ)~\citep{RWalk,lopez2017gradient}} Since AIQ evaluates the historical performance of the model during CL, while in downstream applications we may only care about the final performance of the model (i.e., the performance of $f^{(T)}$), \emph{average final quality} (AFQ) is defined as $AFQ = AQ^{(T)}$.

\paragraph{Forgetting Rate (FR)~\citep{RWalk,lopez2017gradient}} Apart from AIQ and AFQ that measure the learned ``knowledge'' or ``ability'' of the model, measuring the capability to preserve the learned ``knowledge'' or ``ability'' during the continual learning process is also important. The \emph{forgetting rate} (FR) of task $\mathcal T^{(t)}$ can be calculated by the difference between the current performance $m(f^{(T)},\mathcal T^{(t)})$ and the performance when the model first learns this task $m(f^{(t)},\mathcal T^{(t)})$:
\vspace{-0.5ex}
\begin{align}
    FR = \begin{cases}\frac1{T-1}\sum_{t=1}^{T-1} \left(m(f^{(t)}, \mathcal T^{(t)}) - m(f^{(T)}, \mathcal T^{(t)})\right) &(\text{if\ larger\ }m\text{\ is\ better})\\
    \frac1{T-1}\sum_{t=1}^{T-1} \left(m(f^{(T)}, \mathcal T^{(t)})-m(f^{(t)}, \mathcal T^{(t)})\right) &(\text{if\ smaller\ }m\text{\ is\ better}) \\
    \end{cases}
\end{align}

For label-conditioned CLoG, we choose Fréchet inception distance (FID)~\citep{heusel2017gans} as quality metric $m(f,\mathcal T)$ (smaller $m$ is better), which is commonly used to assess the generation quality of image generation models~\citep{ddim,ddpm}.\footnote{Some existing CLoG works used pre-trained classifiers to compute the accuracy of conditional generation, while we find it unsuitable for CLoG and do not adopt it. For example, a pre-trained classifier is not always available for some datasets, and the classifiers often assign wrong prediction for OOD generated images.} For concept-conditioned CLoG, we follow DreamBooth~\citep{ruiz2023dreambooth} to compute the CLIP alignment score~\citep{radford2021learning} between generated image and the provided concept (image alignment score), and between generated image and the text prompts (text alignment score), respectively. The two scores are averaged to obtain a single quality metric for easy comparison.

\subsection{Training specifics}
\label{sec:training_specifics}
To ensure a fair comparison across different CLoG methods, given the diverse training specifics (e.g., image augmentation techniques, network configurations, and other training tricks), a unified protocol is necessary. \emph{The key idea of this paper is to fix the specifics that are irrelevant to CL performance (which might otherwise affect the generation performance) in implementing CLoG baselines}. Specifically, we fix the backbone for GAN and Diffusion Models to StyleGAN2~\citep{Karras2019stylegan2} and DDIM~\citep{ddim} for label-conditioned CLoG, DreamBooth~\citep{ruiz2023dreambooth} and Custom Diffusion~\citep{kumari2023multi} for concept-conditioned CLoG. We fix CL-irrelevant configurations such as DDIM steps, condition encoding, image augmentation, or exponential moving average tricks~\citep{karras2017progressive}, with full details presented in~\Cref{app:implementation_details}. This standardization improves evaluation efficiency by significantly reducing the hyper-parameter space, allowing us to focus on optimizing the hyper-parameters crucial for CL.

\section{Experiments}

\subsection{Implementation Details}
\label{sec:training_details}

To ensure the fair comparison across methods, we follow~\cref{sec:training_specifics} to use unified settings with common hyperparameters and architecture choices. We follow~\citet{heusel2017gans} to compute FID using the entire training dataset as reference images. To achieve the best training performance, we compute the quality metrics on current task every 500 steps and save the best checkpoint. If the method has CL-related hyper-parameters (e.g., regularization weights), we will search for 8 values across different magnitudes and pick the hyper-parameter based on the quality metrics. We found it's hard to train GAN on the long-sequence and large-scale ImageNet-1k benchmark, so we leave it as ``NA'' (Not A Number). We didn't implement C-LoRA on GAN and Custom Diffusion as it is not applicable. We follow~\citet{lin2024class} to use reservoir replay buffer~\citep{vitter1985random} for ER with buffer sizes as 5000 samples for ImageNet-1k, and 200 for the other label-conditioned CLoG benchmarks. Replay-based methods are excluded in Custom-Object as it has very few training samples and thus replay is equivalent to Non-CL. We will present more specific implementation details in~\cref{app:implementation_details}.

\subsection{Result Analysis}

We present the AFQ results as~\cref{tab:MainResult,tab:concept-results}, and postpone the AIQ, FR results in~\cref{app:aiq_fr,app:comprehensive_aiq}. We also conduct additional study on different configurations such as DDIM steps, replay buffer size, different task numbers within the same dataset, different alignment score metrics in~\Cref{app:ablation_study}, and visualize the generation results, compare the method efficiency in~\cref{app:visualize,app:computation_analysis}. We jointly analyze these results and draw some observations as follows.

\begin{table}[t]
\centering
\caption{\label{tab:MainResult}\textbf{AFQ results for label-conditioned CLoG benchmarks.} The best result in each column with the same architecture (StyleGAN2, DDIM) is highlighted in {\textbf{\color{google_red}red}}, while the second best and third best are highlighted in {\textbf{\color{google_blue}blue}} and {\textbf{\color{google_yellow}yellow}}, respectively. The quality metric is FID (\emph{the lower value is better}). We average each AFQ value on 5 class orders and show the standard deviations as superscripts. We use dashlines to split different categories of baselines (Non-CL \& NCL, replay-based, regularization-based, parameter-isolation-based).}
\vspace{-1ex}
\resizebox{\textwidth}{!}{%
\begin{tabular}{cccccccc}
    \toprule
    & \textbf{MNIST} & \textbf{Fashion-MNIST} & \textbf{CIFAR-10} & \textbf{CUB-Birds} & \textbf{Oxford-Flowers} & \textbf{Stanford-Cars} & \textbf{ImageNet-1k}  \\
    \midrule
    \midrule
    \multicolumn{8}{l}{- \textbf{StyleGAN2}} \Bstrut
    \\
    \hdashline\Tstrut
    Non-CL  & \textbf{\color{google_blue} 41.19$^{\pm 3.44}$} & {\textbf{\color{google_blue}66.53$^{\pm 1.46}$}} &  {\textbf{\color{google_blue}63.02$^{\pm 5.18}$}} & 
    \textbf{{\color{google_red}48.36$^{\pm2.16}$}}
    & {\textbf{\color{google_red}99.50$^{\pm 10.4}$}} & {\textbf{\color{google_red}33.68$^{\pm 2.95}$}} & NA\\
     NCL  & 60.98$^{\pm 6.13}$ & 94.10$^{\pm 13.20}$ & 103.34$^{\pm 10.59}$ & {\textbf{\color{google_yellow}112.57$^{\pm 23.05}$}} & 131.98$^{\pm 16.39}$ & {\textbf{\color{google_blue}68.20$^{\pm 2.05}$}} &NA\\
    \hdashline\Tstrut
     ER   & 87.91$^{\pm 24.33}$ & 133.24$^{\pm 35.98}$  & 236.44$^{\pm11.18}$ & 175.99$^{\pm 20.19}$ & 134.94$^{\pm 2.55}$ & 147.88$^{\pm 6.00}$ & NA\\
     GR   & 113.37$^{\pm 38.97}$ & 115.18$^{\pm 26.15}$  & 128.81$^{\pm8.37}$ & 189.27$^{\pm11.55}$ & 161.96$^{\pm10.80}$ & 161.55$^{\pm 27.85}$ & NA\\
     \hdashline\Tstrut
     KD   & 55.04$^{\pm4.88}$ & 86.94$^{\pm 4.05}$  & 105.73$^{\pm 13.27}$ & {\textbf{\color{google_blue}108.68$^{\pm 11.16}$}} & {\textbf{\color{google_blue}120.66$^{\pm 17.47}$}} & 80.45$^{\pm 4.02}$ &NA\\
     L2   & 63.15$^{\pm 13.15}$ & 113.41$^{\pm 7.12}$  & 108.52$^{\pm 6.24}$ & 191.43$^{\pm 17.52}$ & 158.55$^{\pm 11.97}$ & 201.80$^{\pm 32.95}$ &NA\\
     EWC    & 54.73$^{\pm 4.52}$ & 87.20$^{\pm 11.12}$  & 95.33$^{\pm 19.04}$ & 156.06$^{\pm 13.38}$ & 131.62$^{\pm6.00}$ & 100.22$^{\pm 10.81}$ &NA\\
     SI  & 93.12$^{\pm 17.59}$ & 102.29$^{\pm 7.57}$  & 100.13$^{\pm 4.85}$ & 204.44$^{\pm 14.61}$ & 170.53$^{\pm 15.07}$ & 211.72$^{\pm 46.52}$ &NA\\
     MAS   & 57.89$^{\pm 8.53}$ & 86.86$^{\pm 5.29}$  & {\textbf{\color{google_yellow}85.22$^{\pm 2.83}$}} & 
     186.34$^{\pm 17.63}$ & 144.31$^{\pm 14.99}$ & 149.11$^{\pm 19.21}$ &NA\\
     A-GEM   & {\textbf{\color{google_yellow}41.51$^{\pm16.42}$}} & {\textbf{\color{google_yellow}85.37$^{\pm 18.99}$}} & 98.42$^{\pm 11.47}$ & 116.37$^{\pm 13.30}$  & {\textbf{\color{google_yellow}127.93$^{\pm 12.64}$}} & {\textbf{\color{google_yellow}75.46$^{\pm5.54}$}} & NA\\
     \hdashline\Tstrut
     Ensemble  & {\textbf{\color{google_red}8.64$^{\pm 1.74}$}} & {\textbf{\color{google_red}27.76$^{\pm 0.37}$}}  & {\textbf{\color{google_red}45.26$^{\pm 0.61}$}} & 180.71$^{\pm 2.46}$ & 145.59$^{\pm 1.61}$ & 230.74$^{\pm 3.93}$ & NA \\
    \bottomrule
    \toprule
    \multicolumn{8}{l}{- \textbf{DDIM}} \Bstrut
    \\
    \hdashline\Tstrut
    Non-CL  & {\textbf{\color{google_blue}5.59$^{\pm3.67}$}} & {\textbf{\color{google_red}9.02$^{\pm0.23}$}} & {\textbf{\color{google_red}30.19$^{\pm1.29}$}} & {\textbf{\color{google_red}49.30$^{\pm4.43}$}} & {\textbf{\color{google_red}48.81$^{\pm0.84}$}} & {\textbf{\color{google_red}27.97$^{\pm0.42}$}} & {\textbf{\color{google_red}47.27}} \\
     NCL  & 115.47$^{\pm9.30}$ & 139.81$^{\pm19.04}$ & 115.60$^{\pm20.51}$ & {\textbf{\color{google_yellow}98.89$^{\pm6.06}$}} & 102.98$^{\pm16.39}$ &  {\textbf{\color{google_blue}42.81$^{\pm8.91}$}} & {91.46}  \\
    \hdashline\Tstrut
     ER  & 28.64$^{\pm2.74}$ & 52.47$^{\pm2.85}$ &  132.07$^{\pm8.92}$ & {\textbf{\color{google_blue}72.53$^{\pm6.39}$}} & {\textbf{\color{google_yellow}77.03$^{\pm2.62}$}} & 81.26$^{\pm6.44}$ & 101.15  \\
     GR  & 90.28$^{\pm4.72}$ & 34.96$^{\pm6.31}$ &  73.15$^{\pm2.48}$& 106.93$^{\pm4.67}$ &  180.68$^{\pm27.60}$ & 261.59$^{\pm3.24}$ & NA  \\
     \hdashline\Tstrut
     KD  & 149.72$^{\pm13.17}$ & 233.55$^{\pm11.89}$ & 162.13$^{\pm16.11}$ & 181.40$^{\pm8.86}$ & 176.84$^{\pm20.88}$ & 103.06$^{\pm12.55}$ & 107.57  \\
     L2  & 184.05$^{\pm27.14}$ & 190.04$^{\pm5.81}$ & 174.78$^{\pm16.90}$ & 182.79$^{\pm13.50}$ & 191.90$^{\pm33.87}$ & 254.21$^{\pm28.00}$ & 119.22  \\
     EWC & 158.22$^{\pm22.70}$ & 139.52$^{\pm20.07}$ & 127.09$^{\pm19.23}$ & 101.12$\pm^{14.87}$ & 99.34$^{\pm8.27}$ & {\textbf{\color{google_yellow}49.02$^{\pm2.72}$}} & 99.93  \\
     SI  & 182.80$^{\pm25.55}$ & 156.63$^{\pm22.67}$ & 142.32$^{\pm26.74}$ & 113.30$^{\pm15.91}$ & 98.04$^{\pm7.78}$ & 57.06$^{\pm8.39}$  & 100.13  \\
     MAS  & 137.28$^{\pm14.51}$ & 162.25$^{\pm19.61}$ & 124.31$^{\pm10.24}$ & 197.73$^{\pm15.76}$ & 213.12$^{\pm33.11}$ & $282.49^{\pm14.23}$ & 130.21  \\
     A-GEM  & 86.28$^{\pm5.94}$ & 139.46$^{\pm5.21}$ & 129.24$^{\pm27.59}$ & 105.93$^{\pm2.67}$ & 121.27$^{\pm10.92}$ & 50.13$^{\pm2.44}$ & 100.45  \\
     \hdashline\Tstrut
     Ensemble  & {\textbf{\color{google_red}4.12$^{\pm0.14}$}} &  {\textbf{\color{google_blue}10.42$^{\pm0.02}$}} & {\textbf{\color{google_blue}36.52$^{\pm0.55}$}} & 133.32$^{\pm2.07}$ &{\textbf{\color{google_blue}70.16$^{\pm8.67}$}} & 202.15$^{\pm0.52}$ & {\textbf{\color{google_blue}56.97}} \\
     C-LoRA  &  {\textbf{\color{google_yellow}9.45$^{\pm0.38}$}} & {\textbf{\color{google_yellow}24.83$^{\pm5.23}$}} & {\textbf{\color{google_yellow}60.11$^{\pm6.15}$}}  & 148.81$^{\pm1.22}$ & 117.11$^{\pm7.15}$ & 250.90$^{\pm35.87}$ & {\textbf{\color{google_yellow}79.72}}  \\
    \bottomrule
\end{tabular}
}
\vspace{-2ex}
\end{table}

\begin{table}[t]
\centering
\caption{\label{tab:concept-results}\textbf{AFQ results for concept-conditioned CLoG benchmark.} The best result in each row with the same base method (DreamBooth, Custom Diffusion) is highlighted in {\textbf{\color{google_red}red}}, while the second best and third best are highlighted in {\textbf{\color{google_blue}blue}} and {\textbf{\color{google_yellow}yellow}}, respectively. The quality metric is the average of text and image alignment scores (\emph{the higher value is better}). The AFQ is also averaged over 5 orders.}
\vspace{-1ex}
\resizebox{\textwidth}{!}{%
\begin{tabular}{cccccccccc}
    \toprule
     Model & \textbf{NCL} & \textbf{Non-CL} & \textbf{KD} & \textbf{L2} & \textbf{EWC} & \textbf{SI} & \textbf{MAS} & \textbf{Ensemble} & \textbf{C-LoRA} \\
    \midrule
    \midrule
    DreamBooth  & 78.54$^{\pm0.53}$ & {\textbf{\color{google_blue}80.09$^{\pm0.1}$}} & 78.73$^{\pm0.16}$ & 79.00$^{\pm0.38}$ & {\textbf{\color{google_yellow}79.45$^{\pm0.41}$}} & 78.54$^{\pm0.39}$ & 78.00$^{\pm0.46}$  & {\textbf{\color{google_blue}80.09$^{\pm0.25}$}} & {\textbf{\color{google_red}80.42$^{\pm0.25}$}}  \\
    Custom Diffusion & 79.56$^{\pm0.17}$ & {\textbf{\color{google_blue}80.30$^{\pm0.21}$}} & 79.71$^{\pm0.1}$ & 79.92$^{\pm0.14}$ & {\textbf{\color{google_yellow}80.10$^{\pm0.05}$}} & 79.59 $^{\pm0.27}$ & 78.79$^{\pm0.18}$ & {\textbf{\color{google_red}80.39$^{\pm0.24}$}} & - \\
    \bottomrule
\end{tabular}
}
\vspace{-1ex}
\end{table}

\paragraph{NO single method works well across all settings.} We first note that Non-CL is not a CL method since it access all the task data, and is often viewed as the upper bound of CL performance for being forgetting-free and able to transfer knowledge between different tasks. Except for Non-CL, no single method works well on all benchmarks. Specifically, the seemingly best-performing parameter-isolation-based methods ($i.e.$, ensemble, C-LoRA) work well on MNIST, FashionMNIST, CIFAR-10, ImageNet-1k, and Custom-Objects, but fail on Oxford-Flowers, CUB-Birds, and Stanford-Cars. This is because they isolate the parameters for different tasks while knowledge transfer is significant for other baselines on these benchmarks due to the similar features shared across tasks, and each task has too few samples to train a task-specific module. Other than the failure of effective knowledge transfer, parameter-isolation-based methods may also be memory-hungry (see \cref{tab:memory_analysis}) as they have to allocate parameters for each new task. While C-LoRA greatly addresses the issue by only allocating few LoRA weights, the issue will be amplified in scenarios where the task sequence is long (e.g., learning millions of new concepts). Above all, the current methods are not satisfactory enough and our CLoG benchmark remains an open challenge for developing new methods. 

\paragraph{NCL is comparable to regularization-based methods.} Although NCL naively trains on the current task data without any protection to previous learned knowledge, it exhibits similar performance compared to the regularization-based methods on all benchmarks. This indicates that regularization-based methods cannot effectively prevent forgetting. We also found that some of them achieve worse performance than NCL because the regularization makes them hard to learn new tasks, for example, MAS achieves poor AFQ on Oxford-Flowers, Stanford-Cars, and Custom-Objects based on Diffusion Models, but it also has almost zero forgetting (shown in~\cref{tab:fr,tab:CustomFR}) on these benchmarks. Note that we have grid searched the regularization weights from $0.001$ to $10000$ according to the prior works~\citep{ewc,mas,shao2023class} to ensure a faithful implementation. The failure of regularization-based methods also demonstrates the challenge of CLoG due to the use of sophisticated deep generative models.

\paragraph{Replay-based methods face imbalance issue.} Surprisingly, replay-based methods ($i.e.$, ER, GR) don't always outperform non-exemplar methods on CLoG, which contradicts the common observations in classification-based CL where ER is very effective. We relate this phenomenon to the amplification of \emph{CL imbalance}~\citep{guo2023dealing} and \emph{data imbalance}~\citep{Ahn2020SSILSS,huang2021importance}: the limited replayed samples have been seen and trained many times which make them easier to learn than new task data, and leads to mode collapse~\citep{Srivastava2017VEEGANRM} for previous tasks and low plasticity in learning new tasks. The severe mode collapse can be observed for GAN-based ER (see~\cref{app:visualize} for visualizations).
CLoG turns to be more sensitive to these issues than classification-based CL possibly because the modeling of data distribution $p(\bm x)$ is more difficult than classification distribution $p(\bm y|\bm x)$ as discussed in~\cref{sec:formulate_clog}.


\paragraph{Comparison between GAN and Diffusion Models.} Generally, GAN is harder to optimize than Diffusion Models on CLoG, with worse Non-CL and Ensemble performance as shown in~\cref{tab:MainResult}.
This suggests that Diffusion Models are more promising as the base architecture for CLoG. 

\paragraph{Comparison between label-conditioned and concept-conditioned CLoG.} It is clear that with pre-trained backbone and fewer training samples, the results on concept-conditioned CLoG exhibit less forgetting, and NCL performance is also strong. The results align with the recent CL works that pre-trained text generation models resist forgetting better~\citep{shao2023class,cao2024generative}. However, the visualization results shown in~\cref{fig:vis_dreambooth,fig:vis_custom_diffusion} are far from perfect, which means it is still necessary to improve the concept-conditioned CLoG methods in future research.

\paragraph{Method Efficiency.} We also report the training time and memory consumption of the CLoG methods in~\cref{app:computation_analysis}. It is noted that DDIM-based GR is significantly slow, as generating the replay samples for DDIM requires multiple denoising steps (50 in our implementation). It is also noted that the parameter-isolation-based methods have linearly increasing memory consumption when the number of tasks increases, while other methods only consume constant memory budget. 

\section{Discussions and Limitations}
\label{sec:limitation}

\paragraph{Discussions} There has long been a period that CL research focusing on addressing forgetting for classification tasks, and although many advancements have been achieved in the past years, CL methods are rarely applied in real world applications. With the emergence of various “all-in-one” foundation models, the relevance of focusing mainly on classification-based CL is increasingly questionable. Current trends suggest that generative-based foundation models are poised to become the next generation of AI products, integral to everyday life. In this context, CLoG becomes crucial, addressing how these models of diverse architectures, complex learning objectives, and open-ended domains can continuously learn newly emerged knowledge~\citep{ke2023continual,verwimp2023continual,liu2023ai}, cater to personalized needs~\citep{rafieian2023ai,pesovski2024generative,shahin2024harnessing}, and possibly enhance human-AI alignment in an evolving world~\citep{soares2014aligning,christian2021alignment,leike2018scalable}. Our benchmark results reveal disappointing performance with traditional CL methods, highlighting a pressing need for refined CLoG strategies for future applications.

\paragraph{Limitations and outlook} This study primarily presents initial benchmarks and baselines for CLoG, with a focus on traditional representative methods. Future work will include expanding these benchmarks across a wider range of image generation tasks, incorporating various generative conditions~\citep{Zhai2019LifelongGC,Zhang_2023_ICCV}, and extending to additional modalities such as molecules~\citep{xu2022geodiff}. Although we only include baselines from classification-based CL, it is interesting to design methods specifically for generative models by applying techniques such as classifier-guidance~\citep{beatGAN}, and include more generative models~\citep{Liu2022FlowSA,Peebles2022ScalableDM,Liu2023InstaFlowOS} other than GAN and Diffusion Models. 
Our current analysis is based solely on existing datasets; hence, we plan to enhance the scope of concept-conditioned CLoG benchmarks by acquiring and incorporating more diverse datasets and domains. Furthermore, while this paper primarily conducts an empirical investigation, advancing the theoretical framework of CLoG will be crucial for its development and understanding. 

\section{Conclusion}

In this paper, we introduce a foundational framework for studying Continual Learning of Generative Models (CLoG). We explore the challenges that CLoG presents compared to the more popular classification-based CL. We establish unified benchmarks, baselines, evaluation protocols, and specific training guidelines for CLoG. Our findings underscore the necessity for developing more advanced CLoG methods in the future. Additionally, we advocate for a shift in focus from classification-based CL to CLoG, given the growing importance of generative foundation models in current research.

\newpage


% ...


% ...

\bibliographystyle{abbrvnat}
\nobibliography*
\bibliography{main}

\begin{thebibliography}{156}
\providecommand{\natexlab}[1]{#1}
\providecommand{\url}[1]{\texttt{#1}}
\expandafter\ifx\csname urlstyle\endcsname\relax
  \providecommand{\doi}[1]{doi: #1}\else
  \providecommand{\doi}{doi: \begingroup \urlstyle{rm}\Url}\fi

\bibitem[Achiam et~al.(2023)Achiam, Adler, Agarwal, Ahmad, Akkaya, Aleman, Almeida, Altenschmidt, Altman, Anadkat, Avila, Babuschkin, Balaji, Balcom, Baltescu, Bao, Bavarian, Belgum, Bello, Berdine, Bernadett-Shapiro, Berner, Bogdonoff, Boiko, Boyd, Brakman, Brockman, Brooks, Brundage, Button, Cai, Campbell, Cann, Carey, Carlson, Carmichael, Chan, Chang, Chantzis, Chen, Chen, Chen, Chen, Chen, Chess, Cho, Chu, Chung, Cummings, Currier, Dai, Decareaux, Degry, Deutsch, Deville, Dhar, Dohan, Dowling, Dunning, Ecoffet, Eleti, Eloundou, Farhi, Fedus, Felix, Fishman, Forte, Fulford, Gao, Georges, Gibson, Goel, Gogineni, Goh, Gontijo-Lopes, Gordon, Grafstein, Gray, Greene, Gross, Gu, Guo, Hallacy, Han, Harris, He, Heaton, Heidecke, Hesse, Hickey, Hickey, Hoeschele, Houghton, Hsu, Hu, Hu, Huizinga, Jain, Jain, Jang, Jiang, Jiang, Jin, Jin, Jomoto, Jonn, Jun, Kaftan, Kaiser, Kamali, Kanitscheider, Keskar, Khan, Kilpatrick, Kim, Kim, Kim, Kirchner, Kiros, Knight, Kokotajlo, Kondraciuk, Kondrich, Konstantinidis, Kosic,
  Krueger, Kuo, Lampe, Lan, Lee, Leike, Leung, Levy, Li, Lim, Lin, Lin, Litwin, Lopez, Lowe, Lue, Makanju, Malfacini, Manning, Markov, Markovski, Martin, Mayer, Mayne, McGrew, McKinney, McLeavey, McMillan, McNeil, Medina, Mehta, Menick, Metz, Mishchenko, Mishkin, Monaco, Morikawa, Mossing, Mu, Murati, Murk, M'ely, Nair, Nakano, Nayak, Neelakantan, Ngo, Noh, Long, O'Keefe, Pachocki, Paino, Palermo, Pantuliano, Parascandolo, Parish, Parparita, Passos, Pavlov, Peng, Perelman, de~Avila Belbute~Peres, Petrov, de~Oliveira~Pinto, Pokorny, Pokrass, Pong, Powell, Power, Power, Proehl, Puri, Radford, Rae, Ramesh, Raymond, Real, Rimbach, Ross, Rotsted, Roussez, Ryder, Saltarelli, Sanders, Santurkar, Sastry, Schmidt, Schnurr, Schulman, Selsam, Sheppard, Sherbakov, Shieh, Shoker, Shyam, Sidor, Sigler, Simens, Sitkin, Slama, Sohl, Sokolowsky, Song, Staudacher, Such, Summers, Sutskever, Tang, Tezak, Thompson, Tillet, Tootoonchian, Tseng, Tuggle, Turley, Tworek, Uribe, Vallone, Vijayvergiya, Voss, Wainwright, Wang, Wang,
  Wang, Ward, Wei, Weinmann, Welihinda, Welinder, Weng, Weng, Wiethoff, Willner, Winter, Wolrich, Wong, Workman, Wu, Wu, Wu, Xiao, Xu, Yoo, Yu, Yuan, Zaremba, Zellers, Zhang, Zhang, Zhao, Zheng, Zhuang, Zhuk, and Zoph]{Achiam2023GPT4TR}
O.~J. Achiam, S.~Adler, S.~Agarwal, L.~Ahmad, I.~Akkaya, F.~L. Aleman, D.~Almeida, J.~Altenschmidt, S.~Altman, S.~Anadkat, R.~Avila, I.~Babuschkin, S.~Balaji, V.~Balcom, P.~Baltescu, H.~Bao, M.~Bavarian, J.~Belgum, I.~Bello, J.~Berdine, G.~Bernadett-Shapiro, C.~Berner, L.~Bogdonoff, O.~Boiko, M.~Boyd, A.-L. Brakman, G.~Brockman, T.~Brooks, M.~Brundage, K.~Button, T.~Cai, R.~Campbell, A.~Cann, B.~Carey, C.~Carlson, R.~Carmichael, B.~Chan, C.~Chang, F.~Chantzis, D.~Chen, S.~Chen, R.~Chen, J.~Chen, M.~Chen, B.~Chess, C.~Cho, C.~Chu, H.~W. Chung, D.~Cummings, J.~Currier, Y.~Dai, C.~Decareaux, T.~Degry, N.~Deutsch, D.~Deville, A.~Dhar, D.~Dohan, S.~Dowling, S.~Dunning, A.~Ecoffet, A.~Eleti, T.~Eloundou, D.~Farhi, L.~Fedus, N.~Felix, S.~P. Fishman, J.~Forte, I.~Fulford, L.~Gao, E.~Georges, C.~Gibson, V.~Goel, T.~Gogineni, G.~Goh, R.~Gontijo-Lopes, J.~Gordon, M.~Grafstein, S.~Gray, R.~Greene, J.~Gross, S.~S. Gu, Y.~Guo, C.~Hallacy, J.~Han, J.~Harris, Y.~He, M.~Heaton, J.~Heidecke, C.~Hesse, A.~Hickey, W.~Hickey,
  P.~Hoeschele, B.~Houghton, K.~Hsu, S.~Hu, X.~Hu, J.~Huizinga, S.~Jain, S.~Jain, J.~Jang, A.~Jiang, R.~Jiang, H.~Jin, D.~Jin, S.~Jomoto, B.~Jonn, H.~Jun, T.~Kaftan, L.~Kaiser, A.~Kamali, I.~Kanitscheider, N.~S. Keskar, T.~Khan, L.~Kilpatrick, J.~W. Kim, C.~Kim, Y.~Kim, H.~Kirchner, J.~R. Kiros, M.~Knight, D.~Kokotajlo, L.~Kondraciuk, A.~Kondrich, A.~Konstantinidis, K.~Kosic, G.~Krueger, V.~Kuo, M.~Lampe, I.~Lan, T.~Lee, J.~Leike, J.~Leung, D.~Levy, C.~M. Li, R.~Lim, M.~Lin, S.~Lin, M.~Litwin, T.~Lopez, R.~Lowe, P.~Lue, A.~A. Makanju, K.~Malfacini, S.~Manning, T.~Markov, Y.~Markovski, B.~Martin, K.~Mayer, A.~Mayne, B.~McGrew, S.~M. McKinney, C.~McLeavey, P.~McMillan, J.~McNeil, D.~Medina, A.~Mehta, J.~Menick, L.~Metz, A.~Mishchenko, P.~Mishkin, V.~Monaco, E.~Morikawa, D.~P. Mossing, T.~Mu, M.~Murati, O.~Murk, D.~M'ely, A.~Nair, R.~Nakano, R.~Nayak, A.~Neelakantan, R.~Ngo, H.~Noh, O.~Long, C.~O'Keefe, J.~W. Pachocki, A.~Paino, J.~Palermo, A.~Pantuliano, G.~Parascandolo, J.~Parish, E.~Parparita, A.~Passos,
  M.~Pavlov, A.~Peng, A.~Perelman, F.~de~Avila Belbute~Peres, M.~Petrov, H.~P. de~Oliveira~Pinto, M.~Pokorny, M.~Pokrass, V.~H. Pong, T.~Powell, A.~Power, B.~Power, E.~Proehl, R.~Puri, A.~Radford, J.~Rae, A.~Ramesh, C.~Raymond, F.~Real, K.~Rimbach, C.~Ross, B.~Rotsted, H.~Roussez, N.~Ryder, M.~D. Saltarelli, T.~Sanders, S.~Santurkar, G.~Sastry, H.~Schmidt, D.~Schnurr, J.~Schulman, D.~Selsam, K.~Sheppard, T.~Sherbakov, J.~Shieh, S.~Shoker, P.~Shyam, S.~Sidor, E.~Sigler, M.~Simens, J.~Sitkin, K.~Slama, I.~Sohl, B.~D. Sokolowsky, Y.~Song, N.~Staudacher, F.~P. Such, N.~Summers, I.~Sutskever, J.~Tang, N.~A. Tezak, M.~Thompson, P.~Tillet, A.~Tootoonchian, E.~Tseng, P.~Tuggle, N.~Turley, J.~Tworek, J.~F.~C. Uribe, A.~Vallone, A.~Vijayvergiya, C.~Voss, C.~Wainwright, J.~J. Wang, A.~Wang, B.~Wang, J.~Ward, J.~Wei, C.~Weinmann, A.~Welihinda, P.~Welinder, J.~Weng, L.~Weng, M.~Wiethoff, D.~Willner, C.~Winter, S.~Wolrich, H.~Wong, L.~Workman, S.~Wu, J.~Wu, M.~Wu, K.~Xiao, T.~Xu, S.~Yoo, K.~Yu, Q.~Yuan, W.~Zaremba,
  R.~Zellers, C.~Zhang, M.~Zhang, S.~Zhao, T.~Zheng, J.~Zhuang, W.~Zhuk, and B.~Zoph.
\newblock Gpt-4 technical report, 2023.
\newblock URL \url{https://api.semanticscholar.org/CorpusID:257532815}.

\bibitem[Ahn et~al.(2020)Ahn, Kwak, Lim, Bang, Kim, and Moon]{Ahn2020SSILSS}
H.~Ahn, J.~Kwak, S.~F. Lim, H.~Bang, H.~Kim, and T.~Moon.
\newblock Ss-il: Separated softmax for incremental learning.
\newblock \emph{2021 IEEE/CVF International Conference on Computer Vision (ICCV)}, pages 824--833, 2020.
\newblock URL \url{https://api.semanticscholar.org/CorpusID:227240555}.

\bibitem[Aljundi et~al.(2017)Aljundi, Chakravarty, and Tuytelaars]{aljundi2017expert}
R.~Aljundi, P.~Chakravarty, and T.~Tuytelaars.
\newblock Expert gate: Lifelong learning with a network of experts.
\newblock In \emph{Proceedings of the IEEE Conference on Computer Vision and Pattern Recognition}, pages 3366--3375, 2017.

\bibitem[Aljundi et~al.(2018)Aljundi, Babiloni, Elhoseiny, Rohrbach, and Tuytelaars]{mas}
R.~Aljundi, F.~Babiloni, M.~Elhoseiny, M.~Rohrbach, and T.~Tuytelaars.
\newblock Memory aware synapses: Learning what (not) to forget.
\newblock In \emph{Proceedings of the European conference on computer vision (ECCV)}, pages 139--154, 2018.

\bibitem[Anonymous(2022)]{anonymous2022imagen}
Anonymous.
\newblock Imagen video: High definition video generation with diffusion models.
\newblock \emph{Submitted to Transactions on Machine Learning Research}, 2022.
\newblock URL \url{https://openreview.net/forum?id=ETH1GDbSPw}.

\bibitem[Brooks et~al.(2024)Brooks, Peebles, Holmes, DePue, Guo, Jing, Schnurr, Taylor, Luhman, Luhman, Ng, Wang, and Ramesh]{videoworldsimulators2024}
T.~Brooks, B.~Peebles, C.~Holmes, W.~DePue, Y.~Guo, L.~Jing, D.~Schnurr, J.~Taylor, T.~Luhman, E.~Luhman, C.~Ng, R.~Wang, and A.~Ramesh.
\newblock Video generation models as world simulators.
\newblock 2024.
\newblock URL \url{https://openai.com/research/video-generation-models-as-world-simulators}.

\bibitem[Buzzega et~al.(2020)Buzzega, Boschini, Porrello, Abati, and Calderara]{buzzega2020dark}
P.~Buzzega, M.~Boschini, A.~Porrello, D.~Abati, and S.~Calderara.
\newblock Dark experience for general continual learning: a strong, simple baseline.
\newblock \emph{Advances in neural information processing systems}, 33:\penalty0 15920--15930, 2020.

\bibitem[Cao et~al.(2024)Cao, Lu, Huang, Liu, and Cheng]{cao2024generative}
X.~Cao, H.~Lu, L.~Huang, X.~Liu, and M.-M. Cheng.
\newblock Generative multi-modal models are good class-incremental learners.
\newblock \emph{arXiv preprint arXiv:2403.18383}, 2024.

\bibitem[Chaudhry et~al.(2018{\natexlab{a}})Chaudhry, Dokania, Ajanthan, and Torr]{RWalk}
A.~Chaudhry, P.~K. Dokania, T.~Ajanthan, and P.~H. Torr.
\newblock Riemannian walk for incremental learning: Understanding forgetting and intransigence.
\newblock In \emph{Proceedings of the European conference on computer vision (ECCV)}, pages 532--547, 2018{\natexlab{a}}.

\bibitem[Chaudhry et~al.(2018{\natexlab{b}})Chaudhry, Ranzato, Rohrbach, and Elhoseiny]{chaudhry2018efficient}
A.~Chaudhry, M.~Ranzato, M.~Rohrbach, and M.~Elhoseiny.
\newblock Efficient lifelong learning with a-gem.
\newblock \emph{arXiv preprint arXiv:1812.00420}, 2018{\natexlab{b}}.

\bibitem[Chaudhry et~al.(2019)Chaudhry, Rohrbach, Elhoseiny, Ajanthan, Dokania, Torr, and Ranzato]{chaudhry2019tiny}
A.~Chaudhry, M.~Rohrbach, M.~Elhoseiny, T.~Ajanthan, P.~K. Dokania, P.~H. Torr, and M.~Ranzato.
\newblock On tiny episodic memories in continual learning.
\newblock \emph{arXiv preprint arXiv:1902.10486}, 2019.

\bibitem[Chrabaszcz et~al.(2017)Chrabaszcz, Loshchilov, and Hutter]{chrabaszcz2017downsampled}
P.~Chrabaszcz, I.~Loshchilov, and F.~Hutter.
\newblock A downsampled variant of imagenet as an alternative to the cifar datasets.
\newblock \emph{arXiv preprint arXiv:1707.08819}, 2017.

\bibitem[Christian(2021)]{christian2021alignment}
B.~Christian.
\newblock \emph{The alignment problem: How can machines learn human values?}
\newblock Atlantic Books, 2021.

\bibitem[Cong et~al.(2020{\natexlab{a}})Cong, Zhao, Li, Wang, and Carin]{Cong2020GANMW}
Y.~Cong, M.~Zhao, J.~Li, S.~Wang, and L.~Carin.
\newblock Gan memory with no forgetting.
\newblock \emph{ArXiv}, abs/2006.07543, 2020{\natexlab{a}}.
\newblock URL \url{https://api.semanticscholar.org/CorpusID:219686951}.

\bibitem[Cong et~al.(2020{\natexlab{b}})Cong, Zhao, Li, Wang, and Carin]{ganmemory}
Y.~Cong, M.~Zhao, J.~Li, S.~Wang, and L.~Carin.
\newblock Gan memory with no forgetting.
\newblock \emph{Advances in Neural Information Processing Systems}, 33:\penalty0 16481--16494, 2020{\natexlab{b}}.

\bibitem[De~Lange et~al.(2021)De~Lange, Aljundi, Masana, Parisot, Jia, Leonardis, Slabaugh, and Tuytelaars]{de2021continual}
M.~De~Lange, R.~Aljundi, M.~Masana, S.~Parisot, X.~Jia, A.~Leonardis, G.~Slabaugh, and T.~Tuytelaars.
\newblock A continual learning survey: Defying forgetting in classification tasks.
\newblock \emph{IEEE transactions on pattern analysis and machine intelligence}, 44\penalty0 (7):\penalty0 3366--3385, 2021.

\bibitem[Delange et~al.(2021)Delange, Aljundi, Masana, Parisot, Jia, Leonardis, Slabaugh, and Tuytelaars]{clsurveyMat}
M.~Delange, R.~Aljundi, M.~Masana, S.~Parisot, X.~Jia, A.~Leonardis, G.~Slabaugh, and T.~Tuytelaars.
\newblock A continual learning survey: Defying forgetting in classification tasks.
\newblock \emph{IEEE Transactions on Pattern Analysis and Machine Intelligence}, page 1–1, 2021.
\newblock ISSN 1939-3539.
\newblock \doi{10.1109/tpami.2021.3057446}.
\newblock URL \url{http://dx.doi.org/10.1109/TPAMI.2021.3057446}.

\bibitem[Deng(2012)]{MNIST}
L.~Deng.
\newblock The mnist database of handwritten digit images for machine learning research [best of the web].
\newblock \emph{IEEE Signal Processing Magazine}, 29\penalty0 (6):\penalty0 141--142, 2012.
\newblock \doi{10.1109/MSP.2012.2211477}.

\bibitem[Dhariwal and Nichol(2021{\natexlab{a}})]{beatGAN}
P.~Dhariwal and A.~Nichol.
\newblock Diffusion models beat gans on image synthesis.
\newblock \emph{Advances in neural information processing systems}, 34:\penalty0 8780--8794, 2021{\natexlab{a}}.

\bibitem[Dhariwal and Nichol(2021{\natexlab{b}})]{dhariwal2021diffusion}
P.~Dhariwal and A.~Nichol.
\newblock Diffusion models beat gans on image synthesis, 2021{\natexlab{b}}.

\bibitem[Doan et~al.(2024)Doan, Tran, Tran, Nguyen, Phung, and Le]{doan2024classprototype}
K.~Doan, Q.~Tran, T.~L. Tran, T.~Nguyen, D.~Phung, and T.~Le.
\newblock Class-prototype conditional diffusion model with gradient projection for continual learning, 2024.

\bibitem[Donahue et~al.(2019)Donahue, McAuley, and Puckette]{donahue2019adversarial}
C.~Donahue, J.~McAuley, and M.~Puckette.
\newblock Adversarial audio synthesis, 2019.

\bibitem[Douillard et~al.(2020)Douillard, Cord, Ollion, Robert, and Valle]{douillard2020podnet}
A.~Douillard, M.~Cord, C.~Ollion, T.~Robert, and E.~Valle.
\newblock Podnet: Pooled outputs distillation for small-tasks incremental learning.
\newblock In \emph{Computer vision--ECCV 2020: 16th European conference, Glasgow, UK, August 23--28, 2020, proceedings, part XX 16}, pages 86--102. Springer, 2020.

\bibitem[Fernando et~al.(2017)Fernando, Banarse, Blundell, Zwols, Ha, Rusu, Pritzel, and Wierstra]{fernando2017pathnet}
C.~Fernando, D.~Banarse, C.~Blundell, Y.~Zwols, D.~Ha, A.~A. Rusu, A.~Pritzel, and D.~Wierstra.
\newblock Pathnet: Evolution channels gradient descent in super neural networks.
\newblock \emph{arXiv preprint arXiv:1701.08734}, 2017.

\bibitem[Gao and Liu(2023)]{pmlr-v202-gao23e}
R.~Gao and W.~Liu.
\newblock {DDGR}: Continual learning with deep diffusion-based generative replay.
\newblock In A.~Krause, E.~Brunskill, K.~Cho, B.~Engelhardt, S.~Sabato, and J.~Scarlett, editors, \emph{Proceedings of the 40th International Conference on Machine Learning}, volume 202 of \emph{Proceedings of Machine Learning Research}, pages 10744--10763. PMLR, 23--29 Jul 2023.
\newblock URL \url{https://proceedings.mlr.press/v202/gao23e.html}.

\bibitem[Goodfellow et~al.(2014)Goodfellow, Pouget-Abadie, Mirza, Xu, Warde-Farley, Ozair, Courville, and Bengio]{goodfellow2014generative}
I.~J. Goodfellow, J.~Pouget-Abadie, M.~Mirza, B.~Xu, D.~Warde-Farley, S.~Ozair, A.~Courville, and Y.~Bengio.
\newblock Generative adversarial networks, 2014.

\bibitem[Guo et~al.(2023)Guo, Liu, and Zhao]{guo2023dealing}
Y.~Guo, B.~Liu, and D.~Zhao.
\newblock Dealing with cross-task class discrimination in online continual learning.
\newblock In \emph{Proceedings of the IEEE/CVF Conference on Computer Vision and Pattern Recognition}, pages 11878--11887, 2023.

\bibitem[He et~al.(2016)He, Zhang, Ren, and Sun]{resnet}
K.~He, X.~Zhang, S.~Ren, and J.~Sun.
\newblock Deep residual learning for image recognition.
\newblock In \emph{2016 IEEE Conference on Computer Vision and Pattern Recognition (CVPR)}, pages 770--778, 2016.
\newblock \doi{10.1109/CVPR.2016.90}.

\bibitem[Hemati et~al.(2024)Hemati, Pellegrini, Duan, Zhao, Xia, Masana, Tscheschner, Veas, Zheng, Zhao, Li, Huang, Lomonaco, and van~de Ven]{hemati2024continual}
H.~Hemati, L.~Pellegrini, X.~Duan, Z.~Zhao, F.~Xia, M.~Masana, B.~Tscheschner, E.~Veas, Y.~Zheng, S.~Zhao, S.-Y. Li, S.-J. Huang, V.~Lomonaco, and G.~M. van~de Ven.
\newblock Continual learning in the presence of repetition, 2024.

\bibitem[Heusel et~al.(2017)Heusel, Ramsauer, Unterthiner, Nessler, and Hochreiter]{heusel2017gans}
M.~Heusel, H.~Ramsauer, T.~Unterthiner, B.~Nessler, and S.~Hochreiter.
\newblock Gans trained by a two time-scale update rule converge to a local nash equilibrium.
\newblock \emph{Advances in neural information processing systems}, 30, 2017.

\bibitem[Hinton et~al.(2015)Hinton, Vinyals, and Dean]{KD}
G.~Hinton, O.~Vinyals, and J.~Dean.
\newblock Distilling the knowledge in a neural network.
\newblock \emph{arXiv preprint arXiv:1503.02531}, 2015.

\bibitem[Ho and Salimans(2021)]{ho2021classifierfree}
J.~Ho and T.~Salimans.
\newblock Classifier-free diffusion guidance.
\newblock In \emph{NeurIPS 2021 Workshop on Deep Generative Models and Downstream Applications}, 2021.
\newblock URL \url{https://openreview.net/forum?id=qw8AKxfYbI}.

\bibitem[Ho et~al.(2020{\natexlab{a}})Ho, Jain, and Abbeel]{ddpm}
J.~Ho, A.~Jain, and P.~Abbeel.
\newblock Denoising diffusion probabilistic models.
\newblock \emph{Advances in neural information processing systems}, 33:\penalty0 6840--6851, 2020{\natexlab{a}}.

\bibitem[Ho et~al.(2020{\natexlab{b}})Ho, Jain, and Abbeel]{ho2020denoising}
J.~Ho, A.~Jain, and P.~Abbeel.
\newblock Denoising diffusion probabilistic models, 2020{\natexlab{b}}.

\bibitem[Hou et~al.(2019)Hou, Pan, Loy, Wang, and Lin]{hou2019learning}
S.~Hou, X.~Pan, C.~C. Loy, Z.~Wang, and D.~Lin.
\newblock Learning a unified classifier incrementally via rebalancing.
\newblock In \emph{Proceedings of the IEEE/CVF conference on computer vision and pattern recognition}, pages 831--839, 2019.

\bibitem[Hu et~al.(2022)Hu, yelong shen, Wallis, Allen-Zhu, Li, Wang, Wang, and Chen]{hu2022lora}
E.~J. Hu, yelong shen, P.~Wallis, Z.~Allen-Zhu, Y.~Li, S.~Wang, L.~Wang, and W.~Chen.
\newblock Lo{RA}: Low-rank adaptation of large language models.
\newblock In \emph{International Conference on Learning Representations}, 2022.
\newblock URL \url{https://openreview.net/forum?id=nZeVKeeFYf9}.

\bibitem[Huang et~al.(2018)Huang, Vaswani, Uszkoreit, Shazeer, Simon, Hawthorne, Dai, Hoffman, Dinculescu, and Eck]{huang2018music}
C.-Z.~A. Huang, A.~Vaswani, J.~Uszkoreit, N.~Shazeer, I.~Simon, C.~Hawthorne, A.~M. Dai, M.~D. Hoffman, M.~Dinculescu, and D.~Eck.
\newblock Music transformer, 2018.

\bibitem[Huang et~al.(2021)Huang, Geng, and Li]{huang2021importance}
R.~Huang, A.~Geng, and Y.~Li.
\newblock On the importance of gradients for detecting distributional shifts in the wild.
\newblock \emph{Advances in Neural Information Processing Systems}, 34:\penalty0 677--689, 2021.

\bibitem[Isola et~al.(2018)Isola, Zhu, Zhou, and Efros]{isola2018imagetoimage}
P.~Isola, J.-Y. Zhu, T.~Zhou, and A.~A. Efros.
\newblock Image-to-image translation with conditional adversarial networks, 2018.

\bibitem[Jiang et~al.(2016)Jiang, Liang, Feng, Fan, Pei, Xue, and Guan]{Jiang2016TextCB}
M.~Jiang, Y.~Liang, X.~Feng, X.~Fan, Z.~Pei, Y.~Xue, and R.~Guan.
\newblock Text classification based on deep belief network and softmax regression.
\newblock \emph{Neural Computing and Applications}, 29:\penalty0 61 -- 70, 2016.
\newblock URL \url{https://api.semanticscholar.org/CorpusID:15946343}.

\bibitem[Jing et~al.(2022)Jing, Corso, Chang, Barzilay, and Jaakkola]{NEURIPS2022_994545b2}
B.~Jing, G.~Corso, J.~Chang, R.~Barzilay, and T.~Jaakkola.
\newblock Torsional diffusion for molecular conformer generation.
\newblock In S.~Koyejo, S.~Mohamed, A.~Agarwal, D.~Belgrave, K.~Cho, and A.~Oh, editors, \emph{Advances in Neural Information Processing Systems}, volume~35, pages 24240--24253. Curran Associates, Inc., 2022.
\newblock URL \url{https://proceedings.neurips.cc/paper_files/paper/2022/file/994545b2308bbbbc97e3e687ea9e464f-Paper-Conference.pdf}.

\bibitem[Kalchbrenner et~al.(2014)Kalchbrenner, Grefenstette, and Blunsom]{kalchbrenner2014convolutional}
N.~Kalchbrenner, E.~Grefenstette, and P.~Blunsom.
\newblock A convolutional neural network for modelling sentences, 2014.

\bibitem[Karras et~al.(2017)Karras, Aila, Laine, and Lehtinen]{karras2017progressive}
T.~Karras, T.~Aila, S.~Laine, and J.~Lehtinen.
\newblock Progressive growing of gans for improved quality, stability, and variation.
\newblock \emph{arXiv preprint arXiv:1710.10196}, 2017.

\bibitem[Karras et~al.(2019)Karras, Laine, and Aila]{karras2019stylebased}
T.~Karras, S.~Laine, and T.~Aila.
\newblock A style-based generator architecture for generative adversarial networks, 2019.

\bibitem[Karras et~al.(2020{\natexlab{a}})Karras, Aittala, Hellsten, Laine, Lehtinen, and Aila]{karras2020stylegan2ada}
T.~Karras, M.~Aittala, J.~Hellsten, S.~Laine, J.~Lehtinen, and T.~Aila.
\newblock Training generative adversarial networks with limited data.
\newblock \emph{Advances in neural information processing systems}, 33:\penalty0 12104--12114, 2020{\natexlab{a}}.

\bibitem[Karras et~al.(2020{\natexlab{b}})Karras, Laine, Aittala, Hellsten, Lehtinen, and Aila]{Karras2019stylegan2}
T.~Karras, S.~Laine, M.~Aittala, J.~Hellsten, J.~Lehtinen, and T.~Aila.
\newblock Analyzing and improving the image quality of {StyleGAN}.
\newblock In \emph{Proc. CVPR}, 2020{\natexlab{b}}.

\bibitem[Ke and Liu(2023)]{clsurveybingliu}
Z.~Ke and B.~Liu.
\newblock Continual learning of natural language processing tasks: A survey, 2023.

\bibitem[Ke et~al.(2021)Ke, Liu, Wang, and Shu]{ke2021continual}
Z.~Ke, B.~Liu, H.~Wang, and L.~Shu.
\newblock Continual learning with knowledge transfer for sentiment classification.
\newblock In \emph{Machine Learning and Knowledge Discovery in Databases: European Conference, ECML PKDD 2020, Ghent, Belgium, September 14--18, 2020, Proceedings, Part III}, pages 683--698. Springer, 2021.

\bibitem[Ke et~al.(2022)Ke, Lin, Shao, Xu, Shu, and Liu]{ke2022continual}
Z.~Ke, H.~Lin, Y.~Shao, H.~Xu, L.~Shu, and B.~Liu.
\newblock Continual training of language models for few-shot learning.
\newblock In \emph{Empirical Methods in Natural Language Processing (EMNLP)}, 2022.

\bibitem[Ke et~al.(2023)Ke, Shao, Lin, Konishi, Kim, and Liu]{ke2023continual}
Z.~Ke, Y.~Shao, H.~Lin, T.~Konishi, G.~Kim, and B.~Liu.
\newblock Continual pre-training of language models.
\newblock In \emph{The Eleventh International Conference on Learning Representations (ICLR-2023)}, 2023.

\bibitem[Kemker and Kanan(2017)]{kemker2017fearnet}
R.~Kemker and C.~Kanan.
\newblock Fearnet: Brain-inspired model for incremental learning.
\newblock \emph{arXiv preprint arXiv:1711.10563}, 2017.

\bibitem[Kim et~al.(2022)Kim, Liu, and Ke]{kim2022multi}
G.~Kim, B.~Liu, and Z.~Ke.
\newblock A multi-head model for continual learning via out-of-distribution replay.
\newblock In \emph{Conference on Lifelong Learning Agents}, pages 548--563. PMLR, 2022.

\bibitem[Kingma and Welling(2013)]{kingma2013auto}
D.~P. Kingma and M.~Welling.
\newblock Auto-encoding variational bayes.
\newblock \emph{arXiv preprint arXiv:1312.6114}, 2013.

\bibitem[Kingma and Welling(2022)]{kingma2022autoencoding}
D.~P. Kingma and M.~Welling.
\newblock Auto-encoding variational bayes, 2022.

\bibitem[Kingma et~al.(2023)Kingma, Salimans, Poole, and Ho]{kingma2023variational}
D.~P. Kingma, T.~Salimans, B.~Poole, and J.~Ho.
\newblock Variational diffusion models, 2023.

\bibitem[Kirkpatrick et~al.(2017)Kirkpatrick, Pascanu, Rabinowitz, Veness, Desjardins, Rusu, Milan, Quan, Ramalho, Grabska-Barwinska, et~al.]{ewc}
J.~Kirkpatrick, R.~Pascanu, N.~Rabinowitz, J.~Veness, G.~Desjardins, A.~A. Rusu, K.~Milan, J.~Quan, T.~Ramalho, A.~Grabska-Barwinska, et~al.
\newblock Overcoming catastrophic forgetting in neural networks.
\newblock \emph{Proceedings of the national academy of sciences}, 114\penalty0 (13):\penalty0 3521--3526, 2017.

\bibitem[Kolesnikov et~al.(2021)Kolesnikov, Dosovitskiy, Weissenborn, Heigold, Uszkoreit, Beyer, Minderer, Dehghani, Houlsby, Gelly, Unterthiner, and Zhai]{vit}
A.~Kolesnikov, A.~Dosovitskiy, D.~Weissenborn, G.~Heigold, J.~Uszkoreit, L.~Beyer, M.~Minderer, M.~Dehghani, N.~Houlsby, S.~Gelly, T.~Unterthiner, and X.~Zhai.
\newblock An image is worth 16x16 words: Transformers for image recognition at scale.
\newblock 2021.

\bibitem[Kowsari et~al.(2017)Kowsari, Brown, Heidarysafa, Jafari~Meimandi, Gerber, and Barnes]{hdltex}
K.~Kowsari, D.~E. Brown, M.~Heidarysafa, K.~Jafari~Meimandi, M.~S. Gerber, and L.~E. Barnes.
\newblock Hdltex: Hierarchical deep learning for text classification.
\newblock In \emph{2017 16th IEEE International Conference on Machine Learning and Applications (ICMLA)}, pages 364--371, 2017.
\newblock \doi{10.1109/ICMLA.2017.0-134}.

\bibitem[Krause et~al.(2013)Krause, Stark, Deng, and Fei-Fei]{krause20133d}
J.~Krause, M.~Stark, J.~Deng, and L.~Fei-Fei.
\newblock 3d object representations for fine-grained categorization.
\newblock In \emph{Proceedings of the IEEE international conference on computer vision workshops}, pages 554--561, 2013.

\bibitem[Krizhevsky et~al.(2009)Krizhevsky, Hinton, et~al.]{cifar10}
A.~Krizhevsky, G.~Hinton, et~al.
\newblock Learning multiple layers of features from tiny images.
\newblock 2009.

\bibitem[Krizhevsky et~al.(2012)Krizhevsky, Sutskever, and Hinton]{alexnet}
A.~Krizhevsky, I.~Sutskever, and G.~E. Hinton.
\newblock Imagenet classification with deep convolutional neural networks.
\newblock In F.~Pereira, C.~Burges, L.~Bottou, and K.~Weinberger, editors, \emph{Advances in Neural Information Processing Systems}, volume~25. Curran Associates, Inc., 2012.
\newblock URL \url{https://proceedings.neurips.cc/paper_files/paper/2012/file/c399862d3b9d6b76c8436e924a68c45b-Paper.pdf}.

\bibitem[Kumari et~al.(2023)Kumari, Zhang, Zhang, Shechtman, and Zhu]{kumari2023multi}
N.~Kumari, B.~Zhang, R.~Zhang, E.~Shechtman, and J.-Y. Zhu.
\newblock Multi-concept customization of text-to-image diffusion.
\newblock In \emph{Proceedings of the IEEE/CVF Conference on Computer Vision and Pattern Recognition}, pages 1931--1941, 2023.

\bibitem[Lai et~al.(2015)Lai, Xu, Liu, and Zhao]{rcnn}
S.~Lai, L.~Xu, K.~Liu, and J.~Zhao.
\newblock Recurrent convolutional neural networks for text classification.
\newblock In \emph{Proceedings of the Twenty-Ninth AAAI Conference on Artificial Intelligence}, AAAI'15, page 2267–2273. AAAI Press, 2015.
\newblock ISBN 0262511290.

\bibitem[LeCun et~al.(2010)LeCun, Cortes, and Burges]{lecun2010mnist}
Y.~LeCun, C.~Cortes, and C.~Burges.
\newblock Mnist handwritten digit database, 2010.

\bibitem[Leike et~al.(2018)Leike, Krueger, Everitt, Martic, Maini, and Legg]{leike2018scalable}
J.~Leike, D.~Krueger, T.~Everitt, M.~Martic, V.~Maini, and S.~Legg.
\newblock Scalable agent alignment via reward modeling: a research direction.
\newblock \emph{arXiv preprint arXiv:1811.07871}, 2018.

\bibitem[Li et~al.(2019)Li, Qi, Lukasiewicz, and Torr]{controlgan}
B.~Li, X.~Qi, T.~Lukasiewicz, and P.~Torr.
\newblock Controllable text-to-image generation.
\newblock In H.~Wallach, H.~Larochelle, A.~Beygelzimer, F.~d\textquotesingle Alch\'{e}-Buc, E.~Fox, and R.~Garnett, editors, \emph{Advances in Neural Information Processing Systems}, volume~32. Curran Associates, Inc., 2019.
\newblock URL \url{https://proceedings.neurips.cc/paper_files/paper/2019/file/1d72310edc006dadf2190caad5802983-Paper.pdf}.

\bibitem[Li and Hoiem(2017)]{LwF}
Z.~Li and D.~Hoiem.
\newblock Learning without forgetting.
\newblock \emph{IEEE transactions on pattern analysis and machine intelligence}, 40\penalty0 (12):\penalty0 2935--2947, 2017.

\bibitem[Lin et~al.(2024{\natexlab{a}})Lin, Huang, Ye, Chen, Wang, Li, Ma, Wan, Zou, and Liang]{lin2024selecting}
H.~Lin, B.~Huang, H.~Ye, Q.~Chen, Z.~Wang, S.~Li, J.~Ma, X.~Wan, J.~Zou, and Y.~Liang.
\newblock Selecting large language model to fine-tune via rectified scaling law.
\newblock \emph{arXiv preprint arXiv:2402.02314}, 2024{\natexlab{a}}.

\bibitem[Lin et~al.(2024{\natexlab{b}})Lin, Shao, Qian, Pan, Guo, and Liu]{lin2024class}
H.~Lin, Y.~Shao, W.~Qian, N.~Pan, Y.~Guo, and B.~Liu.
\newblock Class incremental learning via likelihood ratio based task prediction.
\newblock \emph{International Conference on Learning Representations (ICLR)}, 2024{\natexlab{b}}.

\bibitem[Liu et~al.(2023{\natexlab{a}})Liu, Mazumder, Robertson, and Grigsby]{liu2023ai}
B.~Liu, S.~Mazumder, E.~Robertson, and S.~Grigsby.
\newblock Ai autonomy: Self-initiated open-world continual learning and adaptation.
\newblock \emph{AI Magazine}, 44\penalty0 (2):\penalty0 185--199, 2023{\natexlab{a}}.

\bibitem[Liu et~al.(2022)Liu, Gong, and Liu]{Liu2022FlowSA}
X.~Liu, C.~Gong, and Q.~Liu.
\newblock Flow straight and fast: Learning to generate and transfer data with rectified flow.
\newblock \emph{ArXiv}, abs/2209.03003, 2022.
\newblock URL \url{https://api.semanticscholar.org/CorpusID:252111177}.

\bibitem[Liu et~al.(2023{\natexlab{b}})Liu, Zhang, Ma, Peng, and Liu]{Liu2023InstaFlowOS}
X.~Liu, X.~Zhang, J.~Ma, J.~Peng, and Q.~Liu.
\newblock Instaflow: One step is enough for high-quality diffusion-based text-to-image generation.
\newblock \emph{ArXiv}, abs/2309.06380, 2023{\natexlab{b}}.
\newblock URL \url{https://api.semanticscholar.org/CorpusID:261697392}.

\bibitem[Lomonaco et~al.(2021)Lomonaco, Pellegrini, Cossu, Carta, Graffieti, Hayes, De~Lange, Masana, Pomponi, Van~de Ven, et~al.]{lomonaco2021avalanche}
V.~Lomonaco, L.~Pellegrini, A.~Cossu, A.~Carta, G.~Graffieti, T.~L. Hayes, M.~De~Lange, M.~Masana, J.~Pomponi, G.~M. Van~de Ven, et~al.
\newblock Avalanche: an end-to-end library for continual learning.
\newblock In \emph{Proceedings of the IEEE/CVF Conference on Computer Vision and Pattern Recognition}, pages 3600--3610, 2021.

\bibitem[Lopez-Paz and Ranzato(2017)]{lopez2017gradient}
D.~Lopez-Paz and M.~Ranzato.
\newblock Gradient episodic memory for continual learning.
\newblock \emph{Advances in neural information processing systems}, 30, 2017.

\bibitem[Lu et~al.(2023)Lu, Yang, Fei, Huo, Lu, Luo, and Ding]{lu2023vdt}
H.~Lu, G.~Yang, N.~Fei, Y.~Huo, Z.~Lu, P.~Luo, and M.~Ding.
\newblock Vdt: General-purpose video diffusion transformers via mask modeling, 2023.

\bibitem[Madotto et~al.(2021)Madotto, Lin, Zhou, Moon, Crook, Liu, Yu, Cho, Fung, and Wang]{madotto-etal-2021-continual}
A.~Madotto, Z.~Lin, Z.~Zhou, S.~Moon, P.~Crook, B.~Liu, Z.~Yu, E.~Cho, P.~Fung, and Z.~Wang.
\newblock Continual learning in task-oriented dialogue systems.
\newblock In M.-F. Moens, X.~Huang, L.~Specia, and S.~W.-t. Yih, editors, \emph{Proceedings of the 2021 Conference on Empirical Methods in Natural Language Processing}, pages 7452--7467, Online and Punta Cana, Dominican Republic, Nov. 2021. Association for Computational Linguistics.
\newblock \doi{10.18653/v1/2021.emnlp-main.590}.
\newblock URL \url{https://aclanthology.org/2021.emnlp-main.590}.

\bibitem[Masana et~al.(2021)Masana, Tuytelaars, and Van~de Weijer]{masana2021ternary}
M.~Masana, T.~Tuytelaars, and J.~Van~de Weijer.
\newblock Ternary feature masks: zero-forgetting for task-incremental learning.
\newblock In \emph{Proceedings of the IEEE/CVF conference on computer vision and pattern recognition}, pages 3570--3579, 2021.

\bibitem[Masip et~al.(2024)Masip, Rodriguez, Tuytelaars, and van~de Ven]{masip2024continual}
S.~Masip, P.~Rodriguez, T.~Tuytelaars, and G.~M. van~de Ven.
\newblock Continual learning of diffusion models with generative distillation, 2024.

\bibitem[McCloskey and Cohen(1989{\natexlab{a}})]{CF}
M.~McCloskey and N.~J. Cohen.
\newblock Catastrophic interference in connectionist networks: The sequential learning problem.
\newblock volume~24 of \emph{Psychology of Learning and Motivation}, pages 109--165. Academic Press, 1989{\natexlab{a}}.
\newblock \doi{https://doi.org/10.1016/S0079-7421(08)60536-8}.
\newblock URL \url{https://www.sciencedirect.com/science/article/pii/S0079742108605368}.

\bibitem[McCloskey and Cohen(1989{\natexlab{b}})]{McCloskey1989}
M.~McCloskey and N.~J. Cohen.
\newblock {Catastrophic interference in connectionist networks: The sequential learning problem}.
\newblock In \emph{Psychology of learning and motivation}, volume~24, pages 109--165. Elsevier, 1989{\natexlab{b}}.

\bibitem[Minaee et~al.(2021)Minaee, Kalchbrenner, Cambria, Nikzad, Chenaghlu, and Gao]{minaee2021deep}
S.~Minaee, N.~Kalchbrenner, E.~Cambria, N.~Nikzad, M.~Chenaghlu, and J.~Gao.
\newblock Deep learning based text classification: A comprehensive review, 2021.

\bibitem[Nichol and Dhariwal(2021)]{Nichol2021ImprovedDD}
A.~Nichol and P.~Dhariwal.
\newblock Improved denoising diffusion probabilistic models.
\newblock \emph{ArXiv}, abs/2102.09672, 2021.
\newblock URL \url{https://api.semanticscholar.org/CorpusID:231979499}.

\bibitem[Nilsback and Zisserman(2008)]{nilsback2008automated}
M.-E. Nilsback and A.~Zisserman.
\newblock Automated flower classification over a large number of classes.
\newblock In \emph{2008 Sixth Indian conference on computer vision, graphics \& image processing}, pages 722--729. IEEE, 2008.

\bibitem[Oquab et~al.(2023)Oquab, Darcet, Moutakanni, Vo, Szafraniec, Khalidov, Fernandez, Haziza, Massa, El-Nouby, Assran, Ballas, Galuba, Howes, Huang, Li, Misra, Rabbat, Sharma, Synnaeve, Xu, J{\'e}gou, Mairal, Labatut, Joulin, and Bojanowski]{Oquab2023DINOv2LR}
M.~Oquab, T.~Darcet, T.~Moutakanni, H.~Q. Vo, M.~Szafraniec, V.~Khalidov, P.~Fernandez, D.~Haziza, F.~Massa, A.~El-Nouby, M.~Assran, N.~Ballas, W.~Galuba, R.~Howes, P.-Y.~B. Huang, S.-W. Li, I.~Misra, M.~G. Rabbat, V.~Sharma, G.~Synnaeve, H.~Xu, H.~J{\'e}gou, J.~Mairal, P.~Labatut, A.~Joulin, and P.~Bojanowski.
\newblock Dinov2: Learning robust visual features without supervision.
\newblock \emph{ArXiv}, abs/2304.07193, 2023.
\newblock URL \url{https://api.semanticscholar.org/CorpusID:258170077}.

\bibitem[Park et~al.(2019)Park, Liu, Wang, and Zhu]{park2019semantic}
T.~Park, M.-Y. Liu, T.-C. Wang, and J.-Y. Zhu.
\newblock Semantic image synthesis with spatially-adaptive normalization, 2019.

\bibitem[Peebles and Xie(2022)]{Peebles2022ScalableDM}
W.~S. Peebles and S.~Xie.
\newblock Scalable diffusion models with transformers.
\newblock \emph{2023 IEEE/CVF International Conference on Computer Vision (ICCV)}, pages 4172--4182, 2022.
\newblock URL \url{https://api.semanticscholar.org/CorpusID:254854389}.

\bibitem[Pesovski et~al.(2024)Pesovski, Santos, Henriques, and Trajkovik]{pesovski2024generative}
I.~Pesovski, R.~Santos, R.~Henriques, and V.~Trajkovik.
\newblock Generative ai for customizable learning experiences.
\newblock \emph{Sustainability}, 16\penalty0 (7):\penalty0 3034, 2024.

\bibitem[Popescu et~al.(2009)Popescu, Balas, Perescu-Popescu, and Mastorakis]{popescu2009multilayer}
M.-C. Popescu, V.~E. Balas, L.~Perescu-Popescu, and N.~Mastorakis.
\newblock Multilayer perceptron and neural networks.
\newblock \emph{WSEAS Transactions on Circuits and Systems}, 8\penalty0 (7):\penalty0 579--588, 2009.

\bibitem[Radford et~al.(2016)Radford, Metz, and Chintala]{radford2016unsupervised}
A.~Radford, L.~Metz, and S.~Chintala.
\newblock Unsupervised representation learning with deep convolutional generative adversarial networks, 2016.

\bibitem[Radford et~al.(2019)Radford, Wu, Child, Luan, Amodei, Sutskever, et~al.]{radford2019language}
A.~Radford, J.~Wu, R.~Child, D.~Luan, D.~Amodei, I.~Sutskever, et~al.
\newblock Language models are unsupervised multitask learners.
\newblock \emph{OpenAI blog}, 1\penalty0 (8):\penalty0 9, 2019.

\bibitem[Radford et~al.(2021)Radford, Kim, Hallacy, Ramesh, Goh, Agarwal, Sastry, Askell, Mishkin, Clark, et~al.]{radford2021learning}
A.~Radford, J.~W. Kim, C.~Hallacy, A.~Ramesh, G.~Goh, S.~Agarwal, G.~Sastry, A.~Askell, P.~Mishkin, J.~Clark, et~al.
\newblock Learning transferable visual models from natural language supervision.
\newblock In \emph{International conference on machine learning}, pages 8748--8763. PMLR, 2021.

\bibitem[Rafieian and Yoganarasimhan(2023)]{rafieian2023ai}
O.~Rafieian and H.~Yoganarasimhan.
\newblock Ai and personalization.
\newblock \emph{Artificial Intelligence in Marketing}, pages 77--102, 2023.

\bibitem[Ramesh et~al.(2021)Ramesh, Pavlov, Goh, Gray, Voss, Radford, Chen, and Sutskever]{ramesh2021zeroshot}
A.~Ramesh, M.~Pavlov, G.~Goh, S.~Gray, C.~Voss, A.~Radford, M.~Chen, and I.~Sutskever.
\newblock Zero-shot text-to-image generation, 2021.

\bibitem[Rebuffi et~al.(2017)Rebuffi, Kolesnikov, Sperl, and Lampert]{rebuffi2017icarl}
S.-A. Rebuffi, A.~Kolesnikov, G.~Sperl, and C.~H. Lampert.
\newblock icarl: Incremental classifier and representation learning.
\newblock In \emph{Proceedings of the IEEE conference on Computer Vision and Pattern Recognition}, pages 2001--2010, 2017.

\bibitem[Riemer et~al.(2018)Riemer, Cases, Ajemian, Liu, Rish, Tu, and Tesauro]{riemer2018learning}
M.~Riemer, I.~Cases, R.~Ajemian, M.~Liu, I.~Rish, Y.~Tu, and G.~Tesauro.
\newblock Learning to learn without forgetting by maximizing transfer and minimizing interference.
\newblock \emph{arXiv preprint arXiv:1810.11910}, 2018.

\bibitem[Rombach et~al.(2022)Rombach, Blattmann, Lorenz, Esser, and Ommer]{stablediffusion}
R.~Rombach, A.~Blattmann, D.~Lorenz, P.~Esser, and B.~Ommer.
\newblock High-resolution image synthesis with latent diffusion models, 2022.

\bibitem[Roth et~al.(2017)Roth, Lucchi, Nowozin, and Hofmann]{roth2017stabilizing}
K.~Roth, A.~Lucchi, S.~Nowozin, and T.~Hofmann.
\newblock Stabilizing training of generative adversarial networks through regularization.
\newblock \emph{Advances in neural information processing systems}, 30, 2017.

\bibitem[Ruiz et~al.(2023)Ruiz, Li, Jampani, Pritch, Rubinstein, and Aberman]{ruiz2023dreambooth}
N.~Ruiz, Y.~Li, V.~Jampani, Y.~Pritch, M.~Rubinstein, and K.~Aberman.
\newblock Dreambooth: Fine tuning text-to-image diffusion models for subject-driven generation.
\newblock In \emph{Proceedings of the IEEE/CVF Conference on Computer Vision and Pattern Recognition}, pages 22500--22510, 2023.

\bibitem[Russakovsky et~al.(2015)Russakovsky, Deng, Su, Krause, Satheesh, Ma, Huang, Karpathy, Khosla, Bernstein, Berg, and Fei-Fei]{imagenet15russakovsky}
O.~Russakovsky, J.~Deng, H.~Su, J.~Krause, S.~Satheesh, S.~Ma, Z.~Huang, A.~Karpathy, A.~Khosla, M.~Bernstein, A.~C. Berg, and L.~Fei-Fei.
\newblock {ImageNet Large Scale Visual Recognition Challenge}.
\newblock \emph{International Journal of Computer Vision (IJCV)}, 115\penalty0 (3):\penalty0 211--252, 2015.
\newblock \doi{10.1007/s11263-015-0816-y}.

\bibitem[Rusu et~al.(2016)Rusu, Rabinowitz, Desjardins, Soyer, Kirkpatrick, Kavukcuoglu, Pascanu, and Hadsell]{rusu2016progressive}
A.~A. Rusu, N.~C. Rabinowitz, G.~Desjardins, H.~Soyer, J.~Kirkpatrick, K.~Kavukcuoglu, R.~Pascanu, and R.~Hadsell.
\newblock Progressive neural networks.
\newblock \emph{arXiv preprint arXiv:1606.04671}, 2016.

\bibitem[Saharia et~al.(2022)Saharia, Chan, Saxena, Li, Whang, Denton, Ghasemipour, Ayan, Mahdavi, Lopes, Salimans, Ho, Fleet, and Norouzi]{saharia2022photorealistic}
C.~Saharia, W.~Chan, S.~Saxena, L.~Li, J.~Whang, E.~Denton, S.~K.~S. Ghasemipour, B.~K. Ayan, S.~S. Mahdavi, R.~G. Lopes, T.~Salimans, J.~Ho, D.~J. Fleet, and M.~Norouzi.
\newblock Photorealistic text-to-image diffusion models with deep language understanding, 2022.

\bibitem[Seff et~al.(2017)Seff, Beatson, Suo, and Liu]{seff2017continual}
A.~Seff, A.~Beatson, D.~Suo, and H.~Liu.
\newblock Continual learning in generative adversarial nets, 2017.

\bibitem[Seo et~al.(2023)Seo, Kang, and Park]{LFS-GAN}
J.~Seo, J.~Kang, and G.~Park.
\newblock Lfs-gan: Lifelong few-shot image generation.
\newblock In \emph{2023 IEEE/CVF International Conference on Computer Vision (ICCV)}, pages 11322--11332, Los Alamitos, CA, USA, oct 2023. IEEE Computer Society.
\newblock \doi{10.1109/ICCV51070.2023.01043}.
\newblock URL \url{https://doi.ieeecomputersociety.org/10.1109/ICCV51070.2023.01043}.

\bibitem[Serra et~al.(2018)Serra, Suris, Miron, and Karatzoglou]{serra2018overcoming}
J.~Serra, D.~Suris, M.~Miron, and A.~Karatzoglou.
\newblock Overcoming catastrophic forgetting with hard attention to the task.
\newblock In \emph{International Conference on Machine Learning}, pages 4548--4557. PMLR, 2018.

\bibitem[Serrà et~al.(2018)Serrà, Surís, Miron, and Karatzoglou]{serrà2018overcoming}
J.~Serrà, D.~Surís, M.~Miron, and A.~Karatzoglou.
\newblock Overcoming catastrophic forgetting with hard attention to the task, 2018.

\bibitem[Shahin et~al.(2024)Shahin, Chen, and Hosseinzadeh]{shahin2024harnessing}
M.~Shahin, F.~F. Chen, and A.~Hosseinzadeh.
\newblock Harnessing customized ai to create voice of customer via gpt3. 5.
\newblock \emph{Advanced Engineering Informatics}, 61:\penalty0 102462, 2024.

\bibitem[Shao et~al.(2023)Shao, Guo, Zhao, and Liu]{shao2023class}
Y.~Shao, Y.~Guo, D.~Zhao, and B.~Liu.
\newblock Class-incremental learning based on label generation.
\newblock \emph{arXiv preprint arXiv:2306.12619}, 2023.

\bibitem[Shi et~al.(2024)Shi, Xu, Wang, Qin, Wang, Wang, and Wang]{shi2024continual}
H.~Shi, Z.~Xu, H.~Wang, W.~Qin, W.~Wang, Y.~Wang, and H.~Wang.
\newblock Continual learning of large language models: A comprehensive survey.
\newblock \emph{arXiv preprint arXiv:2404.16789}, 2024.

\bibitem[Shi et~al.(2023)Shi, Peng, Xu, Geiger, Liao, and Shen]{shi2023deep}
Z.~Shi, S.~Peng, Y.~Xu, A.~Geiger, Y.~Liao, and Y.~Shen.
\newblock Deep generative models on 3d representations: A survey, 2023.

\bibitem[Shin et~al.(2017)Shin, Lee, Kim, and Kim]{dgr}
H.~Shin, J.~K. Lee, J.~Kim, and J.~Kim.
\newblock Continual learning with deep generative replay.
\newblock \emph{Advances in neural information processing systems}, 30, 2017.

\bibitem[Simonyan and Zisserman(2014)]{vgg}
K.~Simonyan and A.~Zisserman.
\newblock Very deep convolutional networks for large-scale image recognition.
\newblock \emph{CoRR}, abs/1409.1556, 2014.
\newblock URL \url{https://api.semanticscholar.org/CorpusID:14124313}.

\bibitem[Singer et~al.(2023)Singer, Polyak, Hayes, Yin, An, Zhang, Hu, Yang, Ashual, Gafni, Parikh, Gupta, and Taigman]{singer2023makeavideo}
U.~Singer, A.~Polyak, T.~Hayes, X.~Yin, J.~An, S.~Zhang, Q.~Hu, H.~Yang, O.~Ashual, O.~Gafni, D.~Parikh, S.~Gupta, and Y.~Taigman.
\newblock Make-a-video: Text-to-video generation without text-video data.
\newblock In \emph{The Eleventh International Conference on Learning Representations}, 2023.
\newblock URL \url{https://openreview.net/forum?id=nJfylDvgzlq}.

\bibitem[Smith et~al.(2023)Smith, Tian, Halbe, Hsu, and Kira]{l2}
J.~S. Smith, J.~Tian, S.~Halbe, Y.-C. Hsu, and Z.~Kira.
\newblock A closer look at rehearsal-free continual learning.
\newblock In \emph{Proceedings of the IEEE/CVF Conference on Computer Vision and Pattern Recognition}, pages 2409--2419, 2023.

\bibitem[Smith et~al.(2024)Smith, Hsu, Zhang, Hua, Kira, Shen, and Jin]{smith2024continual}
J.~S. Smith, Y.-C. Hsu, L.~Zhang, T.~Hua, Z.~Kira, Y.~Shen, and H.~Jin.
\newblock Continual diffusion: Continual customization of text-to-image diffusion with c-lora, 2024.

\bibitem[Soares and Fallenstein(2014)]{soares2014aligning}
N.~Soares and B.~Fallenstein.
\newblock Aligning superintelligence with human interests: A technical research agenda.
\newblock \emph{Machine Intelligence Research Institute (MIRI) technical report}, 8, 2014.

\bibitem[Sohl-Dickstein et~al.(2015)Sohl-Dickstein, Weiss, Maheswaranathan, and Ganguli]{sohldickstein2015deep}
J.~Sohl-Dickstein, E.~A. Weiss, N.~Maheswaranathan, and S.~Ganguli.
\newblock Deep unsupervised learning using nonequilibrium thermodynamics, 2015.

\bibitem[Song et~al.(2020{\natexlab{a}})Song, Meng, and Ermon]{ddim}
J.~Song, C.~Meng, and S.~Ermon.
\newblock Denoising diffusion implicit models.
\newblock \emph{arXiv preprint arXiv:2010.02502}, 2020{\natexlab{a}}.

\bibitem[Song and Ermon(2020)]{song2020improved}
Y.~Song and S.~Ermon.
\newblock Improved techniques for training score-based generative models.
\newblock \emph{Advances in neural information processing systems}, 33:\penalty0 12438--12448, 2020.

\bibitem[Song et~al.(2020{\natexlab{b}})Song, Sohl-Dickstein, Kingma, Kumar, Ermon, and Poole]{song2020score}
Y.~Song, J.~Sohl-Dickstein, D.~P. Kingma, A.~Kumar, S.~Ermon, and B.~Poole.
\newblock Score-based generative modeling through stochastic differential equations.
\newblock \emph{arXiv preprint arXiv:2011.13456}, 2020{\natexlab{b}}.

\bibitem[Song et~al.(2021)Song, Sohl-Dickstein, Kingma, Kumar, Ermon, and Poole]{song2021scorebased}
Y.~Song, J.~Sohl-Dickstein, D.~P. Kingma, A.~Kumar, S.~Ermon, and B.~Poole.
\newblock Score-based generative modeling through stochastic differential equations, 2021.

\bibitem[Srivastava et~al.(2017)Srivastava, Valkov, Russell, Gutmann, and Sutton]{Srivastava2017VEEGANRM}
A.~Srivastava, L.~Valkov, C.~Russell, M.~U. Gutmann, and C.~Sutton.
\newblock Veegan: Reducing mode collapse in gans using implicit variational learning.
\newblock In \emph{Neural Information Processing Systems}, 2017.
\newblock URL \url{https://api.semanticscholar.org/CorpusID:9302801}.

\bibitem[Sun et~al.(2020)Sun, Ho, and Lee]{sun2020lamal}
F.-K. Sun, C.-H. Ho, and H.-Y. Lee.
\newblock {\{}LAMAL{\}}: {\{}LA{\}}nguage modeling is all you need for lifelong language learning.
\newblock In \emph{International Conference on Learning Representations}, 2020.
\newblock URL \url{https://openreview.net/forum?id=Skgxcn4YDS}.

\bibitem[Sun et~al.(2024)Sun, Liang, Dong, Li, Ding, and Cong]{sun2024create}
G.~Sun, W.~Liang, J.~Dong, J.~Li, Z.~Ding, and Y.~Cong.
\newblock Create your world: Lifelong text-to-image diffusion.
\newblock \emph{IEEE Transactions on Pattern Analysis and Machine Intelligence}, 2024.

\bibitem[Szegedy et~al.(2015)Szegedy, Liu, Jia, Sermanet, Reed, Anguelov, Erhan, Vanhoucke, and Rabinovich]{googlenet}
C.~Szegedy, W.~Liu, Y.~Jia, P.~Sermanet, S.~Reed, D.~Anguelov, D.~Erhan, V.~Vanhoucke, and A.~Rabinovich.
\newblock Going deeper with convolutions.
\newblock In \emph{2015 IEEE Conference on Computer Vision and Pattern Recognition (CVPR)}, pages 1--9, 2015.
\newblock \doi{10.1109/CVPR.2015.7298594}.

\bibitem[Team et~al.(2023)Team, Anil, Borgeaud, Wu, Alayrac, Yu, Soricut, Schalkwyk, Dai, Hauth, et~al.]{team2023gemini}
G.~Team, R.~Anil, S.~Borgeaud, Y.~Wu, J.-B. Alayrac, J.~Yu, R.~Soricut, J.~Schalkwyk, A.~M. Dai, A.~Hauth, et~al.
\newblock Gemini: a family of highly capable multimodal models.
\newblock \emph{arXiv preprint arXiv:2312.11805}, 2023.

\bibitem[Van~de Ven and Tolias(2019)]{van2019three}
G.~M. Van~de Ven and A.~S. Tolias.
\newblock Three scenarios for continual learning.
\newblock \emph{arXiv preprint arXiv:1904.07734}, 2019.

\bibitem[van~den Oord et~al.(2016)van~den Oord, Dieleman, Zen, Simonyan, Vinyals, Graves, Kalchbrenner, Senior, and Kavukcuoglu]{oord2016wavenet}
A.~van~den Oord, S.~Dieleman, H.~Zen, K.~Simonyan, O.~Vinyals, A.~Graves, N.~Kalchbrenner, A.~Senior, and K.~Kavukcuoglu.
\newblock Wavenet: A generative model for raw audio, 2016.

\bibitem[Varshney et~al.(2021)Varshney, Verma, SrijithP., Carin, and Rai]{CAM-GAN}
S.~Varshney, V.~K. Verma, K.~SrijithP., L.~Carin, and P.~Rai.
\newblock Cam-gan: Continual adaptation modules for generative adversarial networks.
\newblock In \emph{Neural Information Processing Systems}, 2021.
\newblock URL \url{https://api.semanticscholar.org/CorpusID:236635024}.

\bibitem[Verwimp et~al.(2023)Verwimp, Ben-David, Bethge, Cossu, Gepperth, Hayes, H{\"u}llermeier, Kanan, Kudithipudi, Lampert, et~al.]{verwimp2023continual}
E.~Verwimp, S.~Ben-David, M.~Bethge, A.~Cossu, A.~Gepperth, T.~L. Hayes, E.~H{\"u}llermeier, C.~Kanan, D.~Kudithipudi, C.~H. Lampert, et~al.
\newblock Continual learning: Applications and the road forward.
\newblock \emph{arXiv preprint arXiv:2311.11908}, 2023.

\bibitem[Vitter(1985)]{vitter1985random}
J.~S. Vitter.
\newblock Random sampling with a reservoir.
\newblock \emph{ACM Transactions on Mathematical Software (TOMS)}, 11\penalty0 (1):\penalty0 37--57, 1985.

\bibitem[Wah et~al.(2011)Wah, Branson, Welinder, Perona, and Belongie]{wah2011caltech}
C.~Wah, S.~Branson, P.~Welinder, P.~Perona, and S.~Belongie.
\newblock The caltech-ucsd birds-200-2011 dataset.
\newblock 2011.

\bibitem[Wang et~al.(2022)Wang, Zhou, Liu, Ye, Bian, Zhan, and Zhao]{wang2022beef}
F.-Y. Wang, D.-W. Zhou, L.~Liu, H.-J. Ye, Y.~Bian, D.-C. Zhan, and P.~Zhao.
\newblock Beef: Bi-compatible class-incremental learning via energy-based expansion and fusion.
\newblock In \emph{The Eleventh International Conference on Learning Representations}, 2022.

\bibitem[Wang et~al.(2023)Wang, Zhang, Su, and Zhu]{wang2023comprehensive}
L.~Wang, X.~Zhang, H.~Su, and J.~Zhu.
\newblock A comprehensive survey of continual learning: Theory, method and application, 2023.

\bibitem[Wang et~al.(2024)Wang, Chen, Ma, Zhou, Huang, Wang, Yang, He, Yu, Yang, Guo, Wu, Si, Jiang, Chen, Loy, Dai, Lin, Qiao, and Liu]{wang2024lavie}
Y.~Wang, X.~Chen, X.~Ma, S.~Zhou, Z.~Huang, Y.~Wang, C.~Yang, Y.~He, J.~Yu, P.~Yang, Y.~Guo, T.~Wu, C.~Si, Y.~Jiang, C.~Chen, C.~C. Loy, B.~Dai, D.~Lin, Y.~Qiao, and Z.~Liu.
\newblock Lavie: High-quality video generation with cascaded latent diffusion models, 2024.
\newblock URL \url{https://openreview.net/forum?id=p09XyFxZkc}.

\bibitem[Wortsman et~al.(2020{\natexlab{a}})Wortsman, Ramanujan, Liu, Kembhavi, Rastegari, Yosinski, and Farhadi]{supsup}
M.~Wortsman, V.~Ramanujan, R.~Liu, A.~Kembhavi, M.~Rastegari, J.~Yosinski, and A.~Farhadi.
\newblock Supermasks in superposition.
\newblock In H.~Larochelle, M.~Ranzato, R.~Hadsell, M.~Balcan, and H.~Lin, editors, \emph{Advances in Neural Information Processing Systems}, volume~33, pages 15173--15184. Curran Associates, Inc., 2020{\natexlab{a}}.
\newblock URL \url{https://proceedings.neurips.cc/paper_files/paper/2020/file/ad1f8bb9b51f023cdc80cf94bb615aa9-Paper.pdf}.

\bibitem[Wortsman et~al.(2020{\natexlab{b}})Wortsman, Ramanujan, Liu, Kembhavi, Rastegari, Yosinski, and Farhadi]{wortsman2020supermasks}
M.~Wortsman, V.~Ramanujan, R.~Liu, A.~Kembhavi, M.~Rastegari, J.~Yosinski, and A.~Farhadi.
\newblock Supermasks in superposition.
\newblock \emph{Advances in Neural Information Processing Systems}, 33:\penalty0 15173--15184, 2020{\natexlab{b}}.

\bibitem[Wu et~al.(2018{\natexlab{a}})Wu, Herranz, Liu, Van De~Weijer, Raducanu, et~al.]{mergan}
C.~Wu, L.~Herranz, X.~Liu, J.~Van De~Weijer, B.~Raducanu, et~al.
\newblock Memory replay gans: Learning to generate new categories without forgetting.
\newblock \emph{Advances in neural information processing systems}, 31, 2018{\natexlab{a}}.

\bibitem[Wu et~al.(2018{\natexlab{b}})Wu, Herranz, Liu, Wang, van~de Weijer, and Raducanu]{Wu2018MemoryRG}
C.~Wu, L.~Herranz, X.~Liu, Y.~Wang, J.~van~de Weijer, and B.~Raducanu.
\newblock Memory replay gans: learning to generate images from new categories without forgetting.
\newblock In \emph{Neural Information Processing Systems}, 2018{\natexlab{b}}.
\newblock URL \url{https://api.semanticscholar.org/CorpusID:55701876}.

\bibitem[Wu et~al.(2024)Wu, Luo, Li, Pan, Vu, and Haffari]{wu2024continual}
T.~Wu, L.~Luo, Y.-F. Li, S.~Pan, T.-T. Vu, and G.~Haffari.
\newblock Continual learning for large language models: A survey.
\newblock \emph{arXiv preprint arXiv:2402.01364}, 2024.

\bibitem[Xiao et~al.(2017{\natexlab{a}})Xiao, Rasul, and Vollgraf]{fashionmnist}
H.~Xiao, K.~Rasul, and R.~Vollgraf.
\newblock Fashion-mnist: a novel image dataset for benchmarking machine learning algorithms.
\newblock \emph{CoRR}, abs/1708.07747, 2017{\natexlab{a}}.
\newblock URL \url{http://arxiv.org/abs/1708.07747}.

\bibitem[Xiao et~al.(2017{\natexlab{b}})Xiao, Rasul, and Vollgraf]{xiao2017fashionmnist}
H.~Xiao, K.~Rasul, and R.~Vollgraf.
\newblock Fashion-mnist: a novel image dataset for benchmarking machine learning algorithms, 2017{\natexlab{b}}.

\bibitem[Xing et~al.(2023)Xing, Dai, Hu, Wu, and Jiang]{xing2023simda}
Z.~Xing, Q.~Dai, H.~Hu, Z.~Wu, and Y.-G. Jiang.
\newblock Simda: Simple diffusion adapter for efficient video generation, 2023.

\bibitem[Xu et~al.(2022)Xu, Yu, Song, Shi, Ermon, and Tang]{xu2022geodiff}
M.~Xu, L.~Yu, Y.~Song, C.~Shi, S.~Ermon, and J.~Tang.
\newblock Geodiff: A geometric diffusion model for molecular conformation generation.
\newblock In \emph{International Conference on Learning Representations}, 2022.
\newblock URL \url{https://openreview.net/forum?id=PzcvxEMzvQC}.

\bibitem[Yan et~al.(2021)Yan, Xie, and He]{yan2021dynamically}
S.~Yan, J.~Xie, and X.~He.
\newblock Der: Dynamically expandable representation for class incremental learning.
\newblock In \emph{Proceedings of the IEEE/CVF Conference on Computer Vision and Pattern Recognition}, pages 3014--3023, 2021.

\bibitem[Yang et~al.(2016)Yang, Yang, Dyer, He, Smola, and Hovy]{yang-etal-2016-hierarchical}
Z.~Yang, D.~Yang, C.~Dyer, X.~He, A.~Smola, and E.~Hovy.
\newblock Hierarchical attention networks for document classification.
\newblock In K.~Knight, A.~Nenkova, and O.~Rambow, editors, \emph{Proceedings of the 2016 Conference of the North {A}merican Chapter of the Association for Computational Linguistics: Human Language Technologies}, pages 1480--1489, San Diego, California, June 2016. Association for Computational Linguistics.
\newblock \doi{10.18653/v1/N16-1174}.
\newblock URL \url{https://aclanthology.org/N16-1174}.

\bibitem[Yin et~al.(2022)Yin, Li, and Xiong]{yin2022contintin}
W.~Yin, J.~Li, and C.~Xiong.
\newblock Contintin: Continual learning from task instructions.
\newblock \emph{arXiv preprint arXiv:2203.08512}, 2022.

\bibitem[Yu et~al.(2021)Yu, Srivastava, and Canales]{Yu_2021}
Y.~Yu, A.~Srivastava, and S.~Canales.
\newblock Conditional lstm-gan for melody generation from lyrics.
\newblock \emph{ACM Transactions on Multimedia Computing, Communications, and Applications}, 17\penalty0 (1):\penalty0 1–20, Feb. 2021.
\newblock ISSN 1551-6865.
\newblock \doi{10.1145/3424116}.
\newblock URL \url{http://dx.doi.org/10.1145/3424116}.

\bibitem[Zając et~al.(2023)Zając, Deja, Kuzina, Tomczak, Trzciński, Shkurti, and Miłoś]{zając2023exploring}
M.~Zając, K.~Deja, A.~Kuzina, J.~M. Tomczak, T.~Trzciński, F.~Shkurti, and P.~Miłoś.
\newblock Exploring continual learning of diffusion models, 2023.

\bibitem[Zeng et~al.(2022)Zeng, Vahdat, Williams, Gojcic, Litany, Fidler, and Kreis]{zeng2022lion}
X.~Zeng, A.~Vahdat, F.~Williams, Z.~Gojcic, O.~Litany, S.~Fidler, and K.~Kreis.
\newblock Lion: Latent point diffusion models for 3d shape generation, 2022.

\bibitem[Zenke et~al.(2017)Zenke, Poole, and Ganguli]{si}
F.~Zenke, B.~Poole, and S.~Ganguli.
\newblock Continual learning through synaptic intelligence.
\newblock In \emph{International Conference on Machine Learning}, pages 3987--3995. PMLR, 2017.

\bibitem[Zhai et~al.(2019)Zhai, Chen, Tung, He, Nawhal, and Mori]{Zhai2019LifelongGC}
M.~Zhai, L.~Chen, F.~Tung, J.~He, M.~Nawhal, and G.~Mori.
\newblock Lifelong gan: Continual learning for conditional image generation.
\newblock \emph{2019 IEEE/CVF International Conference on Computer Vision (ICCV)}, pages 2759--2768, 2019.
\newblock URL \url{https://api.semanticscholar.org/CorpusID:198229709}.

\bibitem[Zhai et~al.(2021)Zhai, Chen, and Mori]{Zhai2021HyperLifelongGANSL}
M.~Zhai, L.~Chen, and G.~Mori.
\newblock Hyper-lifelonggan: Scalable lifelong learning for image conditioned generation.
\newblock \emph{2021 IEEE/CVF Conference on Computer Vision and Pattern Recognition (CVPR)}, pages 2246--2255, 2021.
\newblock URL \url{https://api.semanticscholar.org/CorpusID:232351216}.

\bibitem[Zhang et~al.(2023{\natexlab{a}})Zhang, Rao, and Agrawala]{Zhang_2023_ICCV}
L.~Zhang, A.~Rao, and M.~Agrawala.
\newblock Adding conditional control to text-to-image diffusion models.
\newblock In \emph{Proceedings of the IEEE/CVF International Conference on Computer Vision (ICCV)}, pages 3836--3847, October 2023{\natexlab{a}}.

\bibitem[Zhang et~al.(2023{\natexlab{b}})Zhang, Rao, and Agrawala]{zhang2023adding}
L.~Zhang, A.~Rao, and M.~Agrawala.
\newblock Adding conditional control to text-to-image diffusion models, 2023{\natexlab{b}}.

\bibitem[Zhu et~al.(2021)Zhu, Zhang, Wang, Yin, and Liu]{zhu2021prototype}
F.~Zhu, X.-Y. Zhang, C.~Wang, F.~Yin, and C.-L. Liu.
\newblock Prototype augmentation and self-supervision for incremental learning.
\newblock In \emph{Proceedings of the IEEE/CVF Conference on Computer Vision and Pattern Recognition}, pages 5871--5880, 2021.

\bibitem[Zhu et~al.(2020)Zhu, Park, Isola, and Efros]{zhu2020unpaired}
J.-Y. Zhu, T.~Park, P.~Isola, and A.~A. Efros.
\newblock Unpaired image-to-image translation using cycle-consistent adversarial networks, 2020.

\end{thebibliography}

\newpage

\appendix
\renewcommand\thefigure{\thesection.\arabic{figure}}
\setcounter{figure}{0}

\section{Related Work}
\label{sec:related_work}

\subsection{Generative Models}
\label{app:generative_models}

\paragraph{\textbf{Generative Adversarial Networks. (GAN)}} GAN \citep{goodfellow2014generative} consists of two interacting networks: a generator and a discriminator.  The generator  $G_{\theta_g}$, fed with random noise  $\bm z \sim p_{\bm z}$, is designed to produce images that mimic the true samples from a data distribution  $p_{\text{data}}$  closely enough to deceive the discriminator. Conversely, the discriminator  $D_{\theta_d}$  attempts to discern between the authentic data points $\bm x$ and the synthetic images  $G_{\theta_g}(\bm z)$  produced by the generator. The training objective for this adversarial process is formulated as follows~\citep{goodfellow2014generative} :
\begin{equation}
    \min_{\theta_g}\max_{\theta_d}[\E_{\bm x\sim p_{\text{data}}}\log D_{\theta_d}(\vx)+\E_{\bm z\sim p_{\bm z}}\log (1-D_{\theta_d}(G_{\theta_g}(\bm z))]
\end{equation}

\paragraph{\textbf{Diffusion Models.}} 
Diffusion probabilistic models \citep{sohldickstein2015deep, ho2020denoising, song2021scorebased} generate samples by an iterative denoising process. It defines a gradual process of adding noises, which is called the diffusion process or forward process and generate images by removing the noises step-by-step, which is referred to as the reverse process. In forward process, gaussian noises are added to ${\bm x}_t$, beginning from data ${\bm x}_0$ \citep{ho2020denoising}: 
\begin{equation}
    q({\bm x}_{t+1}|{\bm x}_{t})=\mathcal{N}(\sqrt{1-\beta_t}{\bm x}_t, \beta_t\bm{I}),~ 0\le t < T
\end{equation}
where $\beta_t$ stands for the variance schedule of noise added at time $t$. With diffusion steps $T \to \infty$, ${\bm x}_T$ virtually becomes a random noise sampled from $\mathcal{N}(0, \bm{I})$. In contrast, the reverse process starts from Gaussian noise ${\bm x}_T \sim \mathcal{N}(0, \bm{I})$, during which diffusion network predicts noises $x_t$ \citep{ho2020denoising}: 
\begin{equation}
    p_\theta({\bm x}_{t-1}|{\bm x}_{t})=\mathcal{N}({\bm \mu}_\theta({\bm x}_{t}, t), \bm{\Sigma}_\theta({\bm x}_t, t)), ~0< t \le T
\end{equation}
where ${\bm \epsilon}_\theta({\bm x}_t, t)$ is parameterized by a neural network and can be converted to $\bm {\mu}_\theta({\bm x}_t, t)$ with reparameterization trick \citep{kingma2022autoencoding} and ${\bm \Sigma}_\theta({\bm x}_t, t)=\sigma_t {\bm{I}}$ under the isotropic Gaussian assumption of noises \citep{ho2020denoising}. 
To learn the reverse process, diffusion models are trained by optimizing the variational lower bound \citep{kingma2023variational} of probability $p_\theta({\bm x}_{0:T})=p_\theta({\bm x}_T)\Pi_{t=1}^{T}p_\theta({\bm x}_{t-1}|{\bm x}_{t})$. One commonly used and simple loss equivalent is written as \citep{ho2020denoising}:
\begin{equation}
    L(\theta)=\E_{t,{\bm x}_0, {\bm \epsilon}}\|{{\bm \epsilon}-{\bm \epsilon}_\theta(\sqrt{\Bar{\alpha_t}}{\bm x}_0+\sqrt{1-\Bar{\alpha_t}}{\bm \epsilon}, t}\|^2
\end{equation}
where ${\bm \epsilon} \sim \mathcal{N}(0, \mathbf{I})$ and $\Bar{\alpha_t}=\Pi_{s=1}^t(1-\beta_s)$. The sampling process starts from $x_T\sim \mathcal{N}(0, \bm I)$, iterates $t=T, \dots, 1$ and denoises according to formula \citep{ho2020denoising}:
\begin{equation}
    \vx_{t-1} = \frac{1}{\sqrt{\alpha_t}}(\vx_t-\frac{\beta_t}{\sqrt{1-\Bar{\alpha_t}}}{\bm \epsilon_\theta(\vx_t, t)}) + \sigma_t {\bm z}
\end{equation}
where $\bm z \sim \mathcal{N}(0, \bm I)$ if $t>1$ else $\bm z=0$. 

\subsection{Continual learning of generative models}

Generative models have long been involved in continual learning, however, as an auxiliary measure in generative replay\citep{dgr}. Studies in settings where generative models behave as a CL agent are rather deficient. Relevant works are listed below, classified based on model architecture.

\paragraph{Continual Learning of GAN}
Since GAN \citep{goodfellow2014generative} was proposed, several works have brought out continual learning settings for GANs and incorporated different methods to overcome catastrophic forgetting. 
\citet{seff2017continual} first integrated EWC \citep{ewc} into continual learning for GANs.
\citet{Zhai2019LifelongGC} adopted knowledge distillation to distill knowledge from the previous model to the current model to mitigate catastrophic forgetting.
\citet{Wu2018MemoryRG} implemented deep generative replay \citep{dgr}, e.g., joint retraining and replay alignment, on a conditional GAN to avoid potential accumulate classification errors.
\citet{Cong2020GANMW} prevents forgetting by adding additional parameters to learn newly encountered tasks. Following FiLM and mAdaFM, the authors tailored these modulators for fully connected and convolutional networks to better perceive new information.
Following this strategy, CAM-GAN \citep{CAM-GAN} proposed a combination of group-wise and point-wise convolutional filters to learn novel tasks while further improved CL performance by leveraging task-similarity estimation with Fisher information matrix.
Hyper-LifelongGAN \citep{Zhai2021HyperLifelongGANSL} decomposed convolutional ﬁlters into dynamic task-specific filters generated by a filter generator and task-agnostic fixed weight components. Knowledge distillation techniques were adopted to further reduce forgetting issues.
LFS-GAN \citep{LFS-GAN} introduced newly proposed modulators termed LeFT, a rank-constrained weight factorization method while additional mode-seeking losses are adopted to prevent mode collapsing and enhance generation diversity.

\paragraph{Continual Learning of Diffusion Models}
Diffusion models \citep{sohldickstein2015deep, ho2020denoising, song2021scorebased}, a model that have been proved to be capable of high-quality image generations recent years, have also been experimented in CL. 
\citet{pmlr-v202-gao23e} trained a classifier and diffusion model bi-directional way, where the classifier is used to guide the conditional diffusion sampling. 
\citet{doan2024classprototype} added trainable class prototypes to represent previous classes and utilize gradient projection in diffusion process to alleviate forgetting.
In addition to classifier-guided methods, \citet{zając2023exploring} tested several common forgetting-prevent methods on MNIST \cite{MNIST} and Fashion-MNIST \cite{xiao2017fashionmnist} including experience replay, generative replay and L2 regularization, scratched the surface of continual diffusion model learning. 
\citet{masip2024continual} introduced generative distillation process, aligning predicted noises with previous task models at each step of the reverse sampling trajectory.
\citet{smith2024continual} proposed C-LoRA that trained distinct self-regulated LoRA \cite{hu2022lora} blocks in cross attention layers respectively for different tasks. We extend the C-LoRA method in our benchmarks to more general CLoG settings.

\newpage

\section{Comprehensive Results}

\subsection{Ablation study}
\label{app:ablation_study}

In this section, we conduct a series of ablation studies on the configurations that we fixed in our benchmark experiments. 

\paragraph{Different DDIM steps.} 
Since a larger DDIM step, though may improve the generation quality, will result in significant inference overhead and is irrelevant to CL capability, we fix it to a small value. In our experiments, we set DDIM steps as 50 for all the DDIM-based baselines. We evaluate the CLoG baselines with a larger number of DDIM steps on CIFAR-10, and the results are in~\cref{tab:ddim_step}. It shows that the DDIM step as 50 can already faithfully reflect the performance of CLoG baselines without $2\times$ or $4\times$ computations.

\begin{table}[H]
\centering
\caption{Performance of Different DDIM Steps on CIFAR-10}
\resizebox{0.65\textwidth}{!}{%
\begin{tabular}{cccccc}
    \toprule
    \multicolumn{2}{c}{\textbf{DDIM Step}} & 50 & 100 & 200 \\
    \midrule
    \midrule
     \multirow{3}{*}{NCL} & AFQ & 115.60$^{\pm20.51}$ & 112.41$^{\pm16.62}$ & 105.34$^{\pm13.65}$  \\
      & AIQ & 108.19$^{\pm15.02}$ & 96.82$^{\pm8.82}$ & 95.04$^{\pm6.47}$  \\
      & FR & 107.04$^{\pm27.11}$ & 104.75$^{\pm21.75}$ & 95.94$^{\pm18.22}$  \\
    \midrule
     \multirow{3}{*}{ER} & AFQ & 132.07$^{\pm8.92}$ & 132.94$^{\pm2.91}$ & 131.77$^{\pm2.35}$  \\
      & AIQ & 138.22$^{\pm6.12}$ & 131.59$^{\pm3.84}$ & 136.80$^{\pm3.82}$ \\
      & FR & 93.81$^{\pm13.71}$ & 95.53$^{\pm5.78}$ & 90.00$^{\pm5.49}$  \\
    \midrule
     \multirow{3}{*}{EWC} & AFQ & 127.09$^{\pm19.23}$ & 126.23$^{\pm10.22}$ & 129.14$^{\pm7.79}$  \\
      & AIQ & 113.06$^{\pm8.89}$ & 104.48$^{\pm4.22}$ & 109.01$^{\pm2.16}$  \\
      & FR & 119.74$^{\pm25.80}$ & 120.61$^{\pm13.86}$ & 118.46$^{\pm3.53}$  \\
    \midrule
     \multirow{3}{*}{Ensemble} & AFQ & 36.52$^{\pm0.55}$ & 35.91$^{\pm0.48}$ & 37.73$^{\pm0.95}$  \\
      & AIQ & 36.57$^{\pm1.57}$ & 34.97$^{\pm1.70}$ & 40.70$^{\pm3.52}$  \\
      & FR & 0 & 0 & 0  \\
    \midrule
     \multirow{3}{*}{C-LoRA} & AFQ & 60.11$^{\pm6.15}$ & 61.21$^{\pm5.89}$ & 63.94$^{\pm5.30}$ \\
      & AIQ & 173.43$^{\pm45.28}$ & 56.67$^{\pm6.21}$ & 58.54$^{\pm5.33}$  \\
      & FR & 0 & 0 & 0 \\
    \bottomrule
\end{tabular}\label{tab:ddim_step}
}
\end{table}

\paragraph{Different memory sizes.} In our benchmarks, we follow the existing CL works to set replay buffer sizes as 200 for small-scale CL datasets, and 5000 for large-scale ImageNet-1k.~\cref{tab:memory} shows the results of varying replay buffer sizes on CIFAR-10. It suggests that the performance of ER method is not sensitive to the memory size ranging from $20$ to $400$.  

\begin{table}[H]
\centering
\caption{Performance of Different Memory Size of Experience Replay (ER)}
\resizebox{0.55\textwidth}{!}{%
\begin{tabular}{ccccccc}
    \toprule
    \multicolumn{2}{c}{\textbf{Memory Size}} & 20 & 50 & 100 & 200 & 400  \\
    \midrule
    \midrule
     \multirow{3}{*}{ER} & AFQ & 133.65 & 136.11 & 138.15 & 135.47 & 133.83 \\
     & AIQ & 109.35 & 133.10 & 110.96 & 113.66 & 137.00 \\
     & FR & 96.58 & 100.11 & 102.45 & 97.98 & 95.99 \\
    \bottomrule
\end{tabular}\label{tab:memory}
}
\end{table}

\paragraph{Different class separation.} As a dataset can be split into different numbers of tasks, here we experiment on CIFAR-10 with 2, 5, 10 tasks. The results are shown in~\cref{tab:class_seperate}.

\begin{table}[H]
\centering
\caption{Performance of Different Class Seperation on CIFAR-10}
\resizebox{0.50\textwidth}{!}{%
\begin{tabular}{ccccc}
    \toprule
    \multicolumn{2}{c}{\textbf{Task number}} & 10 & 5 & 2    \\
    \midrule
    \midrule
     \multirow{3}{*}{NCL} & AFQ & 153.32 & 115.60 & 83.50  \\
      & AIQ & 159.97 & 118.19 & 61.09  \\
      & FR & 125.27 & 107.04 & 96.84  \\
    \midrule
     \multirow{3}{*}{ER} & AFQ & 227.92 & 132.07 & 119.95  \\
      & AIQ & 251.29 & 138.22 & 86.94  \\
      & FR & 189.41 & 93.81 & 126.09 \\
    \midrule
     \multirow{3}{*}{EWC} & AFQ & 174.98 & 127.09 & 96.50  \\
      & AIQ & 167.55 & 113.06 & 67.73  \\
      & FR & 149.96 & 119.74 & 123.77  \\
    \midrule
     \multirow{3}{*}{Ensemble} & AFQ & 54.08 & 36.52 &  38.12  \\
      & AIQ & 57.87 & 36.57 &  38.39 \\
      & FR & 0 & 0 & 0 \\
    \midrule
     \multirow{3}{*}{C-LoRA} & AFQ & 93.47 & 60.11 & 44.68  \\
      & AIQ & 87.04 & 173.43 & 41.91 \\
      & FR & 0.01 & 0.31 & 0.21 \\
    \bottomrule
\end{tabular}\label{tab:class_seperate}
}
\end{table}

\paragraph{Different alignment scores.} We also evaluate the concept-conditioned CLoG on another alignment score computed by DINO~\citep{Oquab2023DINOv2LR}. The results are shown in~\cref{tab:DreamBoothResult}.

\begin{table}[H]
\centering

\caption{\textbf{AFQ results for concept-conditioned CLoG benchmark with different alignment scores.} The best result in each row with the same base method (DreamBooth, Custom Diffusion) is highlighted in {\textbf{\color{google_red}red}}, while the second best and third best are highlighted in {\textbf{\color{google_blue}blue}} and {\textbf{\color{google_yellow}yellow}}, respectively. The quality metric is the average of text and image alignment scores (\emph{the higher value is better}). The AFQ is also averaged over 5 orders.}

\resizebox{\textwidth}{!}{%
\begin{tabular}{ccccccccccc}
    \toprule
    Metric & Model & \textbf{NCL} & \textbf{Non-CL} & \textbf{KD} & \textbf{L2} & \textbf{EWC} & \textbf{SI} & \textbf{MAS} & \textbf{Ensemble} & \textbf{C-LoRA} \\
    \midrule
    \midrule
    \multirow{2}{*}{CLIP Avg} & DreamBooth  & 78.54$^{\pm0.53}$ & {\textbf{\color{google_blue}80.09$^{\pm0.1}$}} & 78.73$^{\pm0.16}$ & 79.00$^{\pm0.38}$ & {\textbf{\color{google_yellow}79.45$^{\pm0.41}$}} & 78.54$^{\pm0.39}$ & 78.00$^{\pm0.46}$  & {\textbf{\color{google_blue}80.09$^{\pm0.25}$}} & {\textbf{\color{google_red}80.42$^{\pm0.25}$}}  \\
    & Custom Diffusion & 79.56$^{\pm0.17}$ & {\textbf{\color{google_blue}80.30$^{\pm0.21}$}} & 79.71$^{\pm0.1}$ & 79.92$^{\pm0.14}$ & {\textbf{\color{google_yellow}80.10$^{\pm0.05}$}} & 79.59 $^{\pm0.27}$ & 78.79$^{\pm0.18}$ & {\textbf{\color{google_red}80.39$^{\pm0.24}$}} & - \\
    \midrule
    \multirow{2}{*}{DINO Avg} & DreamBooth  & 70.81$^{\pm0.72}$ & {\textbf{\color{google_yellow}71.76$^{\pm0.2}$}} & 70.55$^{\pm0.71}$ & 70.90$^{\pm0.76}$ & 70.57$^{\pm0.58}$ & 69.69$^{\pm0.81}$ & 64.47$^{\pm2.3}$  & {\textbf{\color{google_blue}72.81$^{\pm0.68}$}} & {\textbf{\color{google_red}73.677$^{\pm0.38}$}} \\
    & Custom Diffusion & 67.43$^{\pm0.71}$ & {\textbf{\color{google_blue}69.96$^{\pm0.4}$}} & 67.64$^{\pm0.28}$ & 67.91$^{\pm0.51}$ & 69.21$^{\pm0.47}$ & {\textbf{\color{google_yellow}69.60$^{\pm1.2}$}} & 64.93$^{\pm1.6}$ & {\textbf{\color{google_red}70.95$^{\pm1.30}$}} & - \\
    \bottomrule
\end{tabular}\label{tab:DreamBoothResult}
}
\end{table}

\newpage

\subsection{AIQ and FR Results}
\label{app:aiq_fr}

In this section, we present the AIQ and FR results on all benchmarks. 

We can observe the average forgetting rate becomes negative in some cases on the CUB-Birds, Oxford-Flowers and Stanford-Cars datasets. This phenomenon suggests the existence of positive knowledge transfer among these datasets. Note that the FR of the ensemble and C-LoRA method is set to zero since we train a separate model for each task.

\begin{table}[H]
\centering
\caption{AIQ results for label-conditioned benchmarks.}
\resizebox{\textwidth}{!}{%
\begin{tabular}{cccccccc}
    \toprule
    & \textbf{MNIST} & \textbf{Fashion-MNIST} & \textbf{CIFAR-10} & \textbf{CUB-Birds} & \textbf{Oxford-Flowers} & \textbf{Stanford-Cars} & \textbf{ImageNet}  \\
    \midrule
    \midrule
    \multicolumn{7}{l}{\textbf{- GAN}} \\
     Non-CL  & 38.33$^{\pm 1.89}$ & 49.87$^{\pm 3.94}$ & 57.13$^{\pm 3.63}$ & 86.62$^{\pm 7.27}$ & 118.11$^{\pm 2.21}$ & 67.97$^{\pm 1.13}$ & NA\\
     NCL  & 45.50$^{\pm 4.41}$ & 73.49$^{\pm 5.27}$ & 80.23$^{\pm 5.46}$ & 125.13$^{\pm 10.88}$ & 127.18$^{\pm 13.54}$ & 97.77$^{\pm 10.44}$ & NA\\
     ER   & 37.23$^{\pm 6.24}$ & 61.62$^{\pm 10.59}$ & 173.08$^{\pm 3.13}$ & 180.04$^{\pm 6.14}$ & 151.53$^{\pm 12.58}$ & 159.86$^{\pm 4.94}$ & NA\\
     GR   & 54.19$^{\pm 12.55}$ & 36.13$^{\pm 12.30}$ & 71.66$^{\pm 0.67}$ & 180.82$^{\pm 2.02}$ & 158.24$^{\pm 2.65}$ & 190.07$^{\pm 19.21}$ & NA\\
     KD   & 39.31$^{\pm 0.34}$ & 69.12$^{\pm 3.48}$ & 80.98$^{\pm 4.71}$ & 135.21$^{\pm 8.06}$ & 131.76$^{\pm 13.04}$ & 102.79$^{\pm 9.81}$ & NA \\
     L2   & 44.01$^{\pm 9.44}$ & 81.90$^{\pm 10.70}$ & 82.65$^{\pm 4.54}$ & 182.36$^{\pm 11.50}$ & 159.24$^{\pm 9.57}$ & 202.29$^{\pm 37.37}$ & NA\\
     EWC  & 38.94$^{\pm 2.48}$ & 58.17$^{\pm 3.96}$ & 67.39$^{\pm 9.77}$ & 155.24$^{\pm 6.27}$ & 134.34$^{\pm 3.98}$ & 150.31$^{\pm 22.91}$ & NA\\
     SI  & 75.24$^{\pm 28.81}$ & 77.05$^{\pm 9.37}$ & 78.55$^{\pm 3.69}$ & 189.11$^{\pm 13.55}$ & 164.99$^{\pm 8.35}$  & 198.58$^{\pm 37.99}$ & NA\\
     MAS & 48.93$^{\pm 2.05}$ & 61.70$^{\pm 2.27}$ & 70.24$^{\pm 4.03}$ & 179.28$^{\pm 9.21}$ & 143.04 $^{\pm 9.29}$ & 169.43$^{\pm 12.43}$ & NA\\
     A-GEM   & 31.99$^{\pm 7.05}$ & 60.94$^{\pm 5.47}$ & 78.23$^{\pm 3.48}$ & 125.17$^{\pm 5.61}$ & 131.95$^{\pm 7.06}$ & 101.79$^{\pm 4.75}$ & NA\\
     Ensemble  & 10.85$^{\pm 3.26}$ & 27.30$^{\pm 1.04}$ & 44.35$^{\pm 2.04}$ & 177.25$^{\pm 3.59}$ & 148.64$^{\pm 6.28}$ & 232.97 $^{\pm 6.27}$ & NA\\
    \bottomrule
    \toprule
    \multicolumn{7}{l}{\textbf{- Diffusion Model}} \\
     Non-CL  & $4.47^{\pm1.30}$ & $9.13^{\pm0.32}$ & $31.08^{\pm2.32}$ & $65.24^{\pm1.60}$ &  $53.76^{\pm2.55}$  & $33.56^{\pm0.39}$ & 46.08 \\
     NCL  &$105.79^{\pm4.02}$ & $128.78^{\pm13.05}$ & $108.19^{\pm15.02}$ & $104.31^{\pm2.03}$ & $101.15^{\pm9.07}$ &  $54.47^{\pm4.27}$ & 92.08 \\
     ER  & $19.76^{\pm1.02}$ & $36.91^{\pm2.13}$ & $138.22^{\pm6.12}$  & $79.46^{\pm4.26}$ & $77.44^{\pm2.09}$  & $77.75^{\pm3.06}$ & 97.16 \\
     GR  & $61.22^{\pm1.27}$ & $27.28^{\pm4.84}$ & $60.58^{\pm1.08}$ &  $194.27^{\pm8.32}$ & $ 98.31^{\pm3.77}$ & $244.96^{\pm7.85}$ & NA \\
     KD  & $150.13^{\pm3.35}$ & $237.93^{\pm10.01}$ & $185.38^{\pm2.31}$ & $178.04^{\pm1.76}$ & $169.74^{\pm10.38}$ & $113.08^{\pm7.12}$ & 110.09 \\
     L2  & $158.51^{\pm14.52}$ & $175.01^{\pm13.47}$ & $164.06^{\pm6.92}$ & $175.35^{\pm10.14}$ & $188.30^{\pm21.85}$ & $267.73^{\pm25.74}$ & 112.21 \\
     EWC & $137.11^{\pm17.60}$ & $131.18^{\pm5.44}$ & $113.06^{\pm8.89}$ & $104.53^{\pm8.63}$ & $101.60^{\pm4.54}$ & $59.11^{\pm3.80}$ & 98.19 \\
     SI  & $149.27^{\pm12.98}$ & $130.66^{\pm12.96}$ & $114.16^{\pm13.86}$ & $115.62^{\pm6.39}$ & $105.92^{\pm3.90}$ & $64.62^{\pm1.77}$ & 102.01 \\
     MAS & $112.17^{\pm10.32}$ & $135.52^{\pm13.56}$ & $109.80^{\pm10.04}$ & $189.30^{\pm13.95}$ & $191.96^{\pm32.86}$ & $227.41^{\pm16.70}$ & 113.23 \\
     A-GEM & $106.25^{\pm6.83}$ & $135.17^{\pm10.41}$ & $115.26^{\pm10.26}$ & $108.94^{\pm2.31}$ & $100.64^{\pm5.55}$ & $56.85^{\pm3.03}$ & 62.99 \\
     Ensemble  & $4.13^{\pm0.21}$ & $10.29^{\pm0.22}$ & $36.57^{\pm1.57}$ & $131.94^{\pm3.45}$ & $72.84^{\pm14.34}$ & $201.71^{\pm2.99}$ & 56.86 \\
     C-LoRA  & $140.51^{\pm4.74}$ & $229.63^{\pm5.09}$ & $173.43^{\pm45.28}$ & $186.01^{\pm23.20}$ & $288.38^{\pm7.12}$ & $269.84^{\pm29.35}$ & 73.16 \\
    \bottomrule
\end{tabular}\label{tab:aiq}
}
\end{table}

\begin{table}[H]
\centering
\caption{{AIQ results for the concept-conditioned CLoG benchmarks.} }

\resizebox{\textwidth}{!}{%
\begin{tabular}{ccccccccccc}
    \toprule
     & Model & \textbf{NCL} & \textbf{Non-CL} & \textbf{KD} & \textbf{L2} & \textbf{EWC} & \textbf{SI} & \textbf{MAS} & \textbf{Ensemble} & \textbf{C-LoRA} \\
    \midrule
    \midrule
    \multirow{2}{*}{CLIP Avg} & DreamBooth  & 78.40 & 79.07 & 78.57 & 78.75 & 79.43 & 78.73  & 77.87 & 0 & 0  \\
    & Custom Diffusion & 79.86 & 79.40 & 79.64 & 79.86 & 79.77 & - & 79.39 & 0 & - \\
    \midrule
    \multirow{2}{*}{DINO Avg} & DreamBooth  & 73.66 & 73.71 & 72.06 & 73.15 & 73.02 & 72.32 & 69.24  & 0 & 0 \\
    & Custom Diffusion & 71.89 & 72.87 & 71.55 & 71.94 & 71.88 & - & 70.46 & 0 & - \\
    \bottomrule
\end{tabular}\label{tab:CustomAIQ}
}
\end{table}

\begin{table}[H]
\centering
\caption{FR results for label-conditioned CLoG benchmarks.}

\resizebox{\textwidth}{!}{%
\begin{tabular}{cccccccc}
    \toprule
    & \textbf{MNIST} & \textbf{Fashion-MNIST} & \textbf{CIFAR-10} & \textbf{CUB-Birds} & \textbf{Oxford-Flowers} & \textbf{Stanford-Cars} & \textbf{ImageNet}  \\
    \midrule
    \midrule
    \multicolumn{7}{l}{\textbf{- GAN}} \\
     Non-CL & 5.13$^{\pm4.30}$ & 10.42$^{\pm 7.03}$ & 5.91$^{\pm 1.29}$ & -44.36$^{\pm 8.99}$ & -18.58$^{\pm 9.94}$ & -38.33$^{\pm 4.21}$ & NA\\
     NCL  & 68.91$^{\pm 7.78}$ & 94.67$^{\pm 17.44}$ & 74.34$^{\pm 13.26}$ & -8.44$^{\pm 20.19}$ & 19.52$^{\pm 10.22}$ & -32.05$^{\pm 7.63}$ & NA\\
     ER   & 94.27$^{\pm 31.22}$ & 128.76$^{\pm 3.26}$ & 209.16$^{\pm 16.74}$ & 15.72$^{\pm 26.08}$ & 13.42$^{\pm 11.76}$ & 56.72$^{\pm 3.45}$ & NA \\
     GR   & 120.05$^{\pm 45.12}$ & 110.12$^{\pm 24.27}$ & 112.96$^{\pm 10.23}$ & 101.54$^{\pm 15.07}$ & 56.11$^{\pm 13.07}$ & 85.51$^{\pm 25.89}$ & NA \\
     KD  & 60.64$^{\pm 5.64}$ & 87.50$^{\pm 4.45}$ & 74.82$^{\pm 16.83}$ & -22.49$^{\pm 13.06}$ & -3.91$^{\pm 15.64}$ & -24.21$^{\pm 6.45}$ & NA\\
     L2   & 49.38$^{\pm 11.48}$ & 66.41$^{\pm 10.29}$ & 40.87$^{\pm 12.91}$ & 12.41$^{\pm 7.87}$ & 0.57$^{\pm 2.88}$ & 5.21$^{\pm 5.22}$ & NA\\
     EWC  & 58.96$^{\pm 6.14}$ & 80.06$^{\pm 13.34}$ & 50.91$^{\pm 32.37}$ & 7.16$^{\pm 12.56}$ & 2.30$^{\pm 4.31}$ & -47.16$^{\pm 11.31}$ & NA \\
     SI  & 4.93$^{\pm 3.61}$ & 0.08$^{\pm 0.64}$ & 2.46$^{\pm 3.16}$ & 14.99$^{\pm 12.91}$ & 1.68$^{\pm 5.34}$ & 17.05$^{\pm 12.49}$ & NA\\
     MAS & 52.57$^{\pm 15.81}$ & 72.31$^{\pm 7.88}$ & 32.45$^{\pm 8.39}$ & 9.18$^{\pm 10.15}$ & 4.18$^{\pm 7.28}$ & -14.71$^{\pm 7.45}$ & NA\\
     A-GEM  & 44.45$^{\pm 20.32}$ & 84.84$^{\pm 23.35}$ & 66.91$^{\pm 14.49}$ & 0.48$^{\pm 11.85}$ & 0.45$^{\pm 12.30}$ & -30.09$^{\pm 7.05}$ & NA \\
     Ensemble  & 0 & 0 & 0 & 0 & 0 & 0 & 0\\
    \bottomrule
    \toprule
    \multicolumn{7}{l}{\textbf{- Diffusion Model}} \\
     Non-CL  & $1.82^{\pm3.74}$ & $-0.72^{\pm1.34}$ & $-2.46^{\pm0.99}$ & $-30.08^{\pm4.29}$ & $-12.73^{\pm6.43}$ & $-20.21^{\pm2.31}$ & 0.19 \\
     NCL  & $139.69^{\pm11.51}$ & $163.36^{\pm23.95}$ & $107.04^{\pm27.11}$ & $6.99^{\pm8.89}$ & $36.76^{\pm13.03}$ & $1.76^{\pm12.15}$ & 62.89  \\
     ER  & $14.70^{\pm30.99}$ & $54.18^{\pm3.92}$ & $93.81^{\pm13.71}$ & $-26.90^{\pm4.34}$ & $8.34^{\pm9.06}$ & $53.60^{\pm7.25}$ & 57.61 \\
     GR  & $100.28^{\pm6.87}$ & $32.37^{\pm5.54}$ & $54.74^{\pm3.98}$ & $75.37^{\pm24.00}$ & $51.59^{\pm12.10}$ & $178.51^{\pm7.15}$ & NA \\
     KD  & $70.52^{\pm11.67}$ & $58.28^{\pm19.35}$ & $17.22^{\pm24.80}$ & $14.89^{\pm8.13}$ & $35.30^{\pm32.65}$ & $-17.80^{\pm18.44}$ & 25.68 \\
     L2  & $202.09^{\pm35.42}$ & $193.35^{\pm15.23}$ & $132.67^{\pm24.35}$ & $25.69^{\pm6.33}$ & $26.03^{\pm34.77}$ & $1.25^{\pm20.09}$ & 21.49 \\
     EWC & $192.15^{\pm29.17}$ & $161.97^{\pm25.18}$ & $119.74^{\pm25.80}$ & $24.14^{\pm15.32}$ & $45.24^{\pm5.50}$ & $0.05^{\pm3.74}$ & 69.17 \\
     SI  & $212.38^{\pm33.81}$ & $179.85^{\pm28.77}$ & $122.83^{\pm35.85}$ & $20.66^{\pm17.76}$ & $35.40^{\pm11.75}$ & $7.57^{\pm8.37}$ & 62.61 \\
     MAS & $-0.16^{\pm0.28}$ & $0.75^{\pm0.83}$ & $0.96^{\pm0.94}$ & $0.72^{\pm0.99}$ & $-0.46^{\pm0.34}$ & $0.35^{\pm0.26}$ & 0.27 \\
     A-GEM & $138.81^{\pm11.12}$ & $163.41^{\pm5.92}$ & $123.51^{\pm34.69}$ & $11.80^{\pm5.91}$ & $57.86^{\pm14.35}$ & $-0.61^{\pm1.22}$ & 62.99 \\
     Ensemble  & 0 & 0 & 0 & 0 & 0 & 0 & 0 \\
     C-LoRA  & 0 & 0 & 0 & 0 & 0 & 0 & 0 \\
    \bottomrule
\end{tabular}\label{tab:fr} 
}
\end{table}

\begin{table}[htbp]
\centering
\caption{{FR results for concept-conditioned CLoG benchmark} }

\resizebox{\textwidth}{!}{%
\begin{tabular}{ccccccccccc}
    \toprule
     & Model & \textbf{NCL} & \textbf{Non-CL} & \textbf{KD} & \textbf{L2} & \textbf{EWC} & \textbf{SI} & \textbf{MAS} & \textbf{Ensemble} & \textbf{C-LoRA} \\
    \midrule
    \midrule
    \multirow{2}{*}{CLIP Avg} & DreamBooth  & 0.6827 & -0.7118 & 1.4377 & 0.4194 & 0.9916 & 1.5471 &  0.2172 & 0 & 0  \\
    & Custom Diffusion & 0.3503 & 0.0846 & 0.256 & 0.0588 & -0.1222 & 0.276 & 0.0049  & 0 & 0  \\
    \multirow{2}{*}{DINO Avg} & DreamBooth  & 2.5382 & 0.6054 & 3.3108 & 2.1836 & 3.3122 & 3.2506  &  0.3461& 0 & - \\
    & Custom Diffusion & 2.5568 & 1.5127 & 2.3811 & 1.5137 & 0.5065 & 1.1772 & -0.0017 & 0 & - \\
    \bottomrule
\end{tabular}\label{tab:CustomFR}
}

\end{table}

\newpage

\subsection{Visualization Results} \label{app:visualize}

\begin{figure}[H]
\centering
\subfigure[NCL (GAN)] {    
\includegraphics[width=0.4\columnwidth]{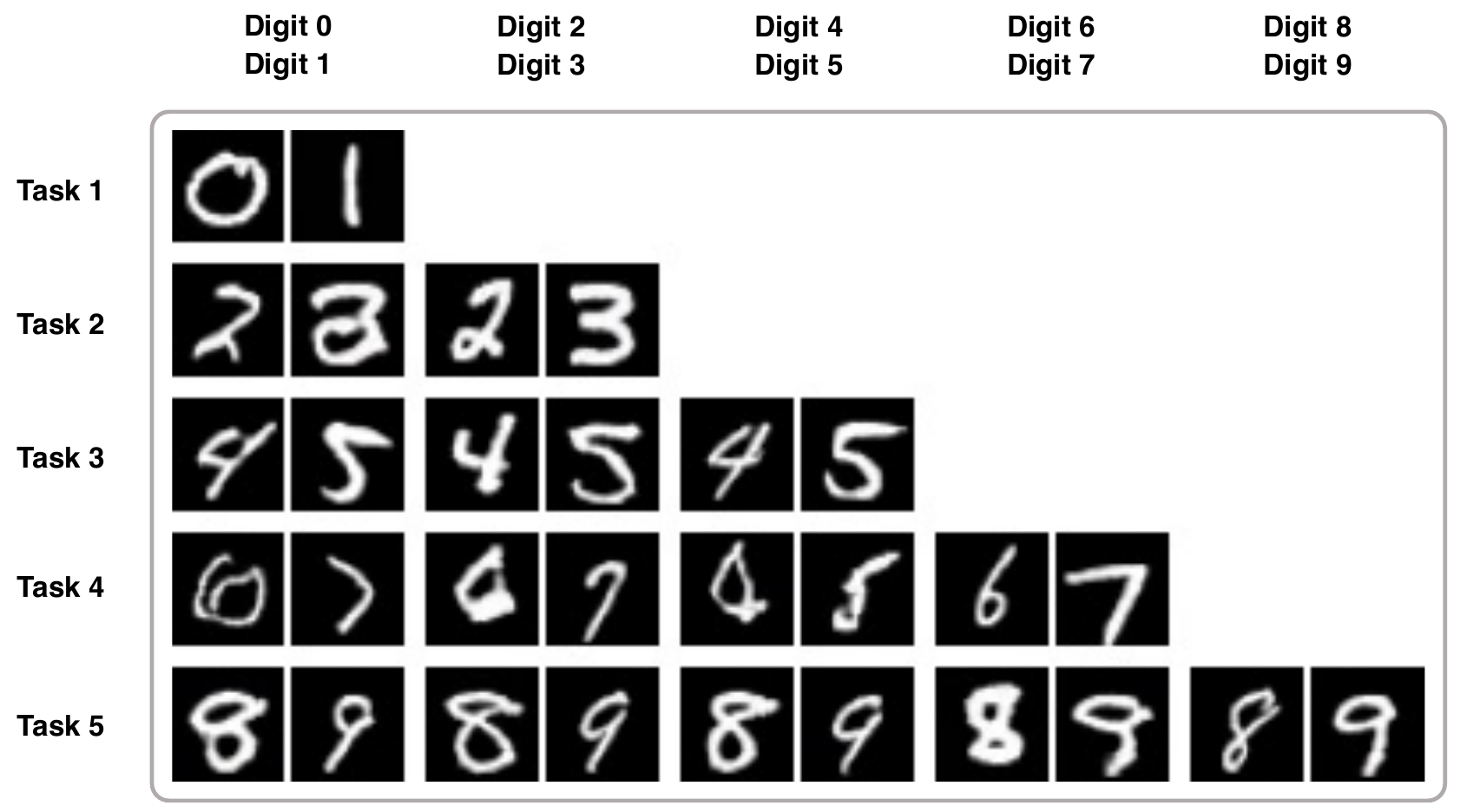}  
}    
\subfigure[Non-CL (GAN)] {    
\includegraphics[width=0.4\columnwidth]{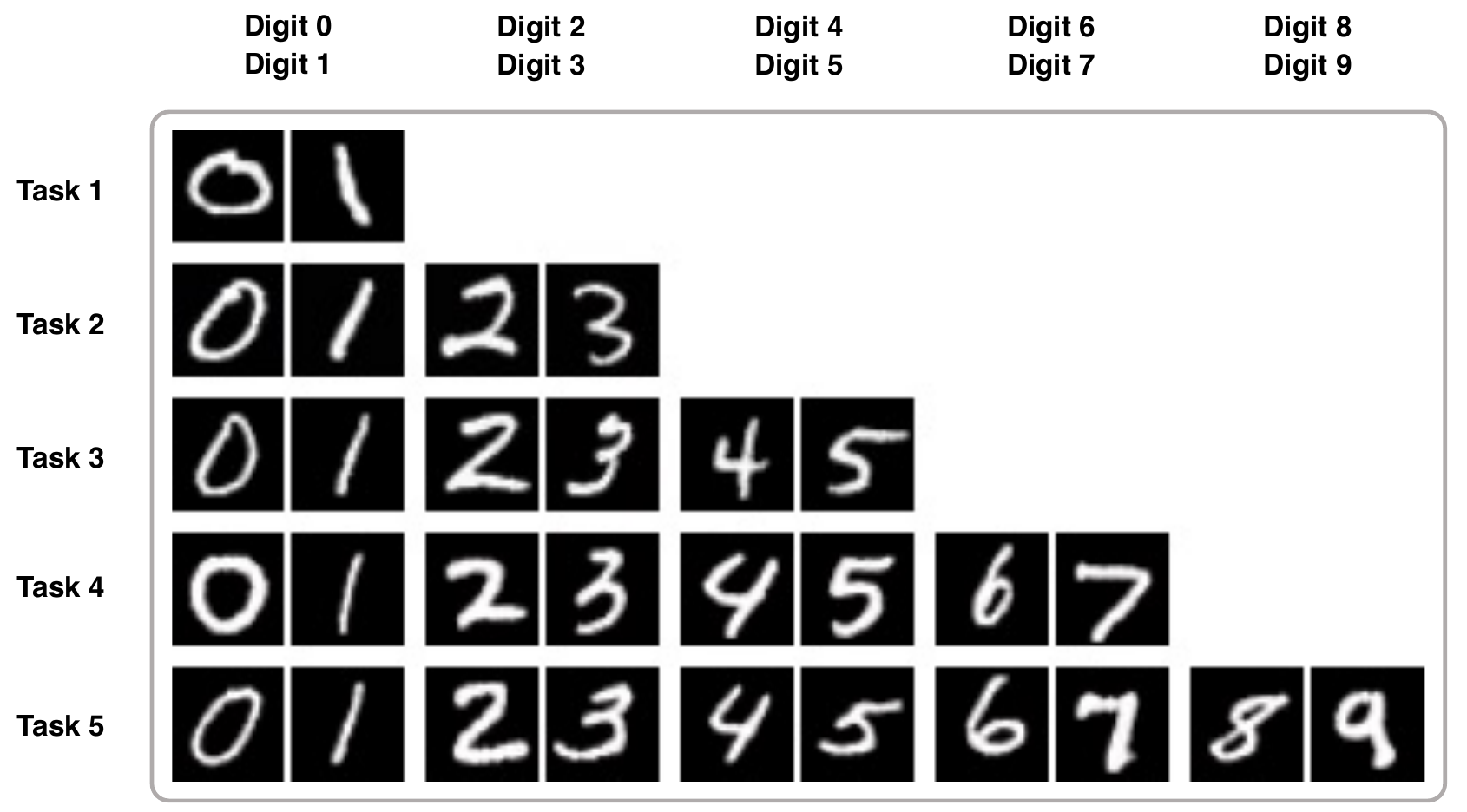}  
}    
\subfigure[ER (GAN)] {    
\includegraphics[width=0.4\columnwidth]{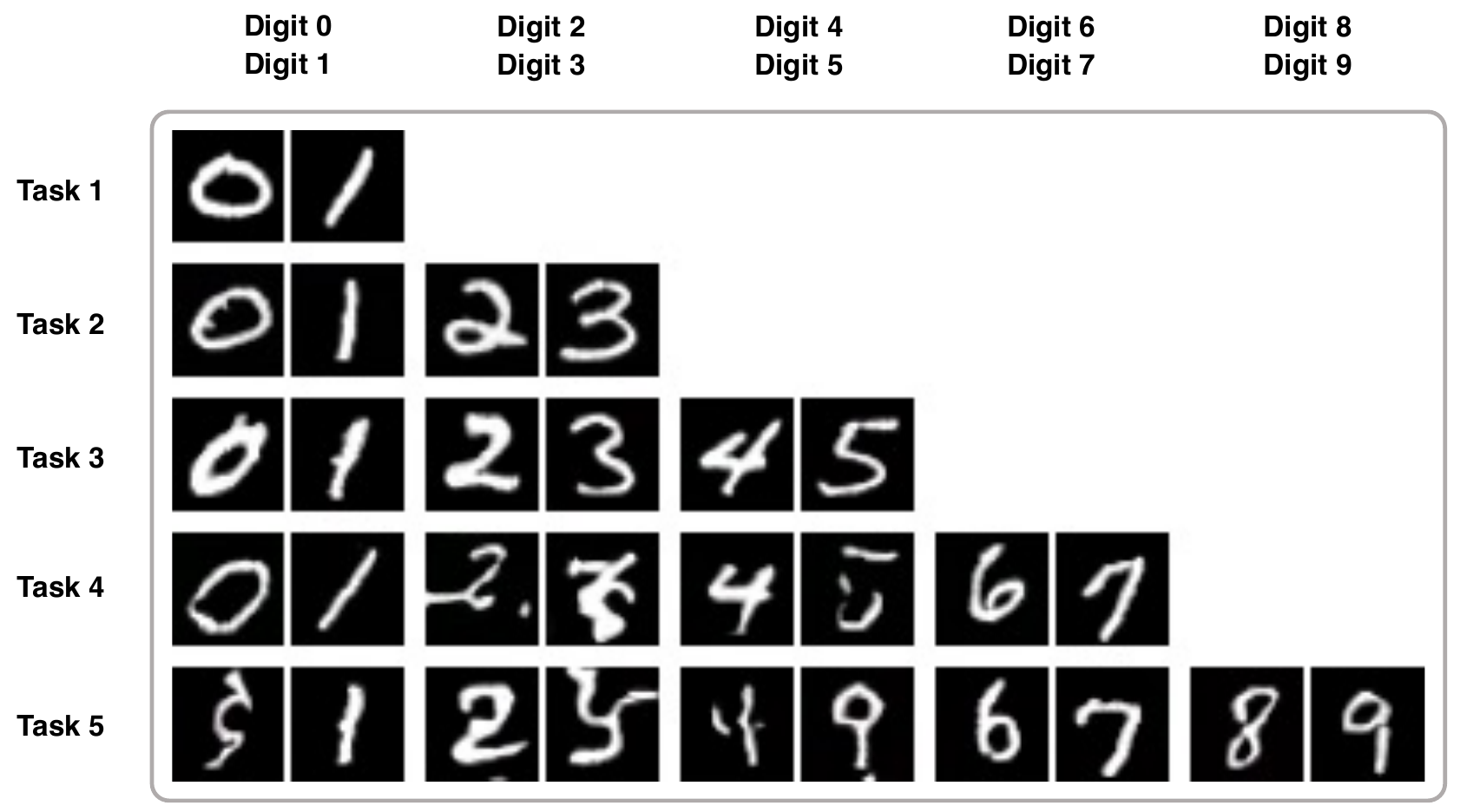}  
}    
\subfigure[EWC (GAN)] {    
\includegraphics[width=0.4\columnwidth]{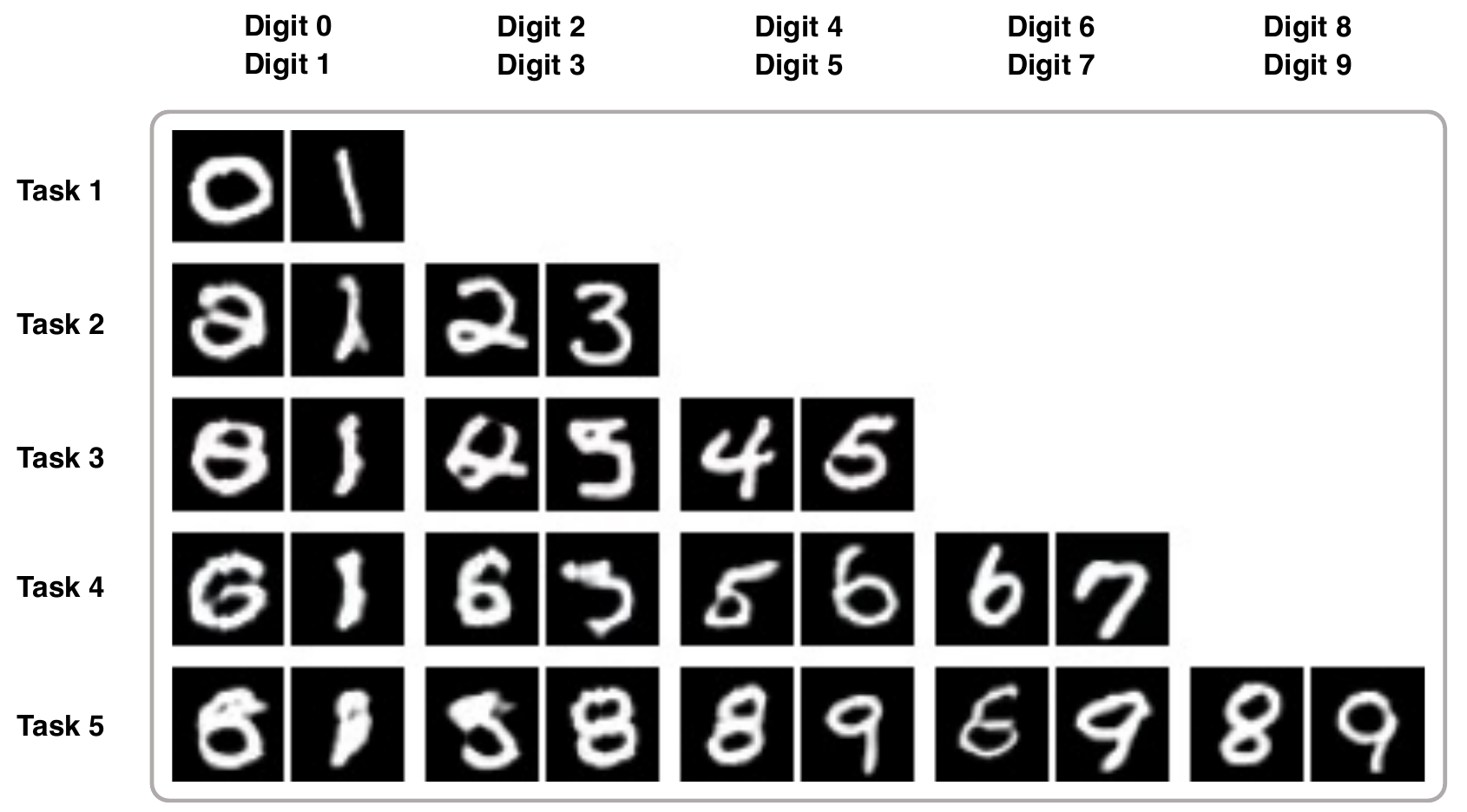}  
} 
\subfigure[Ensemble (GAN)] {    
\includegraphics[width=0.4\columnwidth]{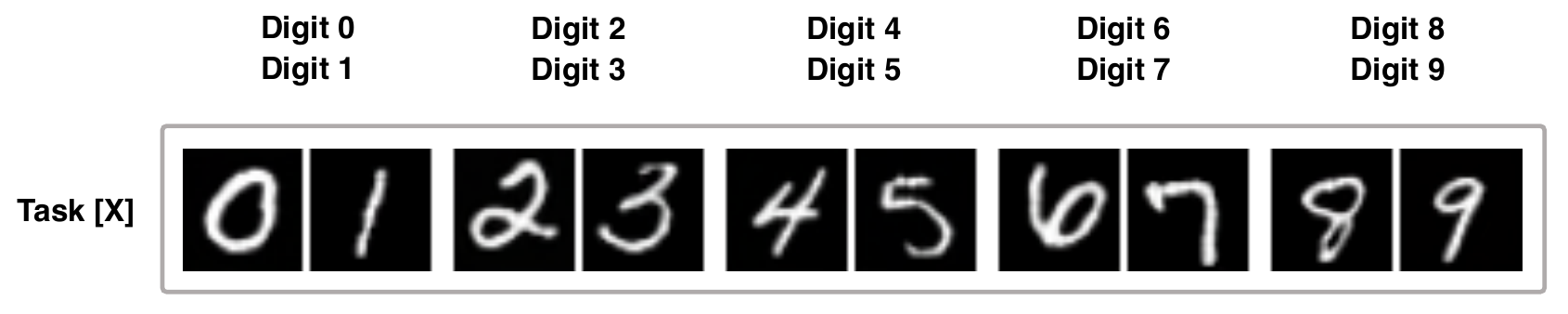}  
} 
\\
\subfigure[NCL (DDIM)] {    
\includegraphics[width=0.4\columnwidth]{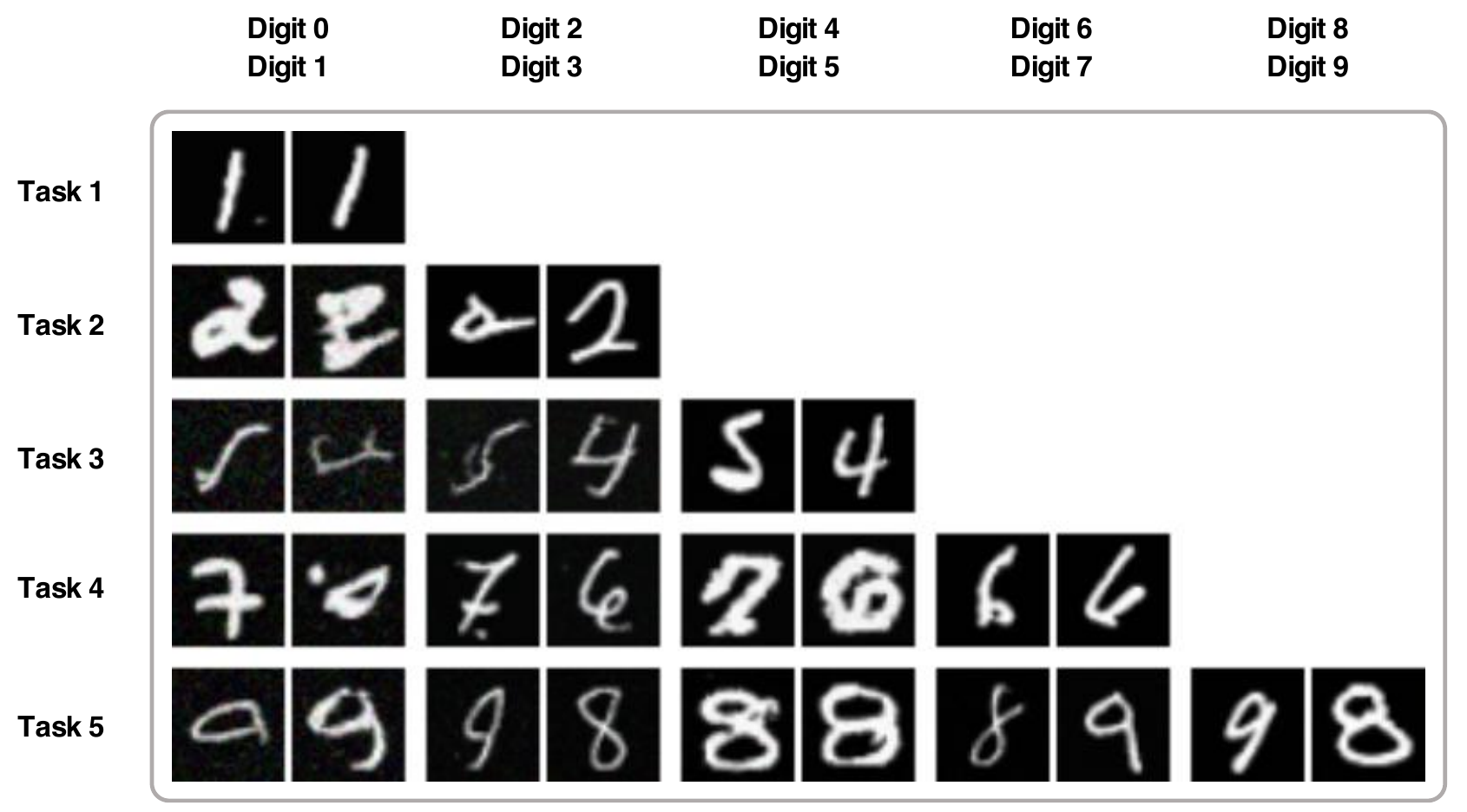}  
}    
\subfigure[Non-CL (DDIM)] {    
\includegraphics[width=0.4\columnwidth]{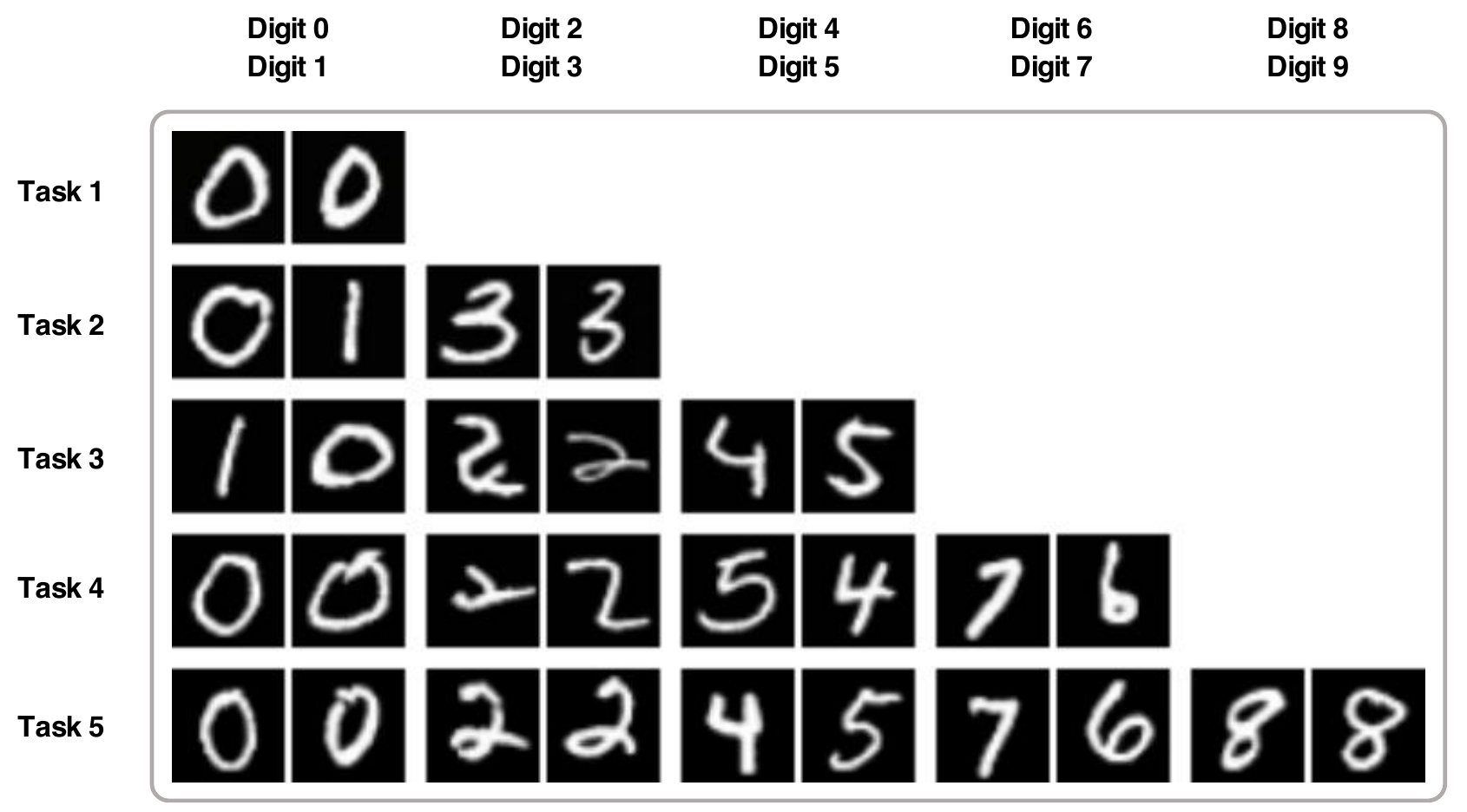}  
}    
\subfigure[ER (DDIM)] {    
\includegraphics[width=0.4\columnwidth]{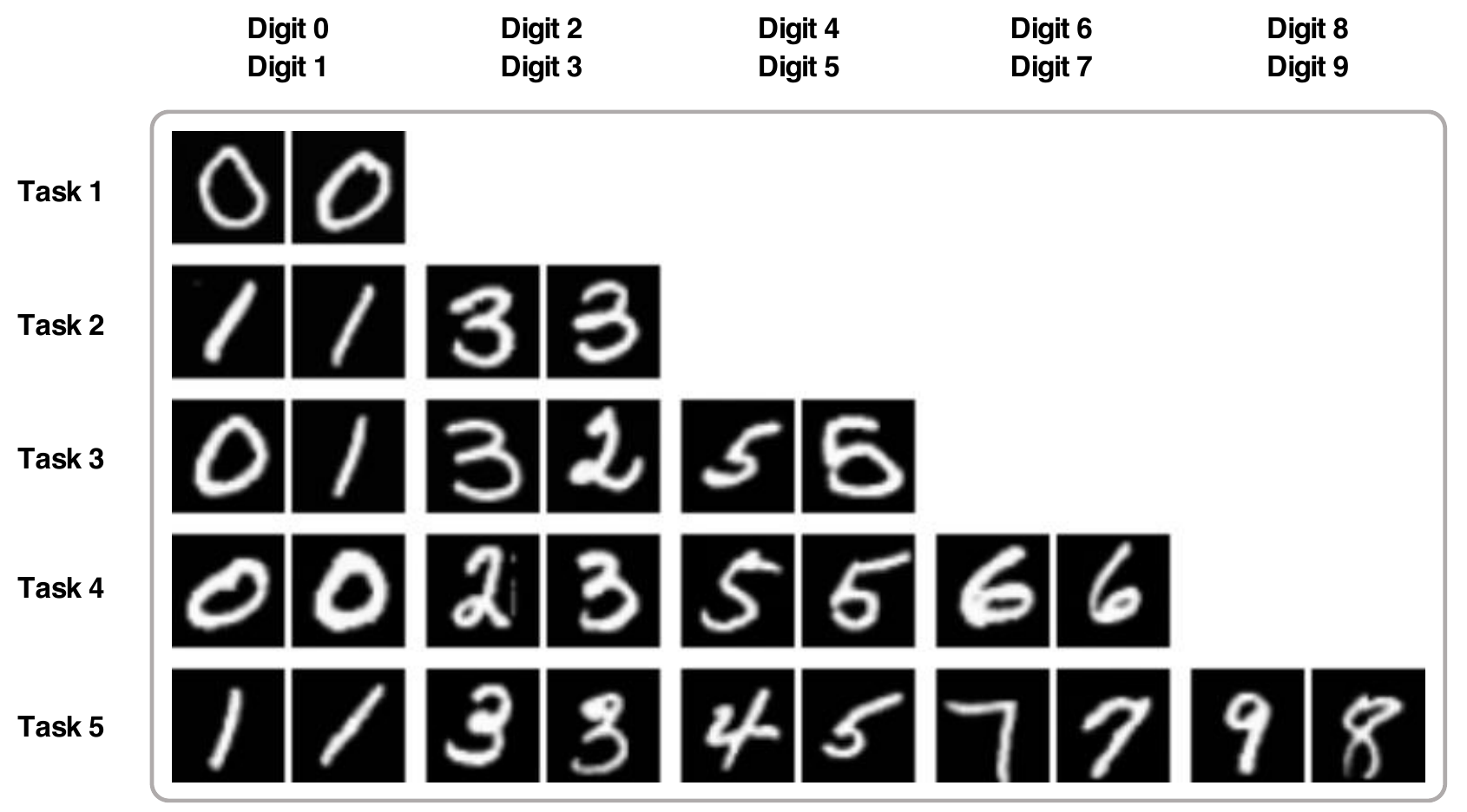}  
}    
\subfigure[EWC (DDIM)] {    
\includegraphics[width=0.4\columnwidth]{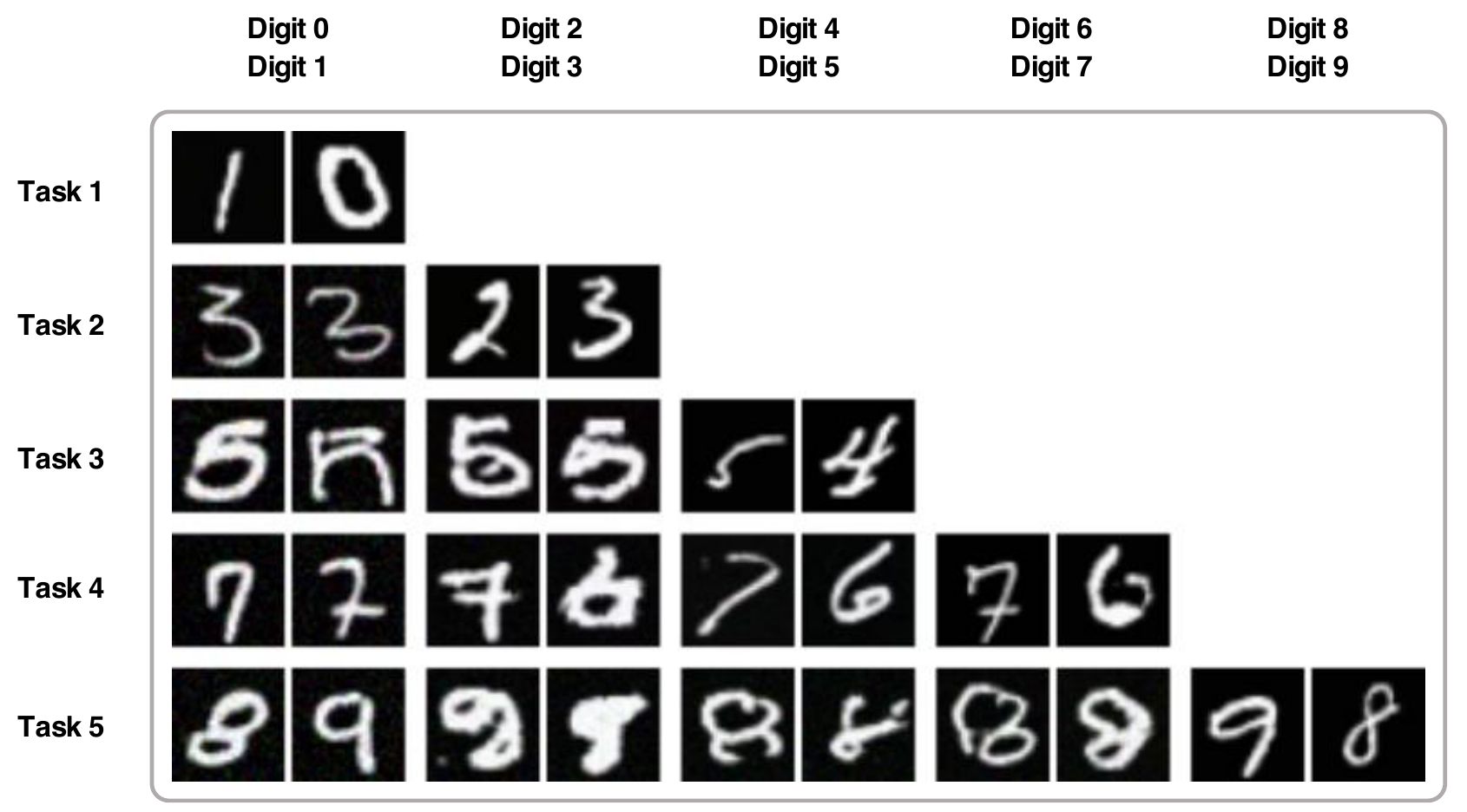}  
} 
\subfigure[Ensemble (DDIM)] {    
\includegraphics[width=0.4\columnwidth]{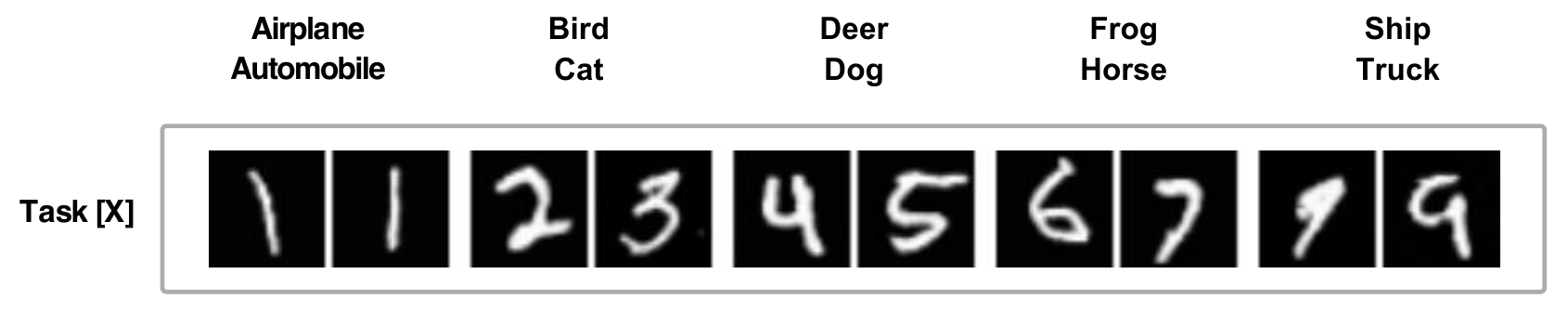}  
} 
\caption{Visualization results of label-conditioned CLoG on the MNIST~\citep{MNIST} dataset.} 
\label{fig:vis_mnist}
\end{figure}

\begin{figure}[H]
\centering
\subfigure[NCL (GAN)] {    
\includegraphics[width=0.4\columnwidth]{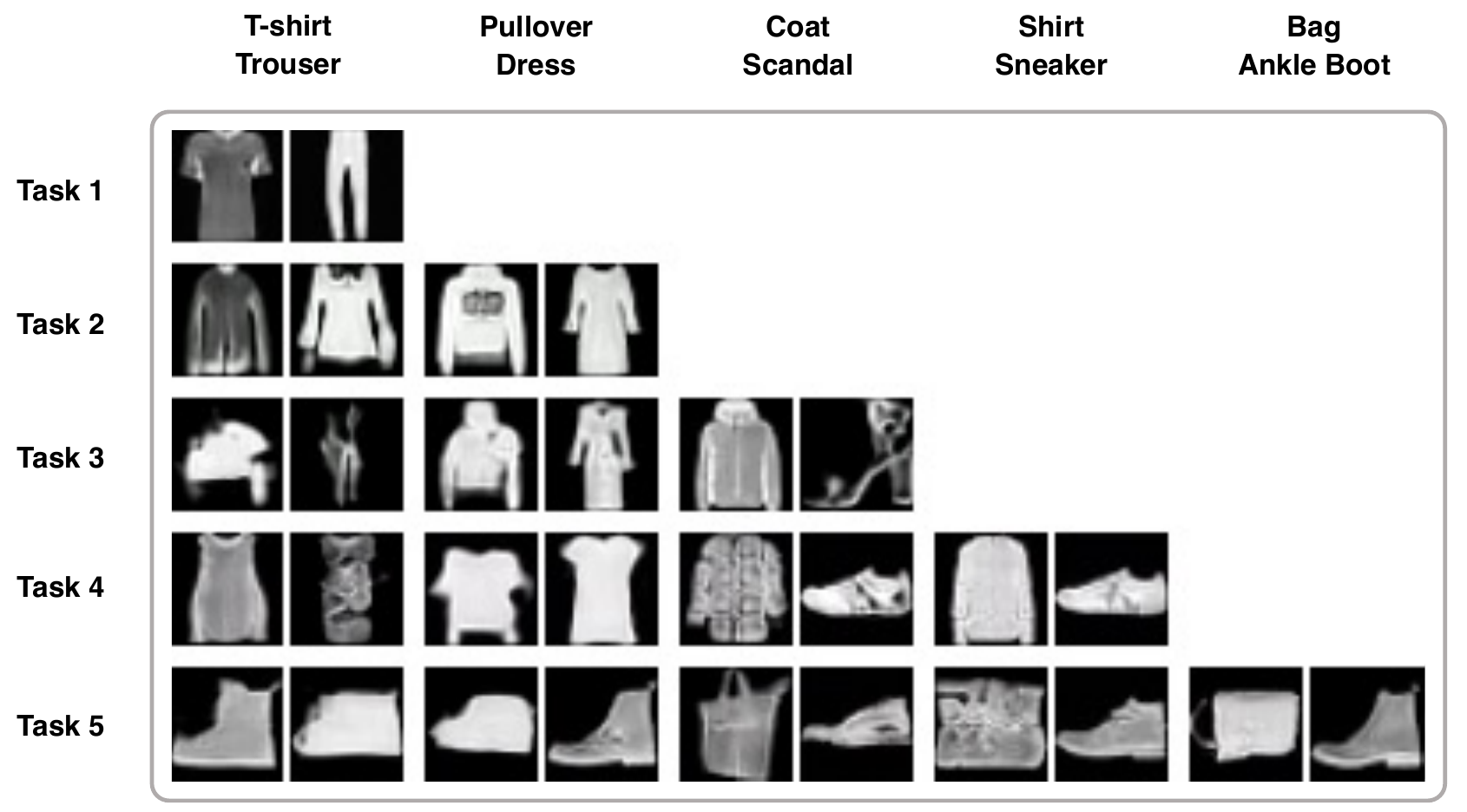}  
}    
\subfigure[Non-CL (GAN)] {    
\includegraphics[width=0.4\columnwidth]{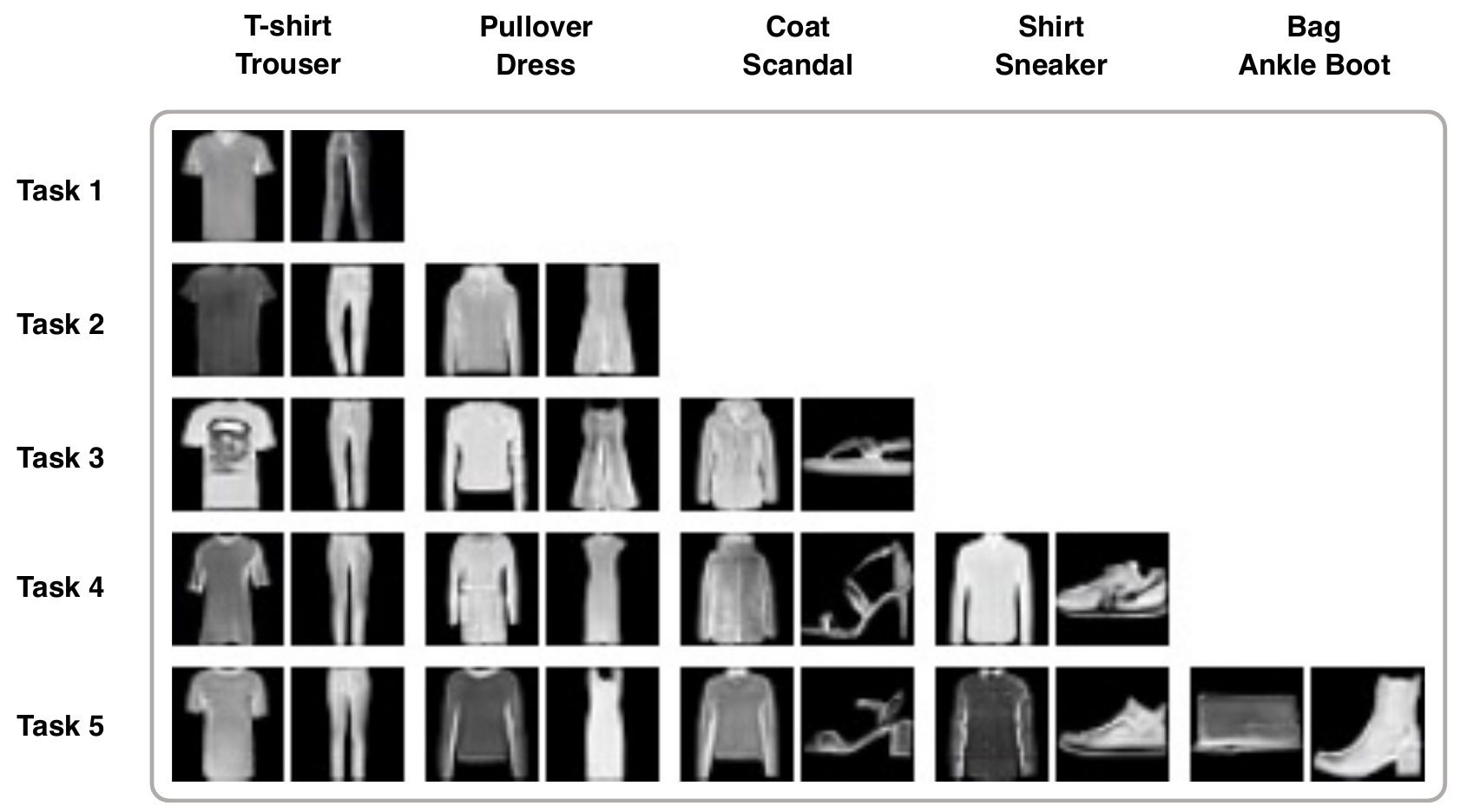}  
}    
\subfigure[ER (GAN)] {    
\includegraphics[width=0.4\columnwidth]{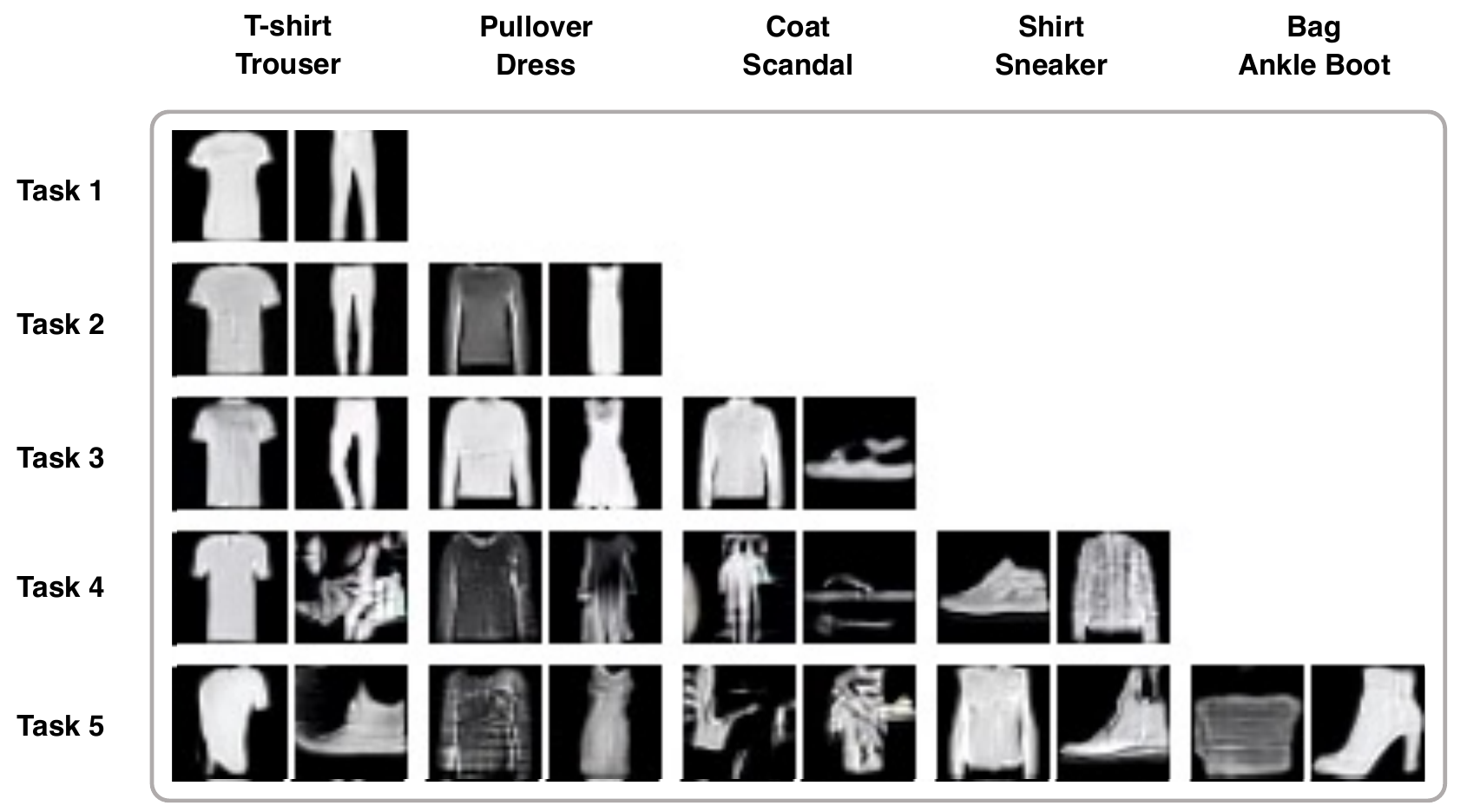}  
}    
\subfigure[EWC (GAN)] {    
\includegraphics[width=0.4\columnwidth]{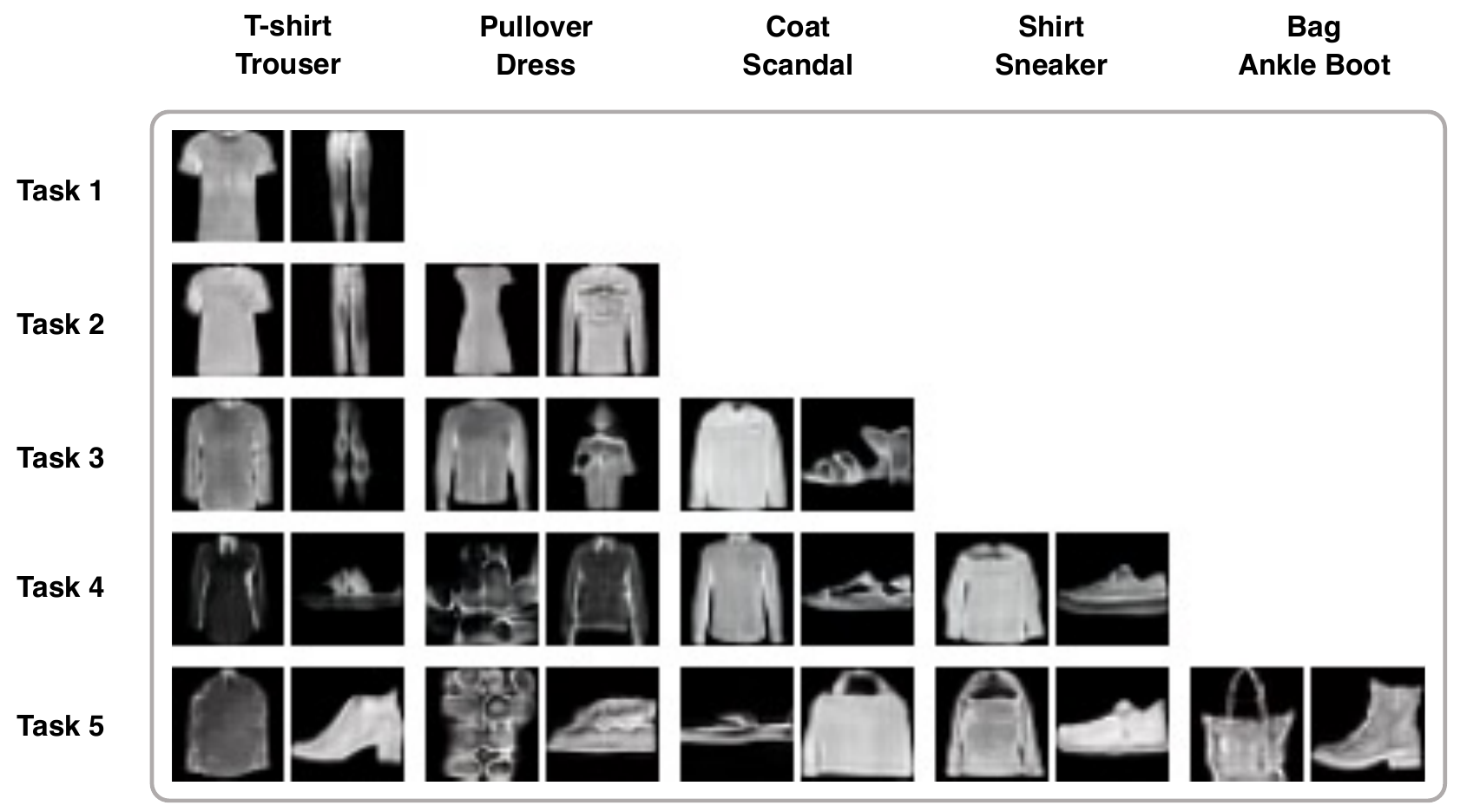}  
} 
\subfigure[Ensemble (GAN)] {    
\includegraphics[width=0.4\columnwidth]{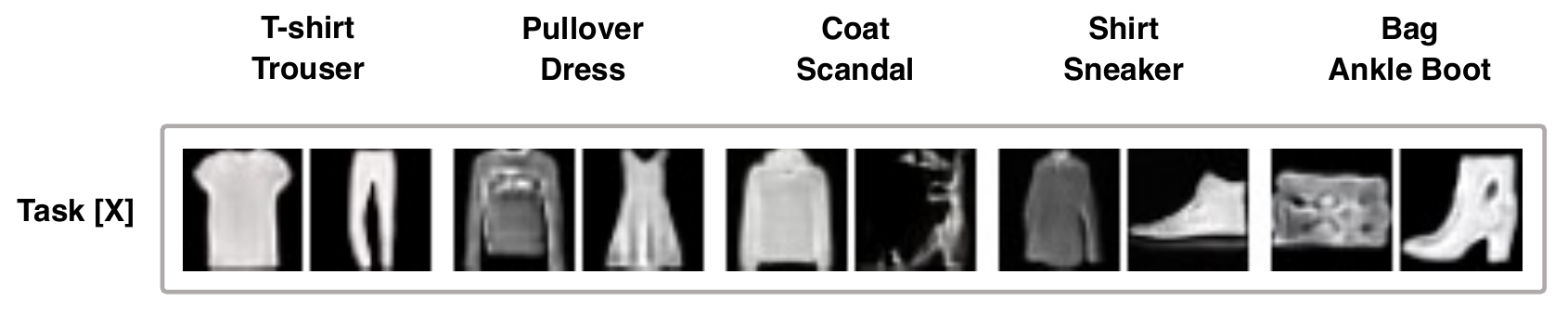}  
} 
\\
\subfigure[NCL (DDIM)] {    
\includegraphics[width=0.4\columnwidth]{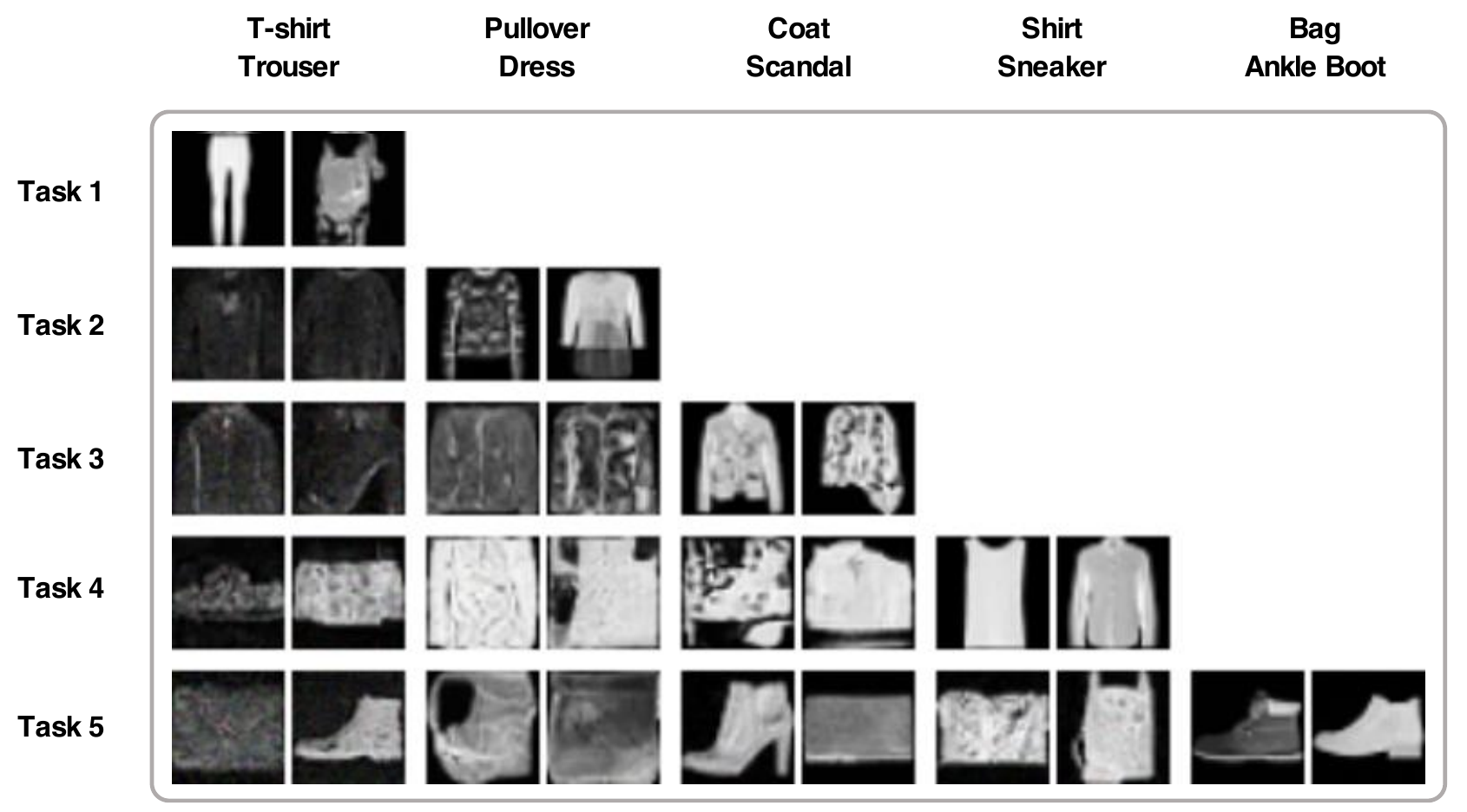}  
}    
\subfigure[Non-CL (DDIM)] {    
\includegraphics[width=0.4\columnwidth]{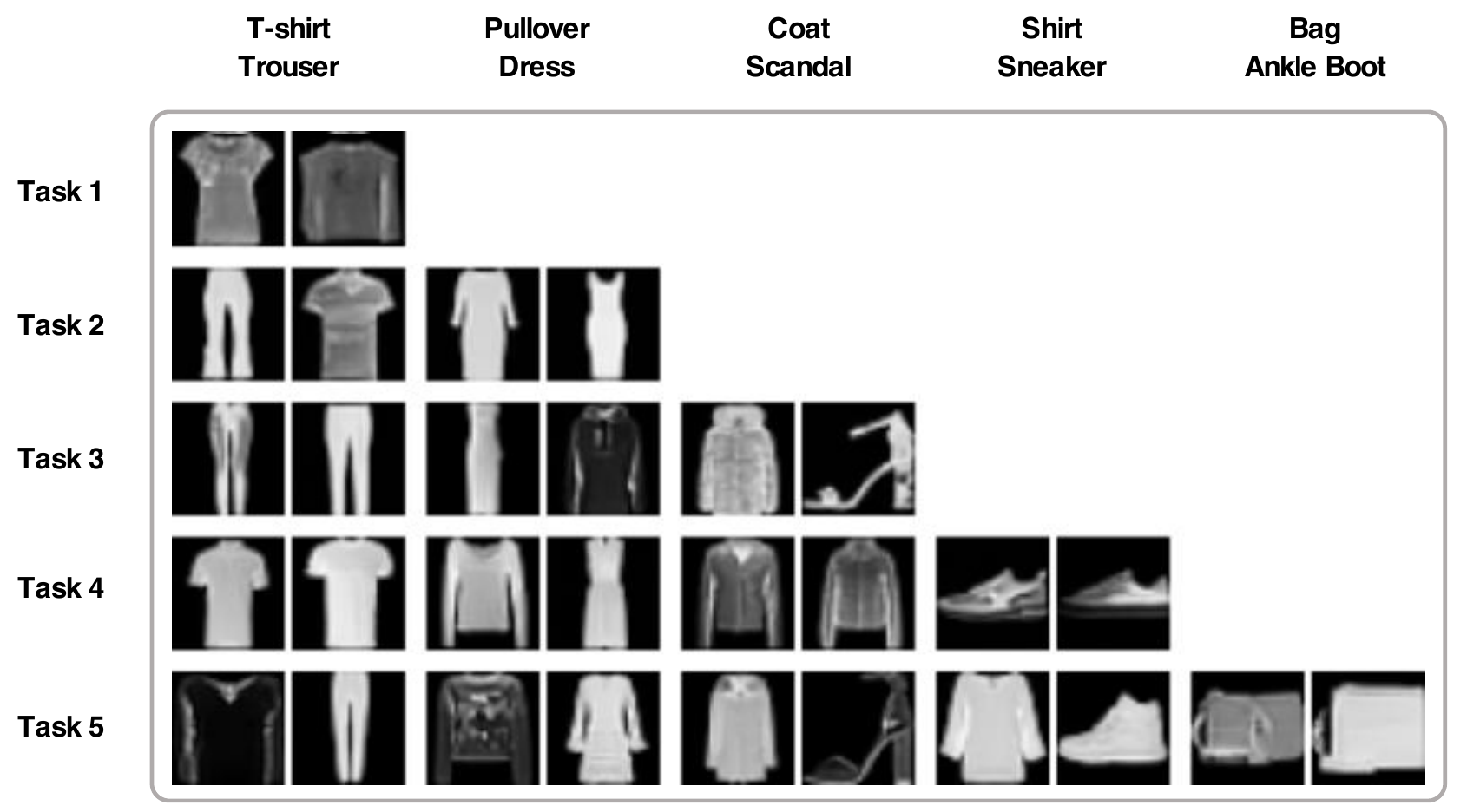}  
}    
\subfigure[ER (DDIM)] {    
\includegraphics[width=0.4\columnwidth]{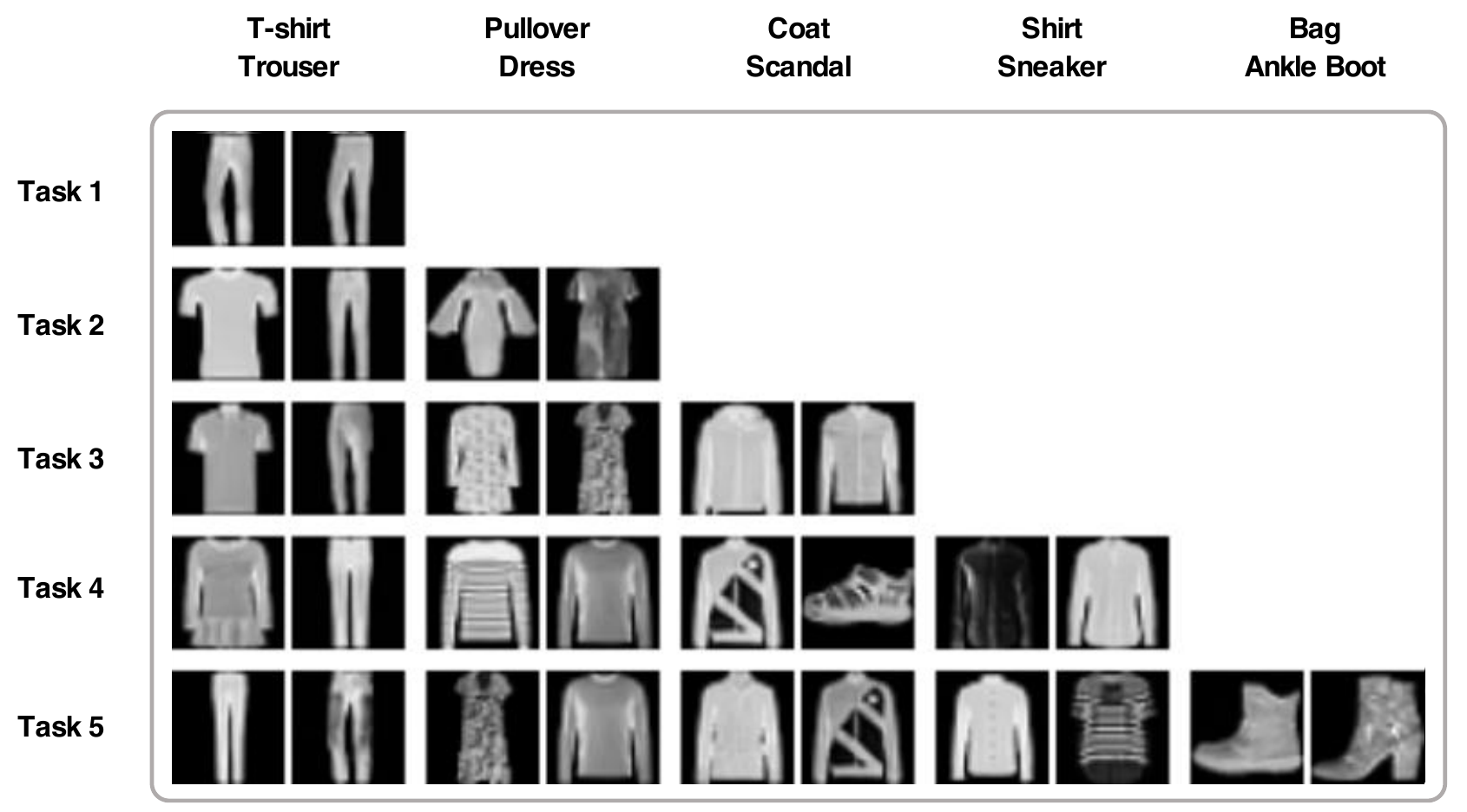}  
}    
\subfigure[EWC (DDIM)] {    
\includegraphics[width=0.4\columnwidth]{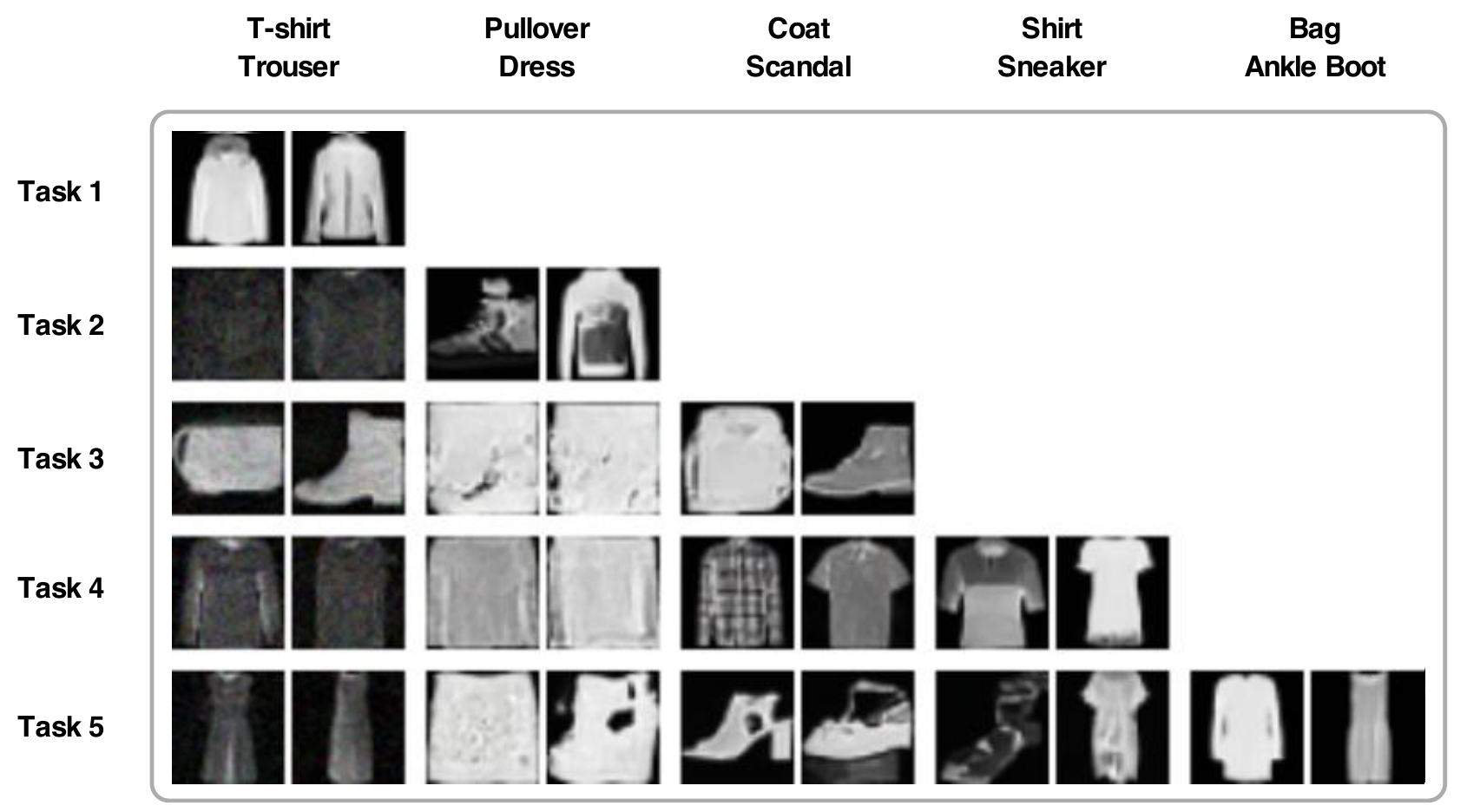}  
} 
\subfigure[Ensemble (DDIM)] {    
\includegraphics[width=0.4\columnwidth]{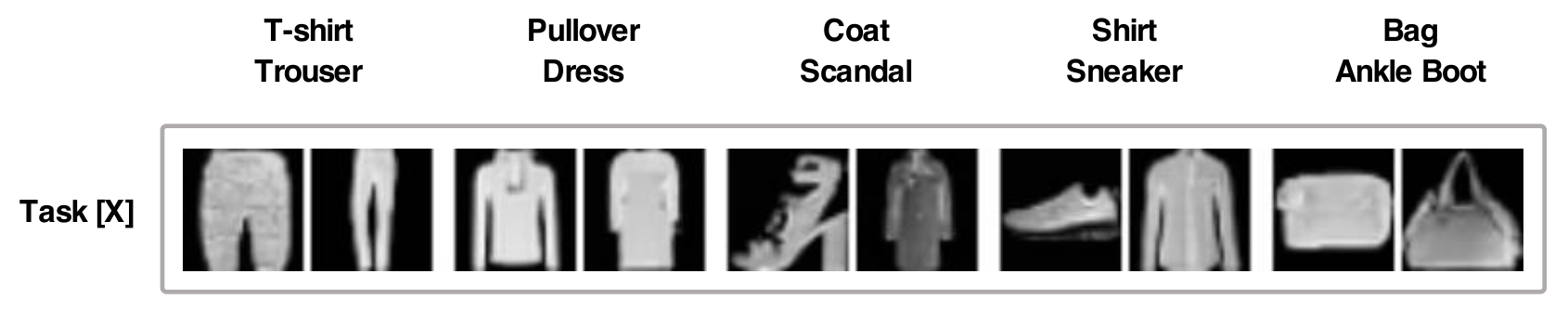}  
}
\caption{Visualization results of label-conditioned CLoG on the Fashion-MNIST~\citep{fashionmnist} dataset.} 
\label{fig:vis_fm}
\end{figure}

\begin{figure}[H]
\centering
\subfigure[NCL (GAN)] {    
\includegraphics[width=0.4\columnwidth]{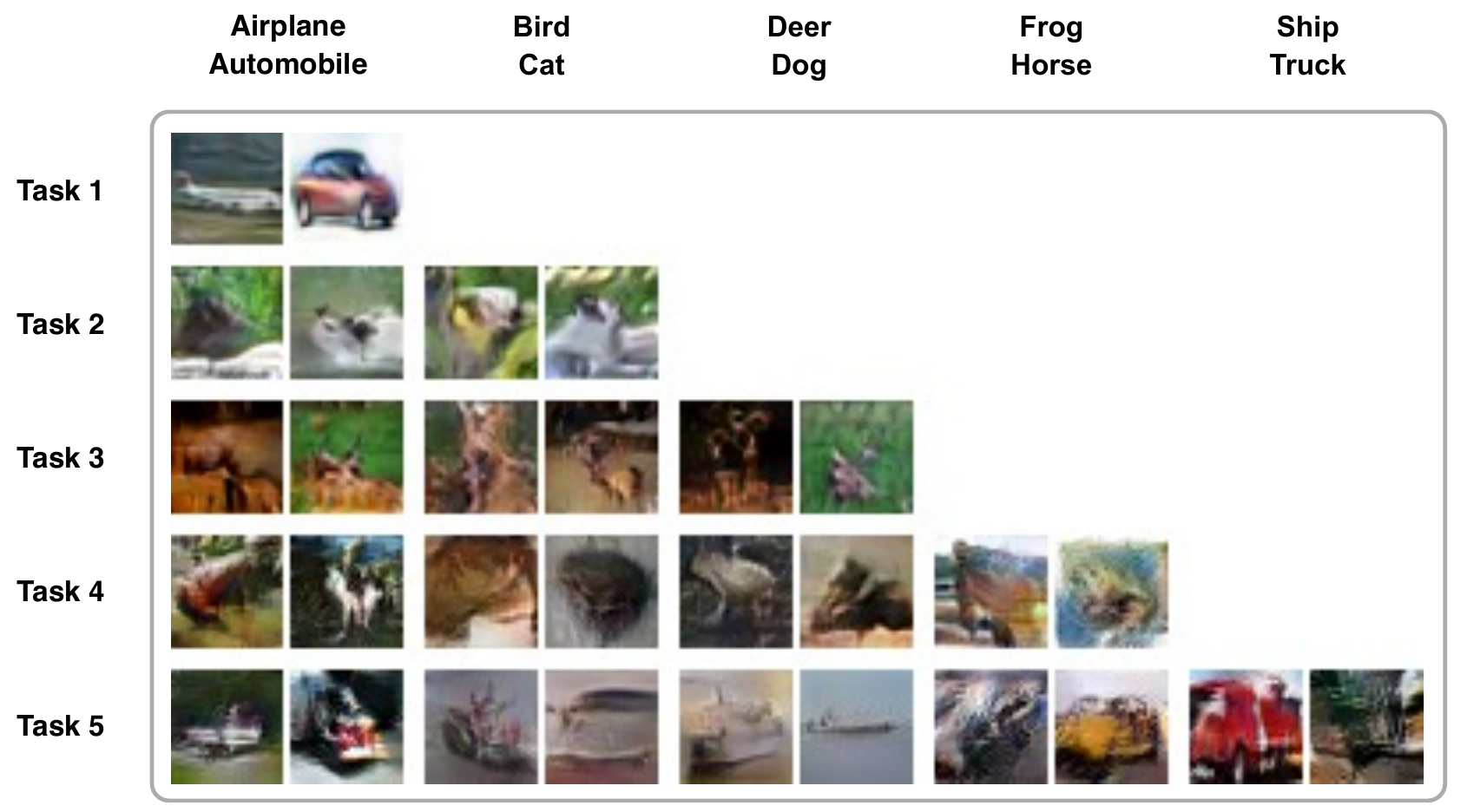}  
}    
\subfigure[Non-CL (GAN)] {    
\includegraphics[width=0.4\columnwidth]{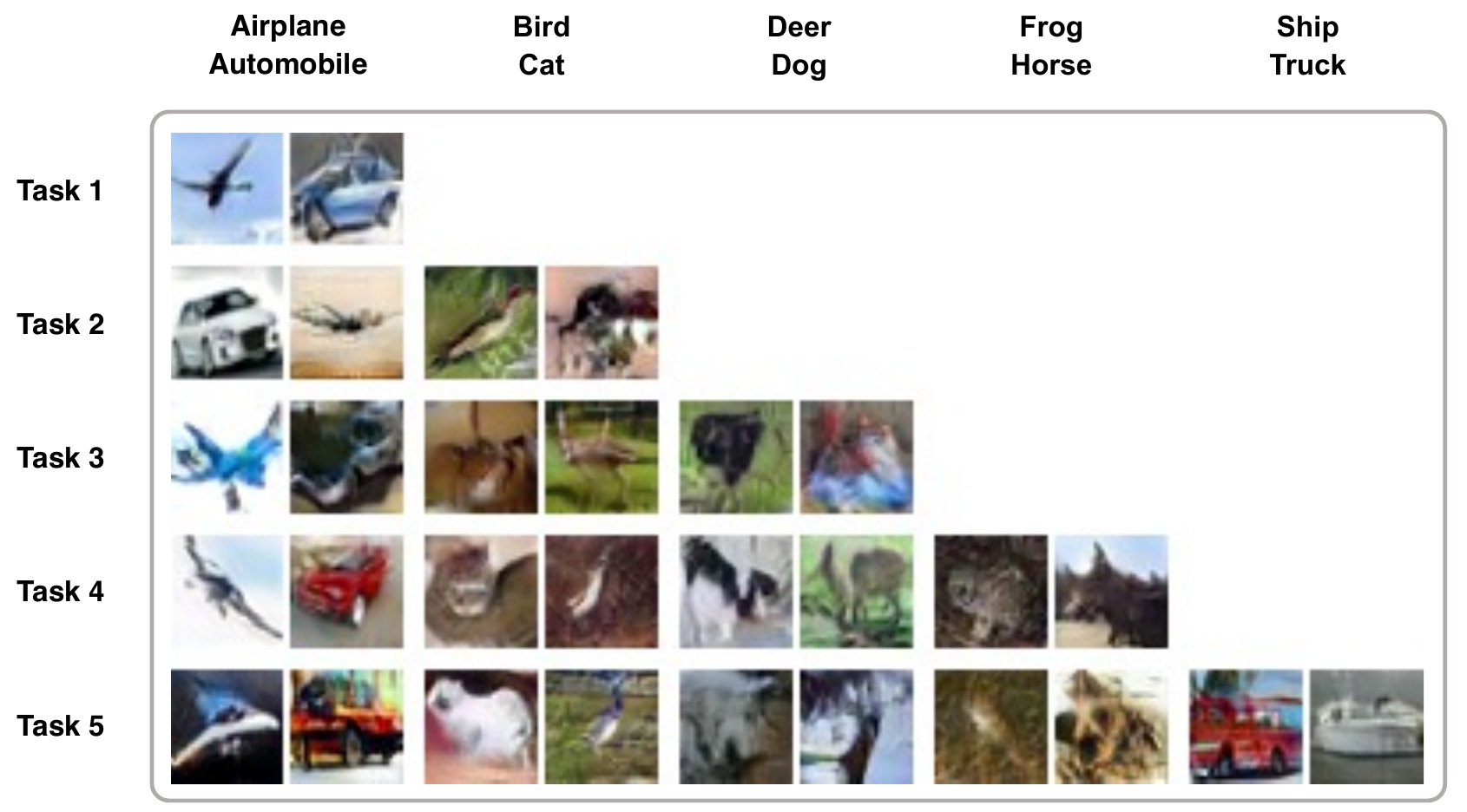}  
}    
\subfigure[ER (GAN)] {    
\includegraphics[width=0.4\columnwidth]{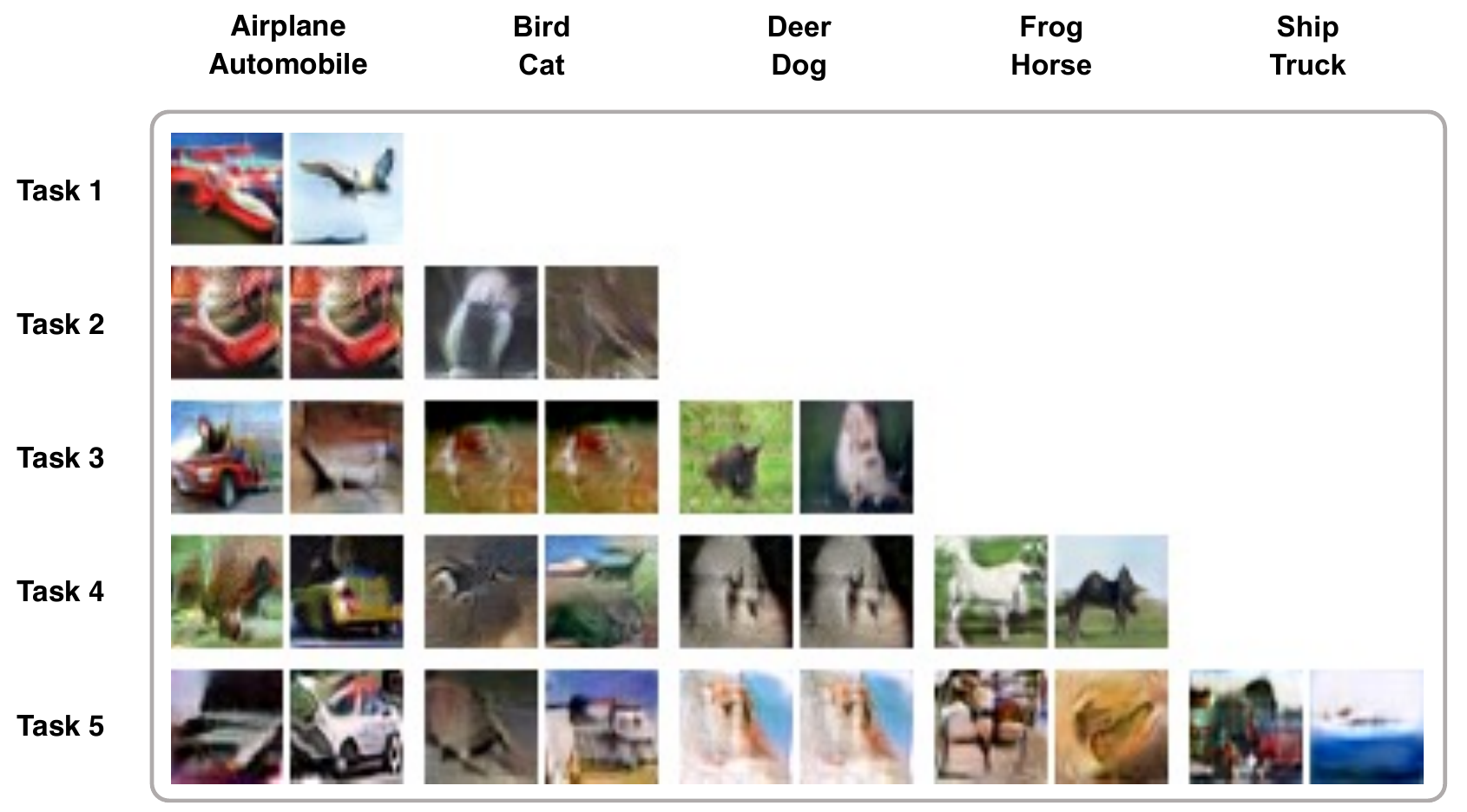}  
}    
\subfigure[EWC (GAN)] {    
\includegraphics[width=0.4\columnwidth]{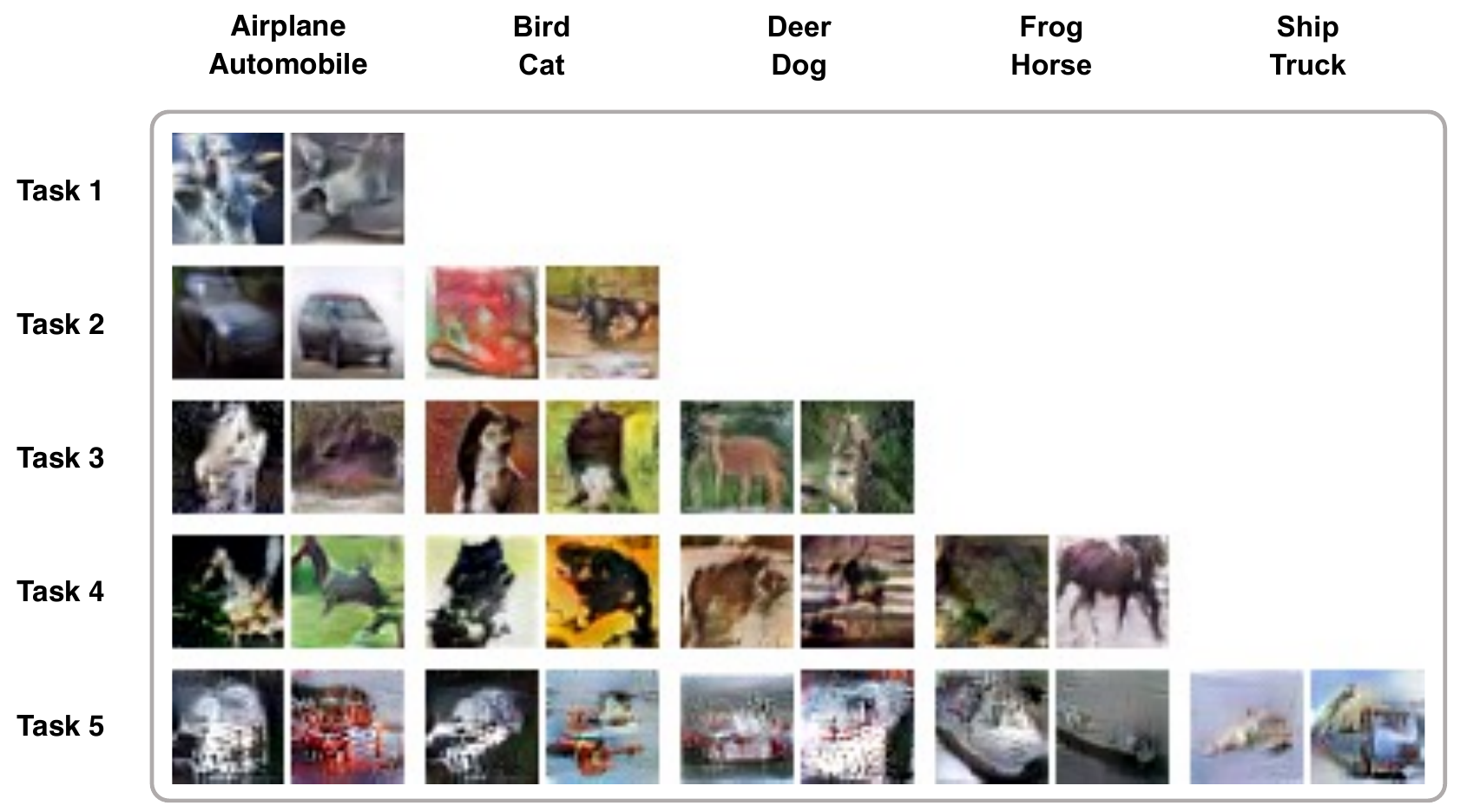}  
} 
\subfigure[Ensemble (GAN)] {    
\includegraphics[width=0.4\columnwidth]{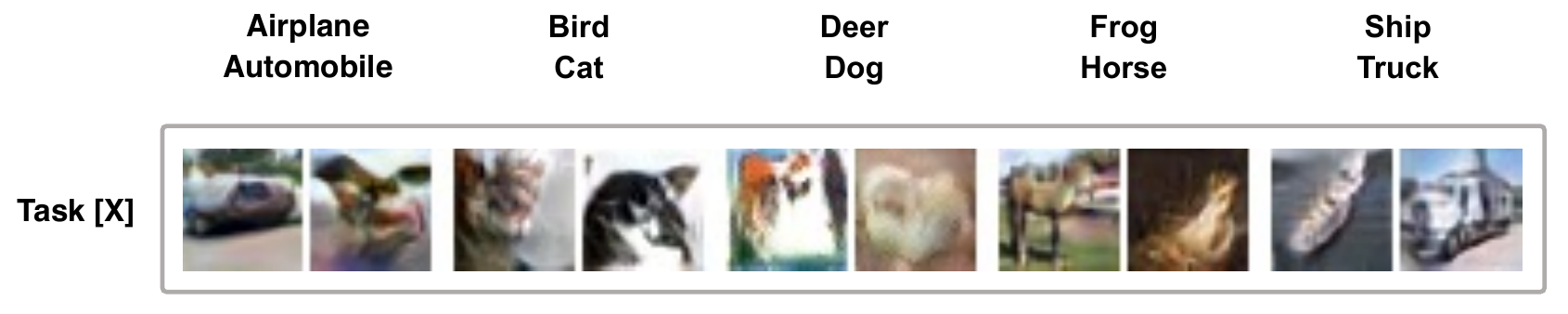}  
} 
\\
\subfigure[NCL (DDIM)] {    
\includegraphics[width=0.4\columnwidth]{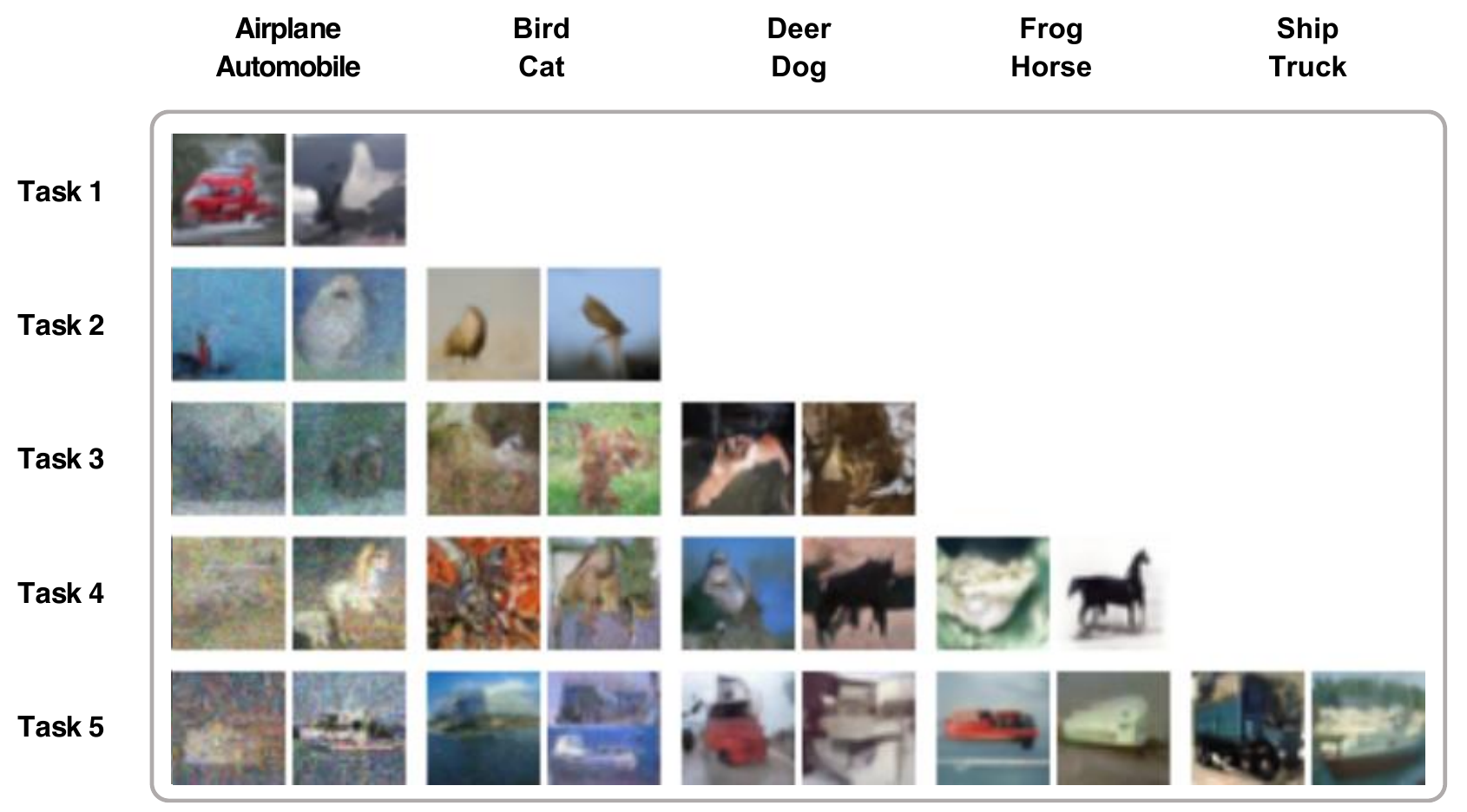}  
}    
\subfigure[Non-CL (DDIM)] {    
\includegraphics[width=0.4\columnwidth]{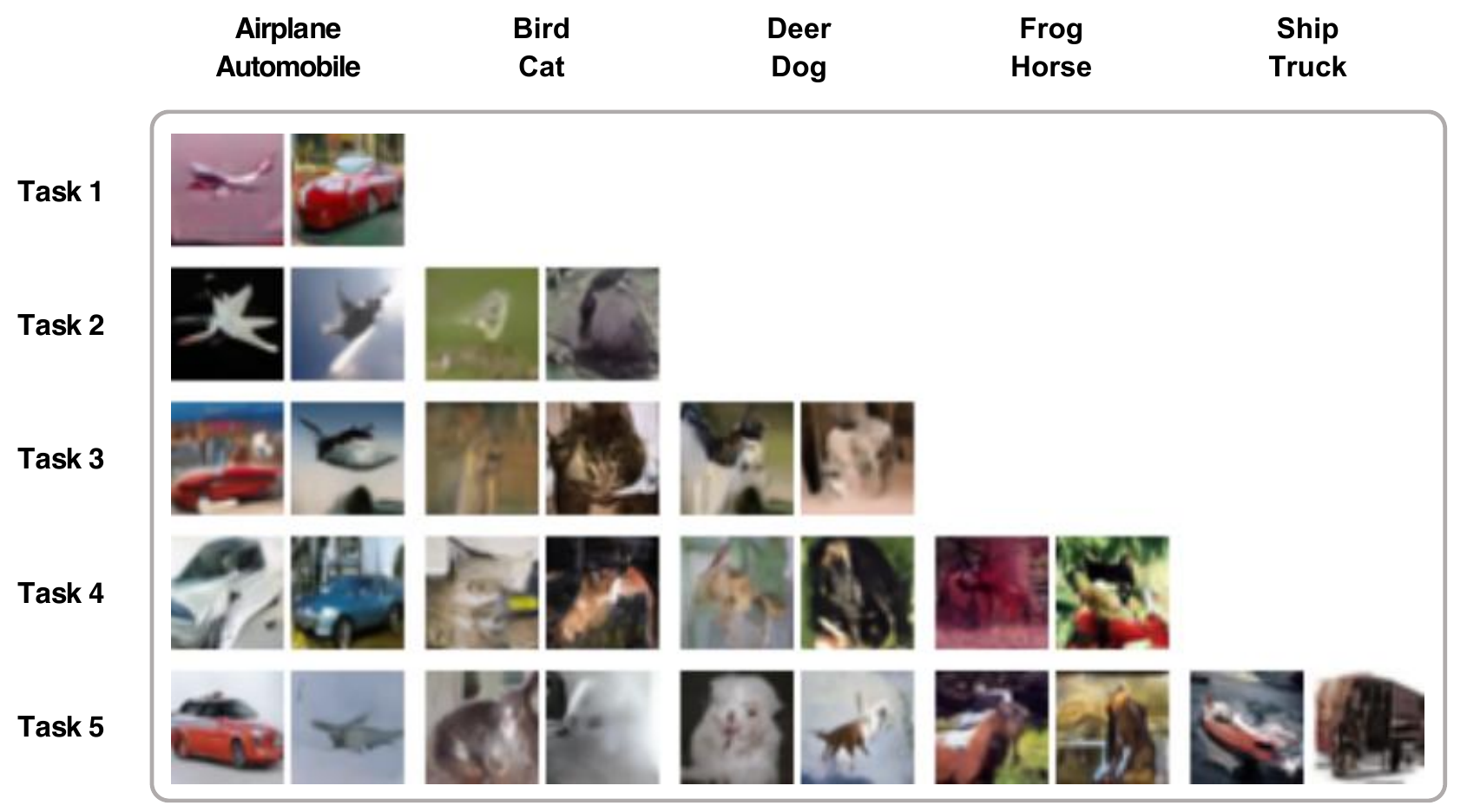}  
}    
\subfigure[ER (DDIM)] {    
\includegraphics[width=0.4\columnwidth]{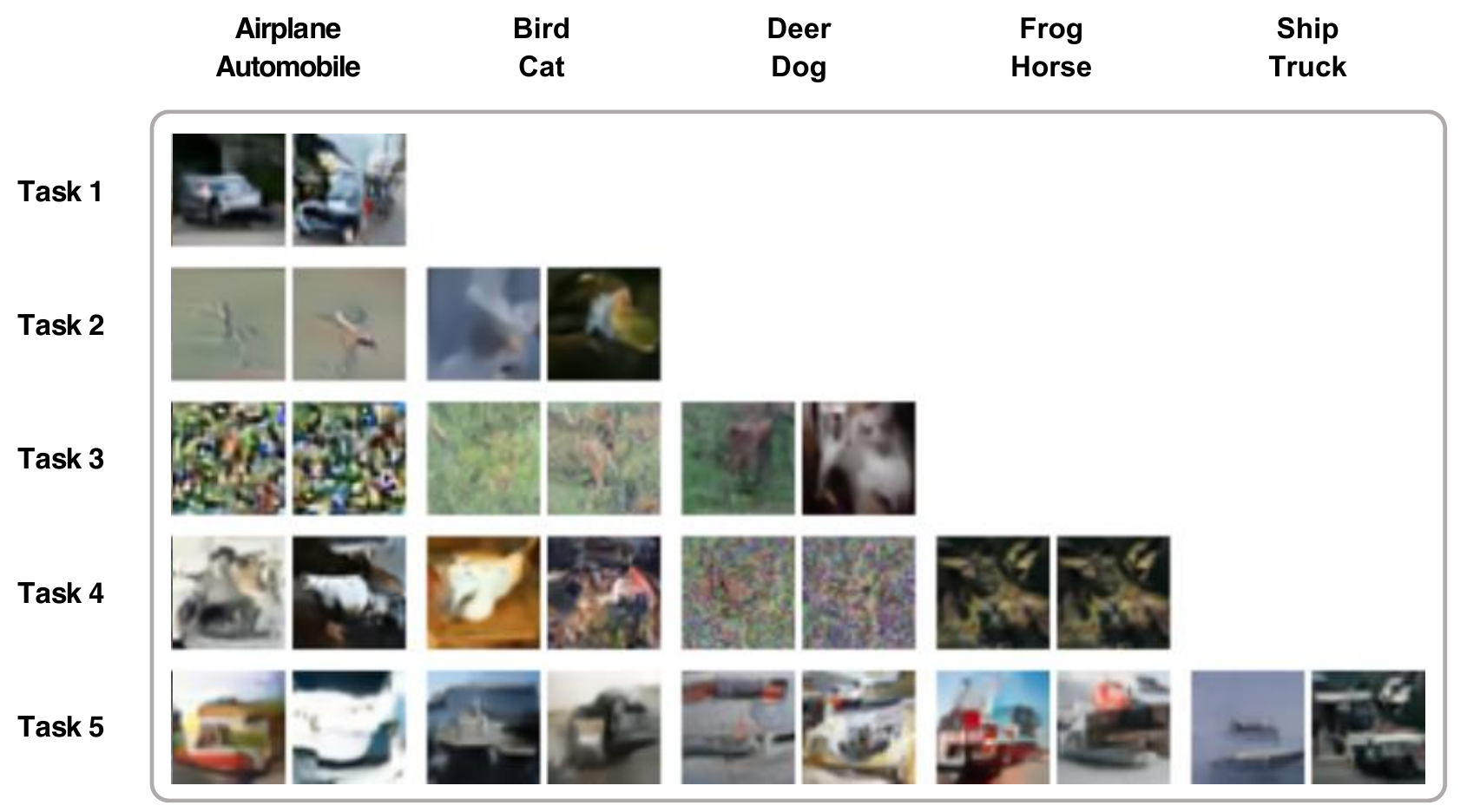}  
}    
\subfigure[EWC (DDIM)] {    
\includegraphics[width=0.4\columnwidth]{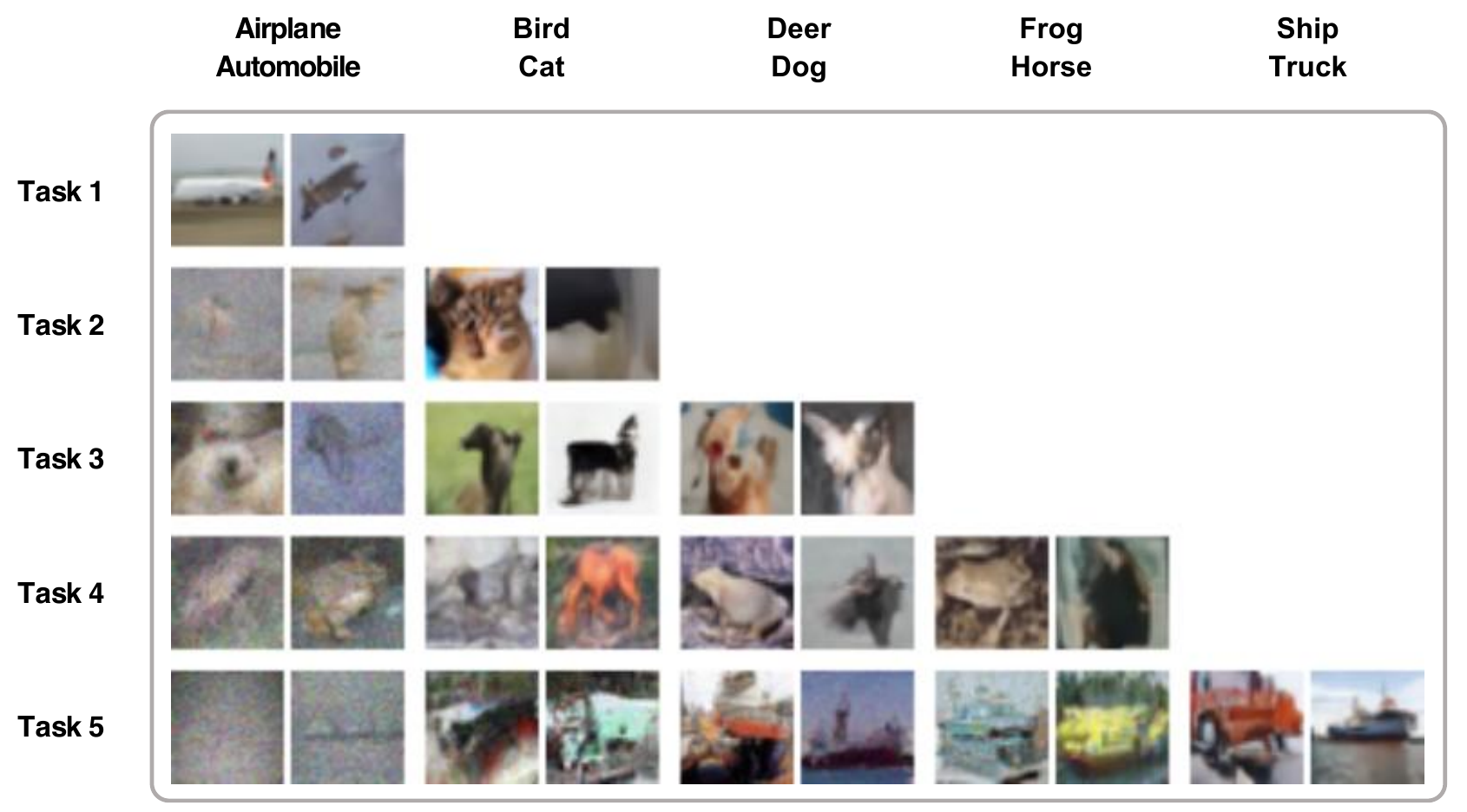}  
} 
\subfigure[Ensemble (DDIM)] {    
\includegraphics[width=0.4\columnwidth]{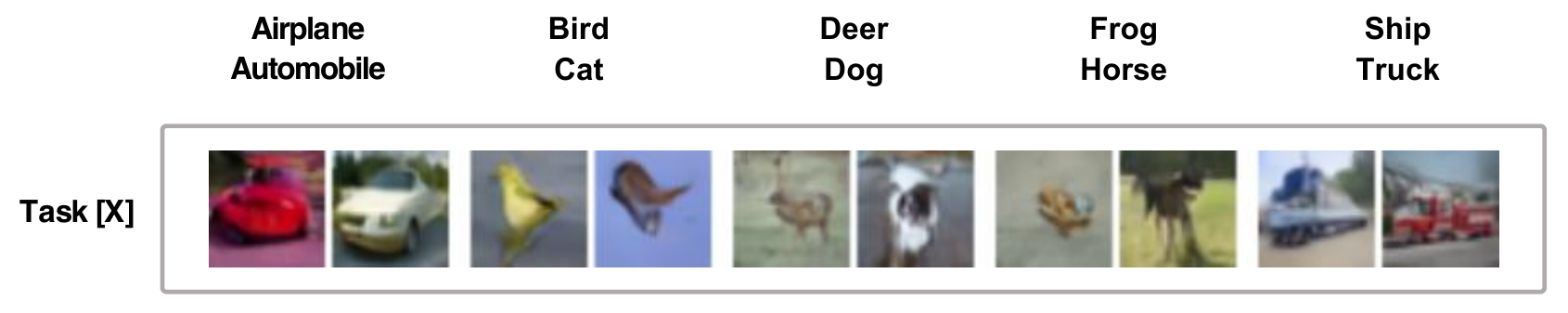}  
} 
\caption{Visualization results of label-conditioned CLoG on the CIFAR-10~\citep{cifar10} dataset.} 
\label{fig:vis_cifar}
\end{figure}

\begin{figure}[H]
\centering
\subfigure[NCL (GAN)] {    
\includegraphics[width=0.4\columnwidth]{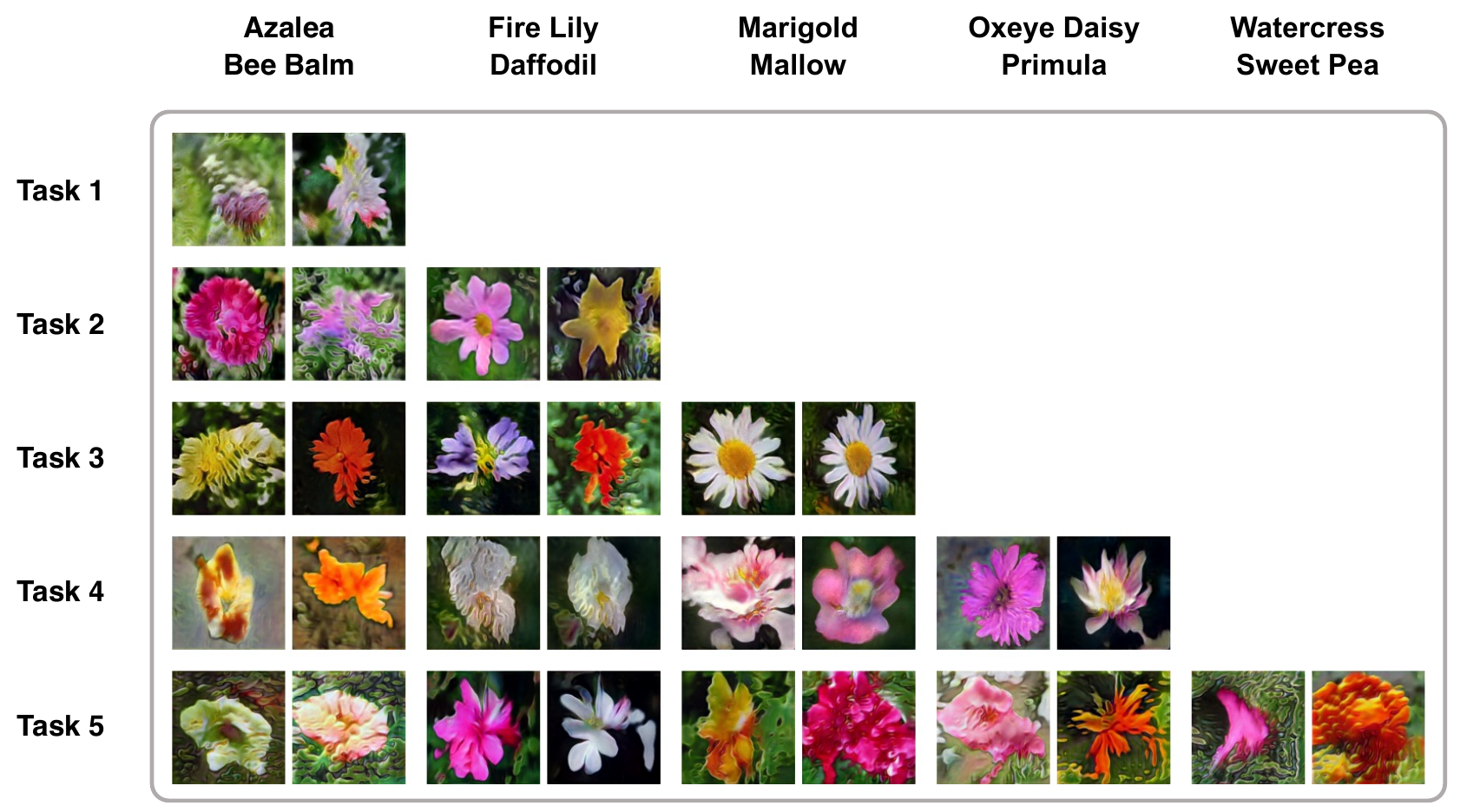}  
}    
\subfigure[Non-CL (GAN)] {    
\includegraphics[width=0.4\columnwidth]{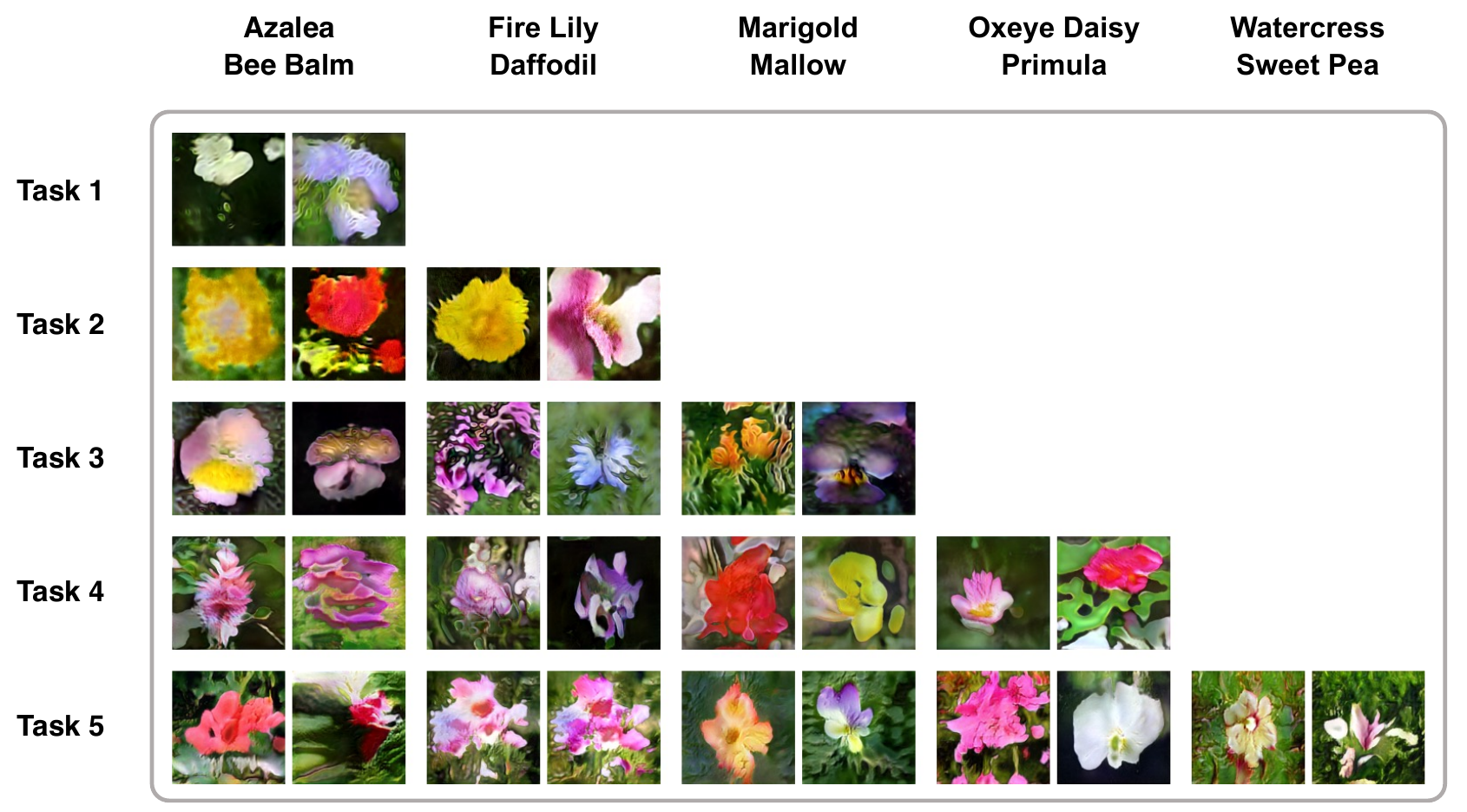}  
}    
\subfigure[ER (GAN)] {    
\includegraphics[width=0.4\columnwidth]{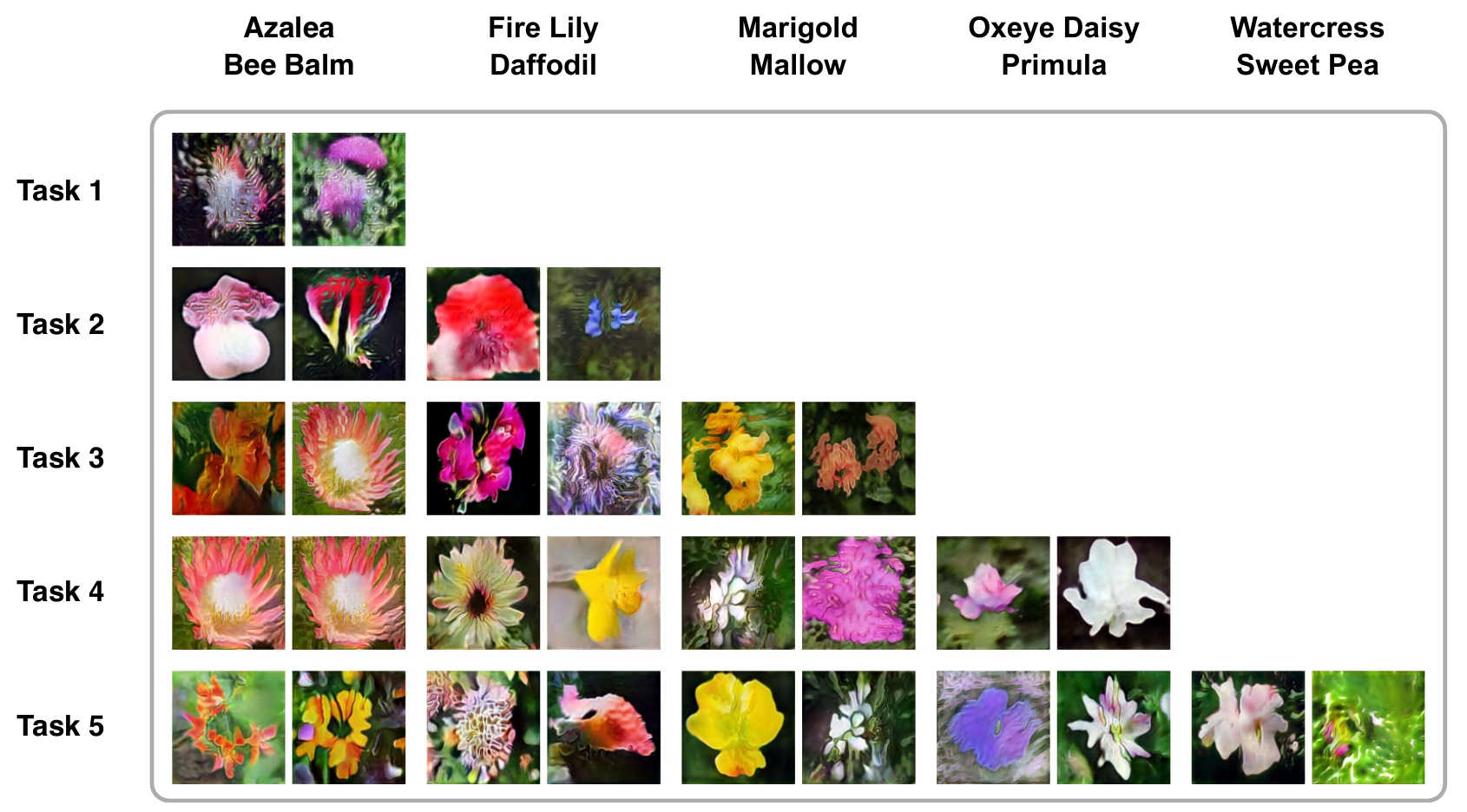}  
}    
\subfigure[EWC (GAN)] {    
\includegraphics[width=0.4\columnwidth]{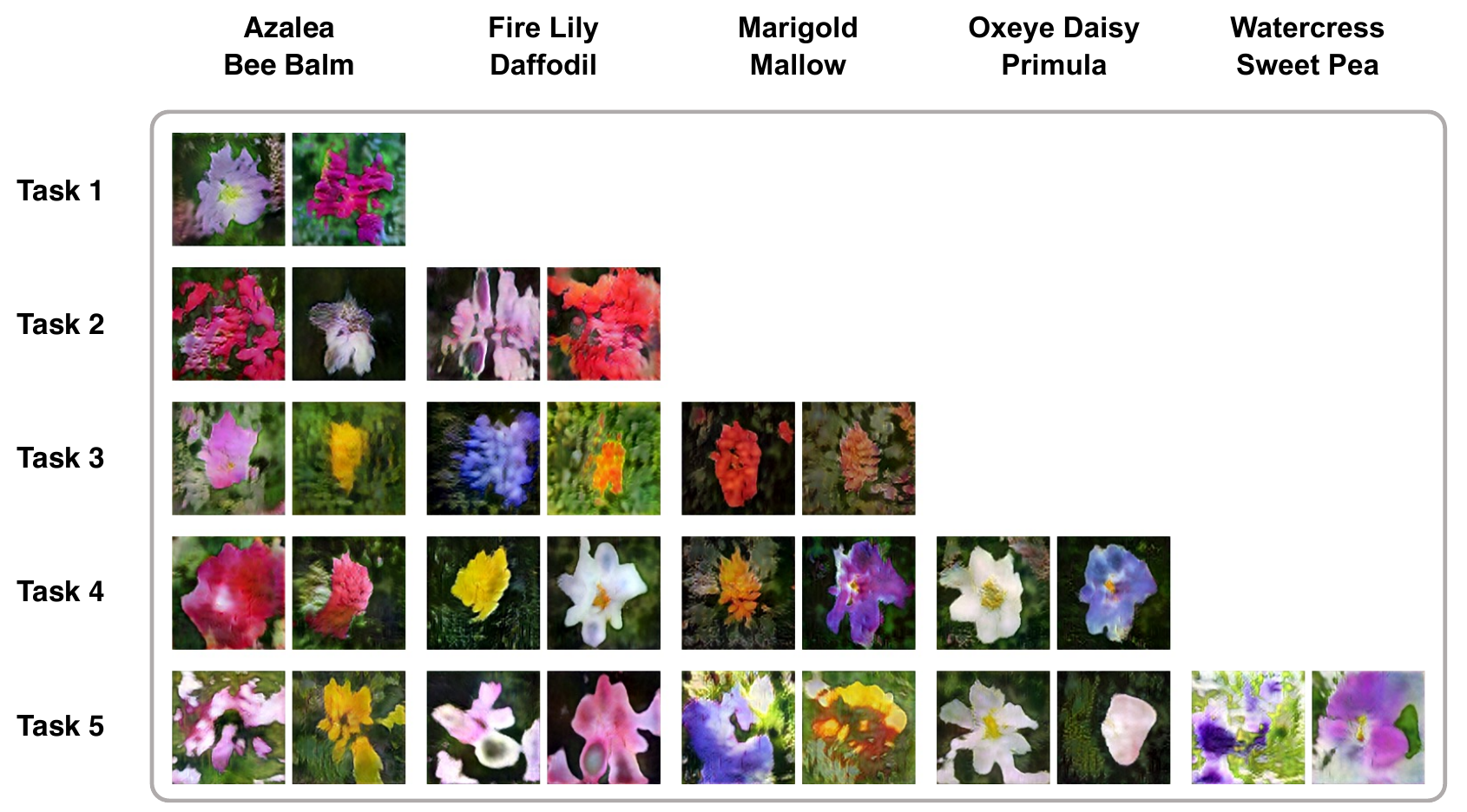}  
} 
\subfigure[Ensemble (GAN)] {    
\includegraphics[width=0.4\columnwidth]{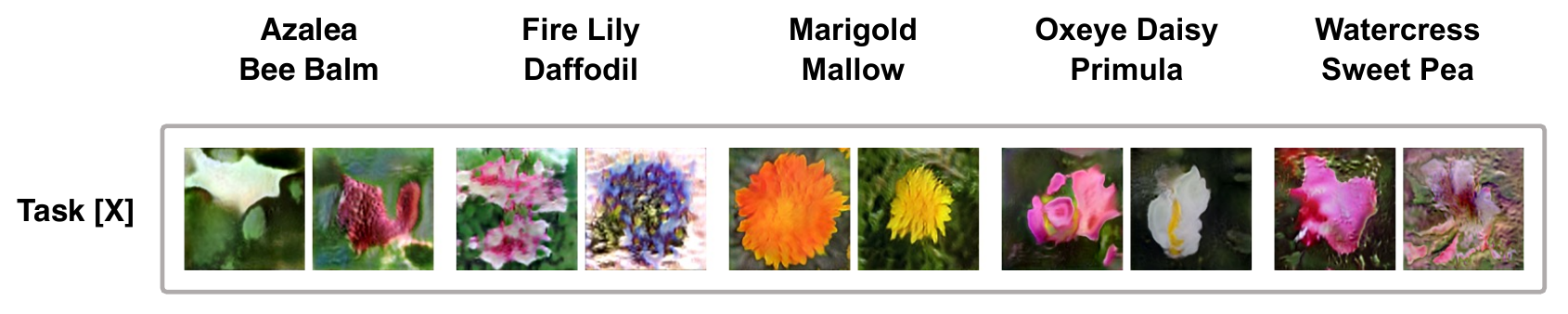}  
} 
\\
\subfigure[NCL (DDIM)] {    
\includegraphics[width=0.4\columnwidth]{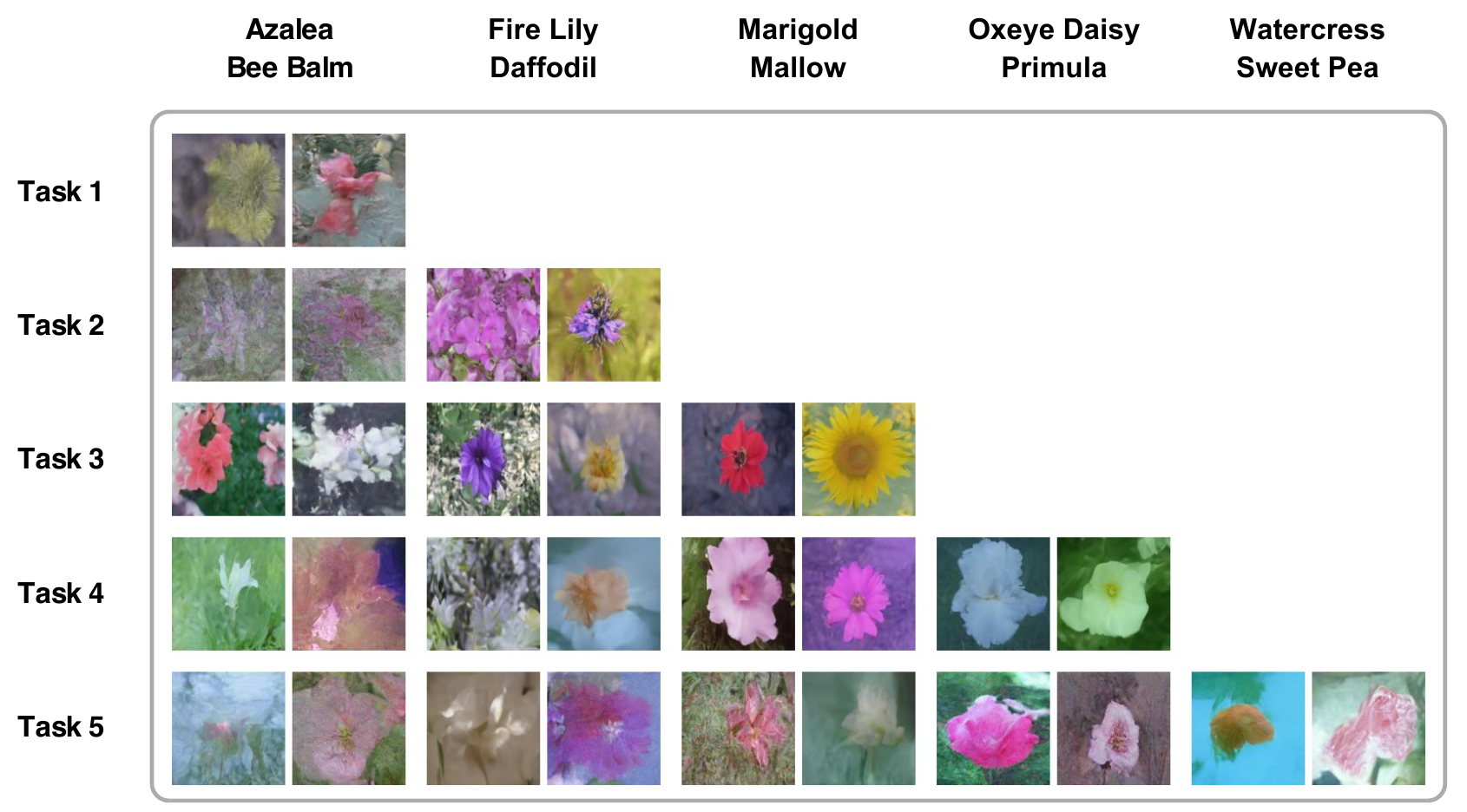}  
}    
\subfigure[Non-CL (DDIM)] {    
\includegraphics[width=0.4\columnwidth]{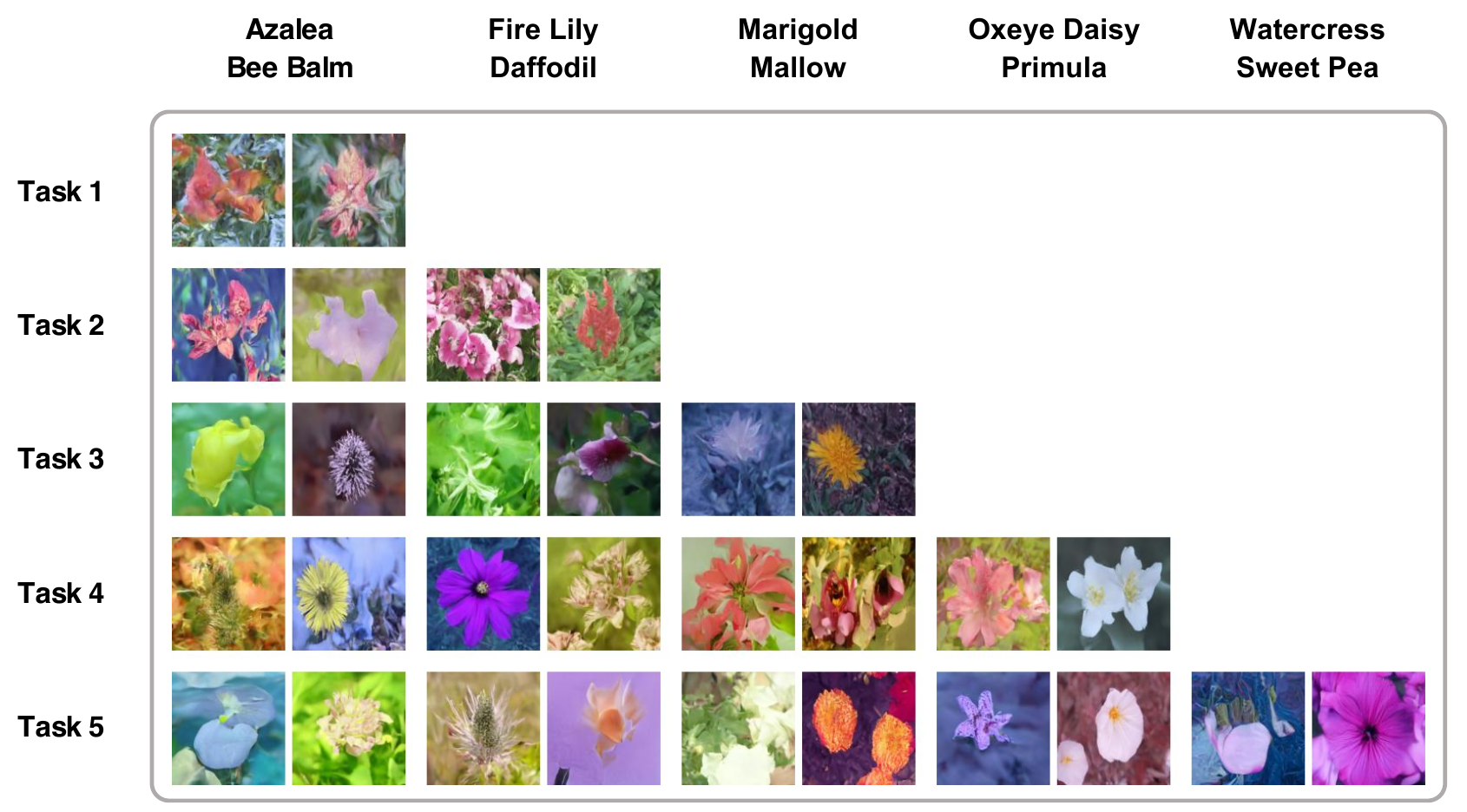}  
}    
\subfigure[ER (DDIM)] {    
\includegraphics[width=0.4\columnwidth]{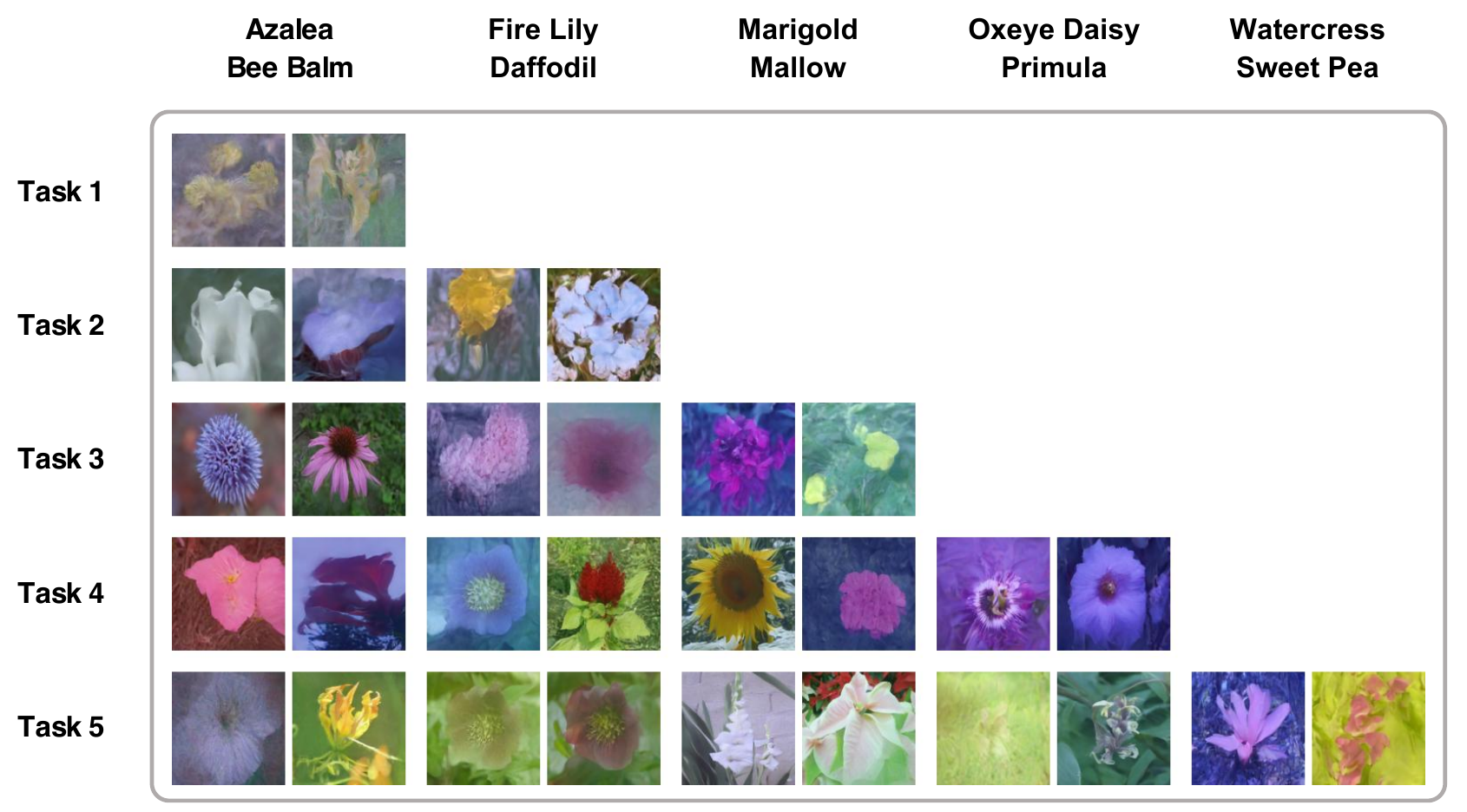}  
}    
\subfigure[EWC (DDIM)] {    
\includegraphics[width=0.4\columnwidth]{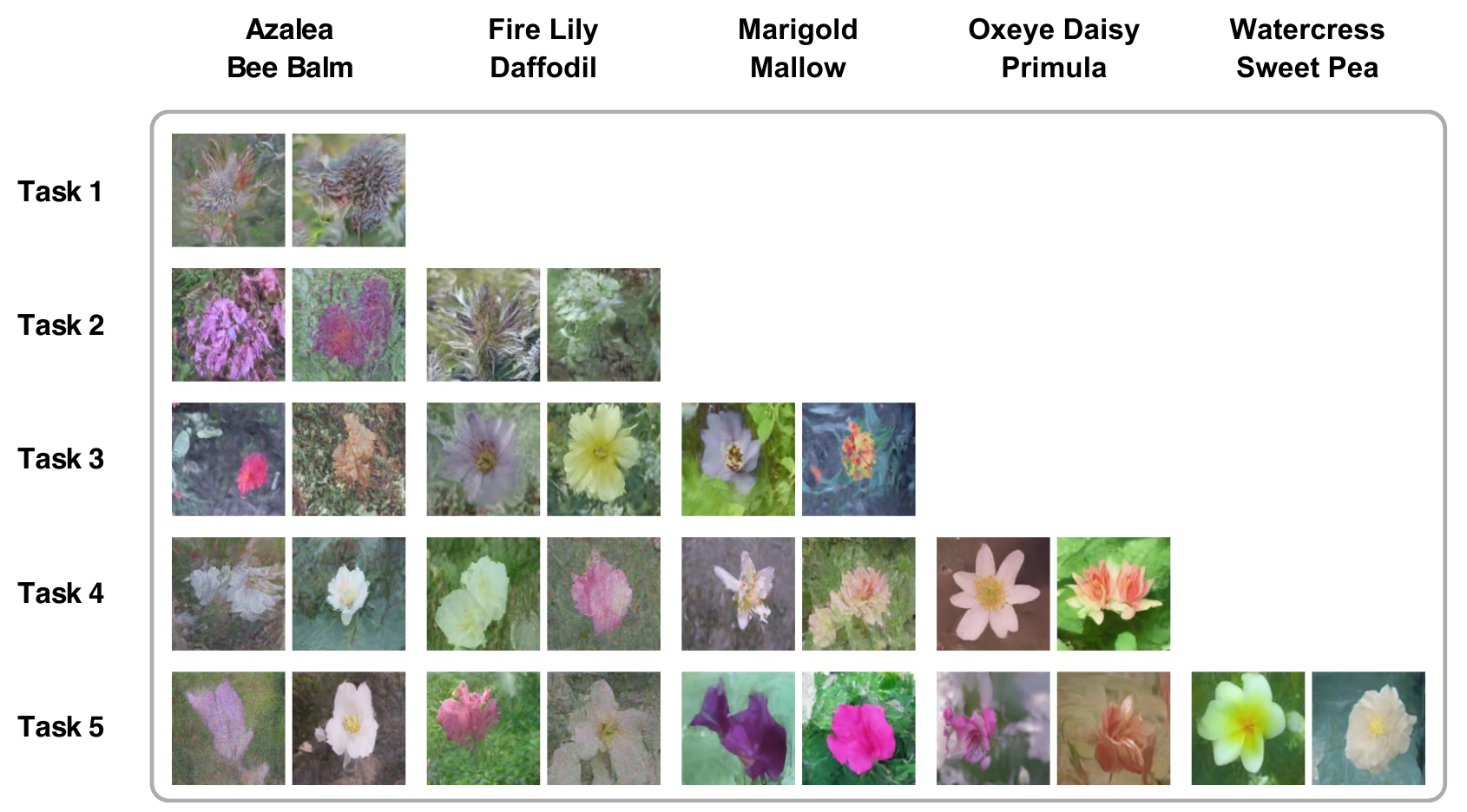}  
} 
\subfigure[Ensemble (DDIM)] {    
\includegraphics[width=0.4\columnwidth]{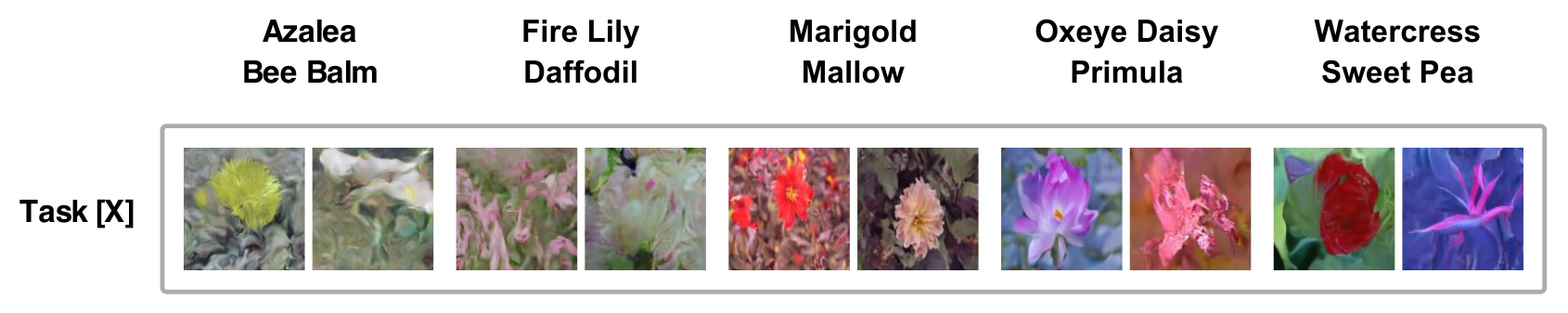}  
} 
\caption{Visualization results of label-conditioned CLoG on the Oxford-Flowers~\citep{nilsback2008automated} dataset.} 
\label{fig:vis_flower}
\end{figure}

\begin{figure}[H]
\centering
\subfigure[NCL (GAN)] {    
\includegraphics[width=0.4\columnwidth]{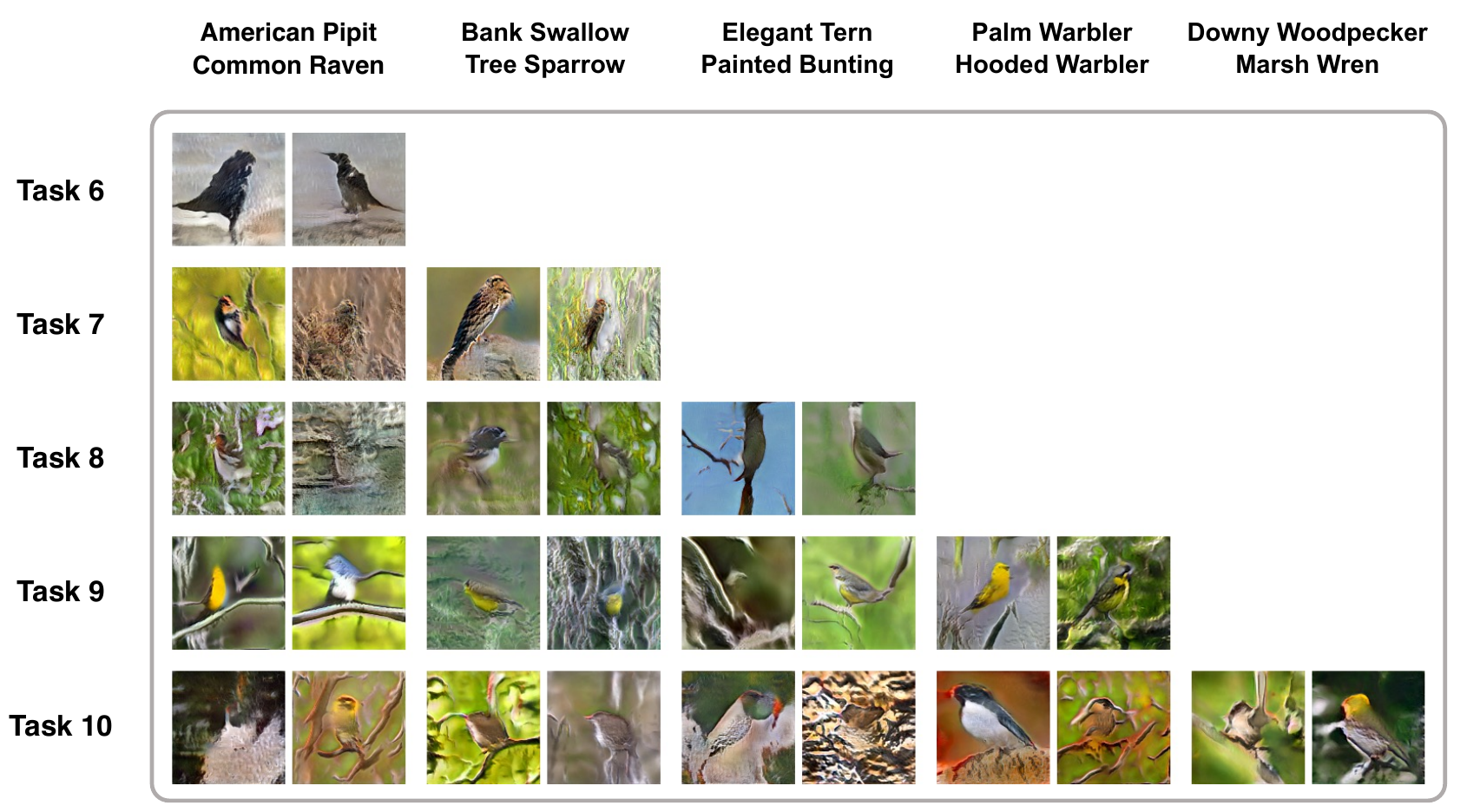}  
}    
\subfigure[Non-CL (GAN)] {    
\includegraphics[width=0.4\columnwidth]{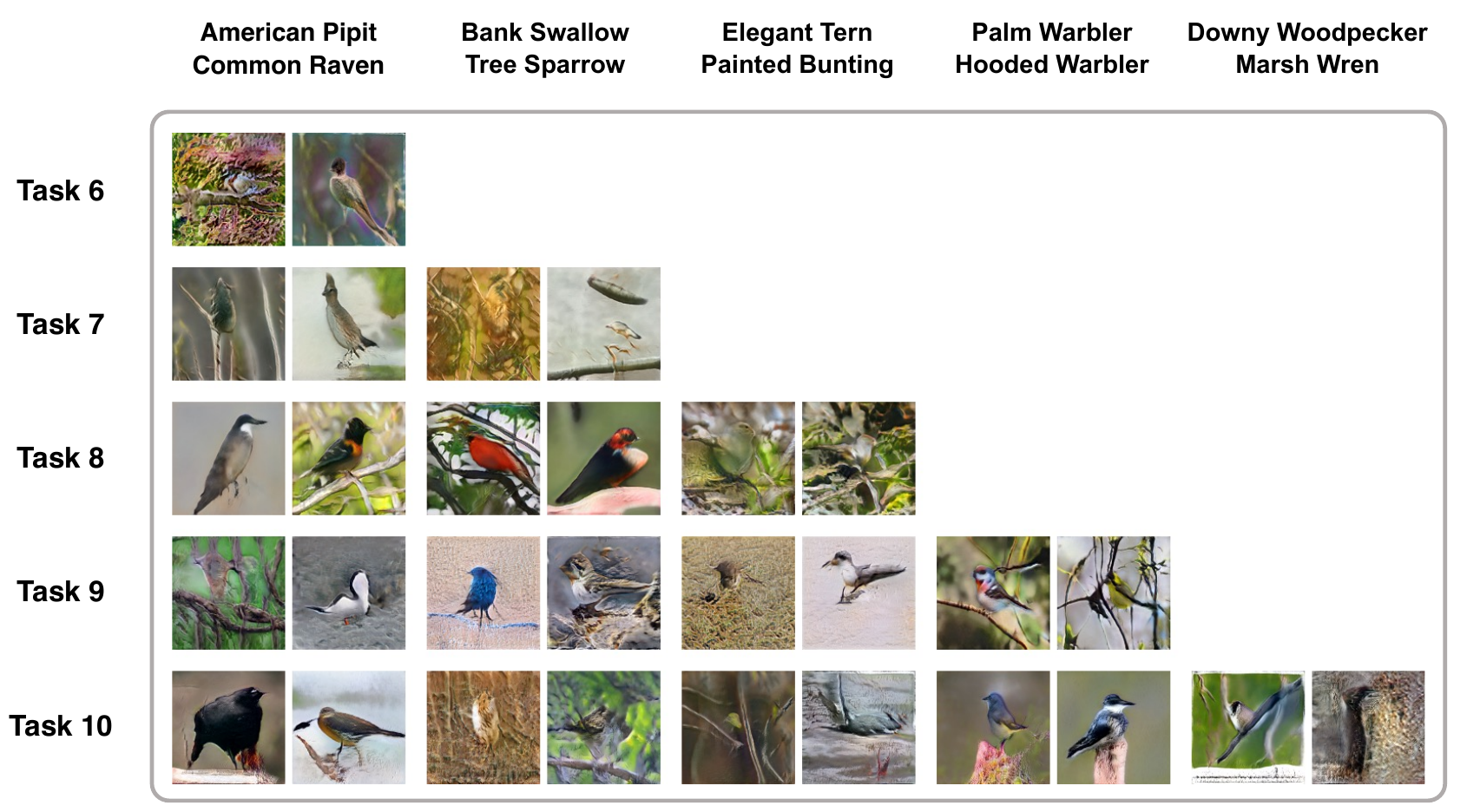}  
}    
\subfigure[ER (GAN)] {    
\includegraphics[width=0.4\columnwidth]{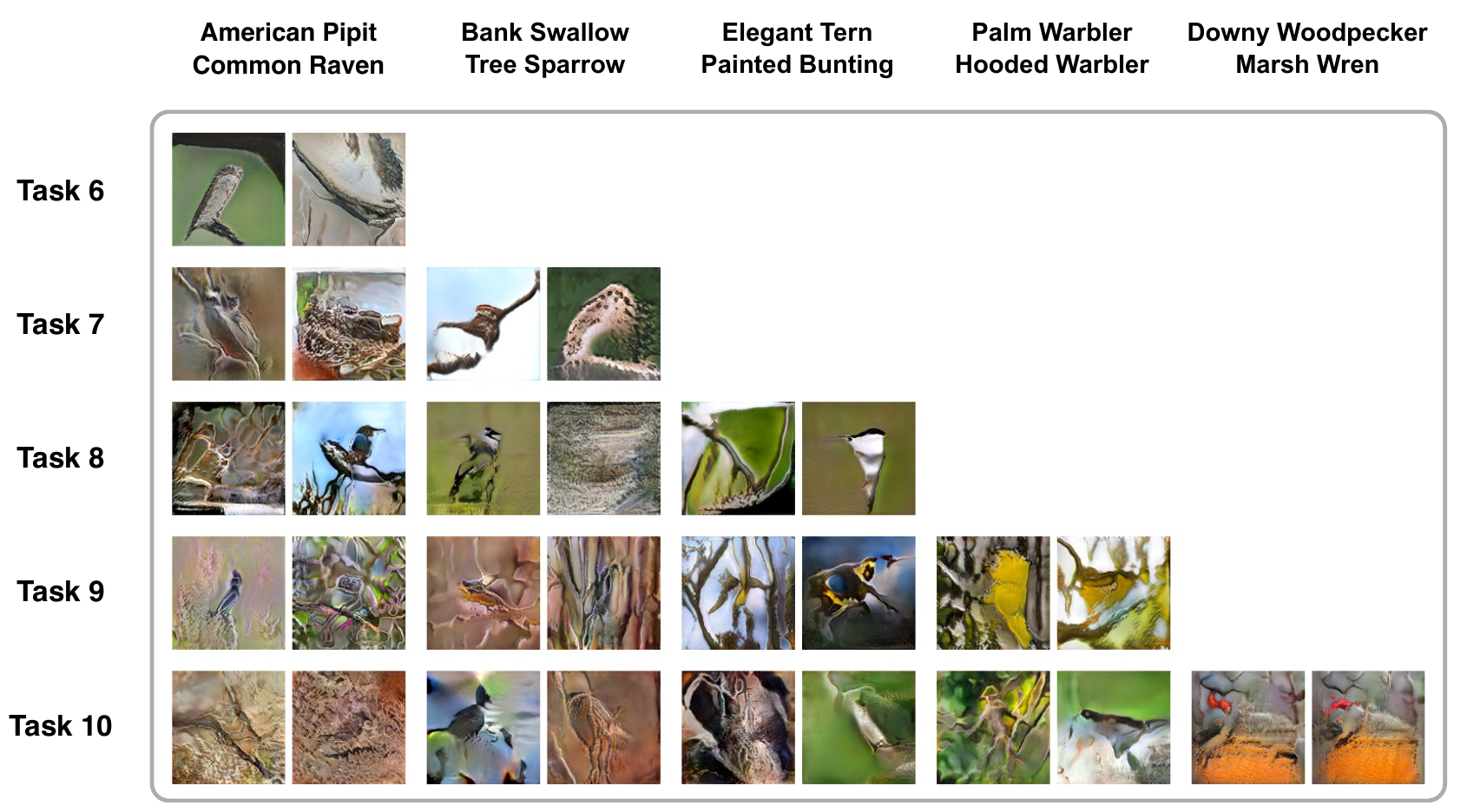}  
}    
\subfigure[EWC (GAN)] {    
\includegraphics[width=0.4\columnwidth]{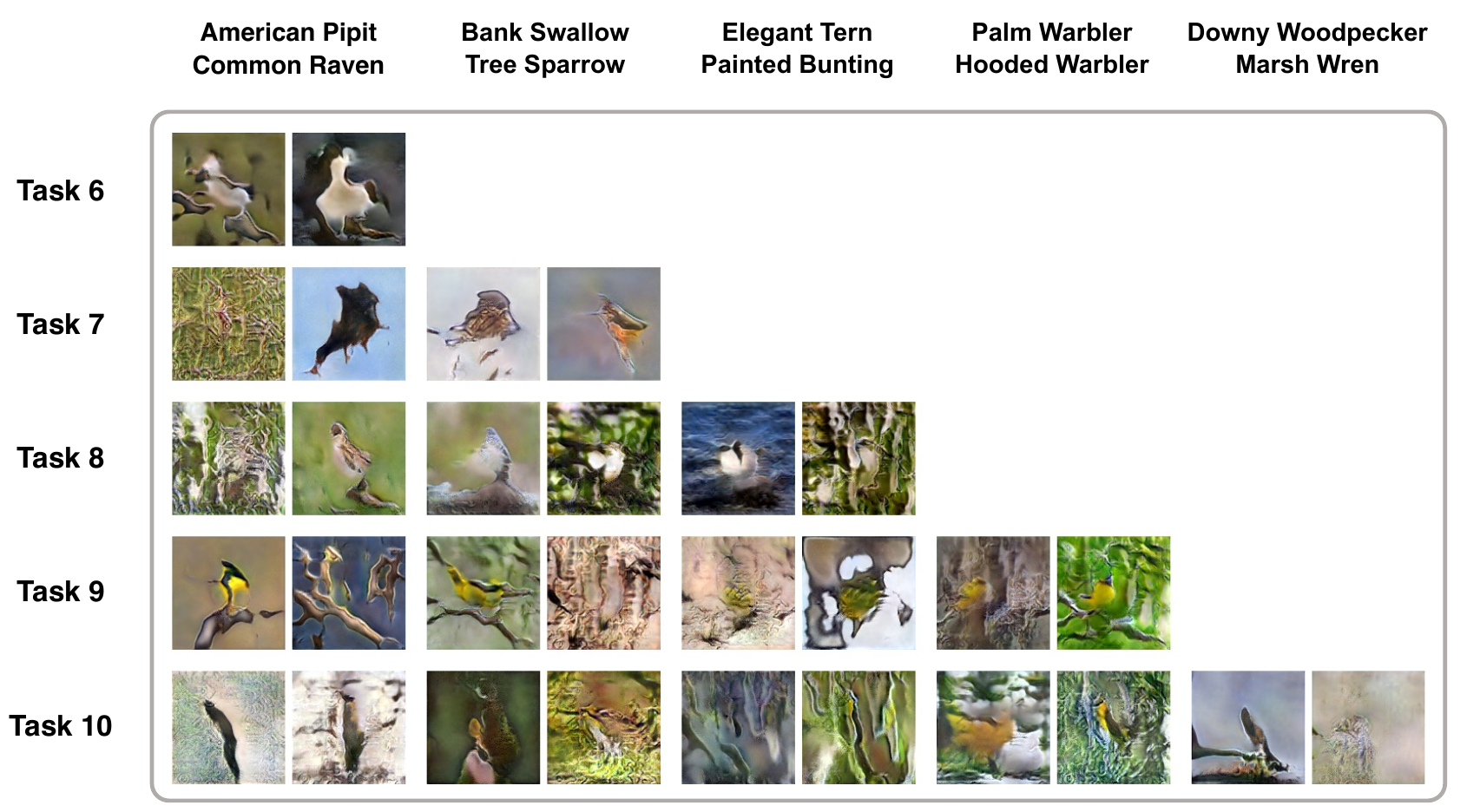}  
} 
\subfigure[Ensemble (GAN)] {    
\includegraphics[width=0.4\columnwidth]{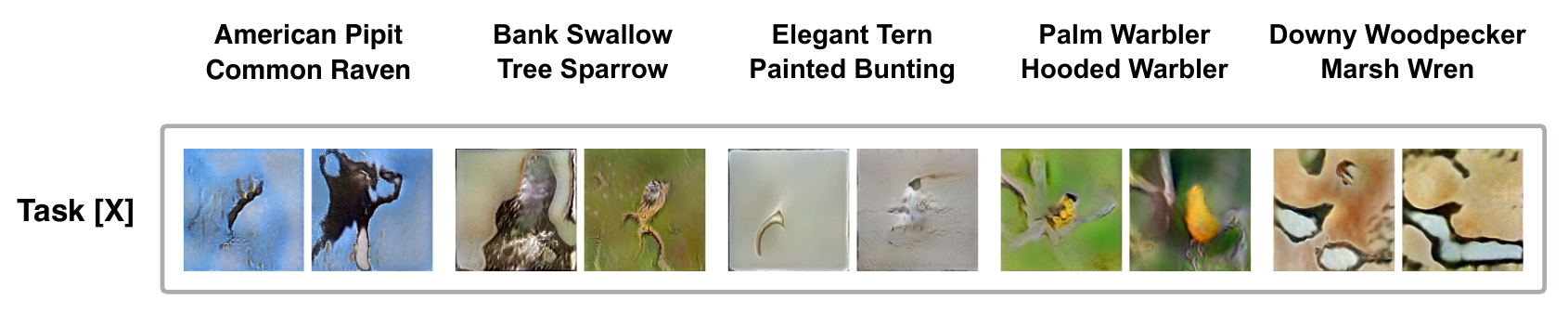}  
} 

\subfigure[NCL (DDIM)] {    
\includegraphics[width=0.4\columnwidth]{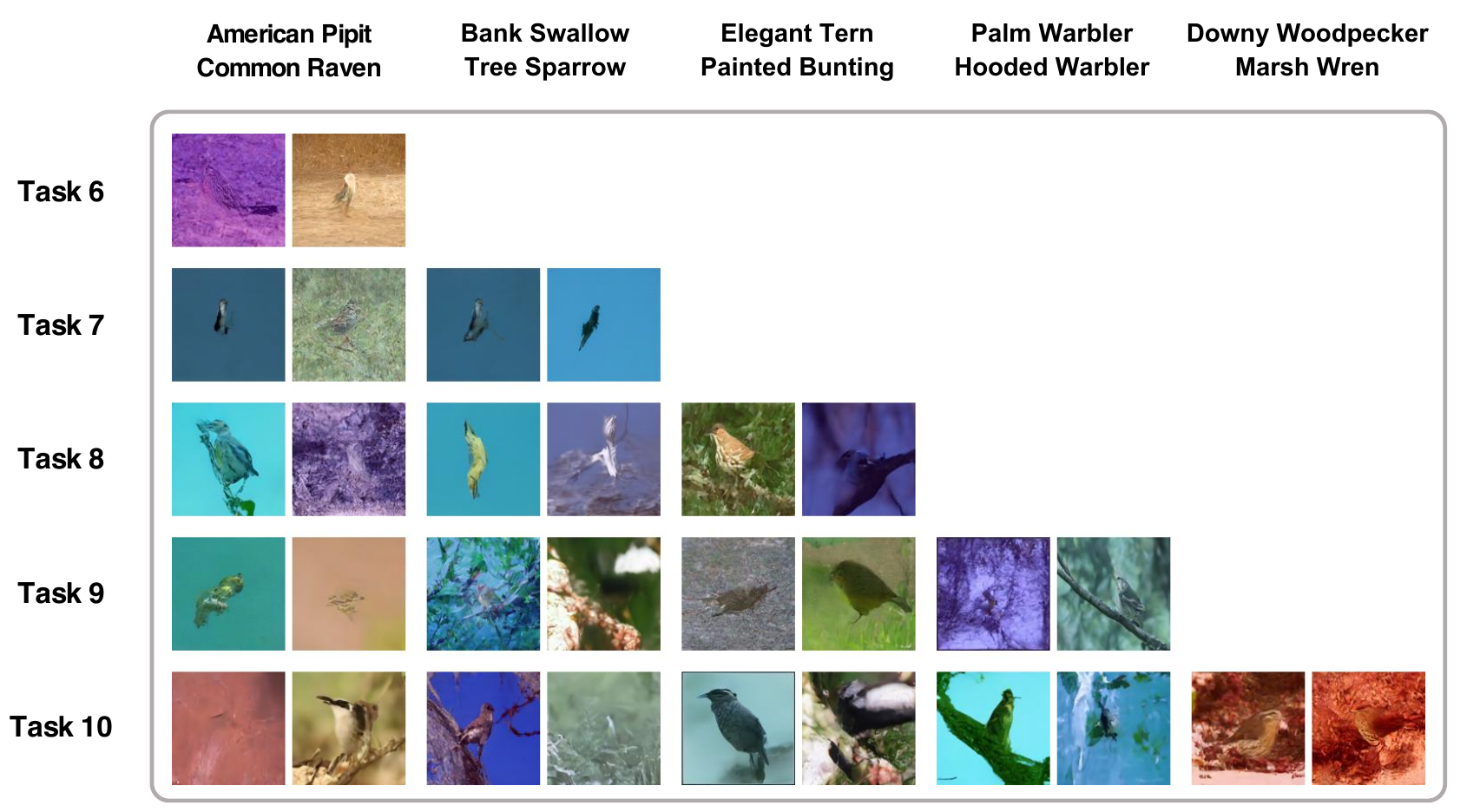}  
}    
\subfigure[Non-CL (DDIM)] {    
\includegraphics[width=0.4\columnwidth]{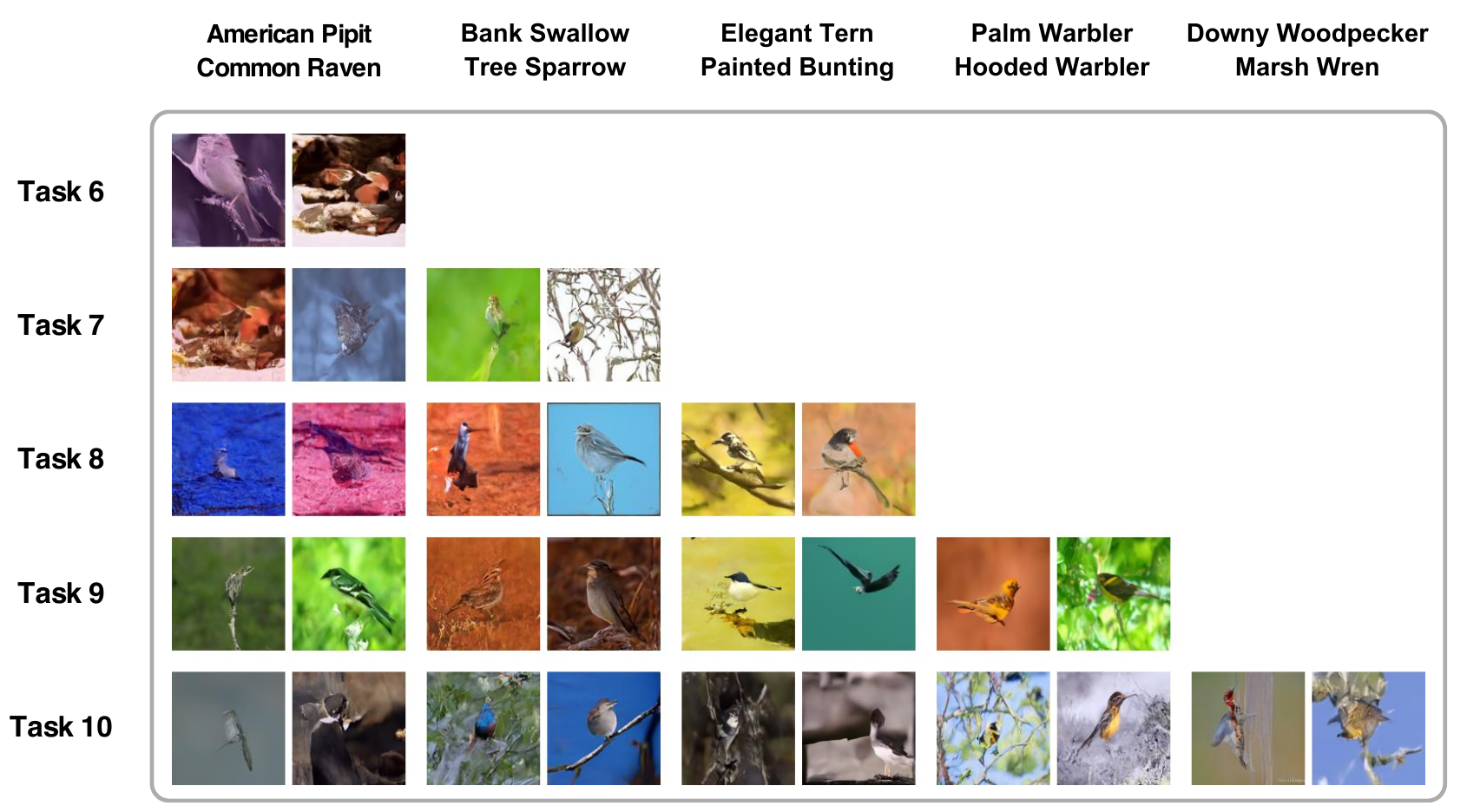}  
}    
\subfigure[ER (DDIM)] {    
\includegraphics[width=0.4\columnwidth]{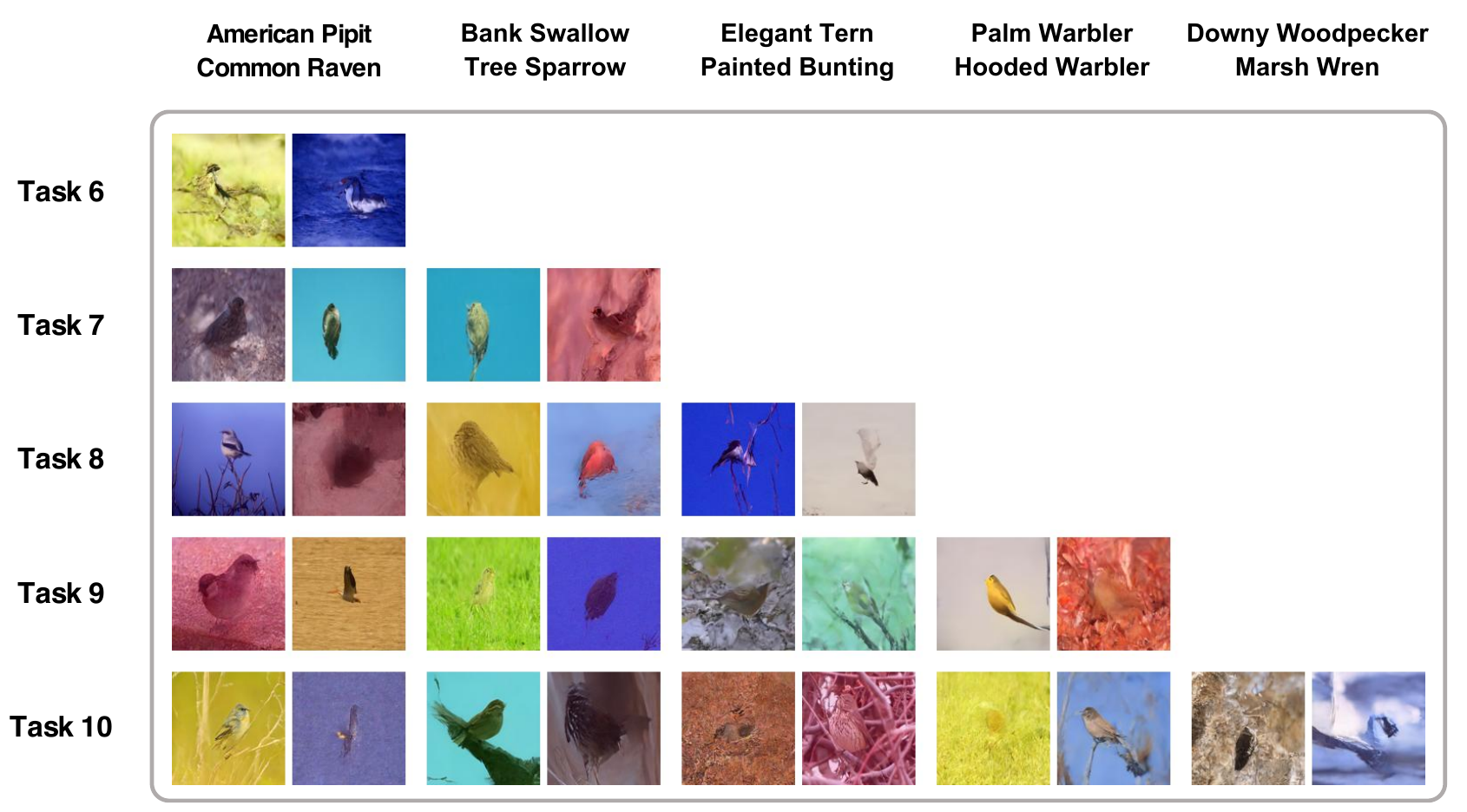}  
}    
\subfigure[EWC (DDIM)] {    
\includegraphics[width=0.4\columnwidth]{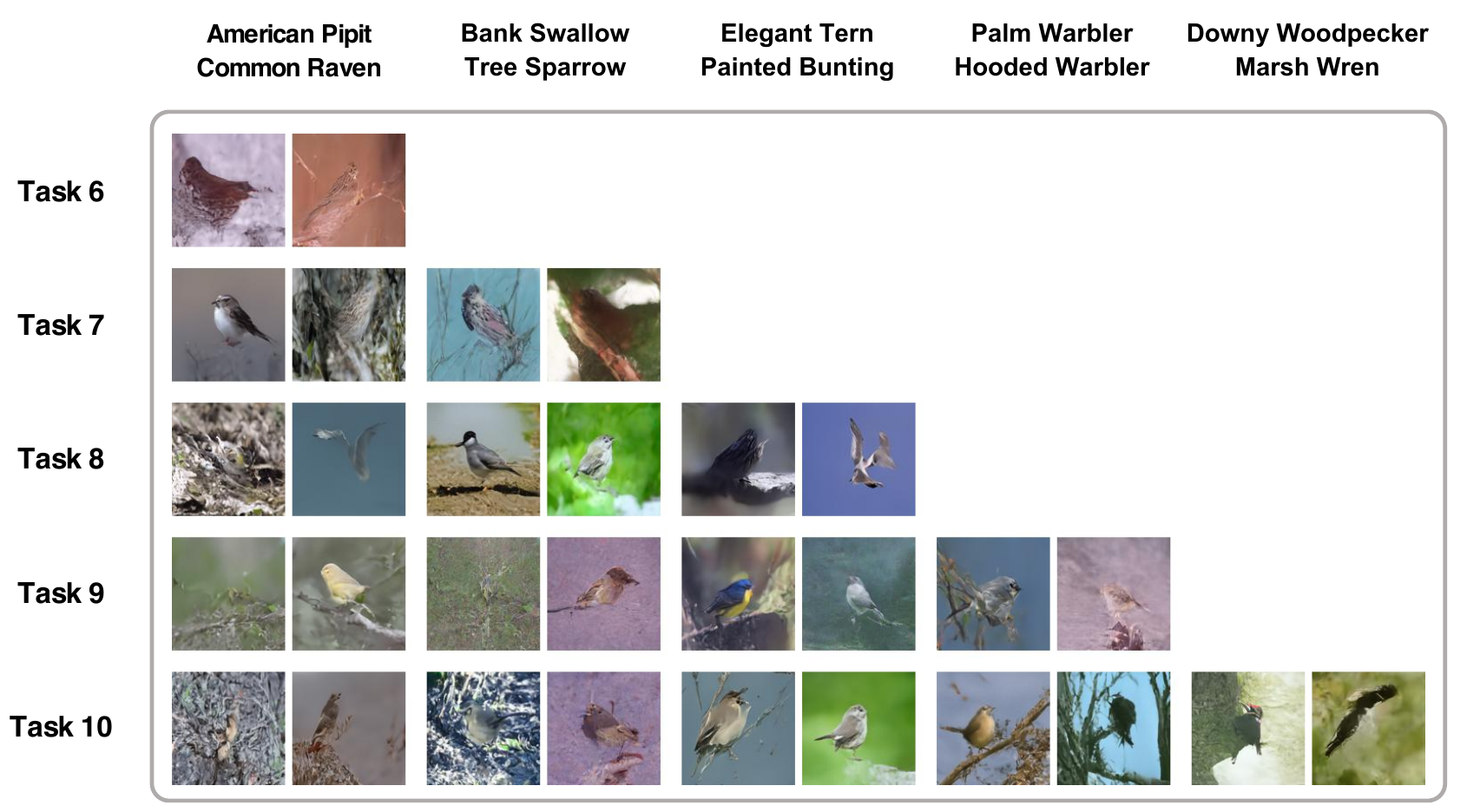}  
} 
\subfigure[Ensemble (DDIM)] {    
\includegraphics[width=0.4\columnwidth]{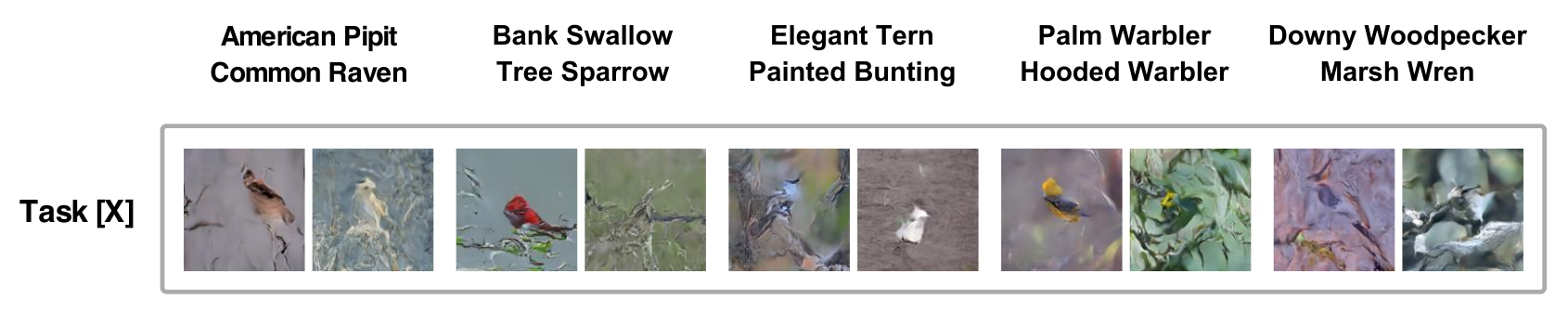}  
} 
\caption{Visualization results of label-conditioned CLoG on the CUB-Birds~\citep{wah2011caltech} dataset.}
\label{fig:vis_bird}
\end{figure}

\begin{figure}[H]
\centering
\subfigure[NCL (GAN)] {    
\includegraphics[width=0.4\columnwidth]{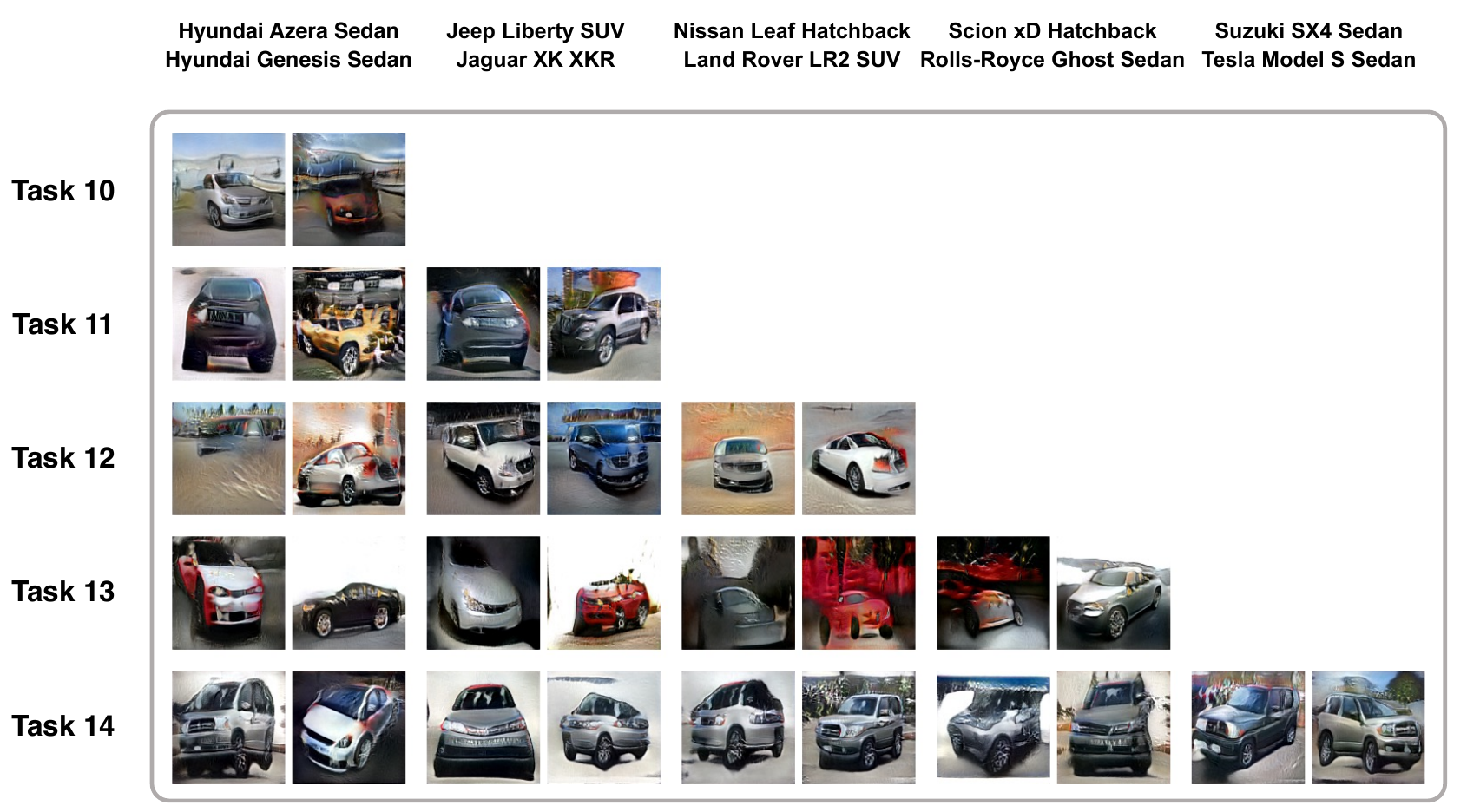}  
}    
\subfigure[Non-CL (GAN)] {    
\includegraphics[width=0.4\columnwidth]{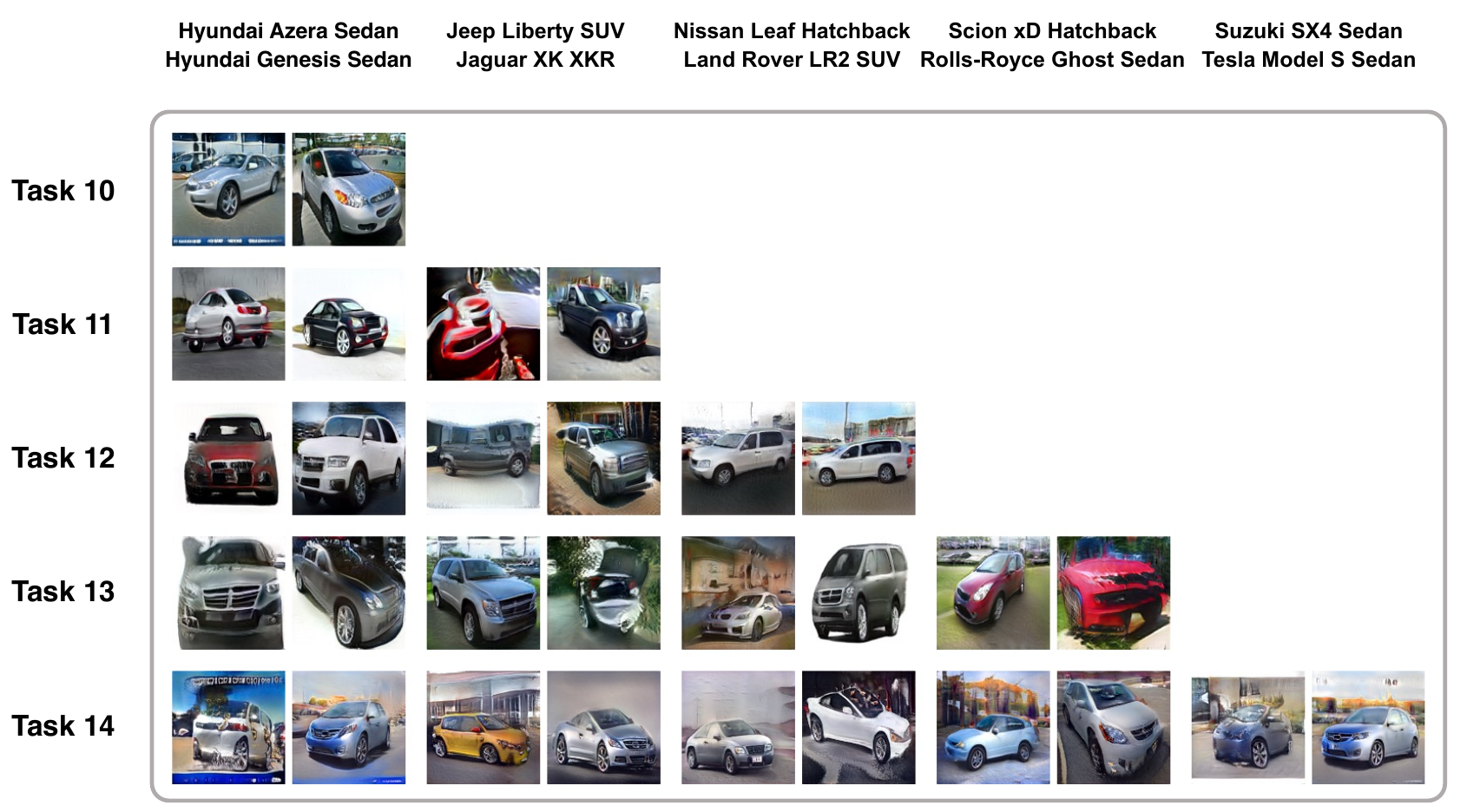}  
}    
\subfigure[ER (GAN)] {    
\includegraphics[width=0.4\columnwidth]{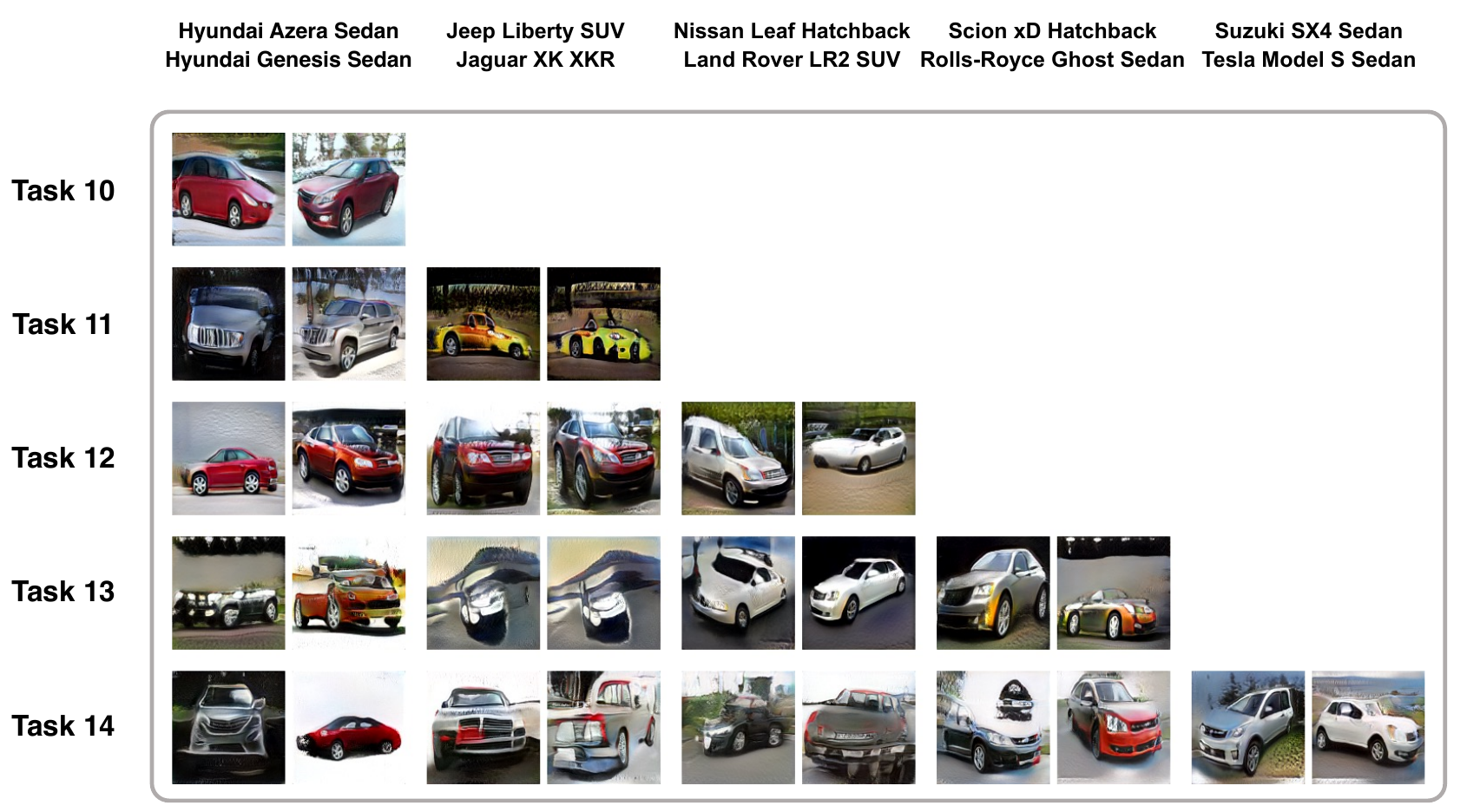}  
}    
\subfigure[EWC (GAN)] {    
\includegraphics[width=0.4\columnwidth]{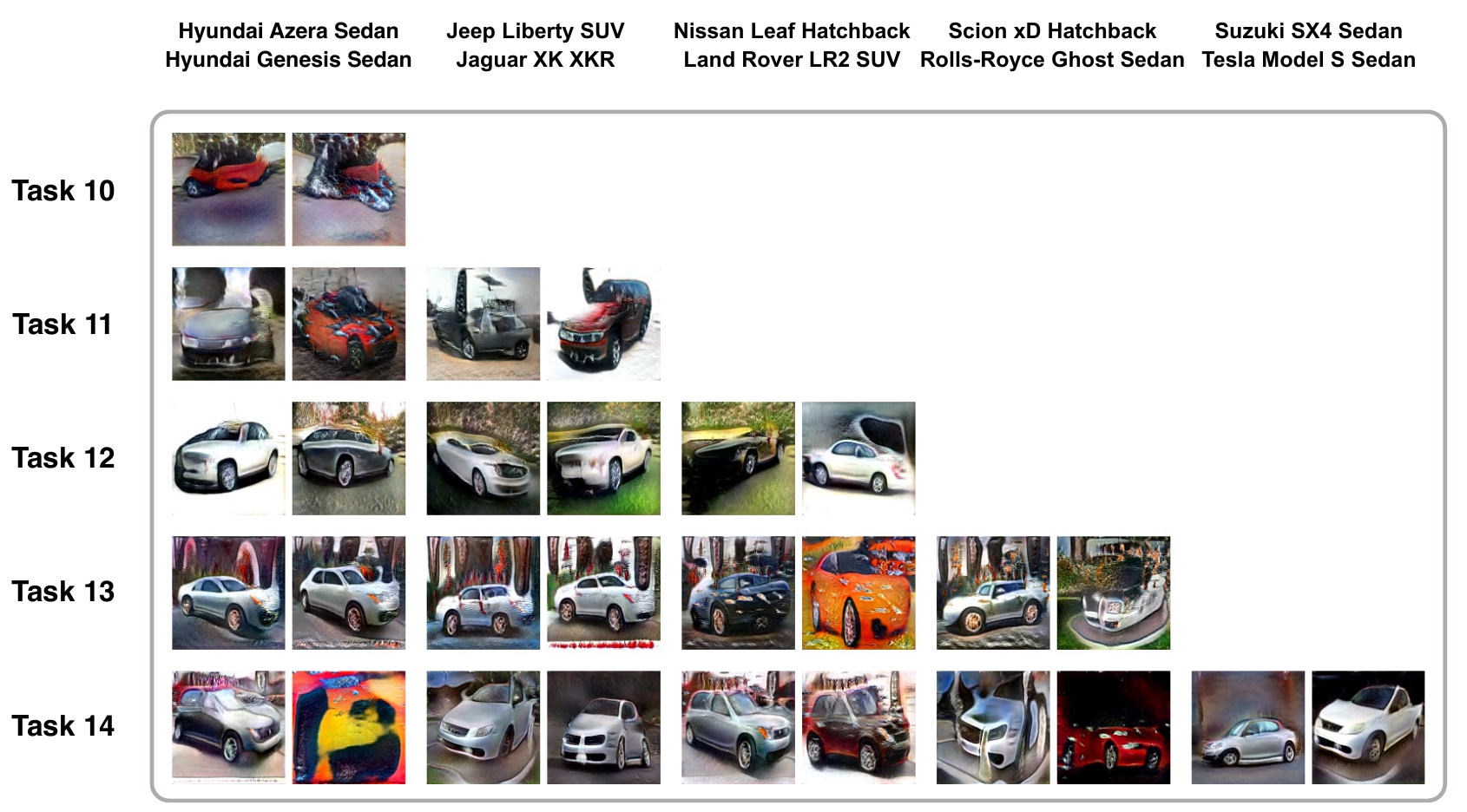}  
} 
\subfigure[Ensemble (GAN)] {    
\includegraphics[width=0.4\columnwidth]{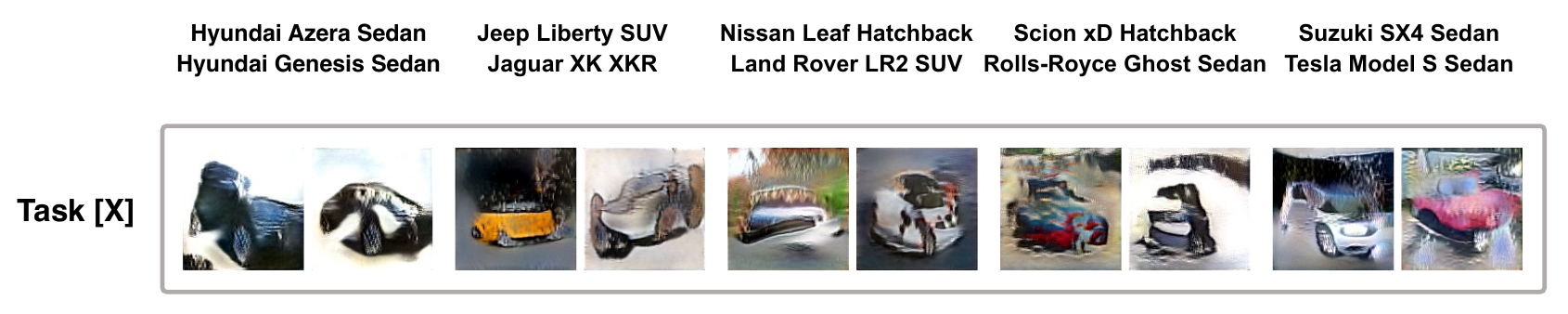}  
} 
\\
\subfigure[NCL (DDIM)] {    
\includegraphics[width=0.4\columnwidth]{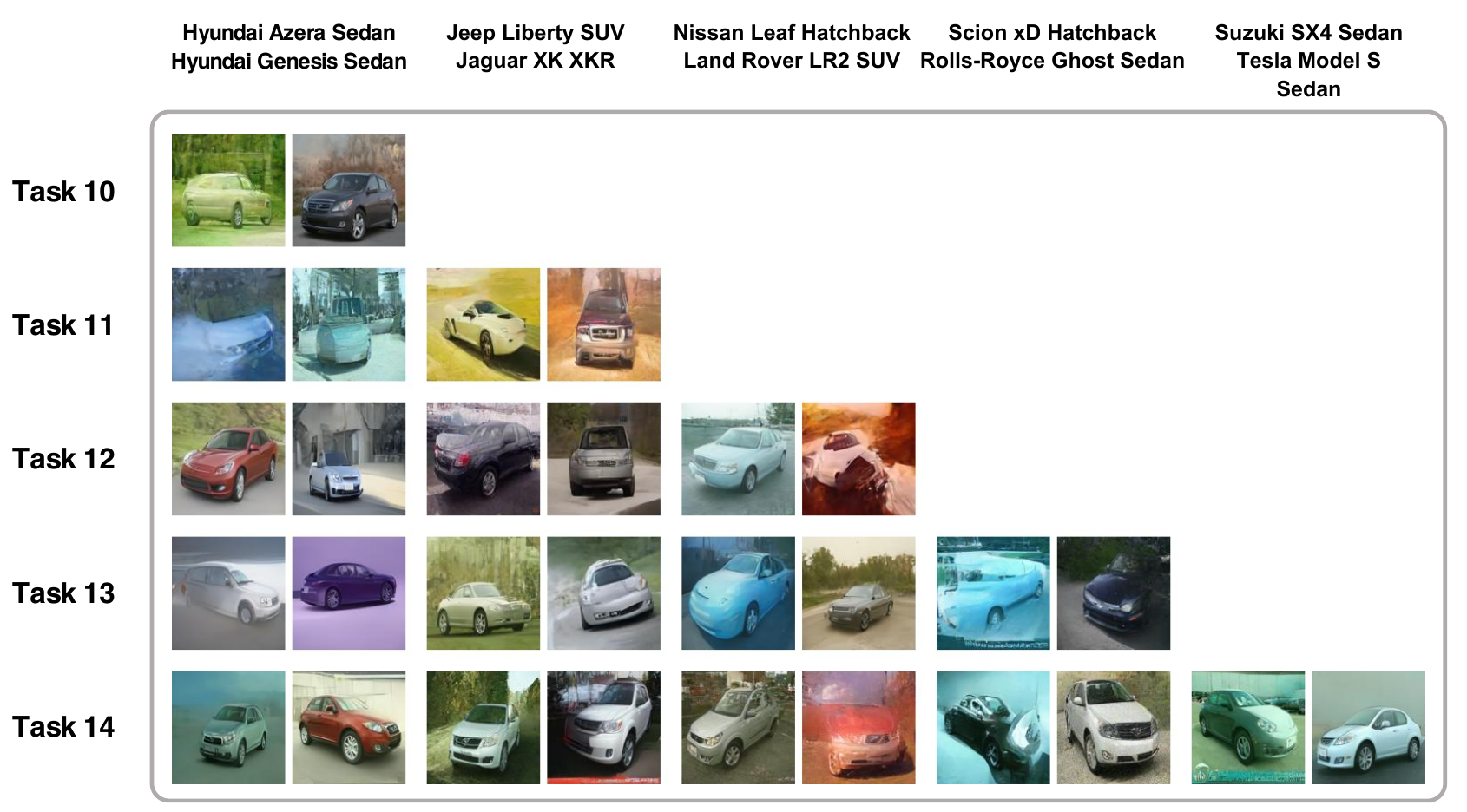}  
}    
\subfigure[Non-CL (DDIM)] {    
\includegraphics[width=0.4\columnwidth]{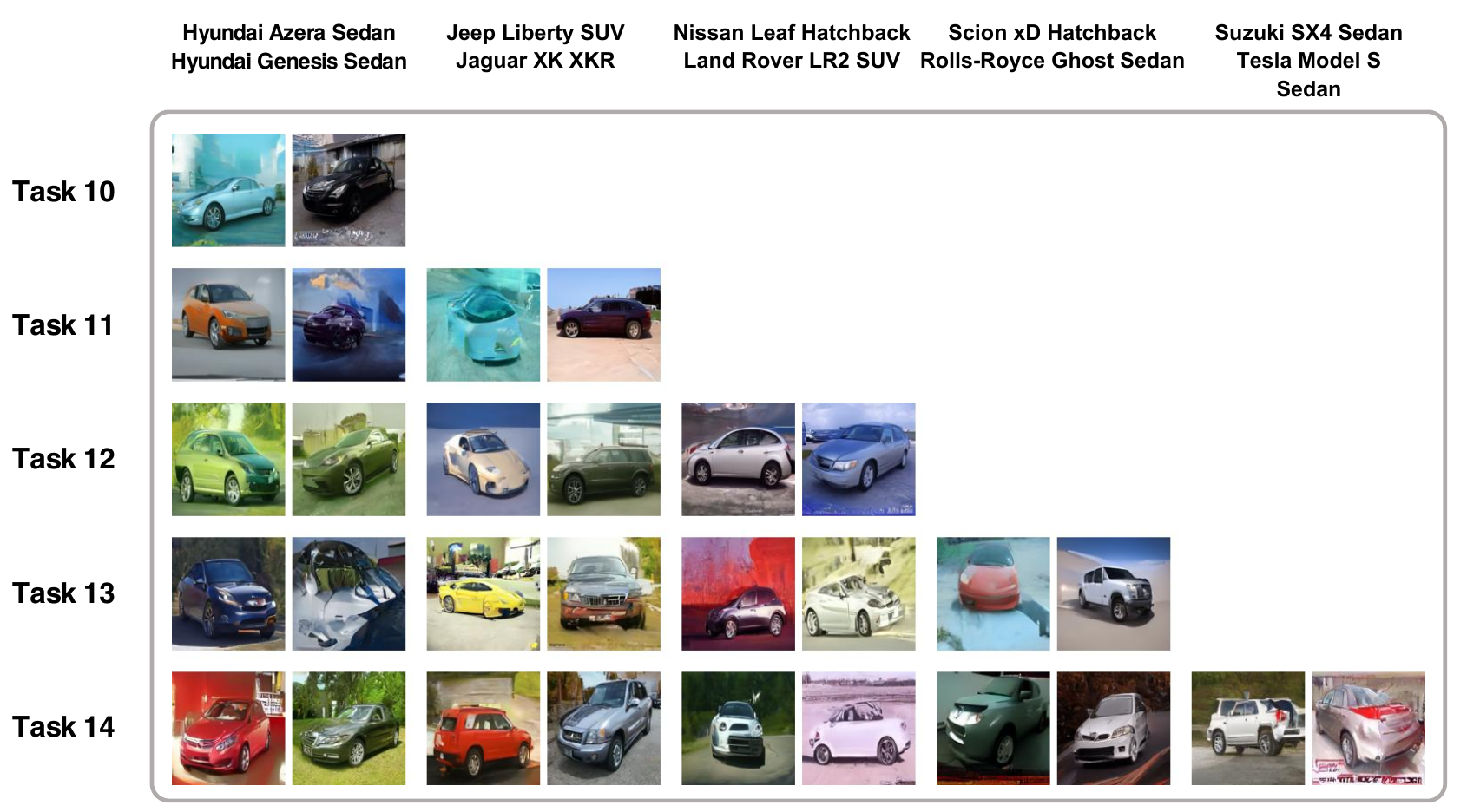}  
}    
\subfigure[ER (DDIM)] {    
\includegraphics[width=0.4\columnwidth]{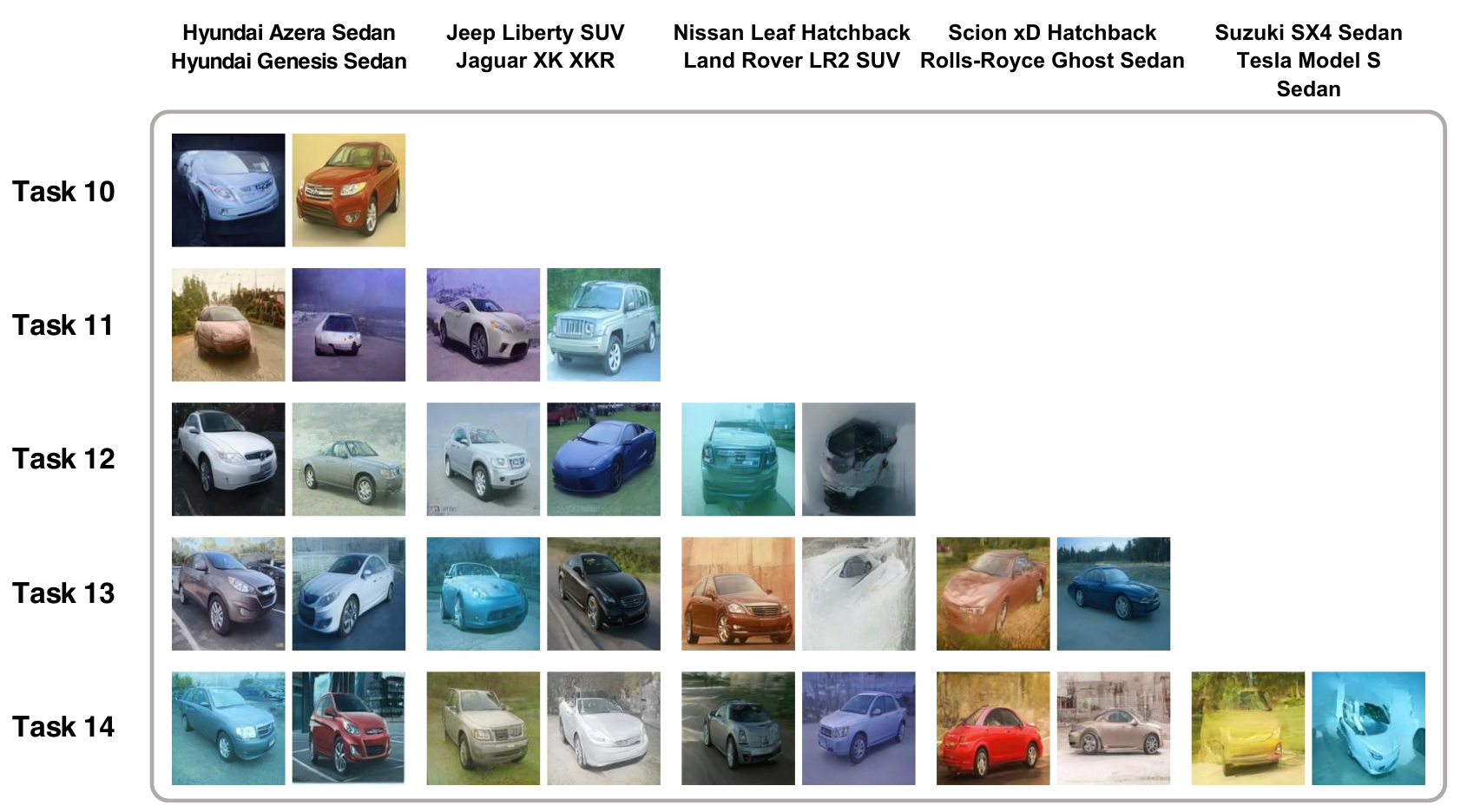}  
}    
\subfigure[EWC (DDIM)] {    
\includegraphics[width=0.4\columnwidth]{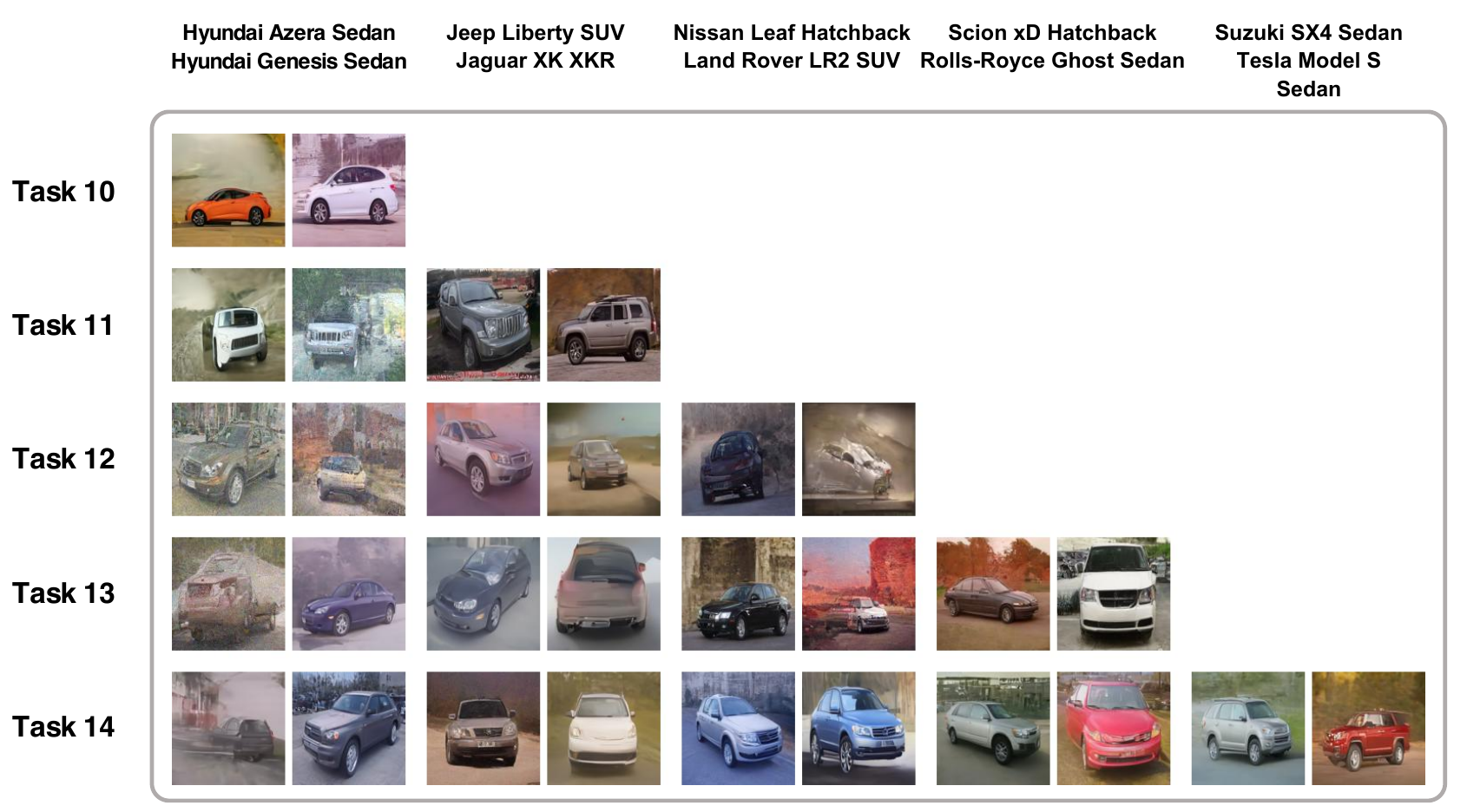}  
} 
\subfigure[Ensemble (DDIM)] {    
\includegraphics[width=0.4\columnwidth]{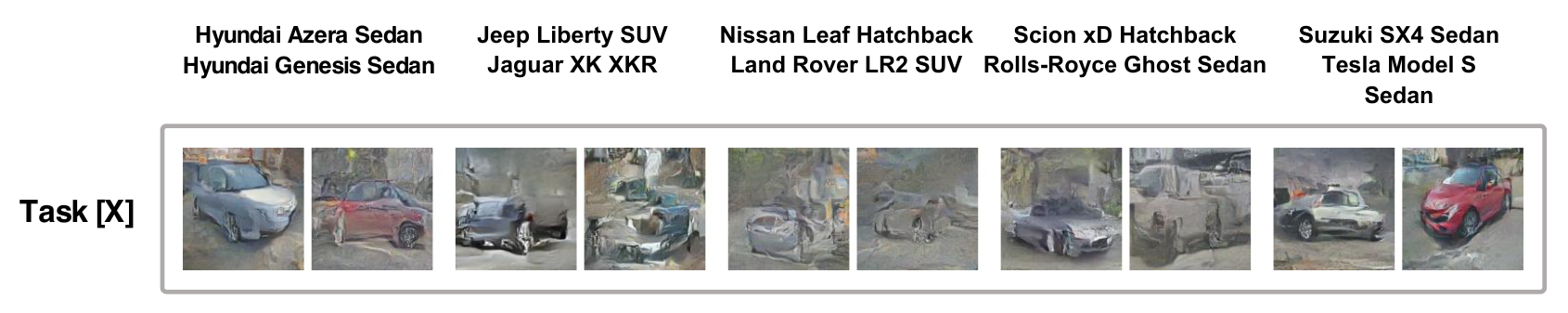}  
} 
\caption{Visualization results of label-conditioned CLoG on the Stanford-Cars~\citep{krause20133d} dataset.}
\label{fig:vis_car}
\end{figure}

\begin{figure}[H]
\label{fig:imagenet}
\centering
\subfigure[NCL (DDIM)] {    
\includegraphics[width=0.4\columnwidth]{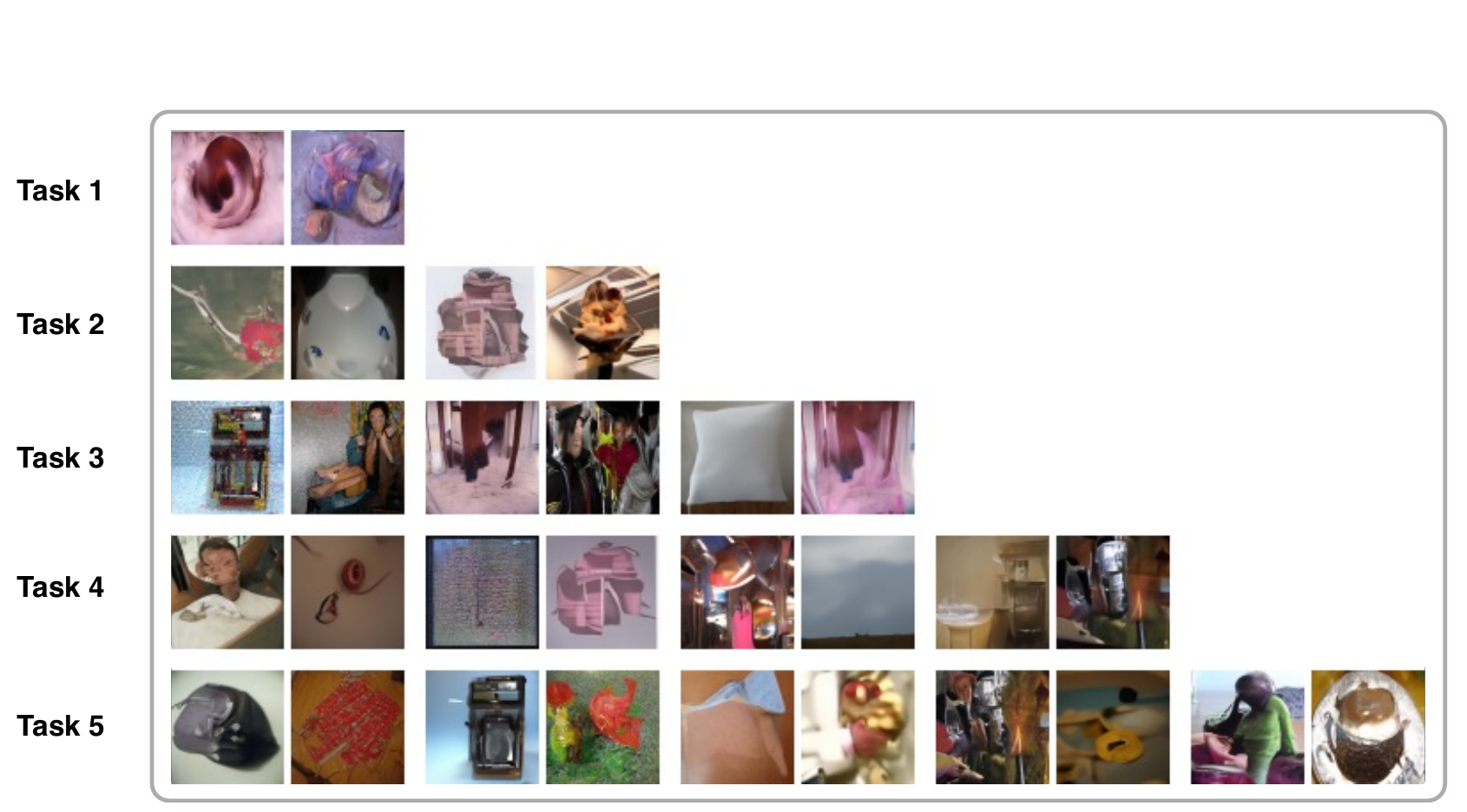}  
}    
\subfigure[Non-CL (DDIM)] {    
\includegraphics[width=0.4\columnwidth]{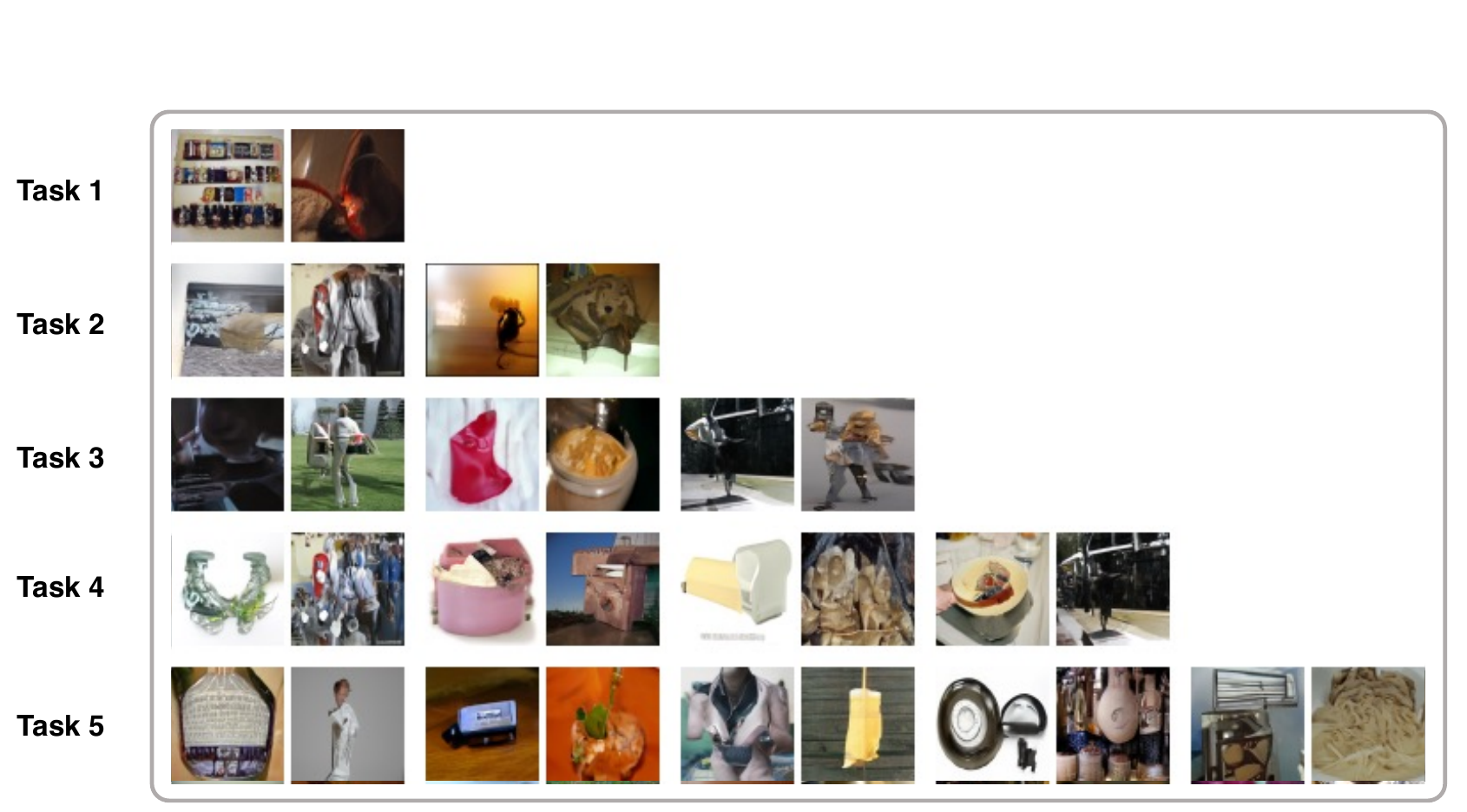}  
}    
\subfigure[KD (DDIM)] {    
\includegraphics[width=0.4\columnwidth]{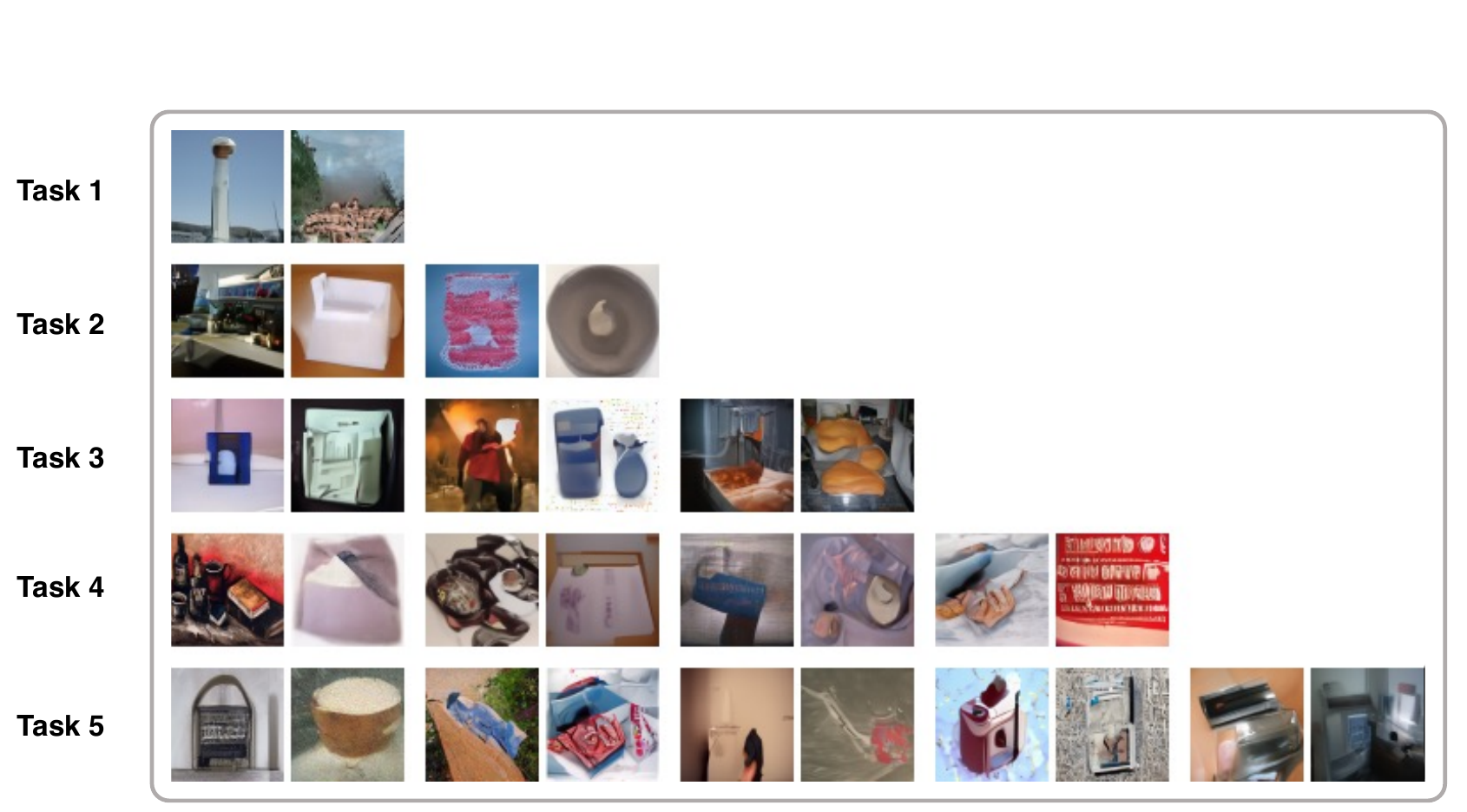}  
}    
\subfigure[EWC (DDIM)] {    
\includegraphics[width=0.4\columnwidth]{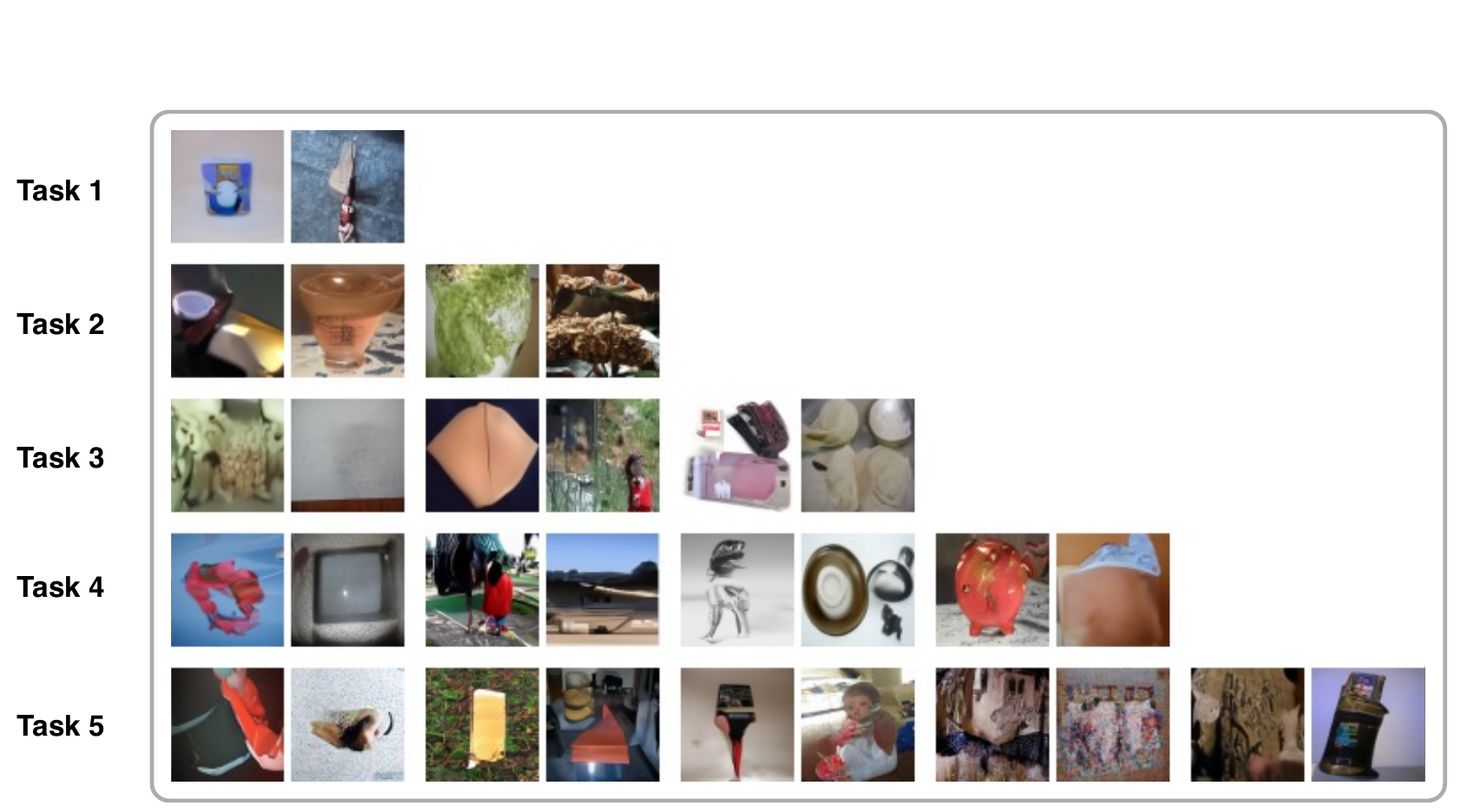}  
} 
\subfigure[Ensemble (DDIM)] {    
\includegraphics[width=0.4\columnwidth]{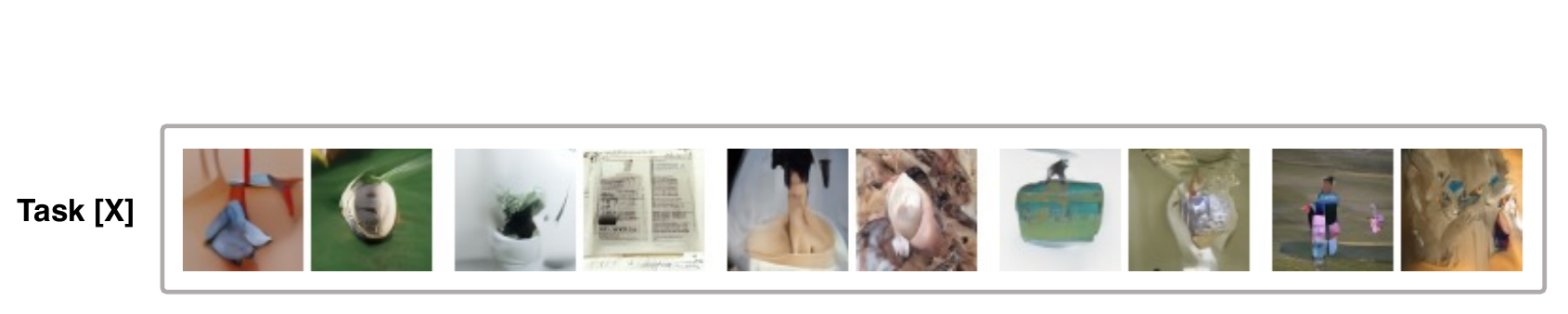}  
} 
\caption{Visualization results of label-conditioned CLoG on the ImageNet-1k~\citep{imagenet15russakovsky} dataset.} 
\label{fig:vis_imagenet}
\end{figure}

We present visualization results of the generated images in this section. Figures~\ref{fig:vis_mnist}, \ref{fig:vis_fm}, \ref{fig:vis_cifar}, \ref{fig:vis_flower}, \ref{fig:vis_bird}, \ref{fig:vis_car}, and \ref{fig:vis_imagenet} showcase synthesized images in the label-conditioned CLoG across the seven datasets in our benchmark.  We visualize the synthesized images from the models over the last five tasks using the first class order, with images selected randomly to avoid cherry-picking.  We select five representative methods to showcase the results: NCL, Non-CL, ER~(replay-based), EWC~(regularization-based), and Ensemble~(parameter-isolation based).

As shown in the figure, a naive way of CL without additional techniques leads to severe forgetting. The regularization-based methods can preserve knowledge of previous tasks to some extent, but the results are still far from satisfying, especially as the number of learning tasks increases. Replay-based methods significantly mitigate the challenges of catastrophical forgetting. However, our empirical studies suggest that they are prone to mode collapse when training GANs, mainly due to the limited size of the replay memory. This may reveal a novel challenge in CLoG compared to traditional classification-based CL. Furthermore, the ensembling method achieves superior performance on each task on the first three datasets, including MNIST, Fashion-MNIST and CIFAR-10. Nevertheless, it synthesizes images with relatively low quality on the other three datasets (see Fig.~\ref{fig:vis_flower}, \ref{fig:vis_bird}, \ref{fig:vis_car}). Take Oxford-Flowers as an example, the separate model trained on each task fail to capture the correct structures of flowers, in contrast to other CL methods. This verifies our analysis that knowledge transfer among different tasks contribute to performance boost on these datasets.

\begin{figure}[H]
\centering
\subfigure[NCL] {    
\includegraphics[width=0.4\columnwidth]{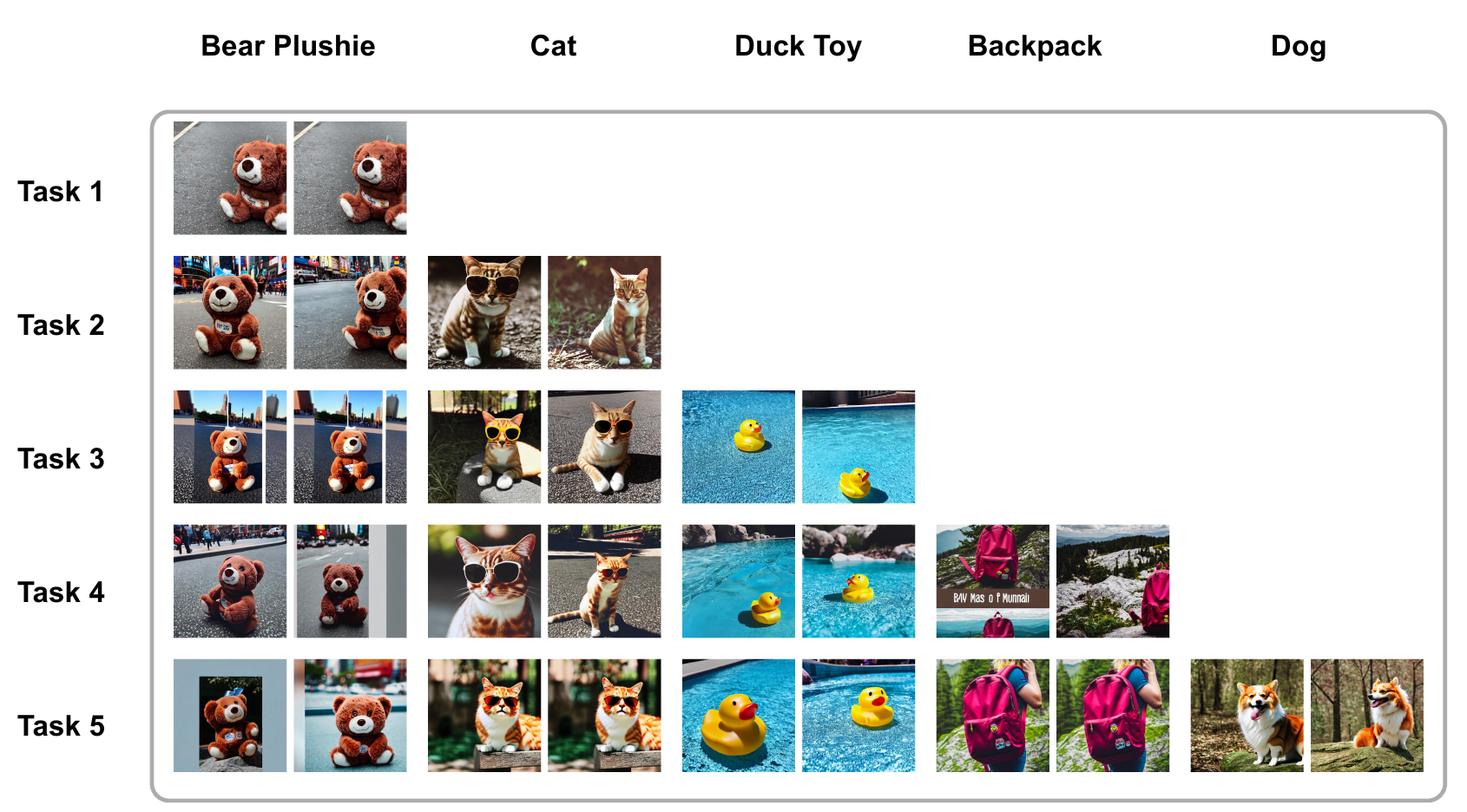}  
}    
\subfigure[Non-CL] {    
\includegraphics[width=0.4\columnwidth]{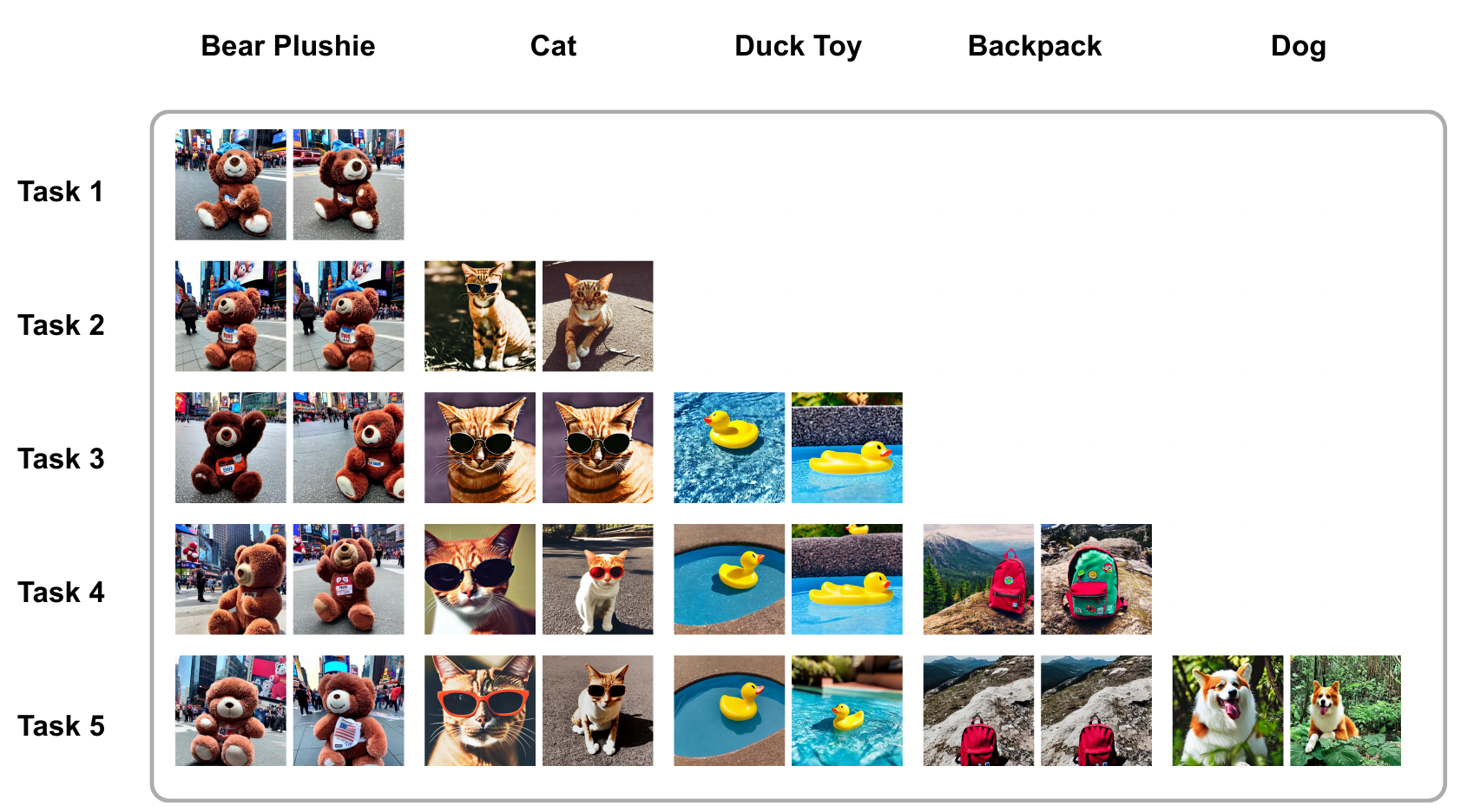}  
}    
\subfigure[KD] {    
\includegraphics[width=0.4\columnwidth]{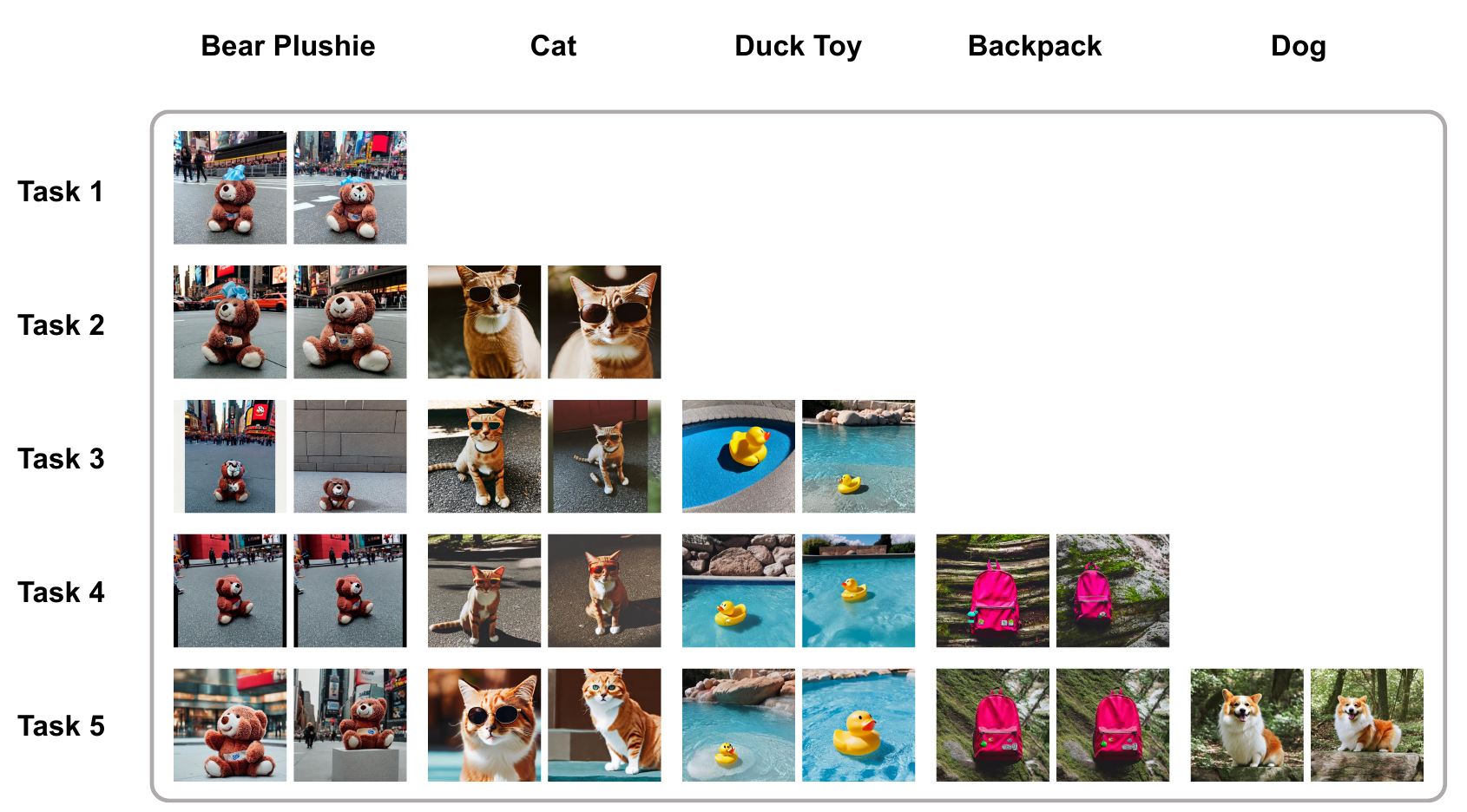}  
}    
\subfigure[EWC] {    
\includegraphics[width=0.4\columnwidth]{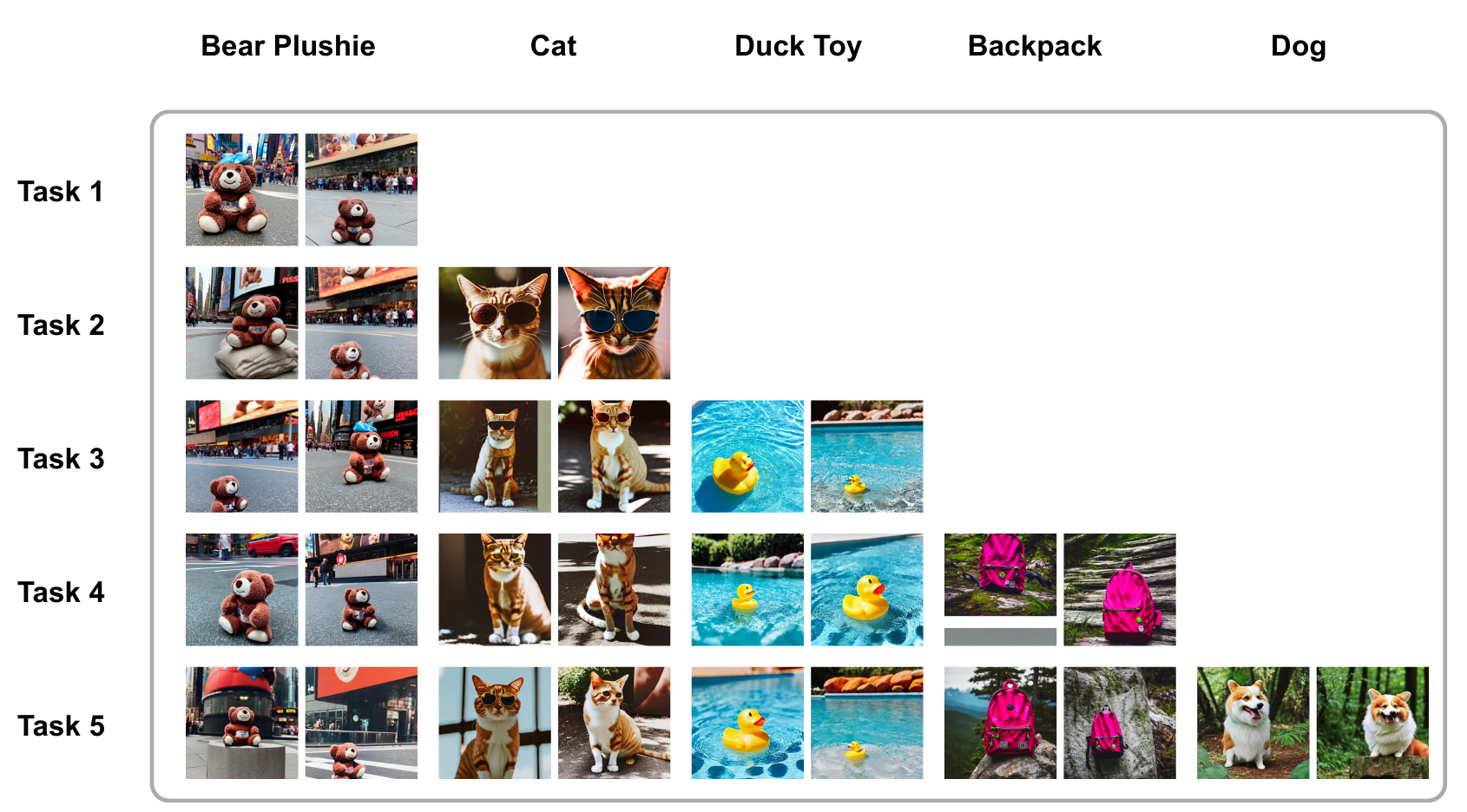}  
} 
\subfigure[Ensemble] {    
\includegraphics[width=0.4\columnwidth]{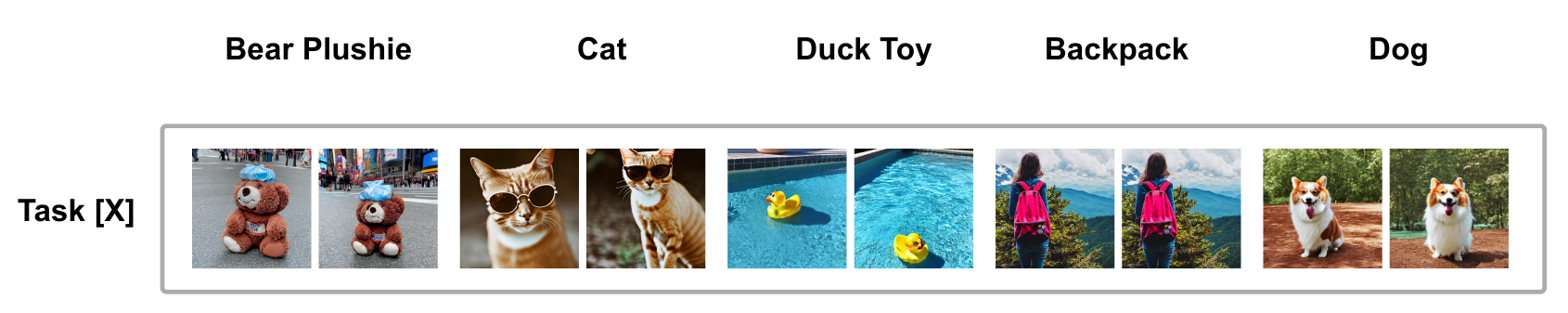}  
} 
\caption{Visualization results of concept-conditioned CLoG on the Custom Objects~\citep{sun2024create} dataset utilizing DreamBooth~\citep{ruiz2023dreambooth}.}
\label{fig:vis_dreambooth}
\end{figure}

Figures~\ref{fig:vis_dreambooth}, and \ref{fig:vis_custom_diffusion}, showcase synthesized images in the concep-conditional CLoG with Custom Diffusion~\citep{kumari2023multi} and DreamBooth~\citep{ruiz2023dreambooth} in our benchmark.  We visualize the synthesized images from the models over the five tasks using the third class order, with images selected randomly to avoid cherry-picking.  We select five representative methods to showcase the results: NCL, Non-CL, KD~(regularization-based), EWC~(regularization-based), and Ensemble~(parameter-isolation-based).

A naive method of continual learning without additional techniques produces relatively high-quality images, particularly when using the DreamBooth method with more training parameters. Regularization-based methods can preserve knowledge from previous tasks to some extent, but the results are still unsatisfactory, especially as the number of learning tasks increases. For example, in Figures \ref{fig:vis_custom_diffusion}, the EWC method shows that by the fifth task, the Custom Diffusion has almost entirely forgotten the color of the bear plushie and the blue hat decoration. Furthermore, Ensemble method achieves superior performance with both Custom Diffusion and DreamBooth. 

\newpage

\newpage

{
\begin{figure}[H]
\centering
\subfigure[NCL] {    
\includegraphics[width=0.4\columnwidth]{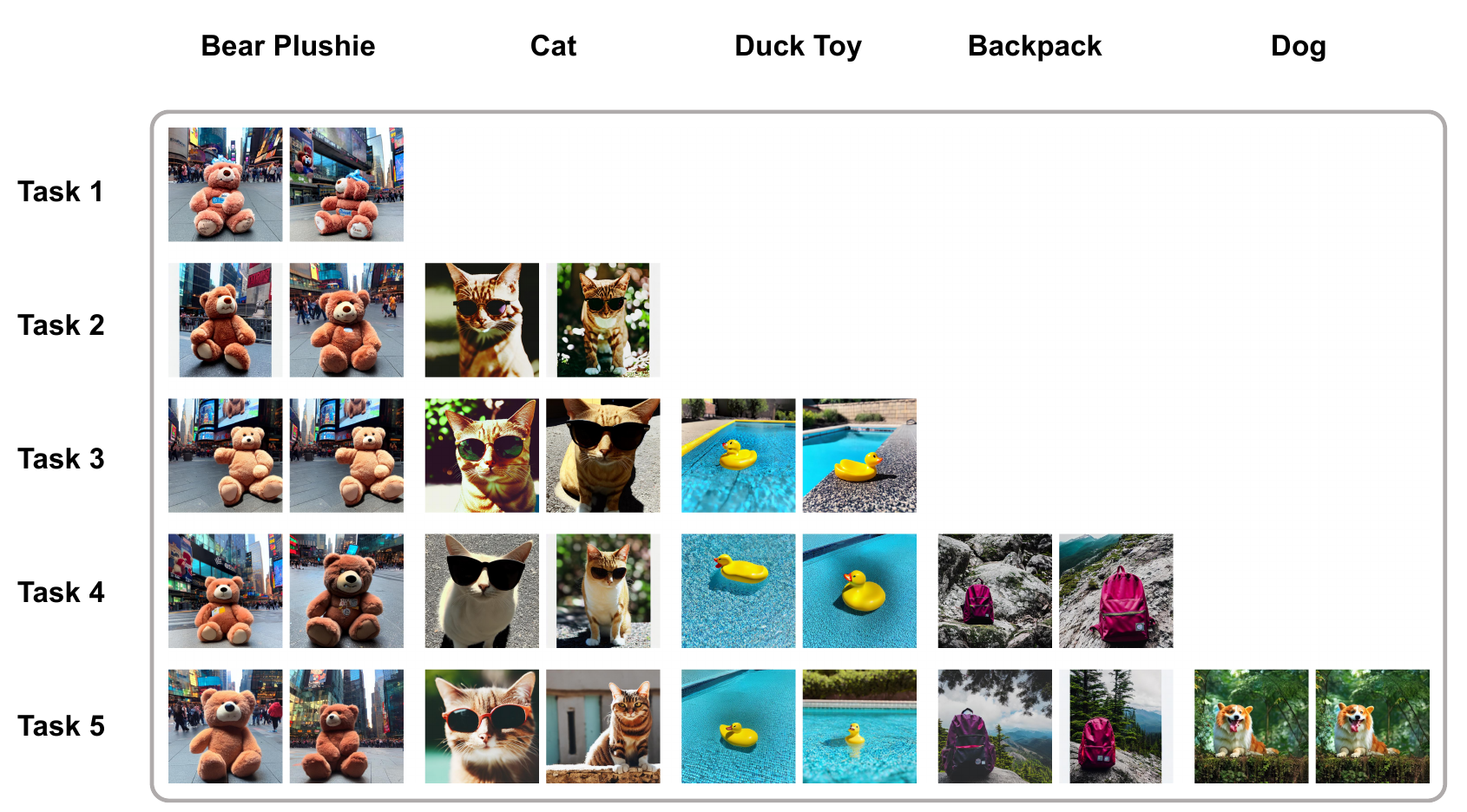}  
}    
\subfigure[Non-CL] {    
\includegraphics[width=0.4\columnwidth]{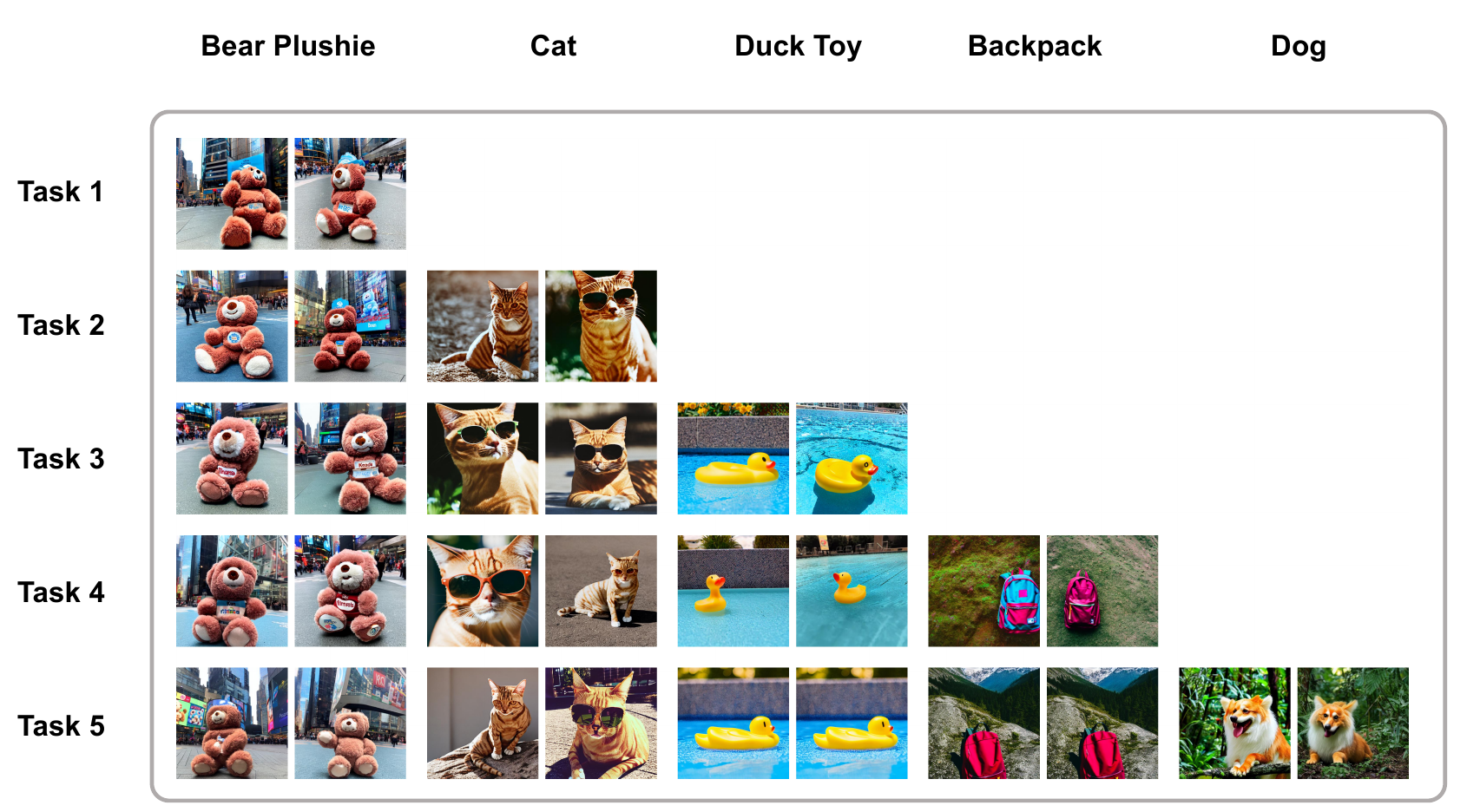}  
}    
\subfigure[KD] {    
\includegraphics[width=0.4\columnwidth]{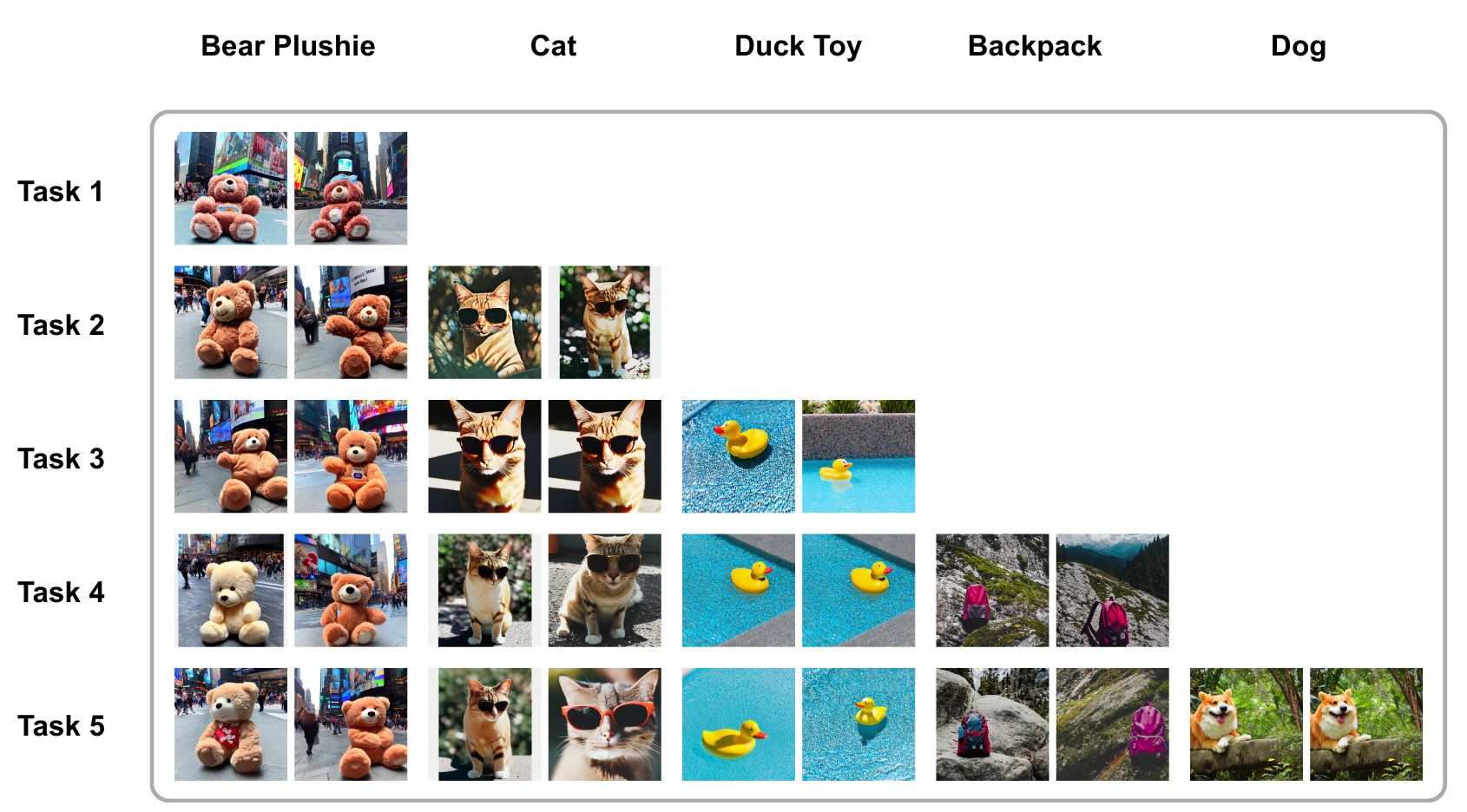}  
}    
\subfigure[EWC] {    
\includegraphics[width=0.4\columnwidth]{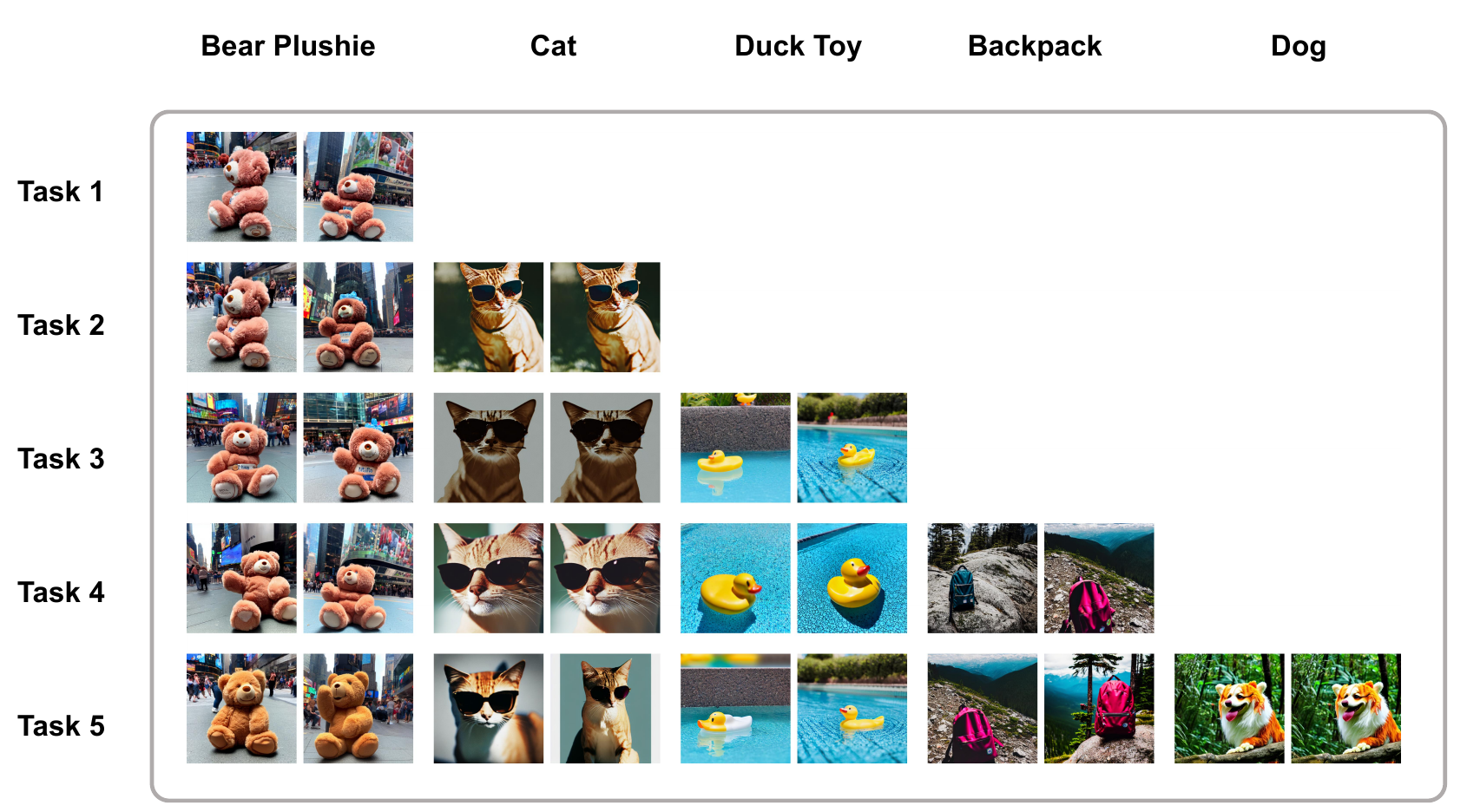}  
} 
\subfigure[Ensemble] {    
\includegraphics[width=0.4\columnwidth]{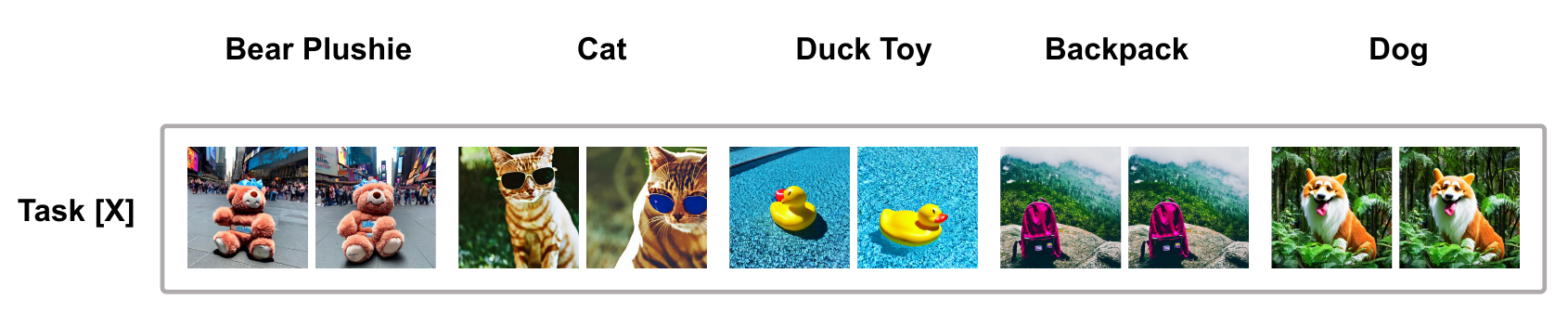}  
} 
\caption{Visualization results of concept-conditioned CLoG on the Custom Objects~\citep{sun2024create} dataset utilizing Custom Diffusion~\citep{kumari2023multi}.}
\label{fig:vis_custom_diffusion}
\end{figure}
}

\newpage

\subsection{Comprehensive AIQ results for each task}
\label{app:comprehensive_aiq}

\begin{figure}[H]
\centering
\subfigure[MNIST] {    
\includegraphics[width=0.4\columnwidth]{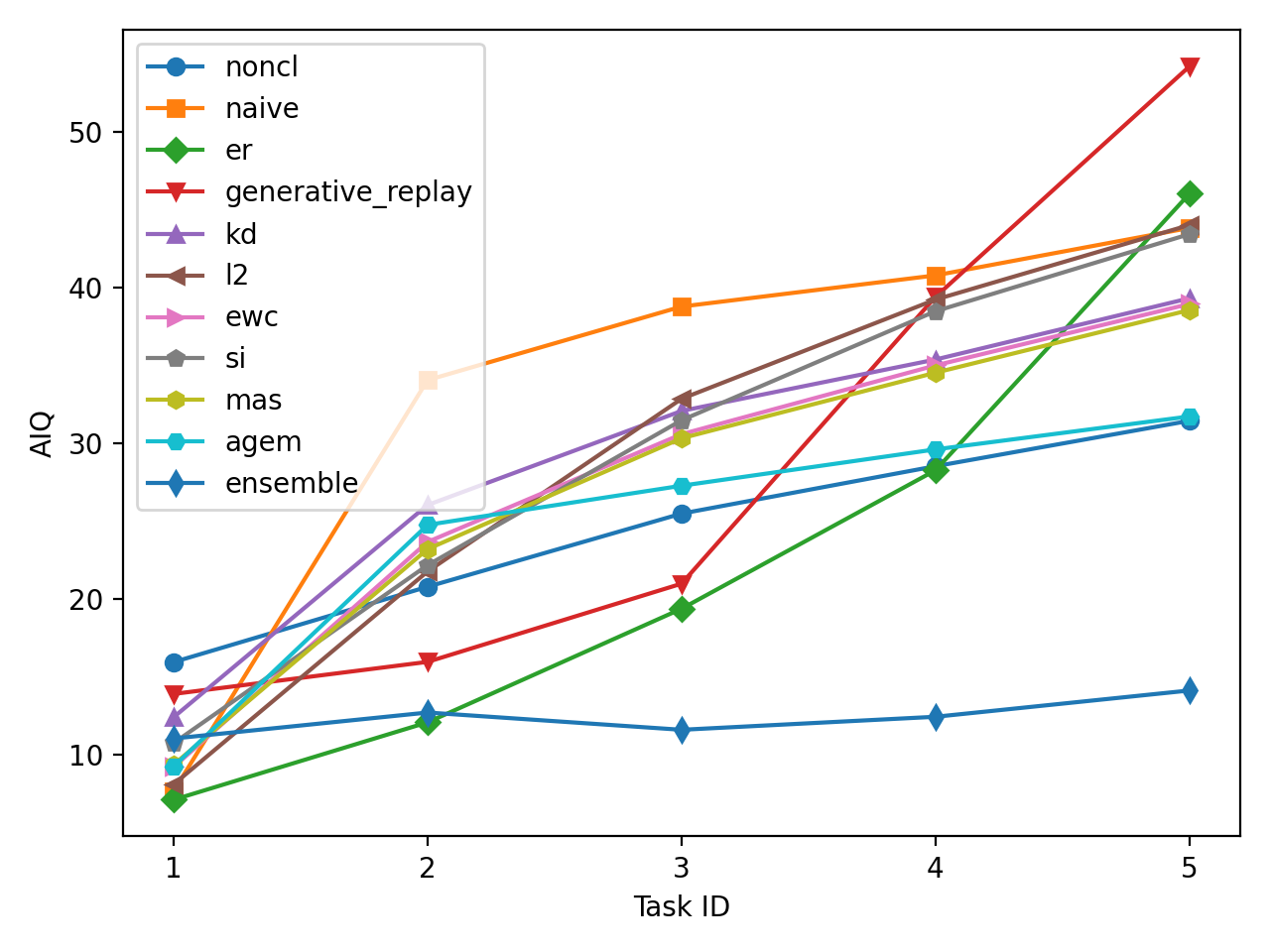}  
}    
\subfigure[Fashion-MNIST] {    
\includegraphics[width=0.4\columnwidth]{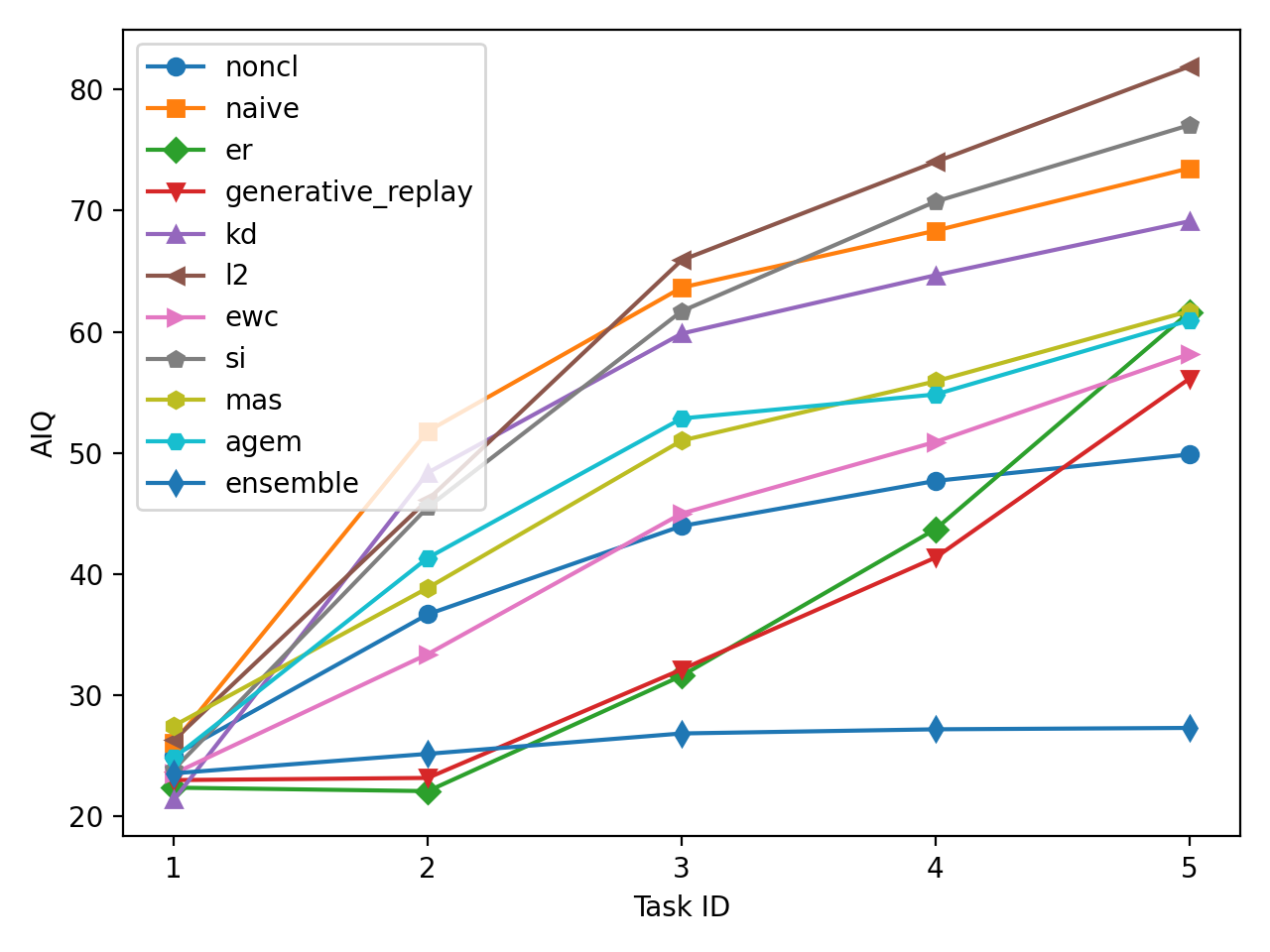}  
}    
\subfigure[CIFAR-10] {    
\includegraphics[width=0.4\columnwidth]{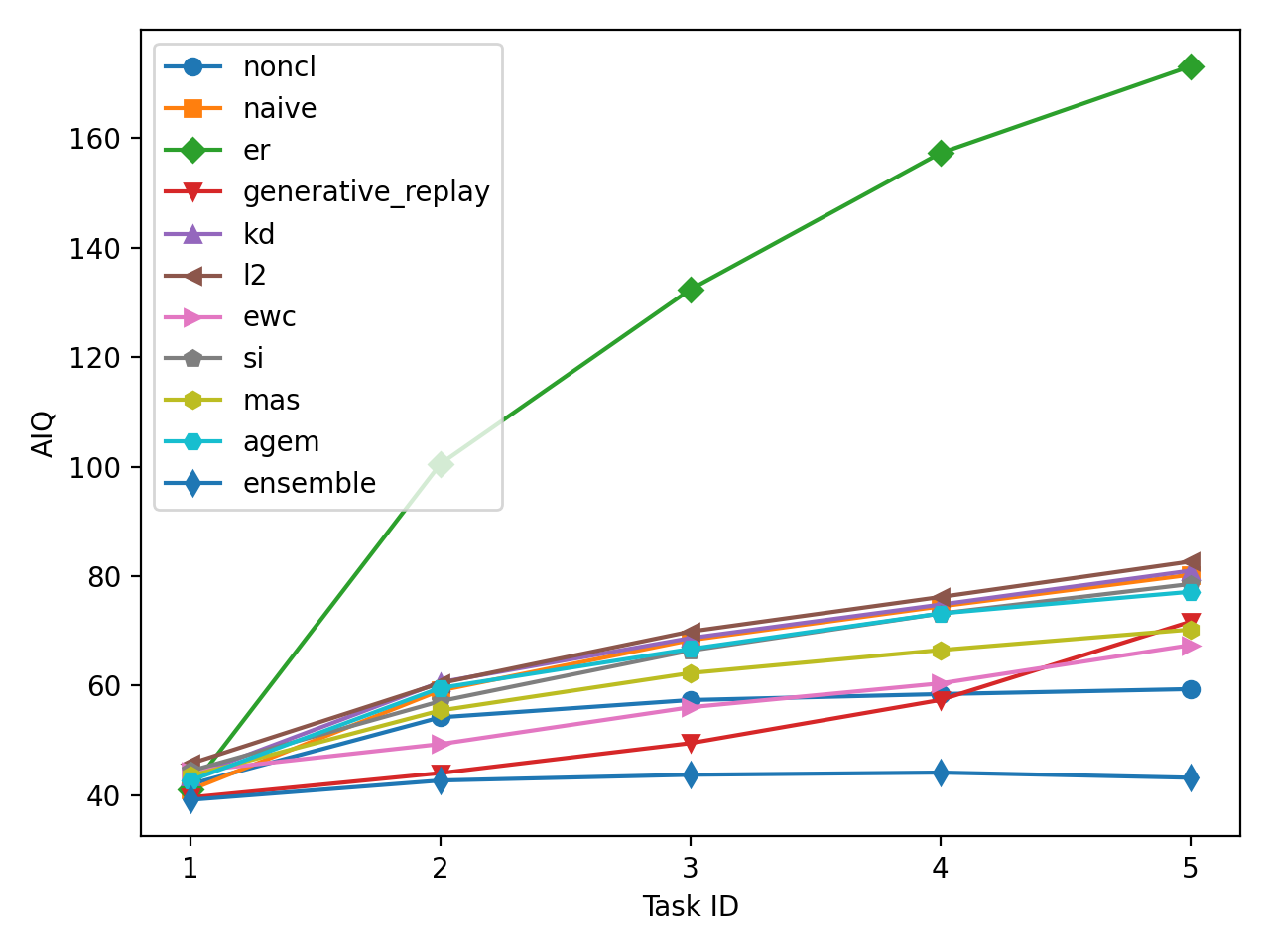}  
}    
\subfigure[Oxford-Flower] {    
\includegraphics[width=0.4\columnwidth]{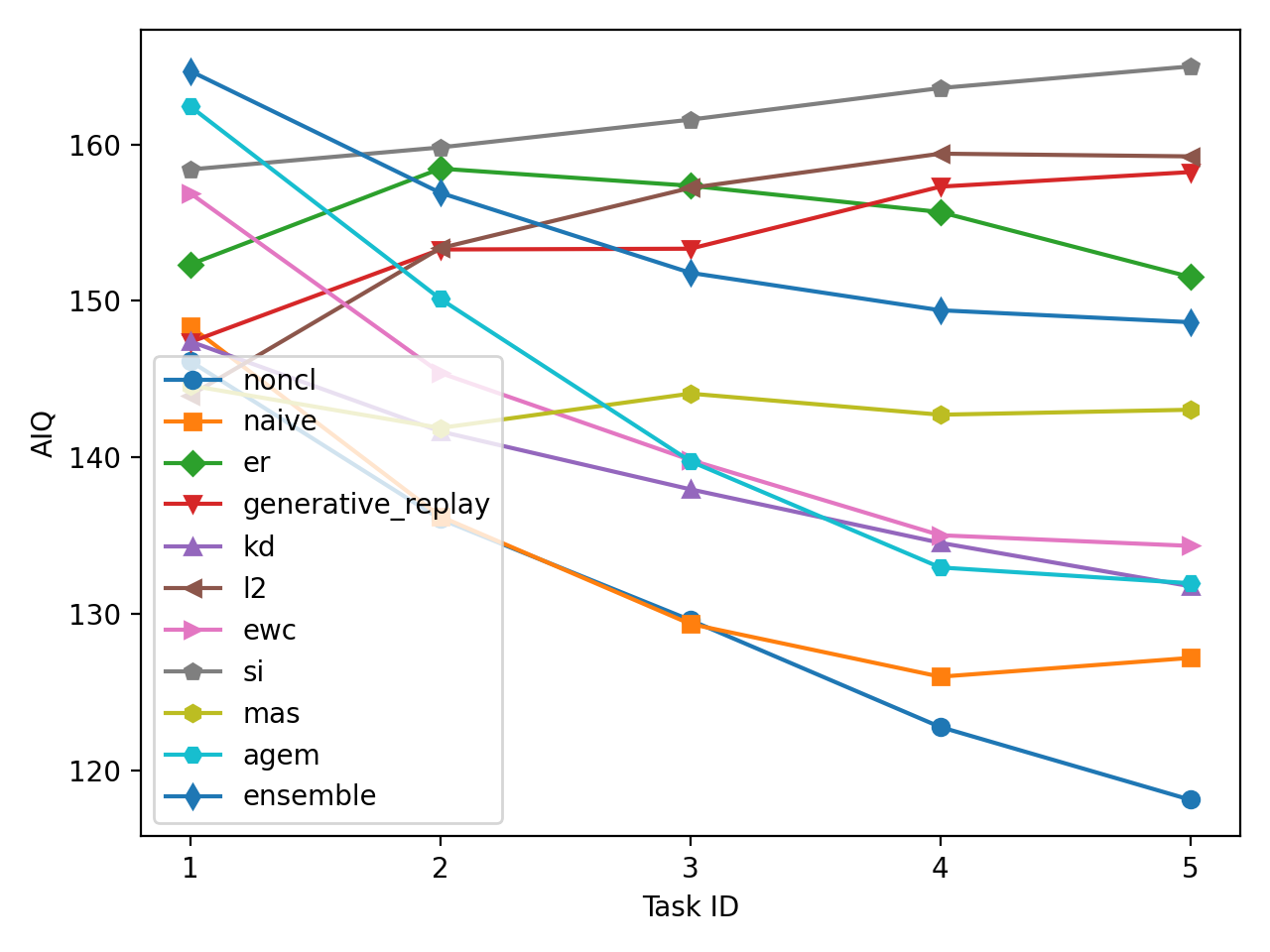}  
} 
\subfigure[CUB-Birds] {    
\includegraphics[width=0.4\columnwidth]{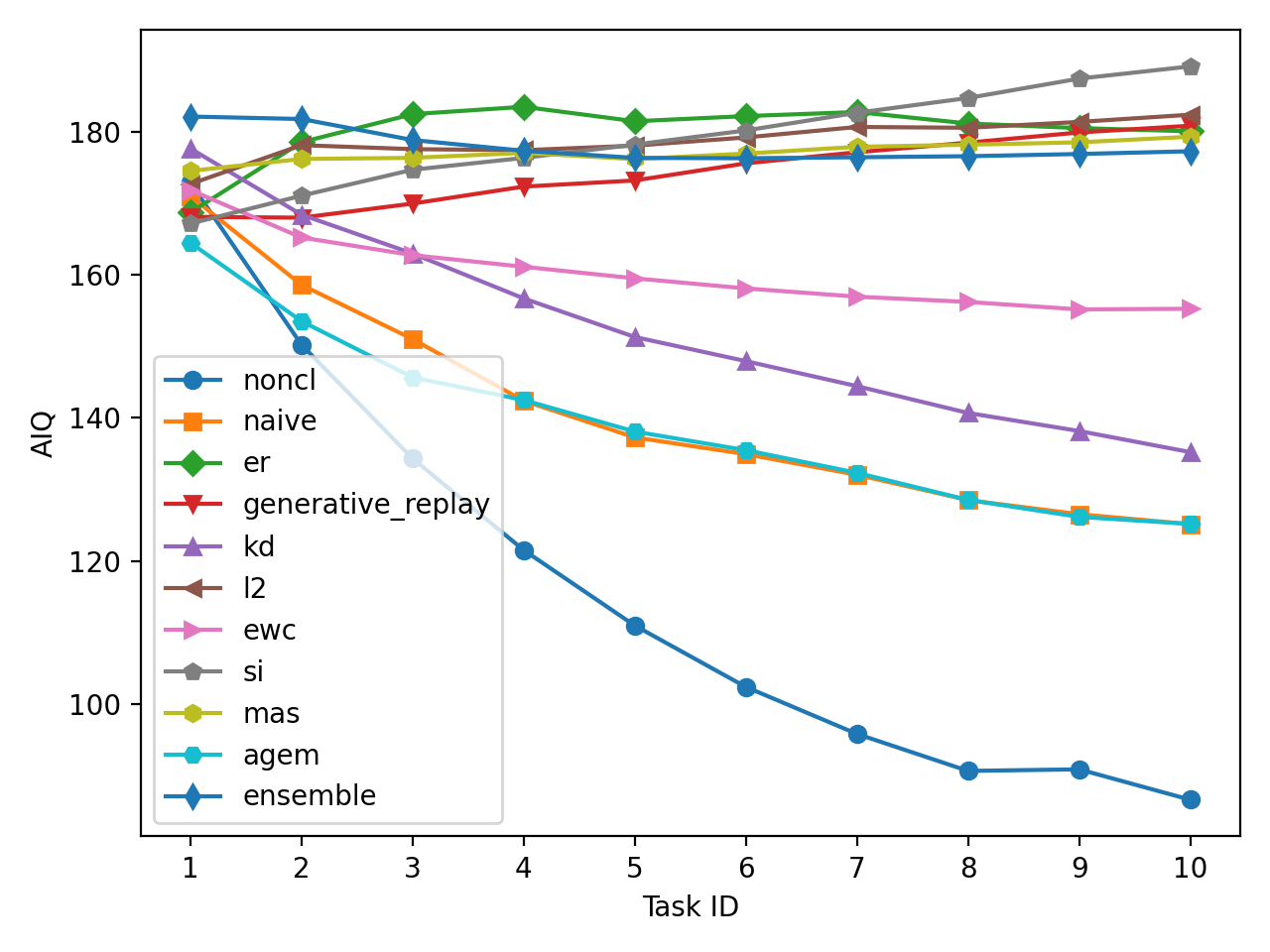}
}
\subfigure[Stanford-Cars] {    
\includegraphics[width=0.4\columnwidth]{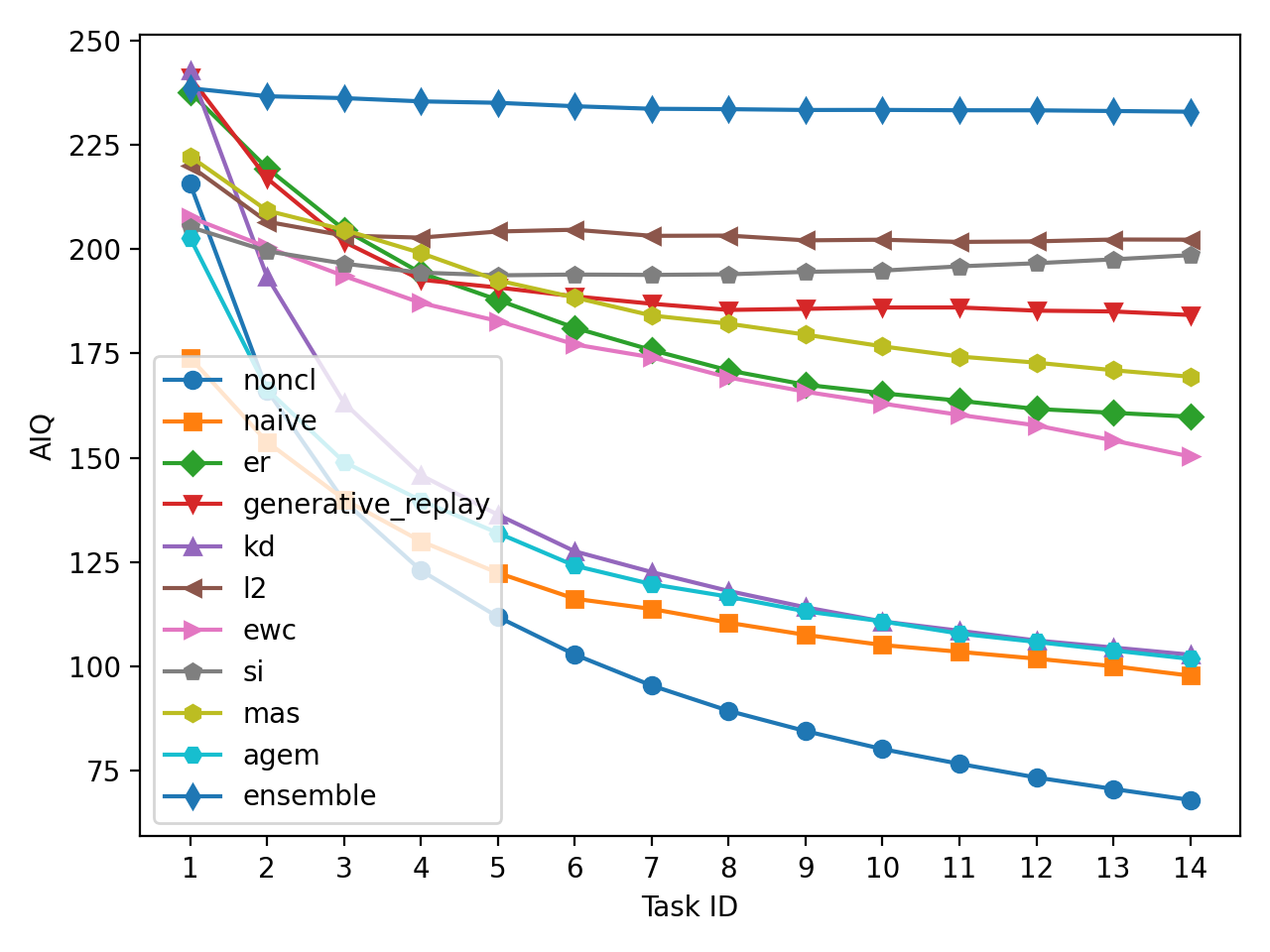}  
} 
\caption{The evolving performance curve of AIQ across various tasks on label-conditioned CLoG benchmarks. Here GANs are employed as the generator backbone.}
\label{fig:aiq_curve_gan}
\end{figure}

\begin{figure}[H]
\centering
\subfigure[MNIST (DDIM)] {    
\includegraphics[width=0.45\columnwidth]{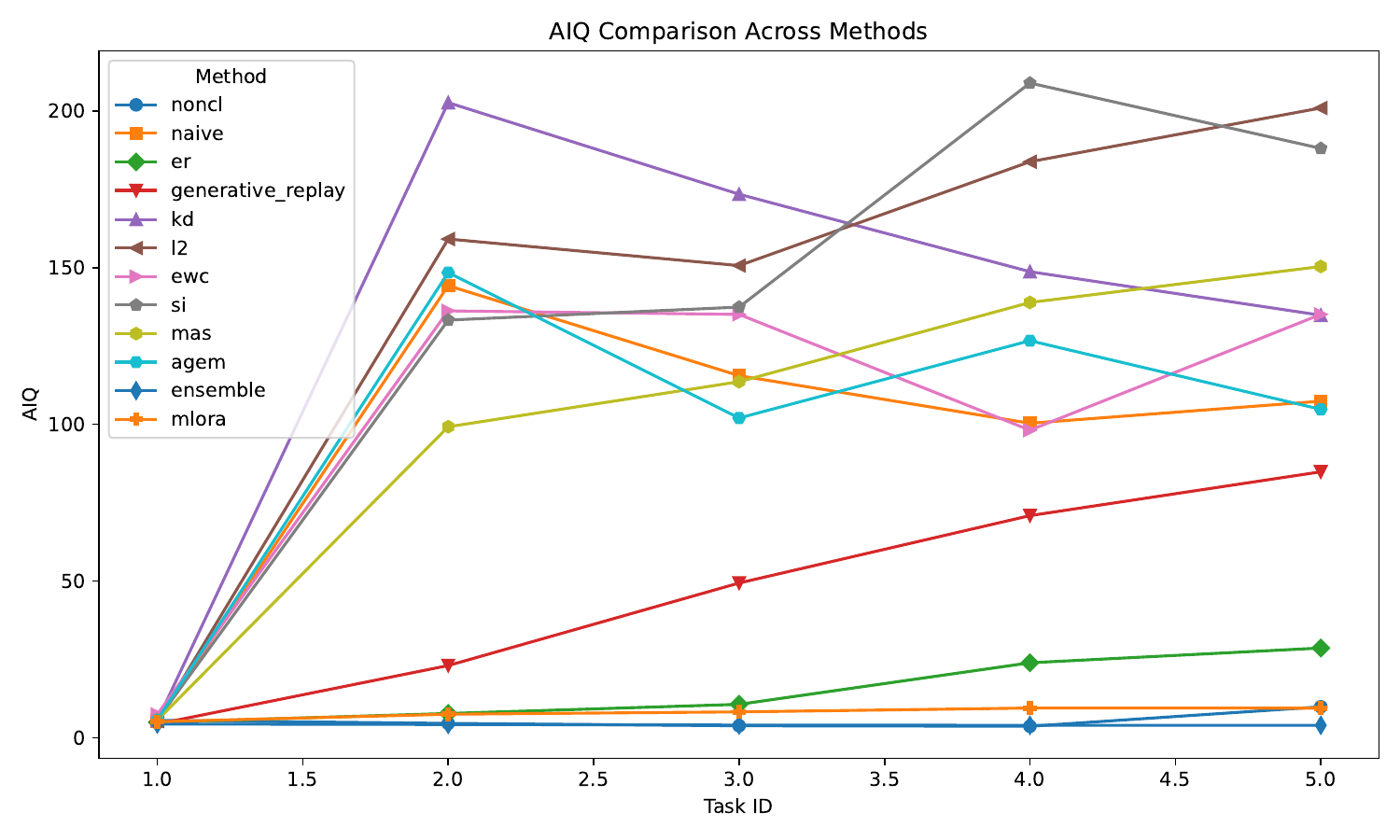}  
}    
\subfigure[Fashion-MNIST (DDIM)] {    
\includegraphics[width=0.45\columnwidth]{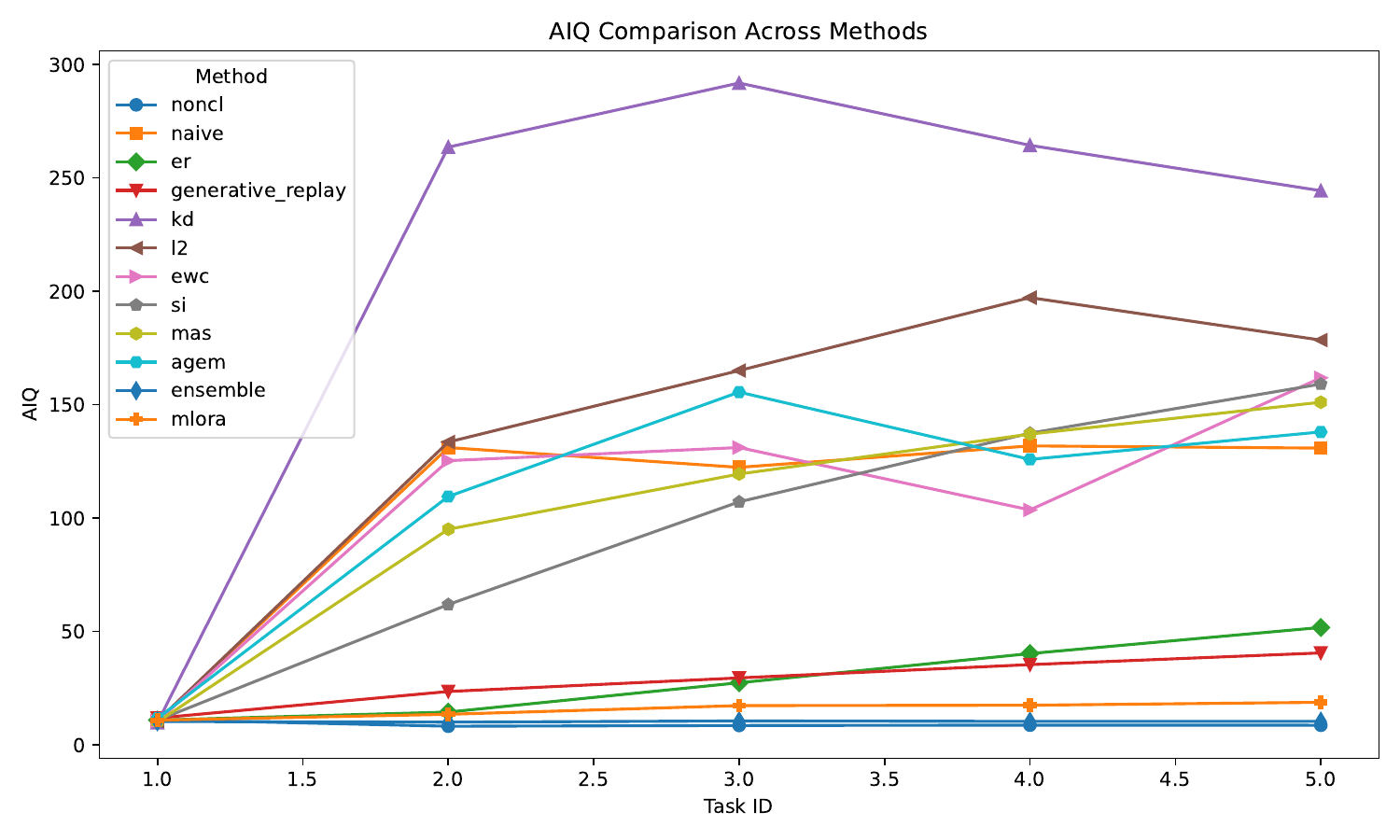}  
}    
\subfigure[CIFAR-10 (DDIM)] {    
\includegraphics[width=0.45\columnwidth]{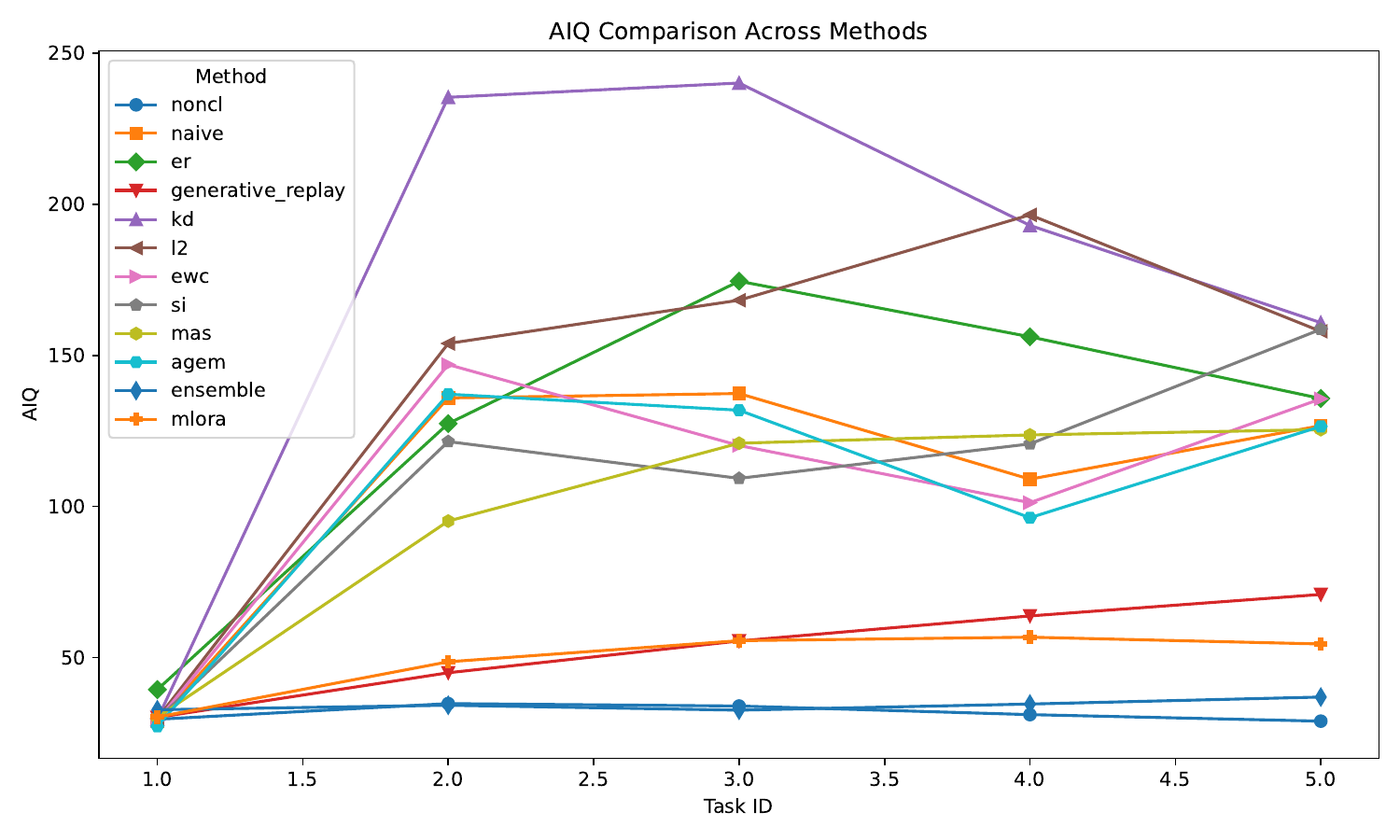}  
}    
\subfigure[Oxford-Flower (DDIM)] {    
\includegraphics[width=0.45\columnwidth]{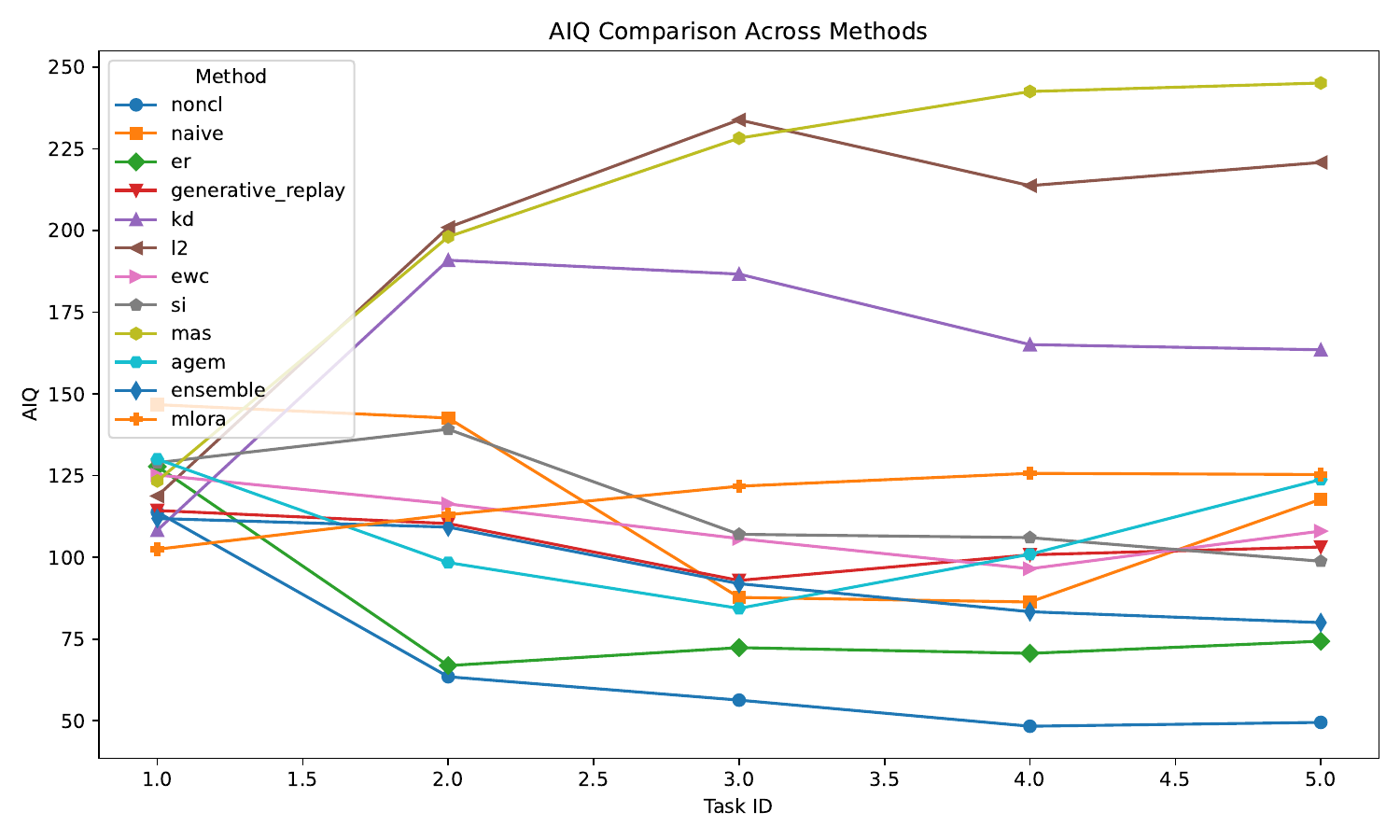}  
} 
\subfigure[CUB-Birds (DDIM)] {    
\includegraphics[width=0.45\columnwidth]{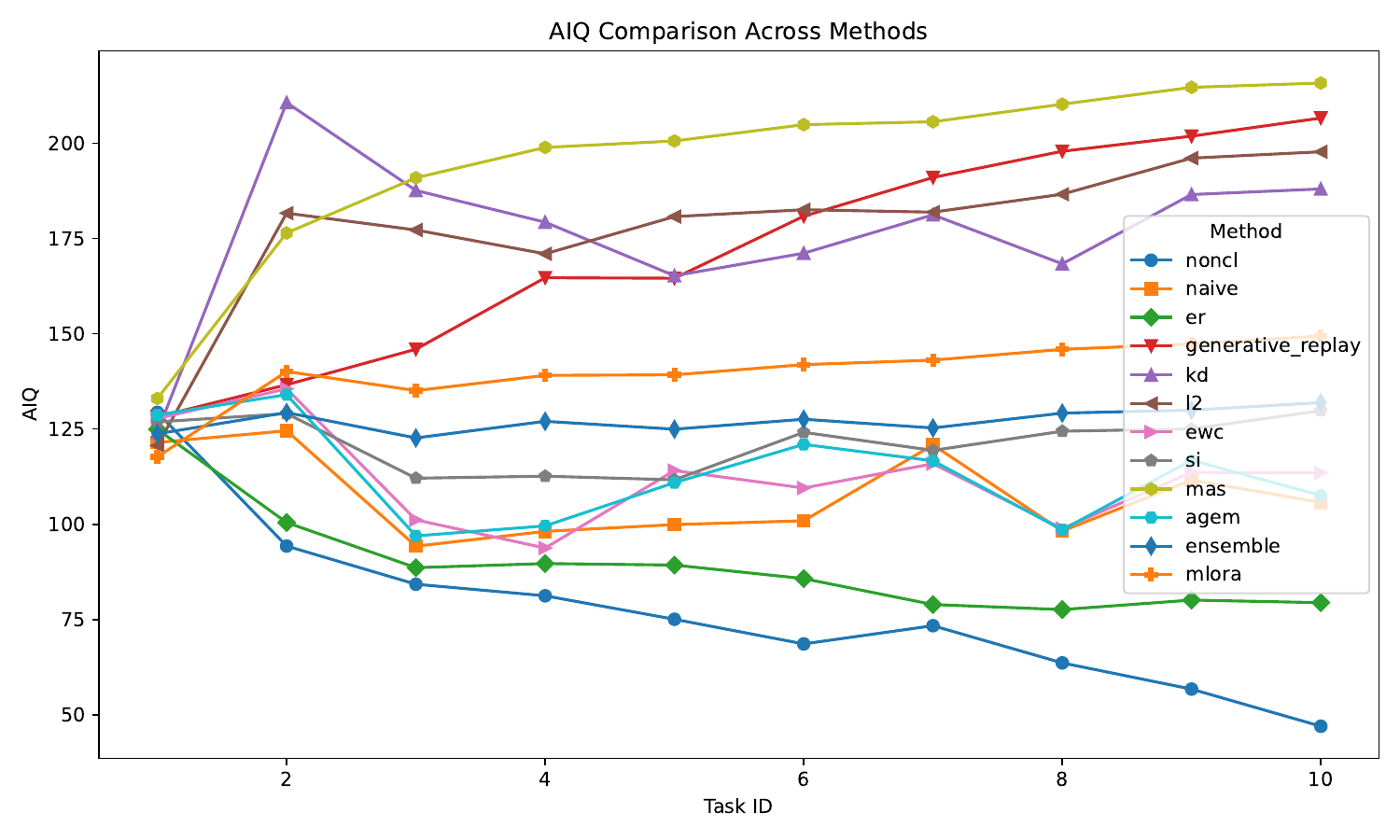}
}
\subfigure[Stanford-Cars (DDIM)] {    
\includegraphics[width=0.45\columnwidth]{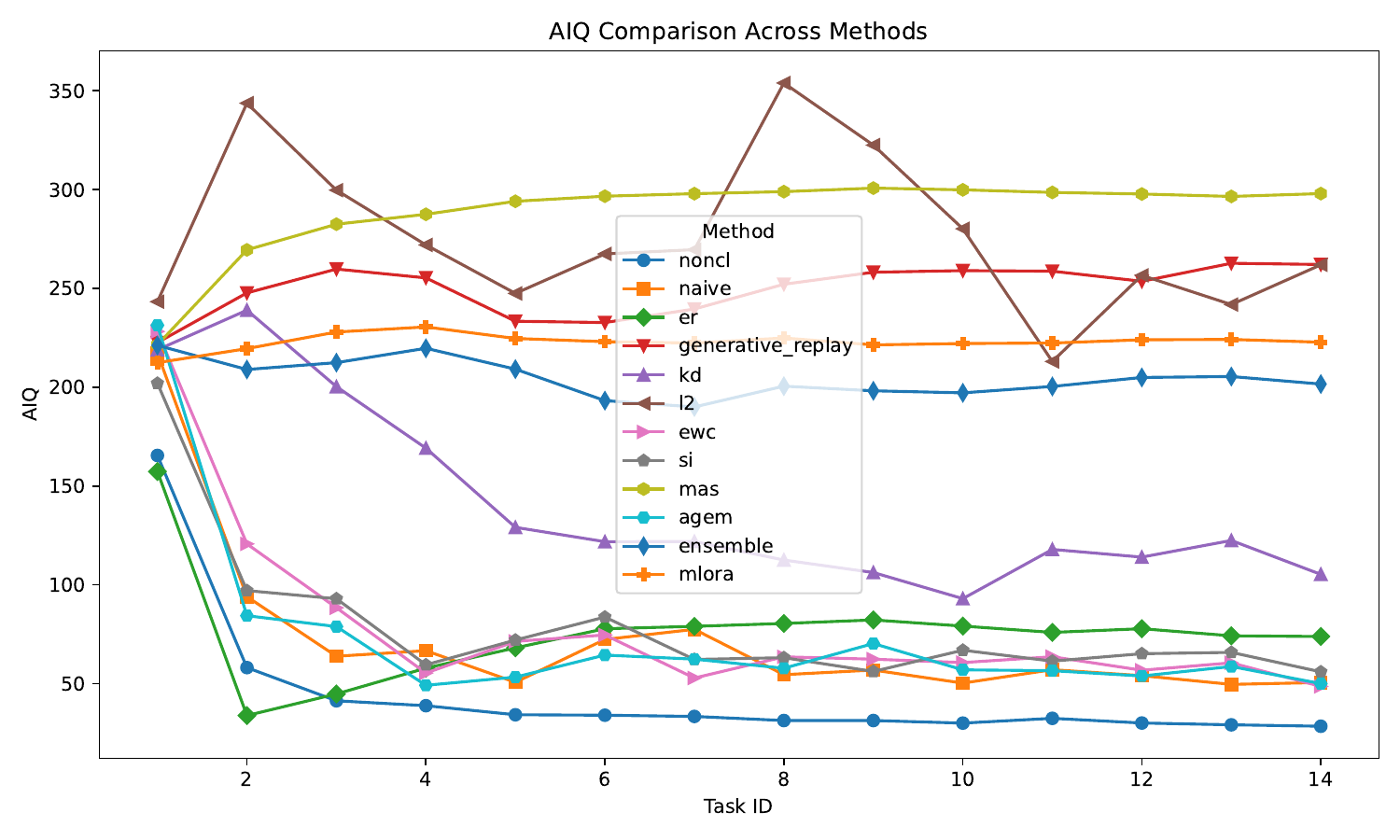}  
} 
\subfigure[ImageNet-1k (DDIM)] {    
\includegraphics[width=0.45\columnwidth]{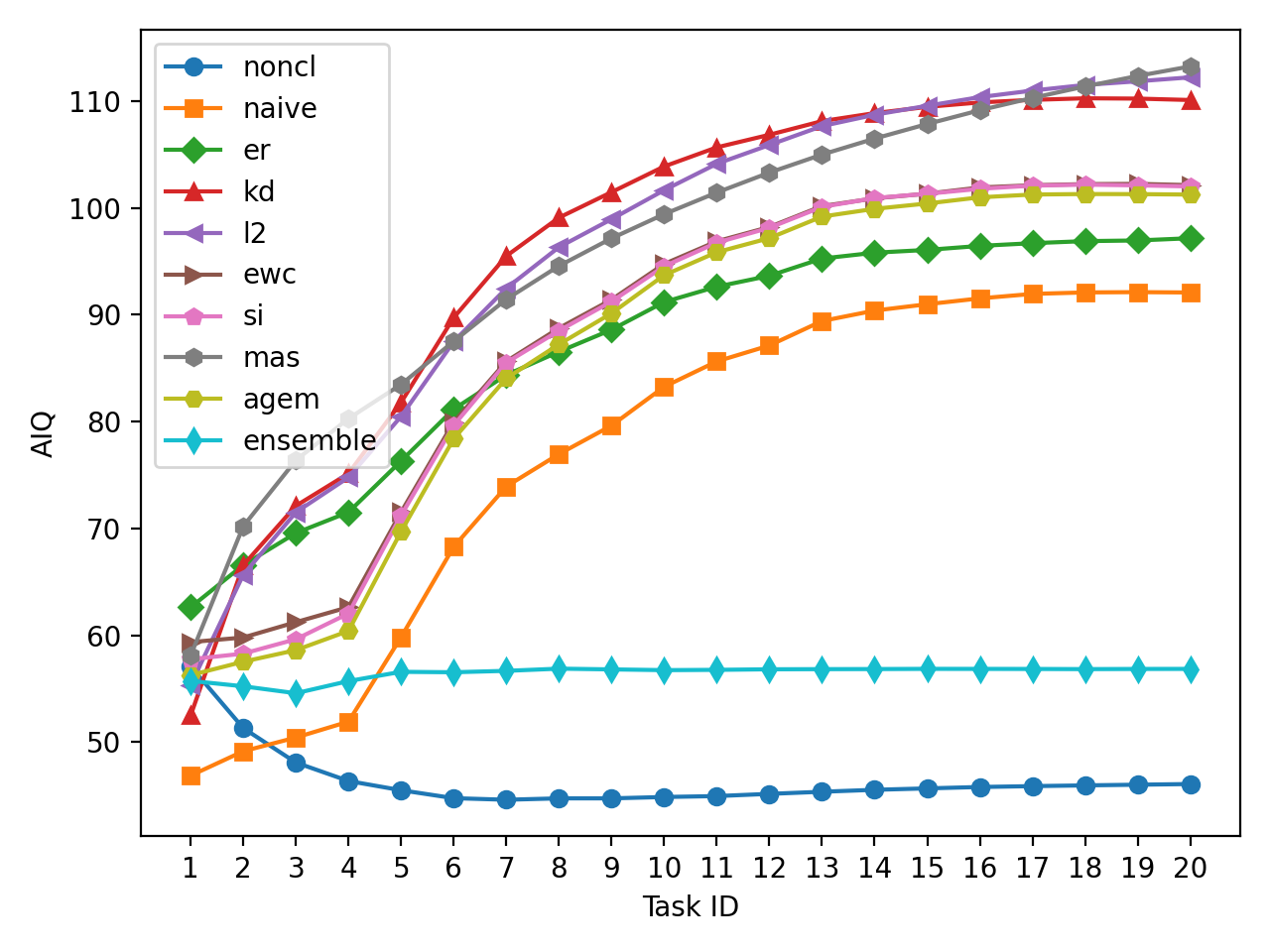}  
} 
\caption{The evolving performance curve of AIQ across various tasks on label-conditioned CLoG benchmarks. Here diffusion models are employed as the generator backbone.}
\label{fig:aiq_curve_ddim}
\end{figure}

To comprehensively investigate the performance of AIQ when increasing the number of learning tasks, we visualize its evolving curve in Fig.~\ref{fig:aiq_curve_gan} and \ref{fig:aiq_curve_ddim}, corresponding to GANs and diffusion models, respectively. Generally, the curve exhibits an upward trend, indicating a tendency to forget the knowledge of previous tasks. However, the AIQ metric gradually decreases on the CUB-Birds, Oxford-Flowers, and Stanford-Cars datasets, demonstrating that incremental learning of similar tasks enhances performance on previous tasks.

\begin{figure}[H]

\centering
\subfigure[DreamBooth CLIP avg] {    
\includegraphics[width=0.4\columnwidth]{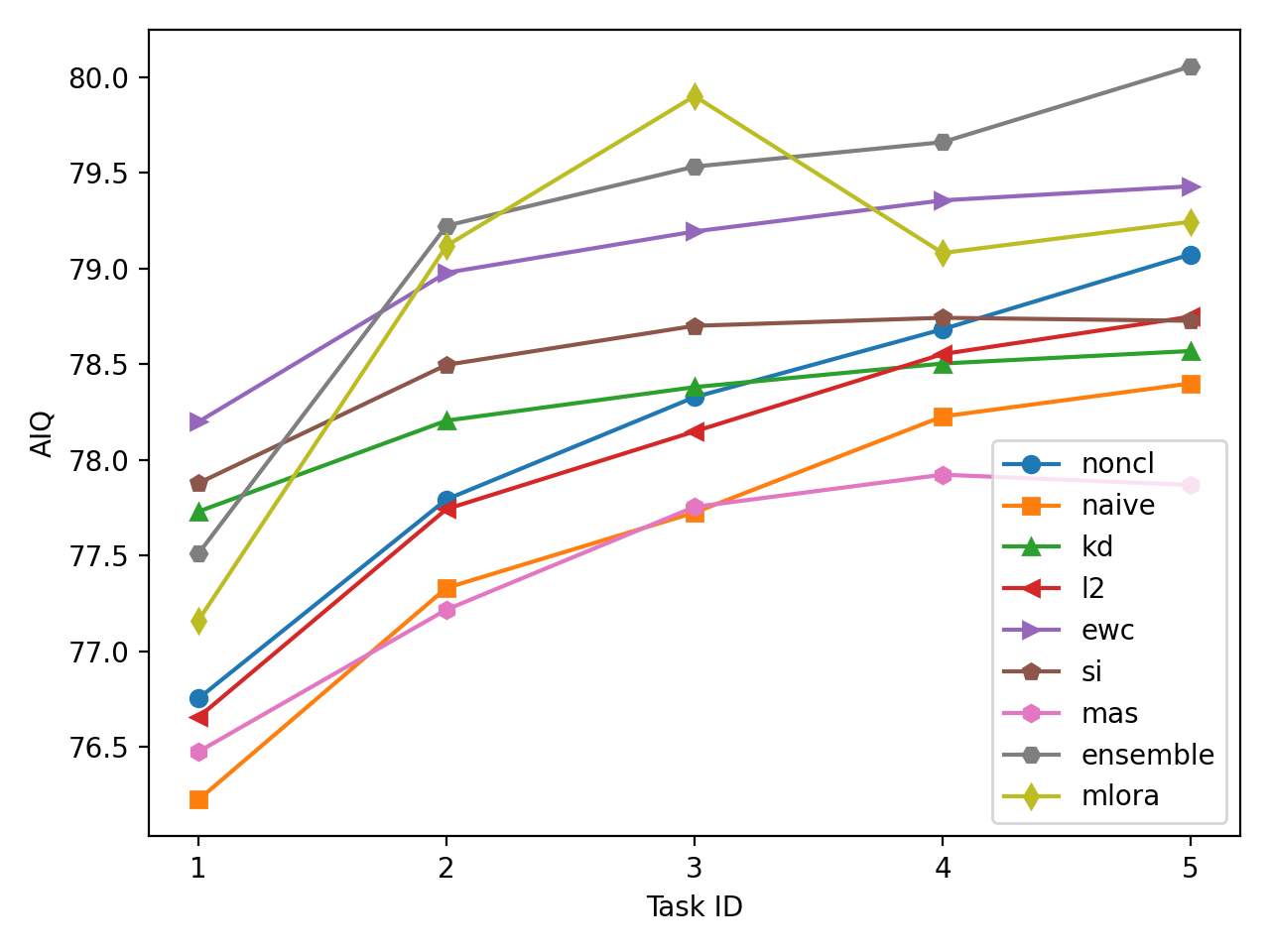}  
}    
\subfigure[Custom Diffusion CLIP avg] {    
\includegraphics[width=0.4\columnwidth]{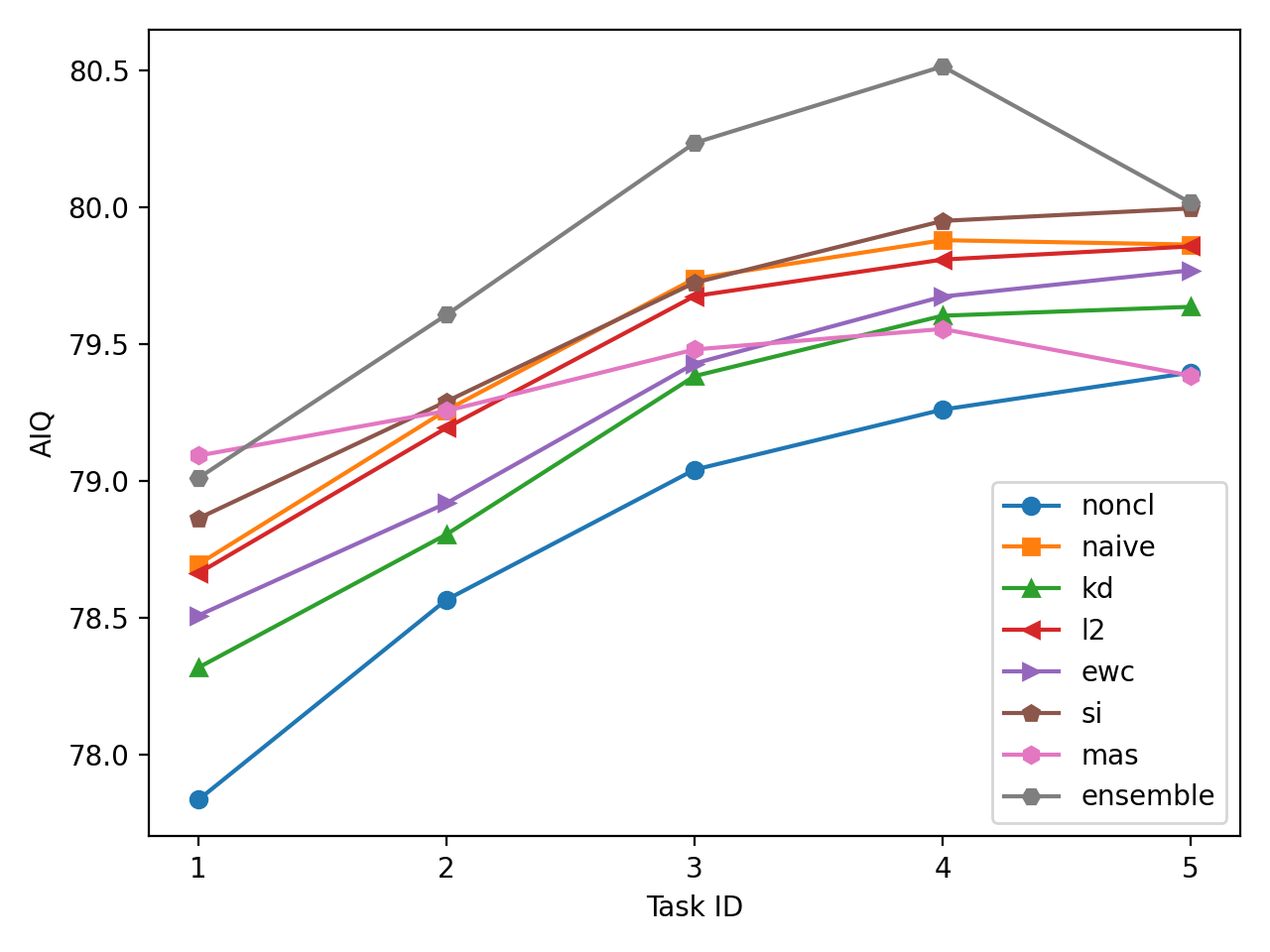}  
}  
\subfigure[DreamBooth DINO avg] {    
\includegraphics[width=0.4\columnwidth]{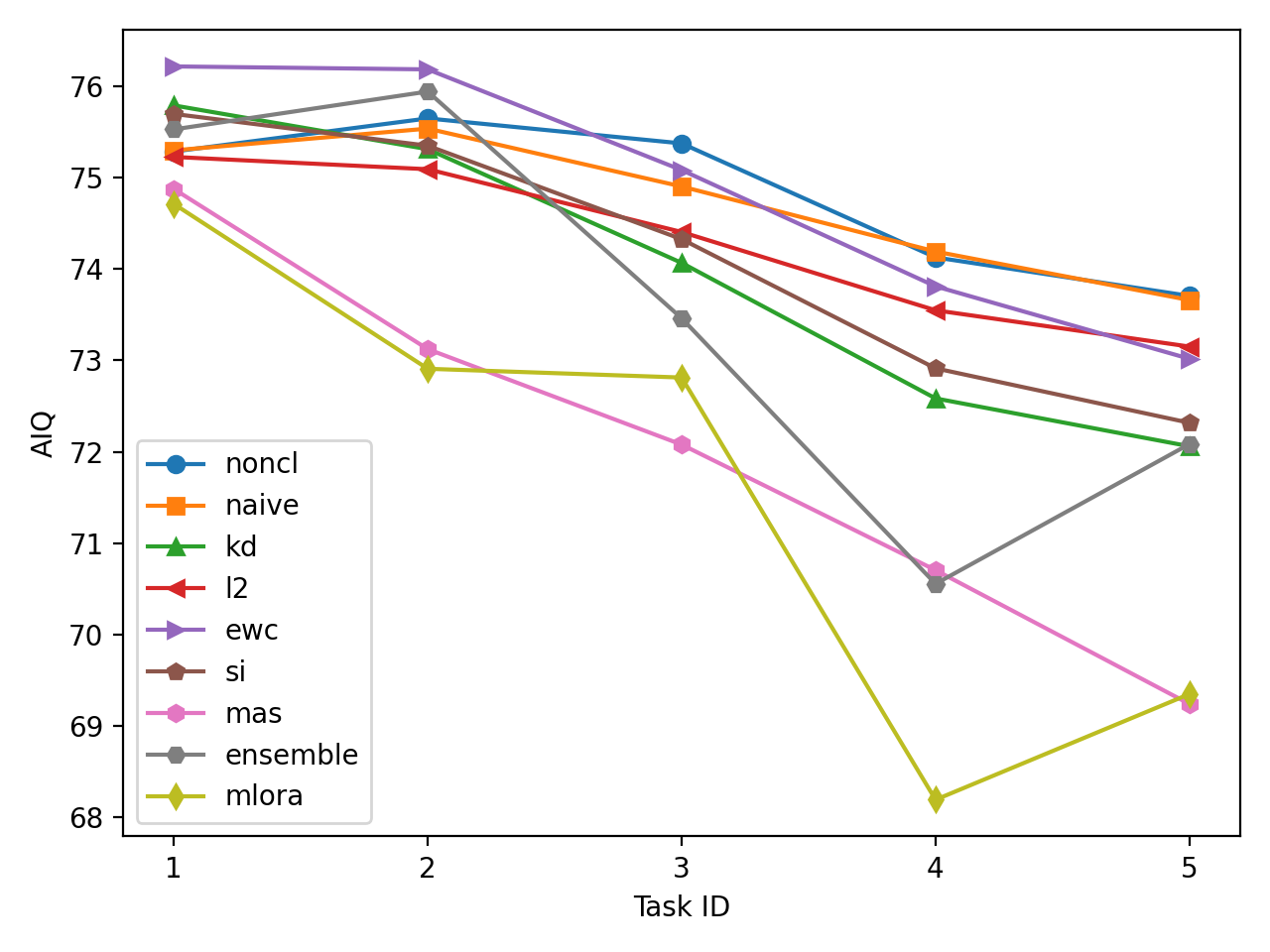}  
}    
\subfigure[Custom Diffusion DINO avg] {    
\includegraphics[width=0.4\columnwidth]{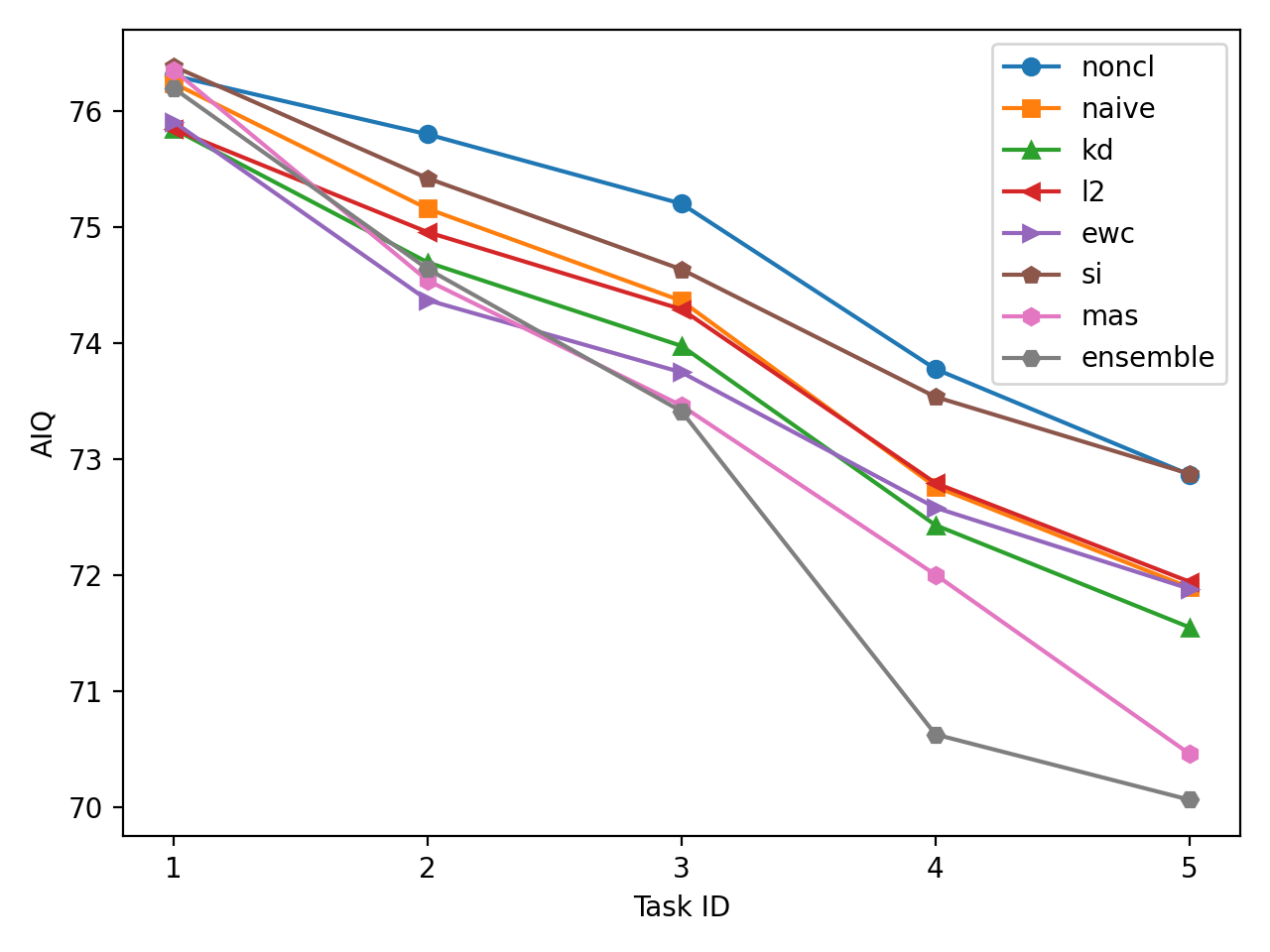}  
} 
\caption{The evolving performance curve of AIQ across various tasks on concept-conditioned CLoG benchmarks. We show the results on the Custom-Objects dataset utilizing DreamBooth~\citep{ruiz2023dreambooth} and Custom Diffusion~\citep{kumari2023multi}.}
\label{fig:custom_obj_aiq_customdiffusion}
\end{figure}

We also visualize evolving curve in Fig.~\ref{fig:custom_obj_aiq_customdiffusion} on DreamBooth and Custom Diffusion models, respectively. If we use CLIP avg to calculate AIQ, the curve exhibits an upward trend, indicating a tendency to forget the knowledge of previous tasks. On the other hand, if we use DINO avg to calculate AIQ, the metric gradually decreases for both the DreamBooth and Custom Diffusion Methods. This demonstrates that incremental learning of similar tasks enhances performance on previous tasks, which is consistent with the actual results of our generated images in Figures~\ref{fig:vis_dreambooth}, and \ref{fig:vis_custom_diffusion}. We prefer the AIQ calculated by DINO avg because DINO is not trained to ignore differences between subjects of the same class. Instead, its self-supervised training objective encourages the distinction of unique features of a subject or image.

\newpage

\subsection{Computational Budget Analysis}
\label{app:computation_analysis}

In this section, we present the memory consumption and training time for different baselines. Notice that we use a mix of different types of GPUs for different set of experiments: We use a single NVIDIA V100 GPU for GAN, a single NVIDIA RTX4090 GPU for DDIM, a single NVIDIA A100 GPU for ImageNet-1k, and a single NVIDIA A800 GPU for Costom-Objects. 

Now we present a detailed analysis of memory consumption of each baseline method. Methods that require replay samples (ER and A-GEM) introduce an auxiliary replay memory to retain previous data. In addition, all regularization-methods require storing the parameters of a teacher model, doubling the total number of model parameters. Among these techniques, EWC, SI and MAS require additional computation for determining the loss weight of each parameter, resulting in a threefold increase in the model's parameter count. Lastly, the ensemble method increases memory consumption by a factor of $T$ (where $T$ represents the total number of tasks), while C-LoRA introduces $T$ additional trainable weights to facilitate conditional generation.

\begin{table}[H]
\centering
\caption{Memory Consumption of label-conditioned CLoG benchmarks: Measured in number of parameters (M) }
\resizebox{\textwidth}{!}{%
\begin{tabular}{cccccccc}
    \toprule
    & \textbf{MNIST} & \textbf{Fashion-MNIST} & \textbf{CIFAR-10} & \textbf{CUB-Birds} & \textbf{Oxford-Flowers} & \textbf{Stanford-Cars} & \textbf{ImageNet}  \\
    \midrule
    \midrule
    \multicolumn{7}{l}{\textbf{- GAN}} \\
     Non-CL & 43.30 & 43.30  & 43.30  & 60.55 & 60.55 & 60.55  & NA  \\
     NCL  & 43.30  & 43.30 & 43.30  & 60.55 & 60.55 & 60.55 & NA \\
     ER   & 43.92  & 43.92  & 43.92  & 70.38 & 70.38  & 70.38 & NA \\
     GR   & 86.60  & 86.60  & 86.60  & 121.10  & 121.10  & 121.10 & NA \\
     KD   & 86.60  & 86.60  & 86.60   & 121.10  & 121.10  & 121.10 &NA  \\
     L2   & 86.60  & 86.60  & 86.60  & 121.10  & 121.10  &121.10  &NA \\
     EWC  & 129.91  &  129.91 &  129.91 & 181.65  & 181.65  & 181.65 &NA \\
     SI  & 129.91  & 129.91  & 129.91  & 181.65  &  181.65  & 181.65 &NA  \\
     MAS & 129.91  &  129.91 & 129.91  & 181.65  & 181.65 & 181.65 &NA \\
     A-GEM   & 43.92  & 43.92  & 43.92  & 70.38  & 70.38  & 70.38 &NA \\
     Ensemble  & 216.52  & 216.52  & 216.52  & 605.49  & 302.75 & 847.69 & NA \\
    \bottomrule
    \toprule
    \multicolumn{7}{l}{\textbf{- Diffusion Model}} \\
     Non-CL  & 37.20 & 37.20 &   37.20 &  85.51  &  85.51  & 85.51 &   346.09  \\
     NCL  & 37.20  &  37.20  &  37.20  &  85.51  & 85.51   &   85.51  &  346.09  \\
     ER  &  37.40  & 37.40   &   37.40  &  88.71  &   88.71  &   88.71 &  348.55  \\
     GR  &  74.40  &  74.40  &  74.40  &  171.02   & 171.02 &   171.02 & NA \\
     KD  &  74.40  &  74.40  &  74.40  &  171.02  &  171.02  & 171.02   &   692.18 \\
     L2  &  74.40  & 74.40   &   74.40 &  171.02  &  171.02  &   171.02 &  692.18   \\
     EWC &  111.60  &  111.60  & 111.60   &  256.53  &  256.53  &  256.53  &  778.71   \\
     SI  &  111.60  &  111.60  &  111.60  &  256.53  & 256.53   &  256.53  &  778.71   \\
     MAS &  111.60  &  111.60  &  111.60  &  256.53  &  256.53  &  256.53  &  778.71  \\
     A-GEM &  37.40  &  37.40  &  37.40  &  88.71  & 88.71   &   88.71 &  348.55   \\
     Ensemble  &  186.00  & 186.00  &  186.00  &  855.10  &  427.55  & 1197.74   &  6921.84  \\
     C-LoRA  &  43.00  &  43.00  &  43.00  &  103.11  &  94.31  &  110.15  &  476.45  \\
    \bottomrule
\end{tabular}
}\label{tab:memory_analysis}
\end{table}

\begin{table}[H]
\centering
\caption{\label{tab:memory_custom} Memory Consumption of concept-conditioned CLoG benchmarks: Measured in number of parameters (M)}

\resizebox{\textwidth}{!}{%
\begin{tabular}{ccccccccccc}
    \toprule
    Metric & Model & \textbf{NCL} & \textbf{Non-CL} & \textbf{KD} & \textbf{L2} & \textbf{EWC} & \textbf{SI} & \textbf{MAS} & \textbf{Ensemble} & \textbf{C-LoRA} \\
    \midrule
    \midrule
    \multirow{2}{*}{All Params} & DreamBooth  & 1016.84 & 1016.84 & 1953.9 & 1953.9 & 2890.97 & 2890.97 & 2890.97  & 5084.2 & 1068.84  \\
    & Custom Diffusion & 1035.12 & 1035.12 & 1990.47 & 1089.59 & 1144.06 & 1144.06 & 1144.06 & 5175.6 & - \\
    \midrule
    \multirow{2}{*}{Train Params} & DreamBooth  & 937.06 & 937.06 & 937.06 & 937.06 & 937.06 &937.06 & 937.06  & 4685.3 & 10.4 \\
    & Custom Diffusion & 54.47 & 54.47 & 54.47 & 54.47 & 54.47 & 54.47 & 54.47 & 272.35 & - \\
    \bottomrule
\end{tabular}
}
\end{table}

\begin{table}[H]
\centering
\caption{\label{tab:runtime} \textbf{Training Time of different baselines on the label-conditioned CLoG benchmark}: Measured in hours over all tasks}
\resizebox{\textwidth}{!}{%
\begin{tabular}{cccccccc}
    \toprule
    & \textbf{MNIST} & \textbf{Fashion-MNIST} & \textbf{CIFAR-10} & \textbf{CUB-Birds} & \textbf{Oxford-Flowers} & \textbf{Stanford-Cars} & \textbf{ImageNet}  \\
    \midrule
    \midrule
    \multicolumn{7}{l}{\textbf{- GAN}} \\
    Non-CL & 37.65  & 37.78 & 32.65  & 80.32  & 18.93 & 79.81 & NA  \\
     NCL  & 12.81  & 12.62  & 10.82 & 14.78  & 6.26  & 10.64 & NA \\
     ER   & 16.32  & 15.21  & 12.66 & 16.58  & 7.10 & 11.81 & NA  \\
     GR   & 15.76  & 14.96  & 12.78  & 17.30  & 7.18  & 12.20 & NA \\
     KD   & 15.64 & 15.48  & 13.38  & 16.94  & 7.06  & 12.07 & NA \\
     L2   & 12.92  & 12.69  & 10.74  & 14.96  & 6.38  & 10.51 & NA \\
     EWC  & 15.12  & 14.92  & 12.62  & 20.72  & 8.54  & 15.06 & NA  \\
     SI  & 16.84  & 16.52  & 14.02  & 25.04  &  10.14  & 17.92 & NA \\
     MAS & 15.16  & 15.12  & 12.74  & 21.17  & 8.72 & 14.93 & NA  \\
     A-GEM & 15.68 & 15.52 & 13.42  & 19.19  & 8.10  & 13.58  & NA  \\
     Ensemble & 12.64 & 12.58 & 10.93 & 14.51  & 6.34  & 10.51  & NA \\ 
    \bottomrule
    \toprule
    \multicolumn{7}{l}{\textbf{- Diffusion Model}} \\
     Non-CL  & 16.11   &  15.30  & 12.88   &  55.01  &  13.33   &   58.33 & 3953.34 \\
     NCL  & 2.83  &  3.19  &  3.17  &  3.33  &  3.33  &  6.38   &  103.84  \\
     ER  &  2.89  &  2.83  &   2.55  &  9.83  &   5.61  &  5.53  &  104.44  \\
     GR  &  6.9  &  8.83  &  10.22  &  12.64  & 7.89 & 16.80   & NA \\
     KD  &  3.56  &  2.56  &  3.11  &  7.12  &  6.88   & 8.52 &  135.67  \\
     L2  &  2.94  &  3.72  &  2.73  &  3.98  &  3.22  &  4.89  &  105.33 \\
     EWC &  3.44  &  2.65  &  2.94  &  9.09  &   7.38 &  8.89  &  121.86  \\
     SI  &  3.89  &  4.64  &  5.05  &   10.72    &  6.64  &  4.59  &  145.86  \\
     MAS &  3.96  &  4.14  &  2.89  &  7.37  &   5.94  & 7.22  &  109.44  \\
     A-GEM &  3.87  &  3.89  &  2.87  &  13.92  &  12.60  &  10.50  &  104.94  \\
     Ensemble  &  3.34  &  2.67  &  2.22  &  6.12  &  4.94  &  5.83  &  102.31  \\
     C-LoRA  &  2.04  &  2.37  &  2.72  &  4.10  &  2.78  &   3.29 &  128.88  \\    
    \bottomrule
\end{tabular}
}
\end{table}

\begin{table}[H]
\centering
\caption{\textbf{Training Time of different baselines on the concept-conditioned CLoG benchmark}: Measured in minutes over all tasks}

\resizebox{\textwidth}{!}{%
\begin{tabular}{cccccccccc}
    \toprule
    Model & \textbf{NCL} & \textbf{Non-CL} & \textbf{KD} & \textbf{L2} & \textbf{EWC} & \textbf{SI} & \textbf{MAS} & \textbf{Ensemble} & \textbf{C-LoRA} \\
    \midrule
    \midrule
    DreamBooth  & 26.76 & 6.29 & 32.54 & 33.09 & 69.79 & 172.83 & 38.68  & 31.58 & 25.66  \\
    Custom Diffusion & 12.40 & 4.02 & 16.65 & 12.49 & 13.33 & 15.45 & 12.25 & 10.15 & - \\
    \bottomrule
\end{tabular}
}
\label{tab:runtime_Customobj} 
\end{table}

\newpage

\newpage 

\section{Implementation Details}
\label{app:implementation_details}

\subsection{Class description}

We list the class description for each label index for each dataset as follows.

\begin{itemize}
    \item \textbf{MNIST} (10 classes)
        \begin{itemize}
            \item digit `0', digit `1', digit `2', digit `3', digit `4', digit `5', digit `6', digit `7', digit `8', digit `9'
        \end{itemize}
    \item \textbf{FasionMNIST} (10 classes)
        \begin{itemize}
            \item T-shirt/top, Trouser, Pullover, Dress, Coat, Sandal, Shirt, Sneaker, Bag, Ankle boot
        \end{itemize}
    \item \textbf{CIFAR-10} (10 classes)
        \begin{itemize}
            \item airplane, automobile, bird, cat, deer, dog, frog, horse, ship, truck
        \end{itemize}
    \item \textbf{ImageNet-1k} (1,000 classes)
        \begin{itemize}
            \item tench Tinca tinca, goldfish Carassius auratus, great white shark white shark man-eater man-eating shark Carcharodon carcharias, tiger shark Galeocerdo cuvieri, hammerhead hammerhead shark, electric ray crampfish numbfish torpedo, stingray, cock, hen, ostrich, {\color{google_blue}(\textbf{... 980 classes are omitted)}} coral fungus, agaric, gyromitra, stinkhorn carrion fungus, earthstar, hen-of-the-woods hen of the woods Polyporus frondosus Grifola frondosa, bolete, ear spike capitulum, toilet tissue toilet paper bathroom tissue
        \end{itemize}

        \item \textbf{Oxford-Flowers} (103 classes)
        \begin{itemize}
            \item alpine sea holly, anthurium, artichoke, azalea, ball moss, balloon flower, barbeton daisy, bearded iris, bee balm, bird of paradise, {\color{google_blue}(\textbf{... 980 classes are omitted)}}, toad lily, tree mallow, tree poppy, trumpet creeper, wallflower, water lily, watercress, wild pansy, windflower, yellow iris
        \end{itemize}
    \item \textbf{CUB-Birds} (200 classes)
    \begin{itemize}
        \item Black footed Albatross, Laysan Albatross, Sooty Albatross, Groove billed Ani, Crested Auklet, Least Auklet, Parakeet Auklet, Rhinoceros Auklet, Brewer Blackbird, Red winged Blackbird, {\color{google_blue}(\textbf{... 180 classes are omitted)}}, Red headed Woodpecker, Downy Woodpecker, Bewick Wren, Cactus Wren, Carolina Wren, House Wren, Marsh Wren, Rock Wren, Winter Wren, Common Yellowthroat
    \end{itemize}
    \item \textbf{Stanford-Cars} (196 classes)
    \begin{itemize}
        \item AM General Hummer SUV 2000, Acura RL Sedan 2012, Acura TL Sedan 2012, Acura TL Type-S 2008, Acura TSX Sedan 2012, Acura Integra Type R 2001, Acura ZDX Hatchback 2012, Aston Martin V8 Vantage Convertible 2012, Aston Martin V8 Vantage Coupe 2012, Aston Martin Virage Convertible 2012, {\color{google_blue}(\textbf{... 176 classes are omitted)}} Toyota Camry Sedan 2012, Toyota Corolla Sedan 2012, Toyota 4Runner SUV 2012, Volkswagen Golf Hatchback 2012, Volkswagen Golf Hatchback 1991, Volkswagen Beetle Hatchback 2012, Volvo C30 Hatchback 2012, Volvo 240 Sedan 1993, Volvo XC90 SUV 2007, smart fortwo Convertible 2012
    \end{itemize}
    \item \textbf{Custom-Objects} (5 concepts)
    \begin{itemize}
            \item dog, duck toy, cat, backpack, bear plushie
    \end{itemize}
\end{itemize}

\newpage

\subsection{Random class ordering}
\label{app:random_class_ordering}

\cref{tab:Random_class_order} shows the different class orderings we used on different dataset. Due to space limitation, we only show the ordering of datasets with small class sequences. For large sequences, we refer readers to check our supplemental materials for details. The first class sequence is set as the sequence of class ordering from the original dataset, while the other sequences are generated via random shuffling.

\begin{table}[H]
    \centering
    \caption{The random class ordering used in our benchmarks. The full orderings can be found in our supplemental materials.}
    \vspace{1ex}
    \resizebox{0.8\textwidth}{!}{\begin{tabular}{ccc}
    \toprule
        \textbf{Dataset} &  \textbf{Class order} & \textbf{Class sequence} \\
        \hline\Tstrut
        \multirow{5}{*}{MNIST, FasionMNIST, CIFAR-10} & 1 & 0, 1, 2, 3, 4, 5, 6, 7, 8, 9 \\
        & 2 & 3, 9, 1, 8, 0, 2, 6, 4, 5, 7 \\
        & 3 & 6, 0, 2, 8, 1, 9, 7, 3, 5, 4 \\
        & 4 & 2, 6, 1, 5, 9, 8, 0, 4, 3, 7 \\
        & 5 & 1, 5, 7, 2, 0, 3, 4, 6, 8, 9 \\
        \hline\Tstrut
        \multirow{5}{*}{Custom-Objects} & 1 & 0, 1, 2, 3, 4 \\
        & 2 & 4, 3,  1,  0,  2 \\
        & 3 & 4, 2,  1, 3, 0 \\
        & 4 & 1, 4, 0, 2, 3 \\
        & 5 & 2, 1, 0, 3, 4 \\
        \bottomrule
    \end{tabular}}
    \label{tab:Random_class_order}
\end{table}

\newpage

\subsection{Label-conditional CLoG}

\paragraph{Implementation Details of StyleGAN2}
We employ the official PyTorch implementation of StyleGAN2-ADA~\citep{karras2020stylegan2ada} as our backbone. The detailed hyperparameters used in our experiments are presented in Table~\ref{tab:hyper_stylegan2}. All training runs are performed for 200 epochs using a single NVIDIA Tesla V100 GPU. We utilize six datasets with different image resolutions: 32x32 pixels (MNIST, Fashion-MNIST, CIFAR-10) and 128x128 pixels (CUB-Birds, Oxford-Flowers, Stanford-Cars). Two variants of StyleGAN2 are implemented to generate images at these resolutions, termed Ours-S and Ours-L, respectively.

We use a minibatch size of 64 for Ours-S and 16 for Ours-L. For the replay-based methods in CLoG, we construct a replay memory containing 200 samples from previous tasks, with the replay size set to one-fourth of the minibatch size (16 for Ours-S and 4 for Ours-L). Following the configuration for CIFAR-10 in the original paper~\citep{karras2020stylegan2ada}, we use 512 feature maps for all layers. The weight of the $R_1$ regularization is set to $\gamma = 0.01$ for Ours-S and $\gamma = 1$ for Ours-L. Additionally, we opt for a more expressive model architecture for the mapping network and the discriminator when synthesizing images at 128x128 pixels. Specifically, we increase the depth of the mapping network from 2 to 8 and enable residual connections in the discriminator. For simplicity, we omit several techniques that are irrelevant to CL capability used in the original paper, including adaptive discriminator augmentation (ADA), style mixing, path length regularization, and exponential moving average (EMA).

\paragraph{Implementation Details of DDIM} We employ the Huggingface diffuser\footnote{\url{https://huggingface.co/docs/diffusers}} implementation of DDIM~\citep{ddim} in our codebase. The detailed hyperparameters used in our experiments are presented in~\cref{tab:hyper_ddim}. All training runs are performed for 200 epochs using a single NVIDIA RTX 4090 GPU for MNIST, Fasion-MNIST, CIFAR-10, CUB-Birds, Oxford-Flowers, Stanford-Cars, and a single NVIDIA A100 GPU for the large-scale ImageNet-1k dataset. Three variants of DDIM are implemented to generate images at small resolution (32x32), meddium resolution (64x64), large resolution (128x128), termed Ours-S, Ours-M, Ours-L, respectively. We use a minibatch size of 256 for Ours-S, 320 for Ours-M, 32 for Ours-L. For the replay-based methods, we maintain a replay buffer containing 200 samples from previous tasks with replay size as 64 for Ours-S and Ours-M, and 8 for Ours-L. Following~\citet{Nichol2021ImprovedDD}, we use different numbers of channel and UNet blocks for Ours-S, Ours-M, and Ours-L. 

\begin{table}[H]
    \centering
    \caption{Hyperparameters of StyleGAN2~\citep{karras2020stylegan2ada} used in our CLoG experiments.}
    \vspace{1ex}
        \begin{tabular}{ccc}
            \toprule
            \textbf{Parameter} & \textbf{Ours-S} & \textbf{Ours-L} \\ 
            \midrule
            Resolution & 32$\times$32 & 128$\times$128 \\
            Training epochs & 200 & 200 \\
            Minibatch size & 64 & 16 \\
            Minibatch stddev & 32 & 32 \\
            Replay size & 64 & 16 \\
            Memory size & 200 & 200 \\
            \midrule
            Feature maps & 512 & 512 \\
            Learning rate $\eta \times 10^3$ & 2.5 & 2.5 \\
            $R_1$ regularization$\gamma$ & 0.01 & 1 \\
            \midrule
            Mapping net depth & 2 & 8 \\
            Resnet D & - & \checkmark \\
            \bottomrule
        \end{tabular}
    \label{tab:hyper_stylegan2}
\end{table}

\begin{table}[htbp]
    \centering
    \caption{Hyperparameters of DDIM~\citep{ddim} used in our CLoG experiments.}
    \vspace{1ex}
        \begin{tabular}{cccc}
            \toprule
            \textbf{Parameter} & \textbf{Ours-S} & \textbf{Ours-M} & \textbf{Ours-L} \\ 
            \midrule
            Resolution & 32$\times$32  & 64$\times$64  & 128$\times$128 \\
            Training epochs & 200 & 100 & 200 \\
            Minibatch size & 256 & 320 & 32 \\
            Replay size & 64 & 64 & 8 \\
            Memory size & 200 & 5000 & 200 \\
            \midrule
            Learning rate $\eta \times 10^3$ & 2.0 & 2.0 & 1.0 \\
            Learning rate warm-up steps & 500 & 500 & 500 \\
            Weight decay & 0.0 & 0.0 & 0.0 \\
            \# Unet blocks ($\times$2) & 4 & 4 & 5 \\
            Unet blocks dimension (the largest) & 256 & 512 & 512 \\
            Dropout & 0.1 & 0.1 & 0.1 \\
            Time embedding dimension & 512 & 512 &512 \\
            \bottomrule
        \end{tabular}
    \label{tab:hyper_ddim}
\end{table}

\newpage

\subsection{Concept-conditional CLoG}

\paragraph{Evaluation Metrics}For concept-conditioned CLoG, we follow DreamBooth~\citep{ruiz2023dreambooth} and Custom Diffusion~\citep{kumari2023multi} to evaluate the alignment between generated image and the provided concept, and the text prompts, respectively. To assess subject fidelity, we use two metrics: CLIP Image Alignment and DINO Image Alignment. CLIP Image Alignment measures the average pairwise cosine similarity between the CLIP embeddings of generated and real images. Similarly, the DINO metric calculates the average pairwise cosine similarity between the ViT-S/16 DINO embeddings of generated and real images. To evaluate prompt fidelity, we compute the average cosine similarity between the CLIP embeddings of the text prompt and the images, which we refer to as CLIP Text Alignment. The averages of the image alignment and text alignment scores are combined to derive a single quality metric for straightforward comparison, labeled respectively as DINO avg and CLIP avg. We evaluate each task using 20 text prompts, generating 50 samples per prompt. This results in a total of 1,000 images generated for each task.

\paragraph{Implementation Details} DreamBooth and Custom Diffusion both utilize generated by initial stable-diffusion-v1-4, rather than real, category images to calculate the prior loss for their training processes. 200 regularization images are preemptively created using a DDPM sampler over 50 steps with the prompt 'photo of a \{category\}'. We use DDPM sampling with 50 steps and a classifier-free guidance scale of 6 for both DreamBooth and Custom Diffusion. All training runs are performed using a single NVIDIA A800 GPU. More details can be found in 
~\cref{tab:hyper_CloG}

DreamBooth adheres to the same data augmentation strategies as Custom Diffusion, which will be introduced later, to ensure a balanced comparison. It trains by fine-tuning both a text transformer and a U-net diffusion model. This training uses a batch size of 1 and a learning rate of 2e-6, which is maintained constant regardless of the number of GPUs or batch size. For generating target images, DreamBooth employs a text prompt formatted as 'photo of a [V] \{category\}', where '[V]' is replaced with a rarely used token from a specific set ('sks', 'phol', 'oxi', 'mth', 'nigh').  Each training task undergoes 800 steps. Conversely, Custom Diffusion uses a slightly different approach by setting the batch size at 2 and a scaled learning rate of 2e-5, adjusted according to the batch size to an effective rate of 4e-5. It trains each task for only 250 steps. During training, target images undergo random resizing: they are enlarged to between 1.2 and 1.4 times their original size every third iteration, with phrases like 'zoomed in' or 'close up' added to the text prompts. Other times, images are resized to between 0.4 and 1.0 times their original size; when the resizing ratio is below 0.6, terms like 'far away' or 'very small' are incorporated into the prompts, focusing loss propagation only within the valid image regions. The training captions, such as 'photo of a V* dog', incorporate a rare token ('ktn', 'pll', 'ucd', 'mth', 'nigh'), with both the token embedding and the cross-attention parameters being optimized during the training process.

\begin{table}[htbp]
    \centering
    \caption{Hyperparameters used in our Concept-conditional CLoG.}
    \vspace{1ex}
    \resizebox{0.8\textwidth}{!}{
        \begin{tabular}{cccc}
            \toprule
            \textbf{Parameter} & \textbf{DreamBooth} & \textbf{DreamBooth-C-LoRA} & \textbf{Custom Diffsuion} \\ 
            \midrule
            Resolution & 512$\times$512  & 512$\times$512  & 512$\times$512 \\
            Training steps & 800 & 800 & 250 \\
            Minibatch size & 1 & 1 & 2 \\
            Inference steps & 50 & 50  & 50  \\
            \midrule
            Learning rate & 2e-6 & 5e-5 & 2e-5 \\
            Learning rate scheduler  & constant & constant & constant \\
            Learning rate warm-up steps  & 0 & 0 & 0 \\
            Prior loss & \checkmark &\checkmark & \checkmark \\
            Prior class images & 200 & 200 & 200 \\
            Data Augmentation & \checkmark &\checkmark & \checkmark \\
            \bottomrule
        \end{tabular}
    }
    \label{tab:hyper_CloG}
\end{table}
\newpage

\section{Impact Statement}
\label{sec:impact_statement}

Our work is essential as it contributes to the advancement of generative models' continuous learning, potentially benefiting human lives and society. 
Our method approaches a general problem and will not have any direct negative impact or be misused in specific domains as long as the task itself is safe, ethical, and fair. 
The risks of these models should be evaluated based on the specific deployment context, including training data, existing guardrails, deployment environment, and authorized access.

\end{document}